\documentclass[nonacm, acmsmall]{acmart}   
\AtBeginDocument{%
  }

\authorsaddresses{}
\usepackage{author-style}

\begin{document}

\title[An Interpretable Recommendation Model for Psychometric Data]{An Interpretable Recommendation Model for Psychometric Data, With an Application to Gerontological Primary Care}

\author{Andre Paulino de Lima}
\email{andre.p.lima@usp.br}
\orcid{0000-0002-3148-6686}
\affiliation{
  \institution{Institute of Mathematics and Computer Science, University of S\~ao Paulo}
  \city{S\~ao Carlos}
  \country{Brazil}
}

\author{Paula Castro}
\authornote{The author led the effort to collect the WHOQOL dataset.}
\email{castro@ufscar.br}
\orcid{0000-0002-0363-0871}
\affiliation{%
  \institution{Department of Gerontology, S\~ao Carlos Federal University}
  \city{S\~ao Carlos}
  \country{Brazil}
}

\author{Suzana Carvalho Vaz de Andrade}
\orcid{0000-0002-8908-6810}
\author{Rosa Maria Marcucci}
\orcid{0000-0002-2238-8750}
\author{Ruth Caldeira de Melo}
\orcid{0000-0002-9713-8617}
\email{ruth.melo@usp.br}
\authornote{These authors led the effort to collect the AMPI-AB dataset. Ruth has revised the manuscript for gerontological concepts.}
\affiliation{%
  \institution{School of Arts, Sciences and Humanities, University of S\~ao Paulo}
  \city{S\~ao Paulo}
  \country{Brazil}
}

\author{Marcelo Garcia Manzato}
\authornote{The author supervised the work and revised the manuscript.}
\email{mmanzato@icmc.usp.br}
\orcid{0000-0003-3215-6918}
\affiliation{
  \institution{Institute of Mathematics and Computer Science, University of S\~ao Paulo}
  \city{S\~ao Carlos}
  \country{Brazil}
}

\renewcommand{\shortauthors}{Lima et al.}

\begin{abstract}
    There are challenges that must be overcome to make recommender systems useful in healthcare settings.
    The reasons are varied: the lack of publicly available clinical data, the difficulty that users may have in understanding the reasons why a recommendation was made, the risks that may be involved in following that recommendation, and the uncertainty about its effectiveness.
    In this work, we address these challenges with a recommendation model that leverages the structure of psychometric data to provide visual explanations that are faithful to the model and interpretable by care professionals.    
    We focus on a narrow healthcare niche, gerontological primary care, to show that the proposed recommendation model can assist the attending professional in the creation of personalised care plans.
    We report results of a comparative offline performance evaluation of the proposed model on healthcare datasets that were collected by research partners in Brazil, as well as the results of a user study that evaluates the interpretability of the visual explanations the model generates.
    The results suggest that the proposed model can advance the application of recommender systems in this healthcare niche, which is expected to grow in demand , opportunities, and information technology needs as demographic changes become more pronounced.
\end{abstract}




\maketitle

\section{Introduction}
\label{section:introduction}

Recommender systems are technology components that typically use behaviour data to learn a person’s preference for an item.
They have been applied to diverse domains with astounding success.
For example, in the ever-growing e-commerce domain, a system provides recommendations based on consumer behaviour data, such as who visited which product web page, who bought which product, or how a product was rated or described by those who bought it \cite{alamdari2020ecommerce}.

However, there are challenges that must be overcome to make them useful in healthcare settings.
For one, application domains can differ markedly in the type of data they accumulate and the predictive tasks they value.
This implies that the learning model, which must bridge the data and the predictive task, can also differ markedly among domains.
Take for instance the application reported by \citeonline{gannod2019machine}.
The \sigladef{Preferences for Everyday Living Inventory - Nursing Homes}{PELI-NH} is a standardised questionnaire that captures the preferences of residents of nursing homes regarding 72 aspects of care provided by these institutions.
It can be used to tailor the services provided by the nursing home to the needs or tastes of their residents.
However, the length of the questionnaire is often seen as a barrier to its use.
To address this issue, the reported system combines the resident's answers given to a shorter version of the questionnaire, with 16 questions, with a dataset of hundreds of fully-answered questionnaires.
Then, the system recommends to the nursing home manager a set of additional preferences from the PELI-NH inventory the resident would probably benefit from changing the default choice.

These differences in data structure and relevant tasks become more evident with closer analysis.
Our community's effort has been mostly directed to develop systems that perform well in domains where voluminous, sparse data are publicly available for evaluating new recommendation models.
Typical benchmark datasets contain millions of data points, each representing the manifest/implicit preference of a user with regard to some item.
Although these datasets hold preference data of thousands of users regarding thousands of items, each user \aspas{rates} only a small and distinct fraction of all available items.
Moreover, operating in a low-stakes domain implies that (a) any consequences of making irrelevant recommendations are limited to the users' attitudes towards the system itself (e.g., decreased trust in the system's ability to provide good recommendations) and (b) the ability to provide explanations is probably not a critical success factor for the application \cite{HandbookRS2022Cap19,rudin2019stop}.

In contrast, some niches in healthcare may be better described as a high-stakes, dense-data domains.
Owing to professional ethics in healthcare, patients must be thoroughly assessed using instruments whose effectiveness has been confirmed and receive the care they need accordingly.
For instance, \premise{geriatric and gerontological practices in Brazil (and elsewhere) encourage the use of \term{standardised instruments}{} (e.g., questionnaires) to guide the comprehensive assessment of patients in primary care}{CGA is a medica procedure}, and professional associations may promote their own inventories \cite{nih1987consensus,gorzoni2017geriatria}.
This policy has important consequences for devising recommenders to operate in this niche.

First, every patient is assessed along health dimensions defined by the instrument.
Thus, if the instrument includes an assessment of cognitive capacity, then a score of cognitive capacity will be available for every patient in a dataset.
This is analogous to having a dataset of movie ratings in which every user has rated exactly the same set of movies.
Moreover, the ratings would have been collected in a more systematic manner, for example, by having each user answer a questionnaire just after watching the movie in a theatre.
In other words, unlike the movie recommendation domain, in which the data describing the users' spontaneous rating behaviour are heavily sparse, the data are dense and structured in this niche, consisting of scores a patient obtains along a set of health dimensions, and the scores are computed following a standardised procedure.

Second, the number of items to be recommended is minute compared to that of more traditional applications of recommender systems.
For example, \citeonline{tavassoli2022ICOPE} collected the assessment of intrinsic capacity\footnote{\sigladef{Intrinsic capacity}{IC} is defined by the \sigladef{World Health Organisation}{WHO} as \aspas{the combination of the individual's physical and mental, including psychological, capacities} that are essential to her everyday functioning. 
An instrument to assess IC evaluates the individual's capacity along five dimensions: cognitive, psychological, sensorial, locomotive, and vitality.
} 
of more than ten thousand older participants (60+ years) and, after a screening process, the recommendations and referrals\footnote{In healthcare, a referral is the process by which a healthcare provider directs a patient to another healthcare professional or specialist for further evaluation, diagnosis, or treatment.
This typically occurs when the referring provider believes that the patient's condition requires expertise or services beyond their scope of practice.
} made for each remaining participant were recorded.
In total, 22 distinct recommendations or referrals were made by the attending primary care team.
In this setting, a recommender could be created to advise the care professional about which \aspas{items} should be included in the patient's care plan, based on the outcome of their assessment.

Last but not least, there is the need to minimise risks of harmful recommendations.
The obvious and probably the safest solution is to adopt an expert-in-the-loop approach, as was done in the nursing home example given earlier \cite{gannod2019machine}.
In this approach, the recommendation is presented to the care professional, who exerts her judgement about its adequacy and decides whether the recommendation should be taken into account when creating the patient's care plan.
To assess whether a recommendation is adequate, the expert would probably benefit from an explanation about each recommendation made \cite{nunes2017systematic}.
Furthermore, any provided explanation should preferably be (a) faithful, in the sense that it reflects what the recommendation model actually computes, and (b) easily interpretable by the experts interacting with the system \cite{rudin2019stop}.

Although this analysis is focused on gerontological primary care, \standing{it suggests that each healthcare niche likely has its relevant data structured differently, and that opportunities to create impact may come from investigating decision-making tasks that are valued by specialists in each niche.}{A research strategy for HRS}
Acting on this belief, this work introduces a new recommendation model that leverages the structural characteristics of the psychometric instruments used by professionals to assess their patients.

More precisely, the proposed recommendation model takes advantage that the psychometric data describing patient assessments are dense, low-dimensional, and meaningful to professionals to produce recommendations and explanations.
The explanations, which are diagrams displayed as an interactive visualisation, are faithful in that they depict what the recommendation model actually computes, and 
scrutable, as the user can inspect the values assigned to the elements of the diagram.

We focus on a narrow healthcare niche, gerontological primary care, to show that the proposed recommendation model can assist the attending professional in the creation of personalised care plans.
In this niche, professionals rely on psychometric instruments to comprehensively assess their patients.
In this expert-in-the-loop use case, the recommender system advises the attending primary care professional about which interventions should be included in the patient's care plan.

\begin{figure}[htpb]
    \centering
    \includegraphics[width=1\linewidth]{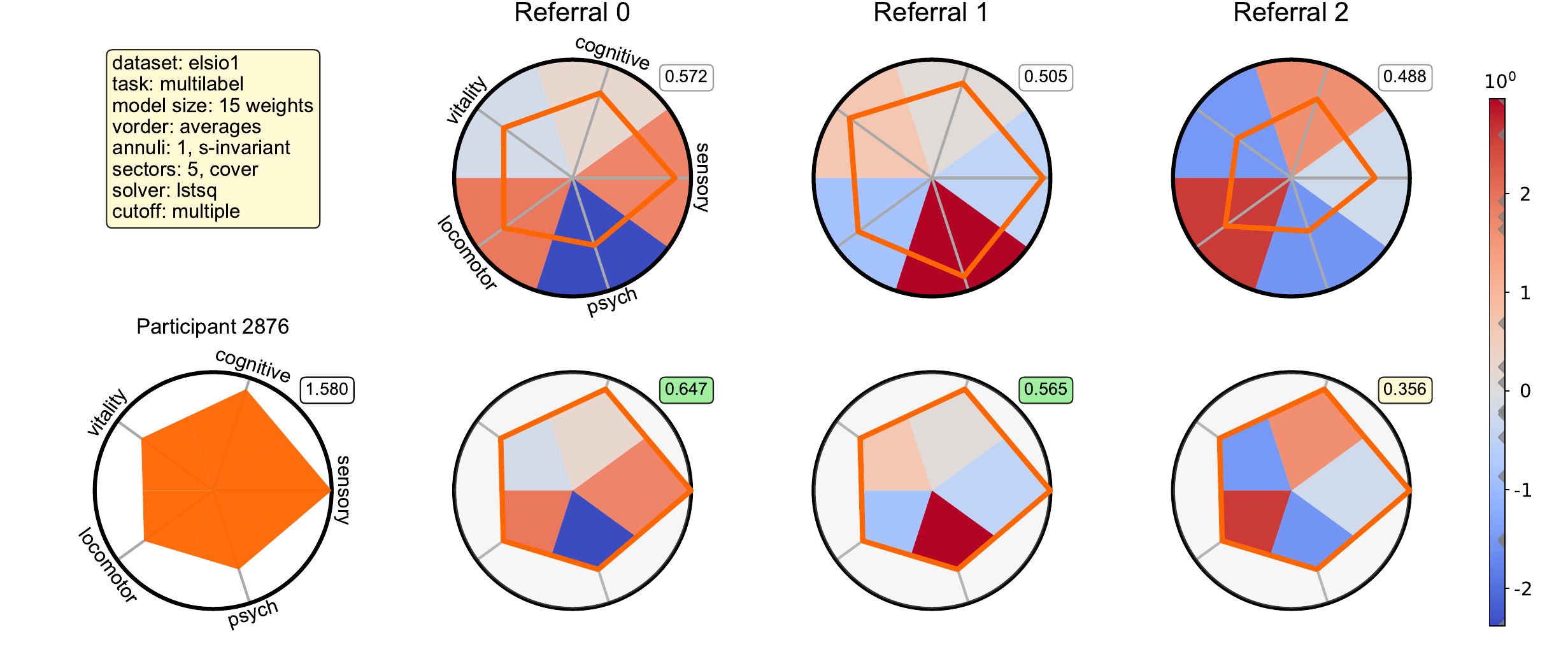}
    \caption{An example of the proposed explanation style. \emph{Legend:} The explanation diagram is composed of a number of radar charts disposed on a rectangular grid.
    There are three types of charts: (a) assignment charts, which are placed on the first row, (b) assessment charts, on the first column, and (c) the matching charts.
    }
    \Description{A diagram composed of radar charts disposed on a 2 x 4 lattice. 
    There is a radar chart in each position, except (1,1), which is occupied by a tag informing the configuration of the Polygrid model that generated this diagram.
    }
    \label{fig:intro:figure_1}
\end{figure}

To give the reader a more concrete grasp of the proposed explanation style, we show an example in Figure \ref{fig:intro:figure_1}.
Building on the study of intrinsic capacity and referrals mentioned earlier \cite{tavassoli2022ICOPE}, the diagram shows an explanation of why some referrals are recommended for a person.
The diagram is composed of a number of radar charts disposed on a grid: (a) the outcome of the patient's assessment is represented by the radar chart in the first column, (b) the referrals are represented in the first row, and (c) the matching between the patient assessment and a referral is shown in a separate chart in the second row, right below the chart of the corresponding referral.

The idea of representing the patient's assessment as a radar chart comes from the literature on geriatrics and gerontology.
As will be seen in Section \ref{section:background:cga:visual}, several authors have highlighted over the years the practical advantages both for professionals and patients of using this diagram to display the results of an assessment made with a psychometric instrument.
We built on this idea by giving the radar chart a footing on psychometric models.
Specifically, we demonstrate a monotonic relationship between the position of an individual on the measurand (i.e., the latent variable being measured by the psychometric instrument) and the area of the assessment polygon.
In Figure \ref{fig:intro:figure_1}, the measurand is intrinsic capacity, and the vertices of the assessment polygon are the scores the patient obtained for the sensory, cognitive, vitality, locomotor, and psychological domains.
The numeric value of the area of the polygon is shown in a tag that appears next to the chart.

Regarding the representation of referrals (or interventions, more generally), each assignment chart contains three elements: a polygon, the colours that fill the disc, and a tag.
The vertices of the polygon are the average scores obtained by a group of patients who were previously assigned to that referral by experts.
The colours represent weights given to different cells of the partition of the disc, and the tag shows the threshold for this referral (more on this in a second).
The colour bar on the right of the diagram maps each colour to its respective numeric weight.

Finally, the matching between the patient's assessment and a referral corresponds to a recommendation decision.
The matching charts have three elements: a polygon, the colours that fill it, and a tag.
The polygon is a copy of the polygon that represents the patient's assessment, which is filled with the colours that appear in the corresponding referral.
The number in the tag is the weighed area of the resulting polygon: if it is greater than the referral's threshold, it appears in green to indicate that the referral is recommended to the patient; otherwise, it appears in yellow.

The recommendation model learns the weights that better replicate the relationship between patient assessments and assigned interventions, as reflected in a dataset with the expert judgements of healthcare professionals.
Similarly to collaborative filtering techniques, both the final users (\aspas{patients}) and recommended items (\aspas{interventions}) are represented in the same abstract space: the unit disc.
The inner product between two such representations, which corresponds to the weighted area of the matching polygon, is the value that appears in the matching chart's tag.
In later sections, we will detail how the many variations of this explanation diagram are supported by the proposed recommendation model, and argue that the proposed model is transparent and interpretable by care professionals who have been trained to use the psychometric instrument.

\subsection{Motivation}
The United Nations has declared the decade of 2021-2030 as the Decade of Healthy Ageing.
This development is partly motivated by the fast demographic changes that are underway worldwide: \aspas{For the first time in history, most people can expect to live into their 60s and beyond} \cite{who2015report}.
Although these changes may bring opportunities for both individuals and societies, an increase in demand for health services is expected.
\premise{A long-term strategy for coping with an increasing demand is to optimise the use of existing resources, which makes the referral process a promising target.}{increased demand can be tackled by optimising existing resources}

Two facts in support of the latter assertion are offered.
First, a recent report by the WHO compiled lessons learned from several EU member states during the course of the COVID-19 pandemic, directed at improving referral systems worldwide \cite{who2023referrals}.
These lessons show that, under an abrupt and large increase in demand, improvements to the referral process led to better outcomes both for individuals and institutions.
Second, many studies in the public health literature suggest that the variation in referral rates among professionals in primary care is significant\footnote{
In this context, referral rate is the ratio $a/b$, where ($a$) is the number of visits to a referring institution that have resulted in the patient being referred to a receiving institution, and ($b$) the number of patient visits to the referring institution \cite{who2023referrals}.}.
This variation has long called the attention of researchers and policymakers, as it may imply financial burden and unfavourable outcomes for patients, and other negative impacts on institutions \cite{odonnel2000variation,wennberg2004interview}.
\future{Although it remains largely unexplained, some evidence suggests that modifiable psychological factors associated with tolerance to ambiguity and uncertainty may play a role in this variation
\cite{shashar2023referrals,pedersene2015attitude}.}{Which modifiable psychological factors are these? Can we show that different visualisations/tasks/instructions can increase or decrease tolerance to uncertainty?}

In summary, the referral process is a promising target because there is room for improvements, and incentives to pursue them.
An interpretable recommendation model that learns from a dataset of standardised patient assessments and referrals made (or reviewed) by experienced practitioners can assist students and novice professionals in developing, improving, or reflecting on their referral practices.
In the long term, this may help reduce unwarranted variation in referrals.

\subsection{Outline}

Section \ref{section:background} briefly discusses how gerontologists foster the well-being of older people in primary care and describes the role of a medical procedure known as a comprehensive geriatric assessment.
This standardised assessment is usually based on a questionnaire developed in the psychometric tradition of factor analysis, which is also briefly presented.
The recurring calls in the literature on gerontology for the use of radar charts to display the patient's assessment are shortly revisited.

Section \ref{section:relatedwork} begins with a review of the literature on healthcare recommender systems to clarify the gap in applications targeting older people in the context of gerontological care.
This is followed by a brief review of machine learning models for multilabel classification and label ranking tasks.
This topic is important because the creation of a personalised care plan in primary care can be formalised by these tasks.
The section concludes with a brief review of the literature on visualisation research focused on the comparative performance of visualisations to support decision-making tasks.

Section \ref{section:proposal} introduces Polygrid, our proposed recommendation model.
The learning pipelines are detailed step by step, first covering multilabel classification tasks and then label ranking tasks.
The effects of different choices for the model's main hyperparameters are illustrated and discussed.
This is followed by two brief discussions.
The first addresses the model's learnability, laying out a theoretical framework to explain why the model successfully learns to perform the target tasks.
The second discussion focuses on the interpretability of the model and the explanations it generates.

The latter discussions, which are conceptual in nature, are complemented with empirical evidence.
Section \ref{section:offlineval} reports the results of a comparative offline performance evaluation, and Section \ref{section:userstudy} reports the results of a user study to evaluate the interpretability of the Polygrid diagram.
Finally, Section \ref{section:conclusion} concludes our work, highlights strengths and limitations, and invites future work.

\section{Background}
\label{section:background}

In this section, we review how gerontologists promote well-being of older people, and some tools they use to provide primary care.
We describe the important role that comprehensive assessments play in the care of older people, and then show that these assessments are founded on psychometric methods.
The first part will be used in the next section, to clarify why current designs of healthcare recommender systems are not aligned with the gerontological practice.
The second part will be used in a later section, to describe the expected structure (or regularities) of psychometric data.

\subsection{How Gerontologists Assess and Foster Well-being of Older Persons}
\label{section:background:cga}

The WHO advances the notion of healthy ageing as \aspas{the process of developing and maintaining the functional ability that enables well-being in older age}\footnote{Functional ability refers to these abilities (among others): autonomy, conceived as the capacity and the right to make own choices, and independence, as the ability to perform tasks of daily life without assistance, such as getting in and out of bed, using the toilet, taking medications, bathing, eating, preparing meals, and shopping for groceries. These functions allow for a person to remain socially relevant and to cope with adverse life events.} \cite{who2015report}.
This is not synonymous with a disease-free state, which is problematic in older age because many individuals develop chronic conditions that, if effectively managed, do not substantially decrease functional ability.
It refers to a state in which individuals maintain the ability to pursue things they value in life.

Gerontology, as a practice, is aligned with this view of healthy ageing, as professionals seek to foster functional ability in older populations so that people can experience well-being for as long as possible \cite{melo2015desafios}.
In a consensus development conference sponsored by the NIH and other institutions in 1987, specialists in the care of older people have agreed that their \sigladef{Comprehensive Geriatric Assessment}{CGA} procedures should be improved.
These improvements should aim how specialists (a) select interventions to restore or preserve health and functional status, (b) predict health outcomes, and (c) monitor clinical change over time (among other goals) \cite{nih1987consensus}.
In Brazil, CGA is currently defined as a medical procedure by the \sigladef{Brazilian Medical Association}{AMB}.
Moreover, the \sigladef{Brazilian Society of Geriatrics and Gerontology}{SBGG}, the main professional association of geriatricians and gerontologists, promotes \href{https://sbgg.org.br/publicacoes-cientificas/avaliacao-geriatrica-ampla/#}{a standard CGA inventory} to its members \cite{gorzoni2017geriatria}.

In general, CGA inventories are structured questionnaires designed to guide care professionals in assessing their geriatric patients.
These inventories typically include items that survey domains such as physical medical conditions, mental health conditions, functioning, and social and environmental circumstances, providing the full biopsychosocial nature of the individual's capacities, problems, and conditions \cite{welsh2014CGA,garrard2020cgareview}.
Based on the patient's responses, a set of numerical scores that reflect the individual's health status can be computed by a standardised procedure, as we shall soon see.

\subsection{The Measurement of Health, Psychometrics, and Factor Analysis}
\label{section:background:measurement-of-health}

Measurement is seen by many as a hallmark of modern science, but \aspas{there is little consensus among philosophers as to how to define measurement, what sorts of things are measurable, or which conditions make measurement possible,} argues \citeonline{tal2020measurement}.
This statement reflects unsettled disputes about measurement in the social sciences and in the part of health sciences that overlaps with psychology.
In these branches, researchers are often faced with the challenge of measuring variables that cannot be directly observed, such as life satisfaction or functional independence \cite{philippi2023challenges}.
The difficulties emerge early in the process of developing a systematic measurement method, in the conceptualisation stage.
For example, most people will take only a couple of seconds to answer \aspas{How satisfied are you with your health?} but take the definition of health that appears in the preamble to the 1947 Constitution of the WHO: \aspas{Health is a state of complete physical, mental and social well-being and not merely the absence of disease or infirmity} \cite{hausman2015valuing}.

This definition of health, while aspirational, presents a challenge to researchers: How should one measure something as intangible as a \aspas{state of complete [...] well-being?}
In the past, social researchers looked to the measurement methods of the natural sciences for principles that could improve their measurement practices, but today many are sceptical about the efficacy of this research programme \cite{michell1999measurement}. 
For historical reasons, the gerontological research community has predominantly subscribed to psychometric methods to meet the challenge of measuring the human attributes that are substantive to their theories, and to the methods of factor analysis in particular \cite{shenk2001teaching}.

\citeonline{borsboom2015psychometrics} define psychometrics as \aspas{a scientific discipline concerned with the question of how psychological constructs (e.g., intelligence, neuroticism, or depression) can be optimally related to observables (e.g., outcomes of psychological tests, genetic profiles, neuroscientific information).}
In other words, the central question of psychometrics, which is of direct relevance to gerontological research, is how to connect theory to observations.
In the factor-analytic approach, this connection is formalised by a class of statistical models called structural equation models \cite{bollen1989semlv}.
There is a vast literature on these models, but we focus on a small subset, the measurement models, because of their importance as building blocks in the development of psychometric scales, a fact that we later exploit to describe the expected structure of data collected with such instruments.

To reduce ambiguity, we clarify our usage of some technical terms in the following.
In Figure \ref{fig:background:selvm:mm}, which shows a path diagram for a measurement model, $\eta$ represents a latent variable.
In the literature, there are several definitions of what a latent variable is, but we use the term in the following sense:
\standing{a latent variable is a random variable that represents a psychological attribute; this attribute is assumed to exist independently of measurement, but (presently) cannot be measured directly without substantial measurement error \cite{bollen2013myths,kappenburg2014comparison}.
Although a latent variable is hardly accessible to direct measurement, the fact that it acts as a common determinant of a set of variables that are more amenable to direct measurement allows us some level of access to it \cite{borsboom2005measuring}.}{LV}
The arrows that depart from $\eta$ towards the indicators $x_1 \ldots x_3$ represent its action as a common determinant.
Note that $x_1 \ldots x_3$ are represented as manifest variables in Figure \ref{fig:background:selvm:mm}, and as latent variables in Figure \ref{fig:background:selvm:cfa2}.
This reflects a subtle duality: once we measure the indicators of a latent variable and gain some access to it, the variable ceases to be \aspas{hidden} and can be treated as a manifest variable.
This latter term has multiple synonyms in the literature (e.g., observed variable, observable variable, indicator), and we use them interchangeably with this meaning: \standing{a manifest variable is a random variable that is more amenable to direct measurement compared to the latent variables in a structural model.}{MV}

An example to embody these ideas: suppose that the latent variable $x_1$ in Figure \ref{fig:background:selvm:cfa2} represents psychological well-being.
Then the variable $q_1$ could represent the response to an item asking the respondent to rate the statement \aspas{How well are you able to concentrate?} on a five-point Likert scale ranging from \aspas{Not at all} to \aspas{Extremely}.
Once the respondent answers the items related to $q_1 \ldots q_4$ (and assuming that the parameters $\gamma_{11} \ldots \gamma_{14}$ are known), the collected responses can be used to compute a score for $x_1$ \cite{mcneish2020sumscores}.
This causes an upstream effect: $x_1$ can now be used as an indicator of $\eta$, as seen in Figure \ref{fig:background:selvm:mm}.
In a sense, the measurement model of the latent variable $x_1$, which comprises $x_1$, $q_1 \ldots q_4$, $\gamma_{11} \ldots \gamma_{14} \mbox{, and } \delta_1 \ldots \delta_4$, works as a scale for $x_1$ in much the same way as the five-point Likert scale ranging from \aspas{Not at all} to \aspas{Extremely} is a scale for $q_1$.
In this usage, 
\standing{the term scale means \aspas{a system for ordering test responses in a progressive series, so as to measure a trait, ability, attitude, or the like}, which corresponds to a definition found in the \href{https://dictionary.apa.org/scale}{\sigladef{American Psychological Association}{APA}'s online dictionary}}{scale}.
In the same vein, we use the term instrument to refer to a system by which researchers assess or gather data about study participants.
For example, suppose that $\eta$ in Figure \ref{fig:background:selvm:cfa2} represents a health-related notion of quality of life.
The questionnaire containing the items related to $q_1 \ldots q_{12}$, together with a scoring procedure, is then an instrument, and the latent variables $x_1 \ldots x_3$ are then called the domains of the instrument.

\begin{figure}[htpb]
    \centering
    \label{fig:background:selvm}
    \begin{subfigure}{1.0\textwidth}
        \centering
	    \begin{tikzpicture}[
	        latvar/.style={ellipse, draw, minimum width=1.5cm, minimum height=1cm},
	        manvar/.style={rectangle, draw, minimum size=0.7cm},
	        error/.style={minimum size=0.5cm},
	        arrow/.style={-Stealth,font=\footnotesize}
	    ]
	            
	    \node[latvar] (eta) at (0,0) {$\eta$};
            
	    \node[latvar, below left=1.5cm and 3.453cm of eta] (x1) {$x_1$};
	    \node[latvar, right=3.0cm of x1] (x2) {$x_2$};
	    \node[latvar, right=3.0cm of x2] (x3) {$x_3$};

	    \draw[arrow] (eta) -- (x1.north east) node[left,  pos=0.4] {$\lambda_{1}$};
	    \draw[arrow] (eta) -- (x2.north)      node[left,  pos=0.4] {$\lambda_{2}$};
	    \draw[arrow] (eta) -- (x3.north west) node[right, pos=0.4] {$\lambda_{3}$};

        \node[error, left=0.8cm of x1] (e1) {$\epsilon_1$};
        \node[error, left=0.8cm of x2] (e2) {$\epsilon_2$};
        \node[error, left=0.8cm of x3] (e3) {$\epsilon_3$};
    
        \draw[arrow] (e1) -- (x1);
        \draw[arrow] (e2) -- (x2);
        \draw[arrow] (e3) -- (x3);
            
	    \node[manvar, below left=1.5cm and 0.65cm of x1] (q1) {$q_1$};
	    \node[manvar, right=0.3cm of q1] (q2) {$q_2$};
	    \node[manvar, right=0.3cm of q2] (q3) {$q_3$};
	    \node[manvar, right=0.3cm of q3] (q4) {$q_4$};

	    \draw[arrow] (x1) -- (q1.north) node[left,  pos=0.4] {$\gamma_{11}$};
	    \draw[arrow] (x1) -- (q2.north) node[left,  pos=0.6] {$\gamma_{12}$};
	    \draw[arrow] (x1) -- (q3.north) node[right, pos=0.6] {$\gamma_{13}$};
	    \draw[arrow] (x1) -- (q4.north) node[right, pos=0.4] {$\gamma_{14}$};
	    
	    \node[error, below=0.8cm of q1] (d1) {$\delta_1$};
	    \node[error, below=0.8cm of q2] (d2) {$\delta_2$};
	    \node[error, below=0.8cm of q3] (d3) {$\delta_3$};
	    \node[error, below=0.8cm of q4] (d4) {$\delta_4$};

	    \draw[arrow] (d1) -- (q1);
	    \draw[arrow] (d2) -- (q2);
	    \draw[arrow] (d3) -- (q3);
	    \draw[arrow] (d4) -- (q4);


	    \node[manvar, below left=1.5cm and 0.65cm of x2] (q5) {$q_5$};
	    \node[manvar, right=0.3cm of q5] (q6) {$q_6$};
	    \node[manvar, right=0.3cm of q6] (q7) {$q_7$};
	    \node[manvar, right=0.3cm of q7] (q8) {$q_8$};

	    \draw[arrow] (x2) -- (q5.north) node[left,  pos=0.4] {$\gamma_{25}$};
	    \draw[arrow] (x2) -- (q6.north) node[left,  pos=0.6] {$\gamma_{26}$};
	    \draw[arrow] (x2) -- (q7.north) node[right, pos=0.6] {$\gamma_{27}$};
	    \draw[arrow] (x2) -- (q8.north) node[right, pos=0.4] {$\gamma_{28}$};
	    
	    \node[error, below=0.8cm of q5] (d5) {$\delta_5$};
	    \node[error, below=0.8cm of q6] (d6) {$\delta_6$};
	    \node[error, below=0.8cm of q7] (d7) {$\delta_7$};
	    \node[error, below=0.8cm of q8] (d8) {$\delta_8$};

	    \draw[arrow] (d5) -- (q5);
	    \draw[arrow] (d6) -- (q6);
	    \draw[arrow] (d7) -- (q7);
	    \draw[arrow] (d8) -- (q8);

	    \node[manvar, below left=1.5cm and 0.65cm of x3] (q9) {$q_9$};
	    \node[manvar, right=0.3cm of q9]  (q10) {$q_{10}$};
	    \node[manvar, right=0.3cm of q10] (q11) {$q_{11}$};
	    \node[manvar, right=0.3cm of q11] (q12) {$q_{12}$};

	    \draw[arrow] (x3) -- (q9.north)  node[left,  pos=0.40] {$\gamma_{39}$};
	    \draw[arrow] (x3) -- (q10.north) node[left,  pos=0.66] {$\gamma_{3,10}$};
	    \draw[arrow] (x3) -- (q11.north) node[right, pos=0.66] {$\gamma_{3,11}$};
	    \draw[arrow] (x3) -- (q12.north) node[right, pos=0.40] {$\gamma_{3,12}$};
	    
	    \node[error, below=0.8cm of q9]  (d9)  {$\delta_9$};
	    \node[error, below=0.8cm of q10] (d10) {$\delta_{10}$};
	    \node[error, below=0.8cm of q11] (d11) {$\delta_{11}$};
	    \node[error, below=0.8cm of q12] (d12) {$\delta_{12}$};

	    \draw[arrow] (d9)  -- (q9);
	    \draw[arrow] (d10) -- (q10);
	    \draw[arrow] (d11) -- (q11);
	    \draw[arrow] (d12) -- (q12);
	        
	    \end{tikzpicture}
        
        \caption{The path diagram of a structural equation model with latent variables. \emph{Legend:} A symbol enclosed by an ellipse represents a latent variable (e.g., $\eta$);
    a symbol enclosed by a square represents an indicator (or manifest variable: the $q_j$ variables);
    an unenclosed symbol represents an error variable (all $\delta$ and $\epsilon$ variables, which are unobserved);
    a straight arrow from A to B indicates causation and should be read as \aspas{A causes B} \cite{bollen1989semlv}.
    A more nuanced interpretation of this arrow is \aspas{variation in the latent variable precedes variation in the indicators} \cite{borsboom2005measuring}.
    The edges connecting latent to manifest variables are annotated with effect coefficients (all $\lambda$ and $\gamma$ coefficients).
    Other visual types commonly used in path diagrams are ignored in our discussion.
    }
        \Description{A path analysis diagram}
        \label{fig:background:selvm:cfa2}
    \end{subfigure}
    \begin{subfigure}{1.0\textwidth}
        \centering
	    \begin{tikzpicture}[
	        latvar/.style={ellipse, draw, minimum width=1.5cm, minimum height=1cm},
	        manvar/.style={rectangle, draw, minimum size=0.7cm},
	        error/.style={minimum size=0.5cm},
	        arrow/.style={-Stealth,font=\footnotesize}
	    ]
	    

      \node (root) at (0,0) {}; 
      \node[latvar, below=0.25cm of root] (eta) {$\eta$};
    
	    \node[manvar, below left=1.5cm and 1.05cm of eta]  (x1) {$x_1$};
	    \node[manvar, below=1.35cm of eta]                 (x2) {$x_2$};
	    \node[manvar, below right=1.5cm and 1.05cm of eta] (x3) {$x_3$};

	    \draw[arrow] (eta) -- (x1.north) node[left,  pos=0.5] {$\lambda_1$};
	    \draw[arrow] (eta) -- (x2.north) node[left,  pos=0.5] {$\lambda_2$};
	    \draw[arrow] (eta) -- (x3.north) node[right, pos=0.5] {$\lambda_3$};
	    
	    \node[error, below=0.8cm of x1] (e1) {$\epsilon_1$};
	    \node[error, below=0.8cm of x2] (e2) {$\epsilon_2$};
	    \node[error, below=0.8cm of x3] (e3) {$\epsilon_3$};

	    \draw[arrow] (e1) -- (x1);
	    \draw[arrow] (e2) -- (x2);
	    \draw[arrow] (e3) -- (x3);

	    \node[below right=0.85cm and 2.6cm of eta] (eq-x1) {$x_{1i} = \lambda_1 \eta_i + \epsilon_{1i}$};
	    \node[below=0.1cm of eq-x1] (eq-x2) {$x_{2i} = \lambda_2 \eta_i + \epsilon_{2i}$};
	    \node[below=0.1cm of eq-x2] (eq-x3) {$x_{3i} = \lambda_3 \eta_i + \epsilon_{3i}$};
	    \draw [decorate,decoration={brace,amplitude=5pt,mirror,raise=0pt}] (eq-x1.north west) -- (eq-x3.south west);

	    \node[right=0.35cm of eq-x1] (cov-q1) {$X := \Lambda H' + \varepsilon \implies X' = H \Lambda' + \varepsilon'$};

        \node[right=0.35cm of eq-x2] (cov-q2) {$XX' = \Lambda H' H \Lambda' + \Lambda H' \varepsilon' + \varepsilon H \Lambda' + \varepsilon \varepsilon'$};
     
        \node[right=0.35cm of eq-x3] (cov-q3) {$\Sigma = \E{XX'} = \Lambda \phi \Lambda' + \Theta$};

      \node[below right=0.05cm and 1.9cm of eta] (heading-eqs) {Structural Equations};
      \node[below right=0.05cm and 6.5cm of eta] (heading-cov) {Covariance Matrix};
	    
      \end{tikzpicture}        
        \caption{A congeneric measurement model for the top latent variable of the previous path diagram, with the domain variables $x_1 \ldots x_3$ \aspas{collapsed} after measurement.
        $\Lambda$ is a $(d,1)$-vector with effect coefficients, $H$ is a $(m,1)$-vector with the true scores, and $X$ is a $(d,m)$-matrix with scores of $m$ individuals on $d$ domains.
        }
        \Description{A path analysis diagram}
        \label{fig:background:selvm:mm}        
    \end{subfigure}
    \caption{A structural equation model represented by a path diagram. \emph{Note:} Figure \ref{fig:background:selvm:cfa2} shows an adaptation of the path diagram for the second order \sigladef{confirmatory factor analysis}{CFA} model from \cite[p. 315]{bollen1989semlv}.
    This model has been employed in prominent models used in gerontological research, including the instruments described in the Annex \ref{section:background:cga:instruments}.
    In this diagram, the latent variable model is composed of the subgraph induced by the latent variables, namely $\eta, x_1 \ldots x_3$.
    It encodes the fact that the second order variable $\eta$ causes variations in the first order variables $x_1 \ldots x_3$, but the latter do not influence each other.
    There are three measurement models, one for each of the first order variables (i.e., $x_1$ to $x_3$).
    A measurement model is composed of the subgraph induced by a given latent variable and its corresponding indicators.
    Figure \ref{fig:background:selvm:mm} focus on the collapsed measurement model for the latent variable $\eta$ to show that its path diagram is a pictorial representation of a system of structural equations, and that the hypothetical covariance matrix $\Sigma$ is derived from this system.}
\end{figure}

Now that we have set up the appropriate terminology, let's have a look at how structural models are estimated.
From the path diagram in Figure \ref{fig:background:selvm:mm}, one can deduce that the score that the $i$-th subject obtains for the $k$-th domain, namely $x_{ki}$, depends on the her position on the measurand ($\eta_i$), the effect the measurand exerts on its indicators ($\lambda_{k}$), and a measurement error ($\epsilon_{ki}$) \cite{kappenburg2014comparison}:
\begin{equation}
    \label{eq:background:sem}
    x_{ki} = \lambda_{k} \eta_i + \epsilon_{ki}.
\end{equation}
In fact, the whole diagram is a pictorial representation of the system of structural equations shown to the right of the diagram.
Suppose that a researcher is creating a new instrument that surveys $d$ domains, each with a mature scale already available.
The researcher gathers data from $m$ subjects, computes their scores for each domain, and organises these scores into a $(d,m)$-matrix $\mathring{X}$.
The structural equations indicate that the matrix $\mathring{X}$, once centred, could be rewritten as $X = \Lambda H' + \varepsilon$, with $H$ (uppercase $\eta$) as a $(m,1)$-vector with the scores of the $m$ subjects on the latent variable $\eta$, $\Lambda$ as a $(d,1)$-vector with the effect coefficients, and $\varepsilon$ as an $(d,m)$-matrix with measurement errors.
From this matrix equation, one can deduce the $(d,d)$-covariance matrix of $X$, namely $\Sigma = \E{XX'} = \Lambda \phi \Lambda' + \Theta$, with $\phi$ as the scalar variance of the latent variable $\eta$, and $\Theta$ as a $(d,d)$-matrix with error covariances.
The estimation process seeks for the values of the model parameters $(\Lambda, \phi, \Theta)$ that minimise the \aspas{distance} between the predicted covariance matrix $\Sigma$ and the sample covariance matrix calculated from $\mathring{X}$, subject to a number of constraints, such as $\E{H} = 0$, $\E{\varepsilon_{k}} = 0$, $\Cov(H,\varepsilon_{k})=0$ for all $k$, and $\Cov(\varepsilon_{k},\varepsilon_{l})=0$ for $k \neq l$, among possibly others \cite[p.10]{kappenburg2014comparison}.

Once the model parameters are estimated, a number of statistical indices of fit can be computed.
Among them, and of special relevance to this work, is the reliability coefficient.
\aspas{A measurement instrument is reliable to the extent that it yields the same outcomes when applied to persons with the same standing of the measured attribute under the same circumstances} \cite{borsboom2015psychometrics}.
However, it must be noted that reliability is \aspas{a property of the data generated by the instrument when applied to a specific population rather than a property of the scale itself} \cite{hayes2020omega}.
For congeneric models such as the one in Figure \ref{fig:background:selvm:mm}, the McDonald's $\omega$ has been recommended as a reliability index \cite{mcneish2018alpha}:
\begin{equation}
    \label{eq:background:omega}
    \omega = \frac{ \big(  \sum_k \lambda_k \big)^2 }{ \big(  \sum_k \lambda_k \big)^2 + \sum_k \Var(\varepsilon_{k})},
\end{equation}
where $\varepsilon_{k}$ is the $k$-th row of the measurement errors matrix.
Thus, $\omega \in (0, 1]$, and the higher the reliability, the more we can trust that the differences between people in their scores on the latent variable are an accurate reflection of actual individual differences in the attribute represented by $\eta$.

Now we can pick up where we left off at the end of Section \ref{section:background:cga}.
If the computed fit indices indicate that the instrument has the desired properties to support its proposed uses, and one of these uses is establishing baseline scores in clinical trials (i.e., sensitive to within-subject difference), 
then the instrument's scoring procedure gains a prominent role.
For the congeneric model in Figure \ref{fig:background:selvm:mm}, there are two scoring procedures that are predominantly used: the optimally weighted score and the sum-score \cite{mcneish2020sumscores}.
The optimally weighted score corresponds to the following weighted sum:
\begin{equation}
    \label{eq:background:weighted-score}
    \hat{\eta}_i := \sum_k \lambda_k \, \mathring{x}_{ki} ,
\end{equation}
and the second, the sum-score, is a special case of Equation \ref{eq:background:weighted-score} with all effect coefficients $\lambda_k = 1$.
The relative advantages of one scoring method over another are a subject of debate even today \cite{mcneish2020sumscores,widaman2023sumscore2,mcneish2023sumscore3,widaman2024sumscore4,sijtsma2024sumscore5,mcneish2024sumscore6,sijtsma2024sumscore8}.
The three instruments described in Appendix \ref{section:background:cga:instruments} use a scoring procedure that is equivalent to sum-scoring, but the result in Equation \ref{eq:background:weighted-score} will be used in Section \ref{section:proposal:interpretable:defence} to show a relationship between the score on a latent variable ($\hat{\eta}_i$) and the geometry of radar charts, a visualisation proposed by many authors to display the results of a patient's CGA, as described next.

Finally, we must highlight a result that will be later used to describe the expected structure of psychometric data: if a measurement model is set so that $\lambda_k > 0$ for all $k$, and the model fits the data well, then the sample covariance matrix should have only positive entries \cite[p.22]{bollen1989semlv}:
\begin{align}
    \label{eq:background:corrpos}
    \Cov(\mathring{X}_{k}, \mathring{X}_{l}) &=  \E{X_{k} \, X_{l}'} = \lambda_k \lambda_l \phi > 0.
\end{align}

\subsection{The Visual Display of the Results of a Patient Assessment}
\label{section:background:cga:visual}

Over the years, several authors have proposed the use of radar charts to organise the visual display of the results of a CGA.
This is puzzling because the proposals were made independently of each other, and the authors point out distinct but convergent advantages of the use of this diagram.

For example, \citeonline{vergani2004polar} proposed the use of a radar chart to show the results of a CGA comprising 12 domain scores.
Each score determines one vertex of the polygon in a radar chart, similar to the assessment chart shown in Figure \ref{fig:intro:figure_1}.
According to the authors, the diagram facilitates communication between clinical and administrative staffs, and also stimulates a dialogue that is useful both for clinical and educational reasons.
In addition, juxtaposition of results from successive assessments of a patient facilitates the identification of the domains in which a change has occurred.

Before that work, \citeonline{shibata1998development} employed a radar chart to summarise a patient's assessment.
The questionnaire they used has about 90 items, and a scoring procedure converts patient responses into five domains scores.
This radar chart is an element of the report produced by an expert system that aimed to promote health-related quality of life of older citizens in Japan.
The patient receives a printed copy of the report, which also contains recommendations based on her  scores.

In a different vein, \citeonline{minemawari1999radar} promote a radar chart to profile disabling conditions of older patients admitted to hospital care.
The diagram seeks to profile a medical condition instead of a patient's assessment, as in previous cases.
Each patient is assessed along six domains, and each profile corresponds to the average scores obtained by a group of patients diagnosed with a given condition.
This is similar to the assignment charts shown in Figure \ref{fig:intro:figure_1}.
The goal of the diagram is to foster integrated care by increasing awareness among professionals working in multidisciplinary care teams about how different factors correlate in inpatients with disabling conditions.

Approaching a more conceptual challenge, \citeonline{jung2020radar} proposes the use of radar charts in the operationalisation of two evasive constructs in the health sciences: physiological reserve and frailty.
The author draws an analogy between a latent variable and the area of the polygon in a radar chart.
The area of the polygon depicts the physiological reserve of an individual, and the area that is external to the polygon (and internal to the outer border of the chart) depicts the notion of frailty.
According to the author, this diagram \aspas{may facilitate communication between healthcare providers to foster shared inter-professional decision-making} and \aspas{can make the interpretation of CGA parameters [i.e., domain scores] ... easier, allowing physicians to tackle those with deficits.}

Finally, \citeonline{cavanaugh2025radar} promote a radar chart to display CGA data in geriatric oncological care.
The polygon in the radar chart represents the subject's actual recorded condition on eleven domains.
The authors argue that the diagram is useful \aspas{to guide supportive interventions} and \aspas{to facilitate communication of the dominant vulnerability-driving individual risks.}

The five studies just described were found in a modest but systematic review.
Inclusion criteria are primary studies that refer both to CGA and radar charts, and selection was based on titles and abstracts.
The Scopus and PubMed databases were searched for such studies, and the submitted query follows the pattern: \aspas{(\texttt{CGA} \texttt{OR} \texttt{elderly questionnaire}) \texttt{AND} \texttt{radar chart}}, with variables replaced by their corresponding expressions in Table \ref{tab:background:cga:query}.
The exclusion criteria were: duplicated studies (multiple proposals by the same author), studies that do not use radar charts to display CGA data, and studies involving populations other than older people.
Of the ten studies included in the scope, five were excluded, and the remaining five studies were the focus of our review.

\begin{table}[t]
    \centering
    \footnotesize
 	\caption{Variables used in searching for CGA studies using radar charts}
    \begin{tabularx}{\linewidth}{@{}P{2.2cm}X@{}}
 		
        \toprule
 		Variable&\makecell[c]{Expression}\\
 		\midrule
 		
        \multirow{6}{\hsize}{\texttt{\makecell[c]{CGA}}} & \aspas{comprehensive geriatric assessment} \texttt{OR} \aspas{comprehensive geriatric evaluation} \texttt{OR} \aspas{comprehensive gerontological assessment} \texttt{OR} \aspas{comprehensive gerontological evaluation} \texttt{OR} \aspas{multidimensional geriatric assessment} \texttt{OR} \aspas{multidimensional geriatric evaluation} \texttt{OR} \aspas{multidimensional gerontological assessment} \texttt{OR} \aspas{multidimensional gerontological evaluation} \texttt{OR} \aspas{multidisciplinary geriatric assessment} \texttt{OR} \aspas{multidisciplinary geriatric evaluation} \texttt{OR} \aspas{multidisciplinary gerontological assessment} \texttt{OR} \aspas{multidisciplinary gerontological evaluation} \\
 		\midrule
        
        \multirow{4}{\hsize}{\texttt{\makecell[c]{elderly \\ questionnaire}}} & \aspas{questionnaire} \texttt{AND} (\aspas{health} \texttt{OR} \aspas{healthcare} \texttt{OR} \aspas{geriatrics} \texttt{OR} \aspas{gerontological} \texttt{OR} \aspas{quality of life} \texttt{OR} \aspas{well-being} \texttt{OR} \aspas{intrinsic capacity} \texttt{OR} \aspas{functional ability} \texttt{OR} \aspas{daily living}) \texttt{AND} (\aspas{elder} \texttt{OR} \aspas{elderly} \texttt{OR} \aspas{senior} \texttt{OR} \aspas{older person} \texttt{OR} \aspas{older adult} \texttt{OR} \aspas{older patient} \texttt{OR} \aspas{older people} \texttt{OR} \aspas{older population}) \\
 		\midrule
        
        \multirow{6}{\hsize}{\texttt{\makecell[c]{radar \\ chart}}} & \aspas{radar chart} \texttt{OR} \aspas{radar diagram} \texttt{OR} \aspas{radar graph} \texttt{OR} \aspas{radar plot} \texttt{OR} \aspas{web chart} \texttt{OR} \aspas{web diagram} \texttt{OR} \aspas{web graph} \texttt{OR} \aspas{web plot} \texttt{OR} \aspas{coweb chart} \texttt{OR} \aspas{coweb diagram} \texttt{OR} \aspas{coweb graph} \texttt{OR} \aspas{coweb plot} \texttt{OR} \aspas{spider chart} \texttt{OR} \aspas{spider diagram} \texttt{OR} \aspas{spider graph} \texttt{OR} \aspas{spider plot} \texttt{OR} \aspas{star chart} \texttt{OR} \aspas{star diagram} \texttt{OR} \aspas{star graph} \texttt{OR} \aspas{star plot} \texttt{OR} \aspas{polar chart} \texttt{OR} \aspas{polar diagram} \texttt{OR} \aspas{polar graph} \texttt{OR} \aspas{polar plot} \texttt{OR} \aspas{Kiviat chart} \texttt{OR} \aspas{Kiviat diagram} \texttt{OR} \aspas{Kiviat graph} \texttt{OR} \aspas{Kiviat plot} \\
        
        \bottomrule
 	\end{tabularx}
	\label{tab:background:cga:query}
\end{table}

\section{Related Work}
\label{section:relatedwork}

In this section, we review the literature on healthcare recommender systems that target older people.
The review aims to clarify the gap between current system designs and applications and the minimal requirements they should meet to function as tools to assist gerontologists create personalised care plans in primary care.
In our view, these requirements include the use of CGA data to describe patients and the provision of faithful explanations to the care professional.

Two other important topics are also covered.
The literature on multilabel classification and label ranking tasks is reviewed because these tasks will be used to formalise the task a gerontologist performs when creating a personalised care plan.
We review the terminology, the main technical approaches to solving the tasks, the benchmark datasets, and the standard evaluation methodologies.
Finally, we also briefly review the literature on visualisation research to summarise current results on the types of visualisation that best support the completion of decision-making tasks.
These results will be used later to inform the design of a user study to assess interpretability.

\subsection{Applications of Recommender Systems in Healthcare}
\label{section:relatedwork:healthrecsys}

The widespread application of recommender systems demonstrates their usefulness.
However, successful applications in healthcare are relatively rare.
In recognition of the importance of this domain and the opportunities it offers, our community has recently organised a series of workshops to exchange ideas on the topic \cite{healthrecsys1,healthrecsys2,healthrecsys3,healthrecsys4,healthrecsys5}.
Another indication of the interest of our community is the increasing number of secondary works dedicated to mapping and describing applications in this domain: at least 15 reviews on the application of recommender systems to healthcare have been published over the years
\cite{sezgin2013review,kamran2015survey,ferretto2017review,horsfraile2018review,azmi2019review,cheung2019review,2019pincayreview,ertugrul2020review,su2020review,decroon2021review,tran2021review,cai2022review,calderon2023review,etemadi2023review,vieira2023review}.

In the following, we report the results of yet another review of the literature on healthcare recommender systems, with a focus on applications targeting older people.
The need for this new review stems from the fact that the research questions approached by existing reviews do not clarify three points of interest to this project, as detailed next.

\subsubsection{Review Planning}

This review aims to establish the prevalence of healthcare recommender systems for older people that: (Q1) use data from standardised assessments, (Q2) provide explanations about provided recommendations, and (Q3) whether these explanations are faithful.
The answers to these questions can show to what extent current blueprints and implementations of health recommender systems align with the way gerontologists assess their patients and create care plans, as seen in Section \ref{section:background:cga}.
The term \aspas{standardised assessment} means that 
a trained professional administers an instrument to a subject,
as discussed in Section \ref{section:background:measurement-of-health}.
The term \aspas{faithful explanation} denotes an explanation that reflects what the recommendation model actually computes \cite{rudin2019stop}.
To answer these questions, the following protocol was adopted:

\begin{itemize}

    \item Stage 1: Identification.
    Primary studies were identified by consulting three databases: Scopus, IEEEXplore, and the ACM/DL.
    The query submitted to the Scopus database follows the pattern \aspas{\texttt{system type} \texttt{AND} \texttt{domain} \texttt{AND} \texttt{population} \texttt{AND} \texttt{task}}, with variables replaced by their corresponding expressions in Table \ref{tab:relatedwork:query}.
    Since submitting this query to the IEEEXplore and the ACM/DL databases did not return any results, we relaxed the search constraints to \aspas{\texttt{system type} \texttt{AND} \texttt{population}}.
    Records from each study were gathered.

    \item Stage 2: Screening.
    Records that appeared multiple times were excluded, as well as studies whose abstracts indicated that the focus was not on healthcare recommender systems for older people; when in doubt, we consulted the full-text to assert the desired focus.
    When multiple reports of the same research project were found (e.g., CARE and +TV4E projects), one of the studies was selected for inclusion, at our discretion.
    Studies presenting early ideas about a healthcare recommender system that did not report results of a working prototype were included at our discretion.
    
    \item Stage 3: Collection.
    The full-text of the selected studies was gathered.
    Studies to which we did not have access through the licencing agreement between our university and the relevant publishers and were not available in the arxiv were excluded.
       
    \item Stage 4: Data extraction.
    The full text of each study was considered.
    We extracted evidence that the proposed recommender system uses data from standardised assessments of the target users to generate recommendations (Q1) and provides faithful explanations for the given recommendations (Q2 and Q3).
    Data about the adopted recommendation approach and the recruited participants were also collected.
    
    \item Stage 5: Inclusion of results from a previous review.
    The results of a review we conducted for a previous unpublished work in January 2021, using a snowballing protocol, were included.
    
\end{itemize}

\begin{table}[htpb]
    \centering
 	\footnotesize
 	\caption{Variables used in searching for HRS studies targeting older adults}
    \begin{tabularx}{\linewidth}{@{}P{1.5cm}X@{}}
 		\toprule
 		Variable&\makecell[c]{Expression}\\
 		\midrule
        \multirow{8}{\hsize}{\texttt{\makecell[c]{system \\ type}}} & \aspas{recommender system}\texttt{ OR }\aspas{recommendation system}\texttt{ OR }\aspas{recommending system}\texttt{ OR }\aspas{recommender model}\texttt{ OR }\aspas{recommendation model}\texttt{ OR }\aspas{recommending model}\texttt{ OR }\aspas{recommender method}\texttt{ OR }\aspas{recommendation method}\texttt{ OR }\aspas{recommending method}\texttt{ OR }\aspas{recommender engine}\texttt{ OR }\aspas{recommendation engine}\texttt{ OR }\aspas{recommending engine}\texttt{ OR }\aspas{recommender framework}\texttt{ OR }\aspas{recommendation framework}\texttt{ OR }\aspas{recommending framework}\texttt{ OR }\aspas{recommender interface}\texttt{ OR }\aspas{recommendation interface}\texttt{ OR }\aspas{recommending interface}\texttt{ OR }\aspas{recommender agents}\texttt{ OR }\aspas{recommendation agents}\texttt{ OR }\aspas{recommending agents}\texttt{ OR }\aspas{collaborative filtering}\texttt{ OR }\aspas{personalised recommendation} \\
 		\midrule
        \multirow{3}{\hsize}{\texttt{\makecell[c]{domain}}} & \aspas{health care}\texttt{ OR }\aspas{healthcare}\texttt{ OR }\aspas{wellness}\texttt{ OR }\aspas{well-being}\texttt{ OR }\aspas{wellbeing}\texttt{ OR }\aspas{active ageing}\texttt{ OR }\aspas{active aging}\texttt{ OR }\aspas{healthy ageing}\texttt{ OR }\aspas{healthy aging}\texttt{ OR }\aspas{healthy lifestyle}\texttt{ OR }\aspas{functional ability}\texttt{ OR }\aspas{functional abilities}\texttt{ OR }\aspas{intrinsic capacity}\texttt{ OR }\aspas{assisted living} \\
 		\midrule
        \multirow{3}{\hsize}{\texttt{\makecell[c]{population}}} & \aspas{geriatrics}\texttt{ OR }\aspas{geriatric}\texttt{ OR }\aspas{gerontology}\texttt{ OR }\aspas{gerontechnology}\texttt{ OR }\aspas{gerontechnologies}\texttt{ OR }\aspas{gerotechnology}\texttt{ OR }\aspas{gerotechnologies}\texttt{ OR }\aspas{gerontological}\texttt{ OR }\aspas{elder}\texttt{ OR }\aspas{elderly}\texttt{ OR }\aspas{senior}\texttt{ OR }\aspas{older person}\texttt{ OR }\aspas{older adult}\texttt{ OR }\aspas{older patient}\texttt{ OR }\aspas{older people}\texttt{ OR }\aspas{older population} \\
 		\midrule
 		\texttt{task} & \aspas{intervention}\texttt{ OR }\aspas{treatment}\texttt{ OR }\aspas{assessment}\texttt{ OR }\aspas{diagnostics} \\
 		\bottomrule
 	\end{tabularx}
	\label{tab:relatedwork:query}
\end{table}

\subsubsection{Review Execution}

\begin{figure}[t]
  \centering
  \includegraphics[scale=0.14]{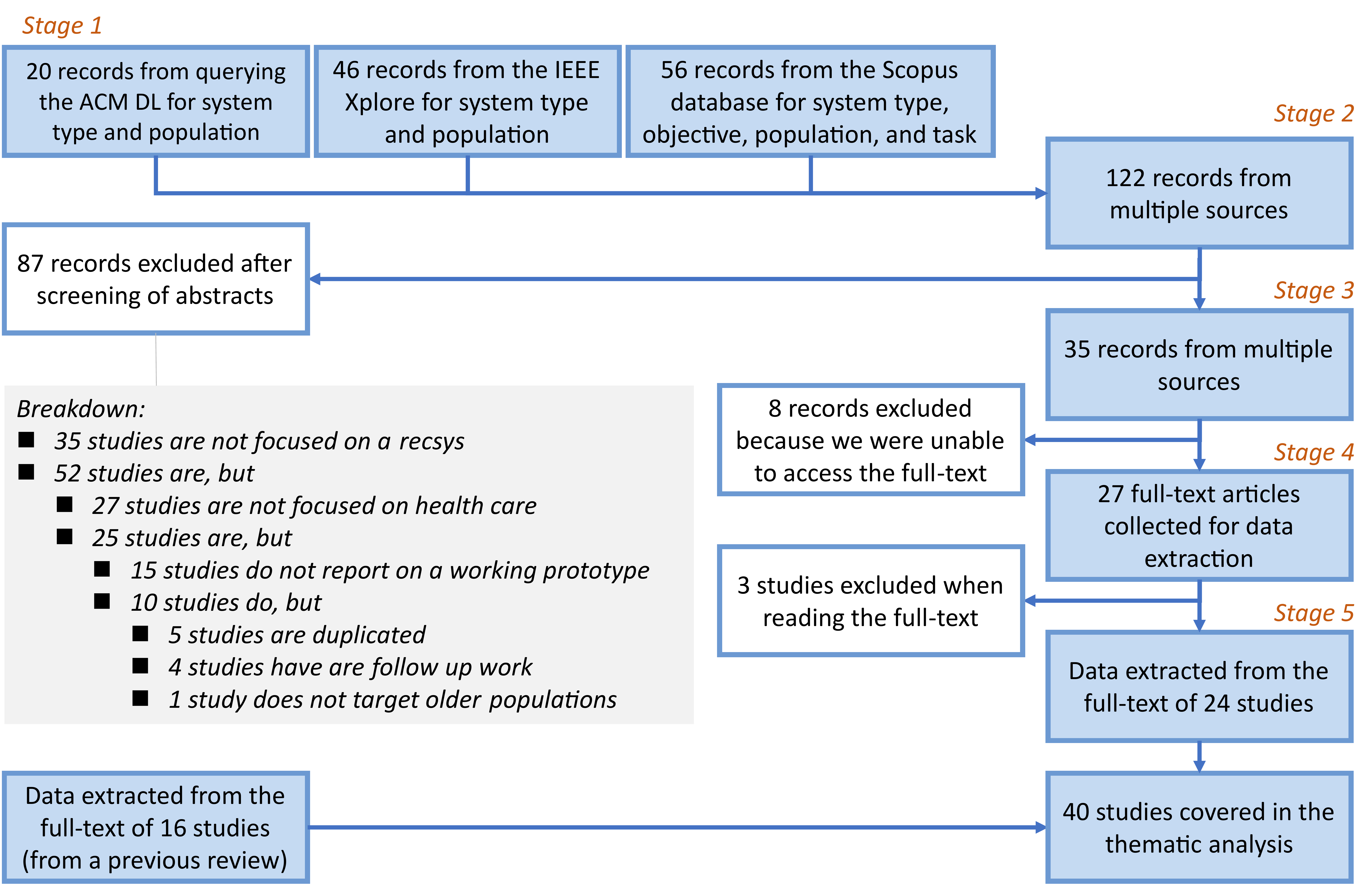}
  \caption{\fix{Results obtained from the execution of the review protocol, by stage}{image looks shitty}}
  \Description{A flow chart illustrating the application of the review protocol.}
  \label{fig:relatedwork:review-protocol}
\end{figure}

The results of the execution are illustrated in Figure \ref{fig:relatedwork:review-protocol}.
In Stage 1, 122 records were identified: 56 records were retrieved from the Scopus database, 46 from the IEEEXplore, and 20 records from the ACM/DL.
In Stage 2, 87 records were removed for varying reasons: 35 studies were excluded because they did not report on a recommender system (e.g., a study focused on some clinical recommendation framework); 27 were not focused on healthcare (e.g. education of older people), 15 were early works (e.g. \aspas{initial ideas of a PhD trajectory}), five were duplicate, four were excluded because they were follow ups of a research project already covered, and one was not focused on older populations. 
Of the 35 studies selected for data extraction (Stage 3), eight were excluded because we were unable to access the full-text of the study.
Of the remaining 27 studies, three were excluded after closer reading.
We extracted data from the remaining 24 studies and combined them with data from 16 studies from a previous review based on a snowballing protocol.


\subsubsection{Results and Discussion}
\label{section:background:healthrecsys:review-results}

The data extracted from the final 40 studies are presented in Table \ref{tab:relatedwork:review1}.
In summary, about one-third of the reviewed studies (13 of 40) use at least one standardised instrument to collect data on the target user.
Most of these instruments have been evaluated for their measurement properties from a psychometric perspective, as is the case of instruments used to measure intrinsic capacity, frailty, and quality of life (QoL) \cite{myself2021interpretable,espin2016nutelcare,vercelli2017myaha,azmi2019collaborative,zacharaki2020frailsafe,frikha2024senselife,kolakowski2025careup}.
However, some of these instruments were developed with econometric or clinimetric methods, traditions that compete with psychometrics in the healthcare space \cite{krabbe2016measurement}, such as the \sigladef{prognosis for patients with multiple pathologies}{PROFUND} index and the \sigladef{clinical dementia rating scale}{CDRS} \cite{rist2015care}.
There are studies that combine mature scales and other data to create inventories that survey typical CGA domains, such as physical medical conditions, mental health conditions, daily functioning, and social and environmental conditions, but the authors do not identify any validation studies \cite{azmi2019collaborative,stiller2010demographic}

Regarding the provision of explanations, about one-sixth of the reviewed studies (6 out of 40) report a recommendation system that can generate explanations.
Most of these systems provide explanations that are faithful (4 out of 6), and half of them have adopted an explanation style that consists in the visual display of quantitative information (3 out of 6), instead of the textual format that is typical of knowledge-based recommender systems.

To organise a closer look, the reviewed studies were classified into five categories:
\begin{itemize}
    \item C1: systems that use standardised assessment and provide faithful explanations, 2 studies.
    \item C2: systems that use standardised assessment but do not provide explanations, 11 studies.
    \item C3: systems that do not use standardised assessment but offer faithful explanations, 2 studies.
    \item C4: systems that do not use standardised assessment but offer post-hoc explanations, 2 studies.
    \item C5: systems that do not use standardised assessment or provide explanations, 23 studies.
\end{itemize}
No studies in which the system uses data from standardised assessments and provides post-hoc explanations were found.
The works in each category are discussed next.

\paragraph{C1}
The studies in this category are relatively recent.
\citeonline{myself2021interpretable} is an early contribution of our research project.
A partner research group interviewed 108 older subjects for quality of life using the WHOQOL-BREF instrument.
These assessments were combined with synthetic interventions to create the dataset used to evaluate the proposed recommendation model.
The idea of representing both assessments and interventions as polygons was already present, as the model recommends an intervention based on a geometric operation between the polygons.
The explanation style, which already resembles the visual style in Figure \ref{fig:intro:figure_1}, corresponds to a faithful depiction of the computation carried out by the recommendation model.
\citeonline{romanvillaran2022pites} reports on a system that prescribes (or deprescribes) drugs to older patients with comorbidities and polypharmacy (i.e., patients that make regular use of five or more medications simultaneously to manage chronic diseases).
This is a knowledge-based system that uses an ontology built on several clinical practice guidelines that have been published over the years.
Both the recommendation and the explanation are presented to a professional in a primary care setting, but unlike the previous study, both recommendations and explanations follow a textual style.
The authors report the use of several standardised assessments (from psychometric and clinimetric traditions), but it is unclear how the system takes their scores into account when generating new recommendations.

\paragraph{C2} The studies in this category make use of standardised assessments but do not provide explanations for given recommendations.
In most cases, the study reports the use of one or more (mature) psychometric scale or instrument (7 out of 11), but there are a few exceptions:
\begin{itemize}

    \item \citeonline{rist2015care} adapted an instrument to assess well-being in older populations developed by the Stanford Center on Longevity \cite{kaneda2011scl}, which seems to follow an econometric model.

    \item \citeonline{azmi2019review} used data obtained from assessments performed by caregivers on nursing home residents.
    The assessment consisted of a battery of instruments for diverse domains, such as physical (e.g., health conditions, functional abilities, nutrition), cognitive, social, and environmental.
    In other words, they combine data from mature scales with their own items.

    \item \citeonline{gannod2019machine} used the PELI-NH instrument, which captures the preference of a resident with regard to the services that are provided by the institution (e.g., the resident wants to choose her clothes) \cite{vanhaitsma2013preferences}.
    This instrument also seems to follow an econometric model.

    \item Although \citeonline{stiller2010demographic} and \cite{orte2018dynamic} did not identify the use of any particular instrument, the authors reported the use of questionnaires to assess a person in all the domains that are usually surveyed in CGAs, such as physical, psychological, social, and environmental.

\end{itemize}
A common feature of the studies in this category is that the recommendation is presented directly to the user (9 out of 11), with the exception of two studies in which the recommendations are directed at the nursing home staff \cite{gannod2019machine,azmi2019collaborative}.
Consequently, the topic of the recommendation falls predominantly in one of these classes: information about health, social, and recreational services available in the neighbourhood of the user, activities for the user to do at home (e.g., physical exercises to maintain or improve health, playing computer games for cognitive improvement), and lifestyle advice (diet plans, guidance to adapt home to avoid adverse events).

\paragraph{C3 and C4}
The studies in these categories report on systems that do not use standardised assessments, but do provide explanations.
The two studies in C2 are knowledge-based systems that provide faithful explanations.
The system devised by \citeonline{robinson2017improving} recommends adaptations to the daily routine in order to reduce energy consumption at home, and the explanation given is based on the amount of money that the system estimates will be saved by adopting the recommendation, while the system designed by \citeonline{rincon2019new} recommends physical activities for older people in nursing homes, and also explains the recommendation in terms of the expected benefit, such as \aspas{because your doctor has recommended it for lowering your cholesterol.}
In contrast, the two studies in C4 provide post-hoc explanations.
These studies, which were conducted by the same research group, propose mechanisms to generate explanations from black-box models applied to the diagnostic of chronic diseases that are highly prevalent in older populations \cite{wu2023lime,wu2023shap}.

\paragraph{C5} The majority of the reviewed studies fall in this category (23 of the 40): studies reporting on recommender systems that do not use standardised assessments and do not provide explanations.
Most studies report on recommender systems in which the recommendation is shown directly to the target user (19 of the 23) \cite{luo2010intelligent, murakami2010prevention, sasaki2013walking, ponce2015quefaire, silva2017tv4e, thakur2018smarthome, martinez2019pharos, allalouf2020music, mishra2020smarthome, shinde2021iotcars, sitparoopan2021homebridge, dhivakar2022cancer, minakata2022walking, martin2023digihealth, huang2024smarthome, prasetyo2024recipe, yuan2024shortvideo, bentlage2025mobile, jiang2025education}, with a few exceptions in which the recommendation is shown to a care professional \cite{bermingham2013recommending,oliva2018health,besik2019rhcs,zhou2024recipe}.
The topic of the recommendations shown to the target user falls predominantly in one of these classes: music or video for educational or therapeutic purposes, safe outdoor walking routes, information about health, social, and recreational services available in the neighbourhood of the user, activities for the user to do at home, and lifestyle advice (diet plans, guidance to adapt home to avoid adverse events).
In contrast, the topics of recommendations directed at care professionals are, for example, recipes to balance nutritional value and tastes of residents in a nursing home \cite{zhou2024recipe}, videos to be used in a group reminiscence therapy for older people with dementia \cite{bermingham2013recommending}, as well as other psychosocial interventions for older people with dementia \cite{oliva2018health}.
The system devised by \citeonline{besik2019rhcs} is an outlier in that it recommends the prescription of drugs to a medical professional, but does not seem to provide explanations for the recommendations given.

\begin{table}[htpb]
    \centering
    \footnotesize
    \caption{Results of the thematic analysis of the reviewed works in categories C1-C4}
    \begin{tabularx}{\linewidth}{@{}P{2.0cm}P{1.8cm}P{2.2cm}P{1.05cm}X@{}}
 		\toprule
	Study & \makecell[c]{Std. Assmt. \\ \hline Measurand} & \makecell[c]{System \\ \hline Region} & \makecell[c]{Exp. \\ \hline Faithful} & \makecell[c]{\emph{What} is recommended to/for whom\\ (and where/when and to what purpose)?}\\

 		\toprule
        \multicolumn{5}{c}{\textbf{(C1) Systems that use standardised assessment and provide faithful explanations}}\\
        \toprule
        
 	\citeonline{myself2021interpretable} & \makecell[t]{Yes\\QoL} & \makecell[t]{Poly\\Brazil (O)} & \makecell[t]{Yes\textdaggerdbl\\Yes} & health \emph{interventions} to care professional during counselling with older patient in primary care \\ \midrule

    \citeonline{romanvillaran2022pites} & \makecell[t]{Yes\\PROFUND} & \makecell[t]{PITeS-TIiSS\\Spain (1)} & \makecell[t]{Yes\textdaggerdbl\\Yes} & drug \emph{prescription} to care professional for older patients with polypharmacy in primary care \\ 

 		\toprule
        \multicolumn{5}{c}{\textbf{(C2) Systems that use standardised assessment but do not provide explanations}}\\
 		\toprule

 	\citeonline{stiller2010demographic} & \makecell[t]{Yes\\CGA} & \makecell[t]{Weitblick\\ Germany (393)} & \makecell[t]{No\\---} & health, social, care, recreational, and household \emph{services} close to older person's home \\ \midrule

    \citeonline{rist2015care} & \makecell[t]{Yes\\CGA} & \makecell[t]{CARE\\EU (21)} & \makecell[t]{No\\---} & physical, mental, and social \emph{activities} for older person living at home\\ \midrule

    \citeonline{espin2016nutelcare} & \makecell[t]{Yes\\malnutrition} & \makecell[t]{NutElCare\\Spain (O)} & \makecell[t]{No\\---} & weekly, varied \emph{dietary plans} to older persons living at home to prevent malnutrition \\ \midrule

    \citeonline{vercelli2017myaha} & \makecell[t]{Yes\\frailty} & \makecell[t]{My-AHA\\EU (600)} & \makecell[t]{No\\---} & activities and lifestyle \emph{advice}  to older adults at home to prevent or manage frailty \\ \midrule

    \citeonline{azmi2019collaborative} & \makecell[t]{Yes\\CGA} & \makecell[t]{---\\Malaysia (139)} & \makecell[t]{No\textdaggerdbl\\---} & health \emph{interventions} to improve well-being of older people living in nursing homes \\ \midrule

    \citeonline{gannod2019machine} & \makecell[t]{Yes\\PELI} & \makecell[t]{---\\USA (255)} & \makecell[t]{No\textdaggerdbl\\---} & suggestions to the nursing home manager to personalise \emph{services} provided to residents \\ \midrule

    \citeonline{zacharaki2020frailsafe} & \makecell[t]{Yes\\frailty} & \makecell[t]{FrailSafe\\EU (479)} & \makecell[t]{No\\---} & lifestyle and behavioural \emph{advice} to older people at home to prevent frailty \\ \midrule

    \citeonline{angelini2022nestore} & \makecell[t]{Yes\\CGA} & \makecell[t]{NESTORE\\EU (60)} & \makecell[t]{No\\---} & physical activities, social \emph{events}, diet plans and \emph{games} to older adults living at home \\ \midrule

    \citeonline{frikha2024senselife} & \makecell[t]{Yes\\IC} & \makecell[t]{Senselife\\EU (O)} & \makecell[t]{No\\---} & Health and social care \emph{services}, cultural and recreational \emph{events} to older person at home \\ \midrule

    \citeonline{llorente2024games} & \makecell[t]{Yes\\dementia} & \makecell[t]{PROCare4Life\\EU (93)} & \makecell[t]{No\\---} & guidance on using \emph{games} for cognitive training to older people with PD/AD living at home \\ \midrule

 	\citeonline{kolakowski2025careup} & \makecell[t]{Yes\\IC} & \makecell[t]{CAREUP\\EU (64)} & \makecell[t]{No\\---} &  \emph{activities} to older people at home to prevent decline in intrinsic capacity (IC)\\ 

 		\toprule
        \multicolumn{5}{c}{\textbf{(C3) Systems that do not use standardised assessment, but offer faithful explanations}}\\
 		\toprule

    \citeonline{robinson2017improving} & \makecell[t]{No \\ ---} & \makecell[t]{---\\UK (10)} & \makecell[t]{Yes \\ Yes} & lifestyle \emph{advice} to older people about how to reduce energy consumption at home \\ \midrule

    \citeonline{rincon2019new} & \makecell[t]{No \\ ---} & \makecell[t]{EmIR\textdagger \\ Portugal (E)} & \makecell[t]{Yes\\Yes} & health promoting \emph{activities} to older person living at a nursing home \\ 
    
 		\toprule
        \multicolumn{5}{c}{\textbf{(C4) Systems that do not use standardised assessment, but offer post-hoc explanations}}\\
 		\toprule
        
    \citeonline{wu2023lime} & \makecell[t]{No\\---} & \makecell[t]{---\\USA (O)} & \makecell[t]{Yes\textdaggerdbl\\No} & health \emph{interventions} to professionals in care of older people to prevent chronic diseases \\ \midrule
    
    \citeonline{wu2023shap} & \makecell[t]{No\\---} & \makecell[t]{---\\USA (O)} & \makecell[t]{Yes\textdaggerdbl\\No} & health \emph{interventions} to professionals in care of older people to prevent chronic diseases \\ 

        

        \bottomrule
 	\end{tabularx}
	\label{tab:relatedwork:review1}
    \justifying \footnotesize \emph{Legend:} The 2nd column indicates whether/which instruments are used. 
    The 3rd column shows the system name and the region in which the study participants reside.
    A dagger (\textdagger) beside the system name indicates that it implements a human-robot interface.
    The number beside the region is the number of participants; (O) indicates an offline evaluation, and (E) an early study which does not report results.
    The 4th column indicates whether explanations are provided, and if they are faithful; a double dagger (\textdaggerdbl) indicates an expert-in-the-loop.
    Topic of recommendation is \emph{highlighted} in the 5th column.
\end{table}

\subsubsection{Conclusion}
\label{section:background:healthrecsys:review-conclusion}

The majority of the surveyed studies report on recommender systems that do not use data from standardised instruments to produce recommendations (i.e., 27 out of 40 studies were negative for Q1).
Moreover, most studies report on systems that do not have explanation facilities (34 out of 40 were negative for Q2), and among the recommender systems that do provide explanations, two-thirds generate faithful explanations (4 out of 6).
This seems to be in agreement with the assumption that, in most cases, the object of the recommendation does not have the potential to produce significantly harmful outcomes: the recommendations are presented directly to the target user, who has the adequate experience to judge their relevance and decide to follow the recommendation only when appropriate.
The results also show a disconnect between current blueprints of healthcare recommender systems for older adults and the practice in gerontological primary care described in Section \ref{section:background:cga}: professionals rely on standardised instruments to perform CGA, and use the assessment results to develop personalised care plans for the patients.
To operate as an assistive technology in this niche, a recommender system would probably benefit from using CGA data and providing faithful explanations, so that the care professional can understand the rationale behind the recommendation and assess its merits and shortcomings.

\subsection{Multilabel classification and label ranking tasks}
\label{section:relatedwork:tasks}


In a nutshell, multilabel classification refers to the task of learning how to map an instance to a set of predefined labels \cite{brinker2006unified}.
For example, the task of predicting what kind of food (Arabic, Chinese, etc.) a person is inclined to consume in food trucks is commercially valuable, and learning how to do it can be framed as a multilabel classification task \cite{rivolli2017foodtruck}.
Label ranking builds on this idea, as it requires not only mapping instances to sets of labels, but also sorting the labels based on some measure.
In the previous example, the ranking could tell us that a person likes Italian food the best.


\subsubsection{Preliminaries}
\label{section:relatedwork:tasks:formalisation}

Let the tuple $D :=(X,Y)$ represent a dataset, with $X$ being an $(m,d)$-matrix holding the data of $m$ subjects regarding $d$ attributes\footnote{This notation conflicts with the one used in Section \ref{section:background:measurement-of-health}, where the preferred notation in the structural equations literature was used.
From now on, we will keep the usual notation in the recommender systems literature, which places user as rows.
}, and $Y$ being an $(m,n)$-matrix holding the data about how each subject is associated with each of the $n$ labels in the label set $\mathcal{L}$.
Let's call $X$ the description matrix, and $Y$ the assignment matrix.
Let $X_i = (x_{i0}, \ldots, x_{i,d-1}) \in \mathcal{X}$ denote the $i$-th row of $X$, with $x_{ik}$ representing the score the $i$-th individual obtained for the $k$-th attribute, and $\mathcal{X}$ is the example space where $x_{ik}$ can take a categorical or numeric value, and no value is missing.

\paragraph{Multilabel Classification Task}
In this task, we say that $Y$ is a multilabel assignment.
\convention{We adopt the usual encoding in benchmark datasets, in which $y_{ij} = 1$ indicates that the $i$-th individual is assigned to the $j$-th label, and $y_{ij} = 0$ indicates the opposite.}{MLC encoding}
In this setup, the solution of a multilabel classification problem $D$ is a function $h: \mathcal{X} \to 2^\mathcal{L}$, which is often called a hypothesis to highlight that $h \in \mathcal{H}$, the hypothesis space that is spanned by a model.
Following \citeonline{madjarov2012multilabel}, let $q$ be a quality criterion that rewards hypotheses with high predictive performance and low complexity.
Then, the aim of a multilabel classification task is to learn a function $h \in \mathcal{H}$ that maximises $q(h, D)$.

\paragraph{Label Ranking Task}
This task encompasses many problem variants, and the latter determine the structure of admissible solutions.
In this work, we focus on problems whose admissible solutions are rankings with no ties.
To that purpose, let the relation $\succ$ denote the preference among two labels: $a \succ b$ means that $a$ is preferred to $b$.
\convention{For a label ranking assignment $Y$, we adopt the usual encoding in benchmark datasets.}{LR encoding}
For example, assume $\mathcal{L} = \{0, 1, 2, 3\}$.
Then, $Y_{i}=(1,2,0,3)$ means that the $i$-th individual is associated with the ranking $1 \succ 2 \succ 0 \succ 3$.
\convention{To encode incomplete rankings, we use $-1$ as a filler}{LR encoding}: $Y_{i}=(1,2,-1,-1)$ encodes the ranking $1 \succ 2$.
In this setup, the solution of a label ranking problem (with no ties and that allows for incomplete rankings) is a function $h: \mathcal{X} \to \mathscr{A}(2^\mathcal{L})$, with $\mathscr{A}(2^\mathcal{L})$ being the union of the permutations of all subsets of $\mathcal{L}$ \cite{brinker2006unified}.
This definition highlights the fact that rankings of varied sizes can be encoded in the same assignment $Y$, e.g., given two subjects $a$ and $b$, it may be the case that $Y_a = (1, 2, 0, 3)$ and $Y_b = (1, 2, -1, -1)$.
As before, the aim of a label ranking task is to learn a function $h \in \mathcal{H}$ that maximises $q(h,D)$.

\paragraph{Description Metrics}
Often, researchers face the need to concisely describe an assignment matrix.
To do this, they resort to the metrics in the literature that have been devised to describe certain aspects of this matrix \cite{herrera2016multilabel}.
For example, the cardinality metric refers to the average number of labels per instance, and the density metric corresponds to the average number of cells that actually map a subject to a label.
These metrics describe an assignment matrix by its rows.
In contrast, the maximum imbalance ratio describes its columns, since it computes the ratio between the most frequent and least frequent labels.
Two other recurring metrics in the literature involve the concepts of labelset and single labelset.
The term \aspas{labelset} (not to be confused with the label set $\mathcal{L}$) refers to the number of distinct rows of an assignment matrix, and the term \aspas{single labelset} refers to a labelset that occurs only once in an assignment matrix.

\subsubsection{Learning to Create Personalised Care Plans}
\label{section:relatedwork:tasks:careplan}

In the study reported by \citeonline{tavassoli2022ICOPE} mentioned earlier, 
assessments of intrinsic capacity of more than ten thousand older participants were collected.
A team of primary care professionals met with the participants with health issues, and a personalised care plan was created for each of these individuals.
The findings of the study indicate that a set of 22 distinct interventions or referrals were made by the attending professionals.

From a computational perspective, one could frame the creation of care plans based on patient assessment data as a multilabel classification problem.
The set of labels $\mathcal{L}$ is the set of interventions or referrals that the regional health system provides.
The patient assessments provide the description of the subjects in $X$, and the recommendations made by the group of attending professionals correspond to the assignment matrix $Y$.
The task corresponds to learning a function $h$ that maps a patient assessment $X_i$ to a care plan $\hat{Y}_i$ that contains exactly the same recommendations made by the attending professional.
The success of this learning task is measured by the degree to which $h$ can replicate the decision-making of the care professionals observed in this particular setting.

Similarly, if the order in which an intervention or referral appears in a care plan is important, a new constraint must be introduced. 
This constraint, although more challenging to satisfy, may be useful, as it can guide the patient and her family on how to prioritise the available resources to obtain the best health outcome for the patient.
For example, in the CGA inventory promoted by the SBGG that we mentioned in Section \ref{section:background:cga}, there is a section in which the care professional records the personalised care plan that was agreed with the patient and family.
The recommended interventions are separated by priority, with interventions that are capable of reducing functional decline having the highest priority.
From a computational modelling perspective, this could be framed as a label ranking task as we defined earlier: solutions are incomplete rankings with no ties.

\subsubsection{Solving Multilabel Classification Tasks}
\label{section:relatedwork:tasks:multilabel}

The methods for multilabel classification tasks are traditionally divided into two broad categories: problem transformation and algorithm adaptation approaches \cite{tsoumakas2007multilabel,madjarov2012multilabel, herrera2016multilabel}.
In the problem transformation approach, one of the simplest methods consists in decomposing a multilabel assignment $Y$ into $n$ single-label assignments, which are then treated as $n$ independent binary classification problems specified as $D_j = (X, Y_{:j})$ for $j \in 0 \ldots (n-1)$.
Accordingly, $n$ separate solutions $h_j : \mathcal{X} \to \{0, 1\}$ are learned and then concatenated to produce a solution for the original problem.
Thus, in prediction time, $\hat{Y}_i = h(X_i) = (h_0, \, \ldots, h_{n-1}) \, (X_i)$.
This is known as the binary relevance method.
Regarding the algorithm adaptation approach, the idea is to choose a model that supports binary classification in single output, and adapt its inner workings to produce multiple outputs.
More precisely, the aim is to adapt a model that generates hypotheses $h \in \mathcal{H} \mbox{ such that } h: \mathcal{X} \to \{0, 1\}$ to the multilabel setting.
For example, ML-C4.5 is an adaptation of the C4.5 algorithm.
The adaptation extends the content of the leaf nodes (to store multiple labels) and replaces the node-splitting measure by a function that considers multiple labels at once \cite{herrera2016multilabel}.

In a comprehensive study that compares the performance of multilabel classifiers, with plenty of representatives of both major approaches, \citeonline{bogatinovski2022multilabel} report that two models, one from each approach, appear at the top of the performance ranking.
These models, identified in their study as \sigladef{Binary Relevance with Random Forests of Decision Trees}{RFDTBR} and \sigladef{Random Forest of Predicting Clustering Trees}{RFPCT}, are both based on random forests.
The study compared 26 methods on 42 datasets across a spectrum of 18 evaluation metrics.
The datasets used in the evaluation have been made publicly available by the curators of the \href{https://cometa.ujaen.es/datasets/}{Cometa repository} \cite{charte2018cometa}.
Regarding the metrics used to evaluate performance, the list includes both example- and label-based metrics.
In the example-based category, one finds the subset-accuracy and the Hamming loss.
The first corresponds to the fraction of predicted labelsets that exactly match their respective ground truth, and the latter measures the fraction of predicted cells that diverge from their respective ground truth.
In the label-based category, one finds the micro- and macro-averaged F1 scores.

\subsubsection{Solving Label Ranking Tasks}
\label{section:relatedwork:tasks:labelranking}

As seen in Section \ref{section:relatedwork:tasks:careplan}, the problem of building a personalised care plan can be modelled as a label ranking task with a modest number of labels.
This makes the literature on preference learning a natural field for looking for solutions \cite{furnkranz2011preference}.
In this tradition, one of the most versatile approaches focuses on learning a separate scoring function for each individual label, which are processed by a ranker.
More precisely, let $g_j: \mathcal{X} \to \mathbb{R}$ for $j \in 0 \ldots n-1$ be a scoring function, and $g(X_i) := (g_0, \, \ldots, g_{n-1}) \, (X_i)$ their concatenation.
Then, a naive solution can be expressed as $g_{\succ}(X_i) := \argsort -g(X_i)$  \cite{fotakis2022labelranking}.
However, this is not a good solution when the ground truth includes incomplete rankings.
A common remedy to this issue is to combine a multilabel classifier and a label ranker so that irrelevant labels are removed from the ranker's output \cite{brinker2007ranking}.
Another popular approach, which is based on decision trees, consists of extending nodes to represent rankings and adapt the node-splitting function to make an efficient use of the ranking data, similar to the algorithm adaptation approach in multilabel classification tasks.

In fact, \citeonline{zhou2018ranking} find that tree-based models achieve highly competitive results compared to models following other traditional approaches to label ranking, and more recently, \citeonline{fotakis2022labelranking} report that random forests usually rank higher than other tree-based models.
The first study compared the performance of an adapted random forest against 11 other models on 16 datasets, and the second study reports a comparison of an adapted random forest against four other tree-based models on 21 datasets (16 semi-synthetic, 5 real-world).
Both studies used the datasets that have been made publicly available by the curators of the \href{https://en.cs.uni-paderborn.de/de/is/research/research-projects/software/label-ranking-datasets}{Paderborn repository} \cite{cheng09icml}.
Regarding the performance metric, both studies report results using a variant of the Kendall's $\tau$ coefficient for tie-free rankings, averaged over the test cases.

\subsubsection{Conclusion}
As a closing thought, it must be noted that two of the top-ranking models for multilabel classification tasks in the extensive evaluation reported by \citeonline{bogatinovski2022multilabel} were based on random forests, as the top-ranking models for label ranking tasks reported by \citeonline{zhou2018ranking} and \citeonline{fotakis2022labelranking}.
Because of its success in both tasks, it will be explored during the evaluation of our model.
For this reason, we highlight some details about the architecture of random forests.
It follows an ensemble strategy to induce multiple decision trees during training.
Each tree is induced from a random subset of the training data and, at each node, only a random subset of features is considered for splitting.
This double-bagging strategy improves the bias-variance trade-off of decision trees that are allowed to grow unpruned.
In analogy to the way the size of multilayer perceptrons is measured, the size of a random forest is computed by summing up the sizes of its individual trees, and the size of a tree is calculated by summing up the number of \aspas{weights} needed to represent its internal and terminal nodes.
For example, if the trees are induced using the CART algorithm, then the resulting trees are binary trees, and each decision node can be represented with two weights (one to indicate a feature, and one to hold its corresponding threshold value), and each leaf can be represented by a single weight (that indicates a labelset or ranking).
The fact that the number of trees is a hyperparameter of this model endows the random forest with the ability to produce models of arbitrarily large size.
Finally, in prediction time, the model merges the results of the individual trees (e.g., by majority vote in classification, by averaging their outputs for regression) to produce a combined prediction.

\subsection{Task-based Effectiveness of Elementary Visualisations}
\label{section:relatedwork:visual}
What is a diagram?
They are part of our daily experience, but giving them an intensional definition is an elusive quest \cite{legg2013diagram}.
The visualisation research community defers this issue by delimiting the scope of interest to visual displays of data that are relevant for and interpretable by users performing certain tasks.
These tasks are then placed at the centre of their research agenda.
The latter point is demonstrated by the many diverse classification schemes (e.g., typologies and taxonomies) for visualisation-supported tasks that the community has developed over the years \cite{dimara2022critical}.
The focus of this brief review is on tasks in which the user is prompted to make a decision, and thus the role of the diagram is to facilitate the successful completion of a decision-making task.

As stated by \citeonline{dimara2022critical}, most classifications omit decision-making tasks.
Instead, they usually focus on analytic tasks, which are tasks in which the user's goal is served by querying a dataset.
An exception is the typology devised by \citeonline{brumar2025typology}, which focuses on decision-making tasks and classifies them into three composable categories: choose, activate, and create.
Given an initial set of options as input, the first corresponds to selecting the top $k$ options by comparing them with each other; the second to independently evaluating the options against a threshold; and the third to combining or generating new options or new information.
Being conceptual by design, this typology does not inform which visualisations (e.g., bar charts or scatterplots) are more effective for a given decision-making task.
However, such guidance exists for low-level analytic tasks, such as those in the taxonomy created by \citeonline{amar2005taxonomy}.
In the latter, one finds (a) the sort tasks, whose goal is \aspas{given a set of data cases, [to] rank them according to some ordinal metric}, (b) the find extrema tasks, aiming at \aspas{find[ing] data cases possessing an extreme value of an attribute over its range within the dataset}, and (c) the filter tasks, whose goal is \aspas{given some concrete conditions on attribute values, [to] find data cases satisfying those conditions.}

One could argue that, in a decision-making scenario in which the user must inspect a dataset, there is a correspondence between these schemes:
(a) the inputs and outputs of the concatenation of a sort task and a find extrema task is described by the choose task, and (b) the activate task describes the inputs and outputs of a filter task.
This circumstantial correspondence between decision-making and analytic tasks can be evoked to extrapolate the empirical results in the literature to attempt an answer to the question about the effectiveness of basic visualisations in decision-making tasks.

For example, \citeonline{saket2019task} conducted a user study to assess five visualisations (bar chart, pie chart, line chart, scatterplot, and tabular presentation) across the ten low-level analytic tasks in the taxonomy by \citeonline{amar2005taxonomy}.
Each participant was randomly assigned to a category (e.g., filter) and, in the main part of the study, was asked to complete 30 tasks (five visualisations, two datasets, and three trials).
The authors report that the bar chart is strongly associated with higher participant performance in general, and associated with highly competitive participant performance in the analytic tasks we singled out earlier, which we associated with decision-making tasks.
This result will be used in Section \ref{section:userstudy} to inform the design of the user study to assess interpretability.

\section{An Interpretable Recommendation Model for Psychometric Data}
\label{section:proposal}

\begin{figure}[htpb]
    \centering
    \includegraphics[width=1\linewidth]{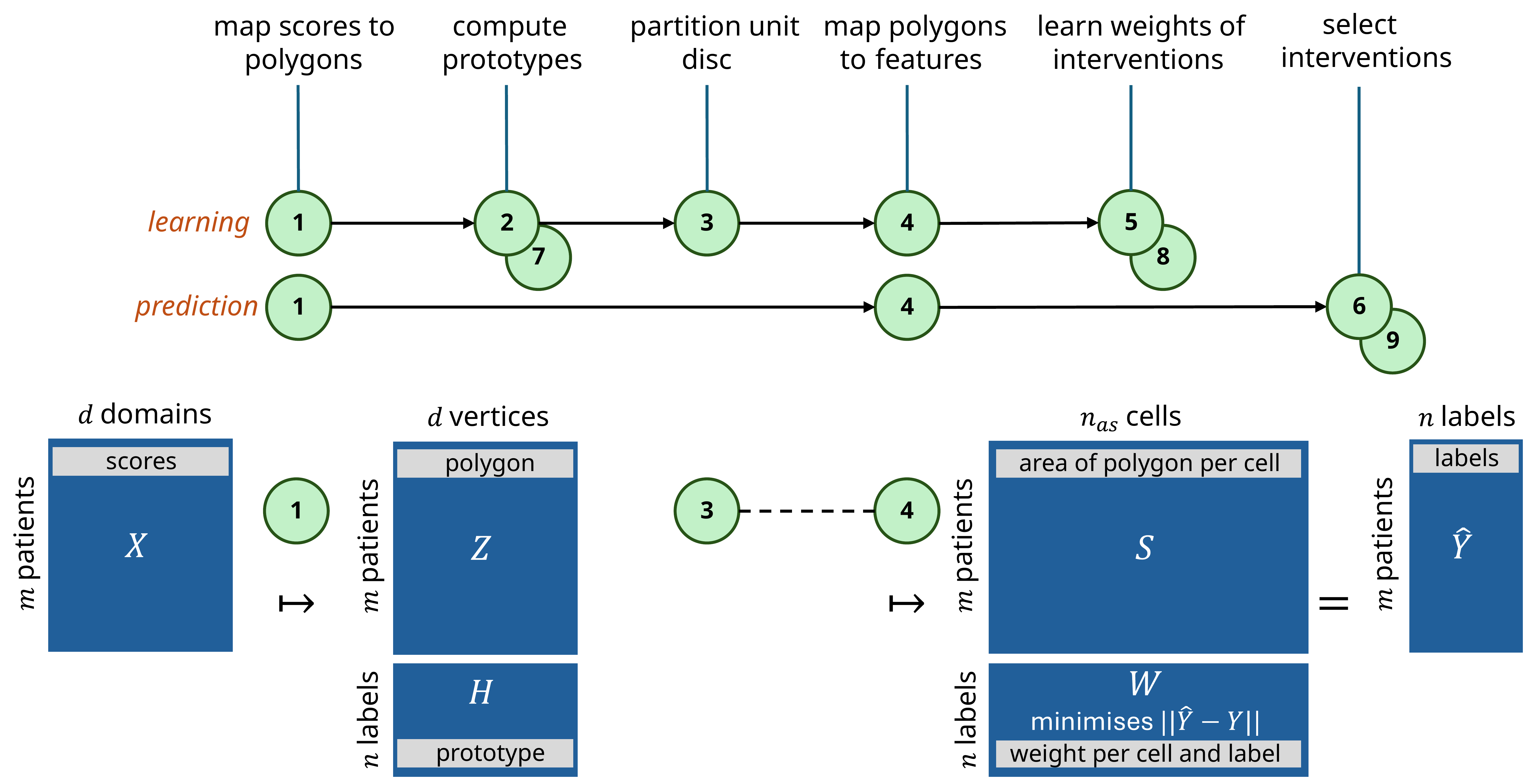}
    \caption{The learning and prediction pipelines of the Polygrid model.
    \emph{Legend:} The pipelines are depicted at the top, as two directed paths.
    Each node corresponds to an algorithm (in the text).
    The steps of the learning pipeline for multilabel classification are indicated by the nodes 1 to 5.
    The learning pipeline for label ranking uses the \aspas{shadow} nodes: 1, 7, 3, 4, and 8.
    The prediction pipeline is encoded similarly.
    The matrices depicted at the bottom of the diagram illustrate how the input data are transformed along the pipelines, e.g., assessments in $X$ are transformed into polygons in $Z$ by the algorithm corresponding to the step 1.
    }
    \Description{}
    \label{fig:proposal:pipelines}
\end{figure}

This section introduces the \textbf{Polygrid} model.
The exposition is organised around its learning and prediction pipelines, which are illustrated in Figure \ref{fig:proposal:pipelines}.
Each pipeline step is described from two perspectives: algorithmic and diagrammatic.
The first informs how the input data are transformed into the output data, and the latter describes how the input and output data are displayed in the diagram.
This choice of exposition aims to argue for the faithfulness of the explanation diagrams.

The remainder of this section is organised as follows.
First, our usage of the term \aspas{psychometric data} is made more precise by presenting the criteria the data must satisfy to be included in this category (Section \ref{section:proposal:data}).
The whoqol dataset is shown to meet the criteria, and a sample which will be used in a running example is drawn and described.
In Section \ref{section:proposal:ml-task}, we describe how Polygrid learns to perform multilabel classification tasks.
The learning and predicting pipelines are described, and the impact of alternative choices of hyperparameters is briefly discussed.
Section \ref{section:proposal:learning-lr} is similarly structured, but details how Polygrid learns to perform label ranking tasks.
Having covered a number of examples of how Polygrid learns, predicts, and produces explanations, we dedicate Section \ref{section:proposal:closerlook} to present a theoretical basis for how Polygrid learns, and Section \ref{section:proposal:interpretable} to discuss in which sense the claim that explanations generated by Polygrid are faithful and interpretable 
is valid.

\subsection{Data Requirements and Data Preparation}
\label{section:proposal:data}

We use the term \aspas{psychometric data} to refer to data collected with psychometric instruments based on the congeneric model in Figure \ref{fig:background:selvm:mm}.
This constraint has two relevant implications for a dataset collected with the instrument: 
the scoring procedure yields numeric values that are quantitative and nonnegative (Eq. \ref{eq:background:weighted-score}), and the pairwise correlation between domain scores is positive (Eq. \ref{eq:background:corrpos}).

Let $\mathring{D}=(\mathring{X},Y)$ represent such a dataset.
We assume that the original scores in $\mathring{X}$ are organised as an $(m,d)$-matrix, in which $m$ is the number of individuals, and $d>2$ is the number of domains of the instrument.
Similarly, the assignments in $Y$ are organised as an $(m,n)$-matrix, with $n > 1$ as the number of labels in the problem space.
The usual encodings described in Section \ref{section:relatedwork:tasks:formalisation} are adopted.

As for data preparation, the original assessments $\mathring{X}$ must be scaled to the unit hypercube $[0, 1]^d$.
In other words, Polygrid requires a dataset $D=(X,Y)$ such that $x_{ik} \coloneq \mathring{x}_{ik}/max(\mathring{X}_{:k})$.
\premise{Once differences in range among domains are removed, namely $\mbox{\textit{rng}}(X_{:k}) \subseteq [0,1] \mbox{ for all } \, k \in 0 \ldots d-1$, all domains are given an equal footing on the assignments, leaving to the learning model the work of identifying their relative relevance.
}{uh}
This also has an important consequence for the explanation diagram.

Having clarified the requirements a dataset must satisfy so that it can be handled by Polygrid, we now show that the whoqol dataset fits the bill.
This dataset was collected by researchers affiliated to the Department of Gerontology at the Federal University of São Carlos, Brazil, in October 2019 \cite{castro2022whoqol}, and contains quality of life assessments of 100 individuals.
The assessments were made with the WHOQOL-BREF instrument (see Annex \ref{section:background:cga:instruments} for details).
This instrument has four domains: physical health, psychological, social relationships, and environment, and its scores range from 4 and 20 inclusive.
The results shown in Table \ref{tab:whoqol-stats} attest that the whoqol dataset satisfies the requirements regarding scores and pairwise correlations (Equations \ref{eq:background:weighted-score} and \ref{eq:background:corrpos}, respectively).

\begin{table}[htb]
    \centering
    \small
    \caption{Statistics of the whoqol dataset. The statistics under the Global heading consider the whole dataset, but the ones under the Correlations heading consider the training set only.}
    \label{tab:whoqol-stats}
    \begin{tabular}{l|rrr|cccc}
         \hline
                &      \multicolumn{3}{c}{Global}      & \multicolumn{3}{c}{Correlations}\\
         Domain & Min. &  Avg. &  Max. & Psychological & Social & Environment\\
         \hline
         Physical      & 10.9 & 13.9 & 17.1 & 0.516 & 0.326 & 0.407 \\
         Psychological & 10.0 & 14.1 & 18.0 & -     & 0.548 & 0.590 \\
         Social        &  5.3 & 14.9 & 20.0 & -     & -     & 0.408 \\
         Environment   &  7.0 & 14.9 & 20.0 & -     & -     & -     \\
         \hline
    \end{tabular}\\
    \noindent
    \justifying \footnotesize 
\end{table}

Originally, this dataset did not have any assignments, so we created two synthetic assignments to extend it.
The first is a multiclass assignment based on a stratification method proposed by \citeonline{silva2014whoqolcut}.
The method employs the sum-score of an assessment (Equation \ref{eq:background:weighted-score}) and a cutoff value to specify a decision boundary.
The latter segments the members of a population into two disjoint groups: those with perceived good quality of life (Good QOL, sum-score $\geq 60$) and those with perceived poor quality of life (Poor QOL, sum-score $< 60$).
As a result, 41 individuals were assigned to \aspas{Good QOL} group, and 59 to the \aspas{Poor QOL} group.
The second assignment is a label ranking assignment.
It was created by applying fuzzy clustering to the assessment scores to obtain four clusters, each identified with a label.
Then, each assessment was assigned to a ranking composed of two labels, representing the clusters to which that assessment obtained the highest degrees of membership.
Table \ref{tab:whoqol-sample} shows the assessments and corresponding assignments for three individuals in the dataset, which will be used in the running example in the next sections.

\begin{table}[htb]
    \caption{\centering Description of the whoqol dataset's instances used in the examples}
    \label{tab:whoqol-sample}
    \begin{tabular}{c|cccc|cccc|cc|cccc}
        \hline
        & \multicolumn{4}{c}{$\mathring{X}$} & \multicolumn{4}{c}{$X$} & \multicolumn{2}{c}{$Y$ (ML)} & \multicolumn{4}{c}{$Y$ (LR)}\\
        $i$ & $\mathring{x}_{i0}$ & $\mathring{x}_{i1}$ & $\mathring{x}_{i2}$ & $\mathring{x}_{i3}$ & $x_{i0}$ & $x_{i1}$ & $x_{i2}$ & $x_{i3}$ & $y_{i0}$ & $y_{i1}$ & $y_{i0}$ & $y_{i1}$ & $y_{i2}$ & $y_{i3}$\\
        \hline
        89 &   13.1 &   11.3 &   12.0 &   10.5 & 0.767 & 0.630 & 0.600 & 0.525 &       0 &       1 &       0 &       2 &      -1 &      -1 \\
        98 &   14.3 &   14.7 &   14.7 &   15.0 & 0.833 & 0.815 & 0.733 & 0.750 &       0 &       1 &       1 &       2 &      -1 &      -1 \\
        30 &   14.9 &   15.3 &   17.3 &   16.0 & 0.867 & 0.852 & 0.867 & 0.800 &       1 &       0 &       3 &       1 &      -1 &      -1 \\
        \hline
    \end{tabular}\\
    \noindent
    \footnotesize \justifying \emph{Legend}: Column \aspas{$i$} shows the numeric key that was assigned to the individual in the dataset.
    Columns grouped under \aspas{$\mathring{X}$} show the original scores of the assessment, and the ones under \aspas{$X$} show the scaled scores.
    For both $\mathring{x}_{ik}$, $x_{ik}$, the order of the domains is the following: physical health ($k\!=\!0)$, psychological ($k\!=\!1$), social relationships ($k\!=\!2$), and environment ($k\!=\!3$).
    Columns grouped under \aspas{$Y$ (ML)} represent the multilabel assignment: ($y_{i0}\!=\!1$) indicates the individual is associated with the \aspas{Good QOL} label, and ($y_{i1}\!=\!1$) indicates association with the \aspas{Poor QOL} label.
    Finally, columns under \aspas{$Y$ (LR)} describe the label ranking assignment: $Y_{i}\!=\!(0, 2, -1, -1)$ indicates that the $i$-th the individual is associated with the incomplete ranking $0 \succ 2$.
    Each individual is assigned to the two labels that best describe her assessment of quality of life.
\end{table}

\subsection{The Polygrid Model in Multilabel Classification Tasks}
\label{section:proposal:ml-task}

This section is divided into three parts.
The first two are dedicated to detail the learning and prediction pipelines illustrated in Figure \ref{fig:proposal:pipelines}.
In brief, during the learning stage, Polygrid goes through five steps to learn representations for the assignment labels (i.e., the assignment weights in $W$).
These steps are detailed in Section \ref{section:proposal:ml-task:learning}.
To make predictions, Polygrid goes through the three steps that are detailed in Section \ref{section:proposal:ml-task:prediction}.
To keep the exposition simple, in both sections, the operation of the model is detailed for a default configuration of hyperparameters.
Alternative configurations, as well as their impact on the performance of the model, are contrasted with the default configuration in Section \ref{section:proposal:ml-task:options}.

\subsubsection{Learning from Data}
\label{section:proposal:ml-task:learning}
\begin{figure}[htpb]
    \centering
    \includegraphics[width=.993\linewidth]{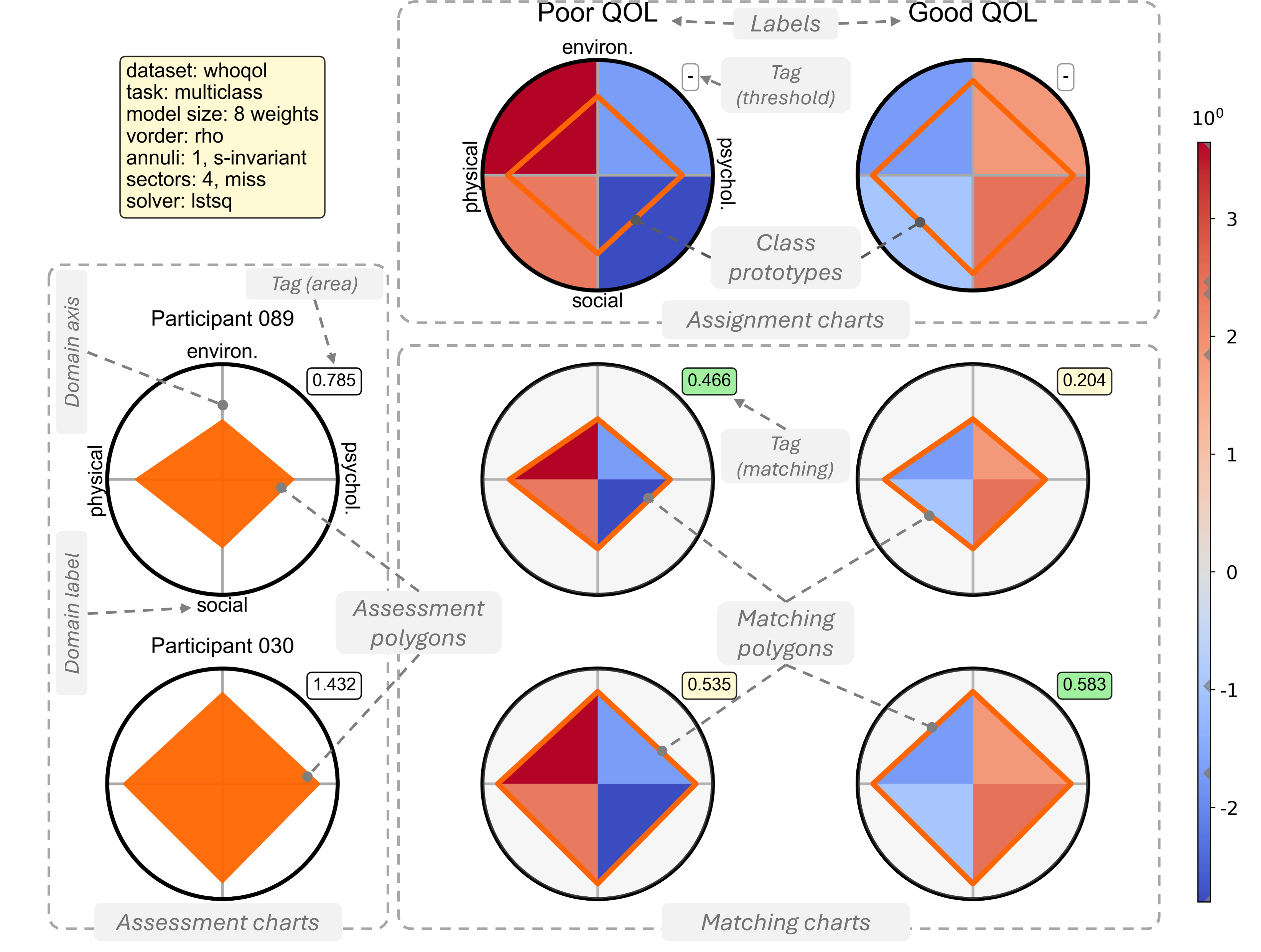}
    \caption{An explanation diagram is composed of a number of radar charts disposed on a rectangular grid.
    The diagram being displayed has been annotated with overlay elements (in grey) to single out its constitutive elements.
    There are three types of charts: (a) the assessment charts, which are placed on the first column, (b) the assignment charts, on the first row, and (c) the matching charts.
    \colorbox{gray!10}{The assessment chart} depicts the results of a patient assessment.    
    The vertices of an assessment polygon represent the scores the patient obtained for each domain of the instrument.
    Each assessment chart has a tag, which informs the area of its corresponding polygon.
    Typically, an explanation diagram displays the assessment of a single individual, but this example includes an extra one to benefit our discussion.
    \colorbox{gray!10}{The assignment chart} depicts the representation of a label in the feature space: the colours in the background correspond to the weights ascribed to different cells of the unit disc.
    The polygon in an assignment chart, whose vertices are the mean scores of all assessments assigned to its respective label, is called the class prototype.
    Each assignment chart has a tag, which informs the threshold value for its respective class.
    In multiclass assignments, this value is omitted because each case is assigned to a single label, as described next.
    \colorbox{gray!10}{In a matching chart,} the polygon is a copy of the assessment polygon on the same row, except that it is filled with \aspas{the colours} of the assignment chart in the same column.
    Each matching chart has a tag, which informs the weighted area of its polygon.
    A green tag indicates that this value is greater than its respective threshold, and the case is assigned to the respective label.
    }
    \Description{}
    \label{fig:proposal:ml-task:default}
\end{figure}

Roughly speaking, during the learning stage, Polygrid computes the data structures that are needed to render an explanation diagram.
These data structures are either related to the assessments in $X$, to the assignments in $Y$, or to the feature space.
We will refer to Figure \ref{fig:proposal:ml-task:default} to point where these structures are visually depicted on the Polygrid diagram.

\paragraph{Step 1}
The first step consists in mapping the assessments in $X$, whose rows are points in the unit hypercube $[0,1]^d$, into polygons in the closed unit disc centred at the origin of the complex plane $\mathbb{\overline{D}}$.
This mapping produces simple, solid polygons whose vertices are scalar multiples of the complex roots of unity\footnote{A simple polygon is a planar shape whose boundary is a simple polygonal chain (i.e., a chain that is not self-intersecting) that is closed, and a solid polygon is a planar shape that includes both the boundary and the interior of a simple polygon, and its interior does not have holes.}, $\zeta^d=1$, confined to the unit disc $\mathbb{\overline{D}}$.
In other words, it maps $X_i = (x_{i0}, \ldots, x_{i,d-1})$ to $Z_i = \poly(x_{i0} \, \zeta_0, \ldots, x_{i,d-1} \, \zeta_{d-1})$, with $\zeta_k$ being ordered in the usual fashion\footnote{The usual ordering of the roots of unit is anticlockwise, from $\zeta_0=1$.
The symbol $\poly$ indicates that the tuple $(z_0, \ldots, z_{d-1})$, taken as the closed chain $(z_0, \ldots, z_{d-1}, z_0)$, forms a solid polygon.}.
Example: for $d=4$, there are four roots of unit: $(\zeta_0, \zeta_1, \zeta_2, \zeta_3) = (1, i, -1, -i)$. 
Thus, $X_a = (x_{a0}, x_{a1}, x_{a2}, x_{a3})$ is mapped to $Z_a = \poly(x_{a0}, ix_{a1}, -x_{a2}, -ix_{a3})$.
Recall that $x_{ik}\!\geq\!0$ by design and $x_{ik}\!>\!0$ by circumstance because the whoqol dataset has only strictly positive scores (see Table \ref{tab:whoqol-stats}).
These constraints ensure that $Z_i$ is not a degenerate polygon.
\future{Since there is no standard name for this family of polygons}{Let's create math for these!}, and most polygons handled by Polygrid are of this family, we will refer to them as polygons and reserve the term \aspas{polygonal shape} to describe a closed polygonal chain (and the area it delimits) in general.
Note that the angle between two consecutive roots $\zeta_k$ and $\zeta_{k+1}$ is constant and equal to $2\pi/d$.
This fact is explored in Algorithm \ref{alg:learning:ml:step1}, which details this step.
The two assessment charts in Figure \ref{fig:proposal:ml-task:default} correspond to the assessments described in Table \ref{tab:whoqol-sample} for individuals $i=89$ (labelled as \aspas{Poor QOL}) and $i=30$ (as \aspas{Good QOL}). 
Thus, the polygons in the assessment charts correspond to $Z_{89}$ and $Z_{30}$.

\begin{algorithm}
\caption{Map assessments to polygons (general)}
\label{alg:learning:ml:step1}
\KwData{$(m,d)$-array $X$}
\KwResult{$(m,d)$-array $Z$, $d$-array $\zeta$}
\Fn{uh-to-ud$(X)$}{
    $\theta \gets 2\pi/d$ \;
    \For{$k \in 0 \ldots d-1$}{
            $\zeta_k \gets \cos(k\theta) + i\sin(k\theta)$ \Comment*[r]{finds d-th roots of unity}
        }
    \For{$i \in 0 \ldots m-1$}{
            $Z_i \gets \poly(x_{i0} \, \zeta_0, \ldots, x_{i,d-1} \, \zeta_{d-1})$ \Comment*[r]{polygon on the complex plane}
        }
        return $(Z, \zeta)$ \;
}
\end{algorithm}

\paragraph{Step 2}
This step consists in computing a class prototype for each label in the assignments.
The class prototype of the $j$-th label is defined as the mean scores of all assessments that are associated with the label $j$, as detailed in Algorithm \ref{alg:learning:ml:step2}.
Class prototypes appear as polygons (with no interior) in the assignment charts in Figure \ref{fig:proposal:ml-task:default}.
Note that the chart corresponding to the label \aspas{Poor QOL} ($j \!=\! 1$) is shown at the left of the one for label \aspas{Good QOL} ($j\!=\!0$).
This inversion occurs because the charts were ordered by the size (area) of the  class prototypes, from smallest to largest.
This choice of layout is arbitrary, but useful if the labels have an ordinal structure.

\begin{algorithm}
\caption{Compute the class prototypes (multilabel)}
\label{alg:learning:ml:step2}
\KwData{$(m,d)$-array $X$, $(m,n)$-array $Y$, $d$-array $\zeta$}
\KwResult{$(n,d)$-array $H$}
\Fn{compute-class-prototypes$(X,Y,\zeta)$}{
    \For{$j \in 0 \ldots n-1$}{
            $\mathcal{I} \gets \{i \mid y_{ij} = 1\}$ \Comment*[r]{collects the rows assigned to label $j$}
            $A_j \gets \frac{1}{\mid \mathcal{I} \mid} \sum_{i \in \mathcal{I}} X_{i}$ \;
            $H_j \gets \poly(a_{j0} \, \zeta_0, \ldots, a_{j,d-1} \, \zeta_{d-1})$ \;
        }
        return $H$ \;
}
\end{algorithm}

\begin{figure}[htpb]
    \centering
    \begin{subfigure}{.48\textwidth}
        \centering
        \includegraphics[width=.87\linewidth]{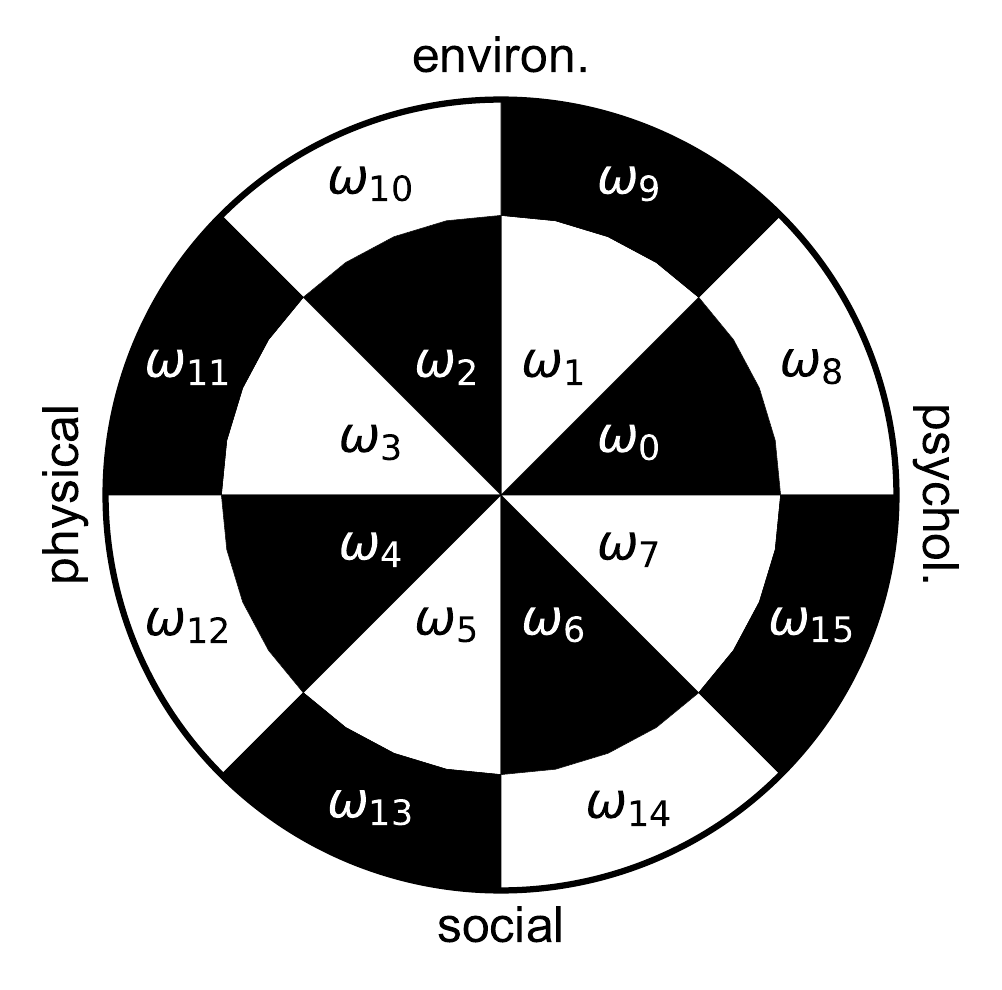}
        \caption{Partition using Algorithm \ref{alg:learning:ml:step3}}
        \Description{}
        \label{fig:proposal:disc1}
    \end{subfigure}
    \hfill
    \begin{subfigure}{.48\textwidth}
        \centering
        \includegraphics[width=.87\linewidth]{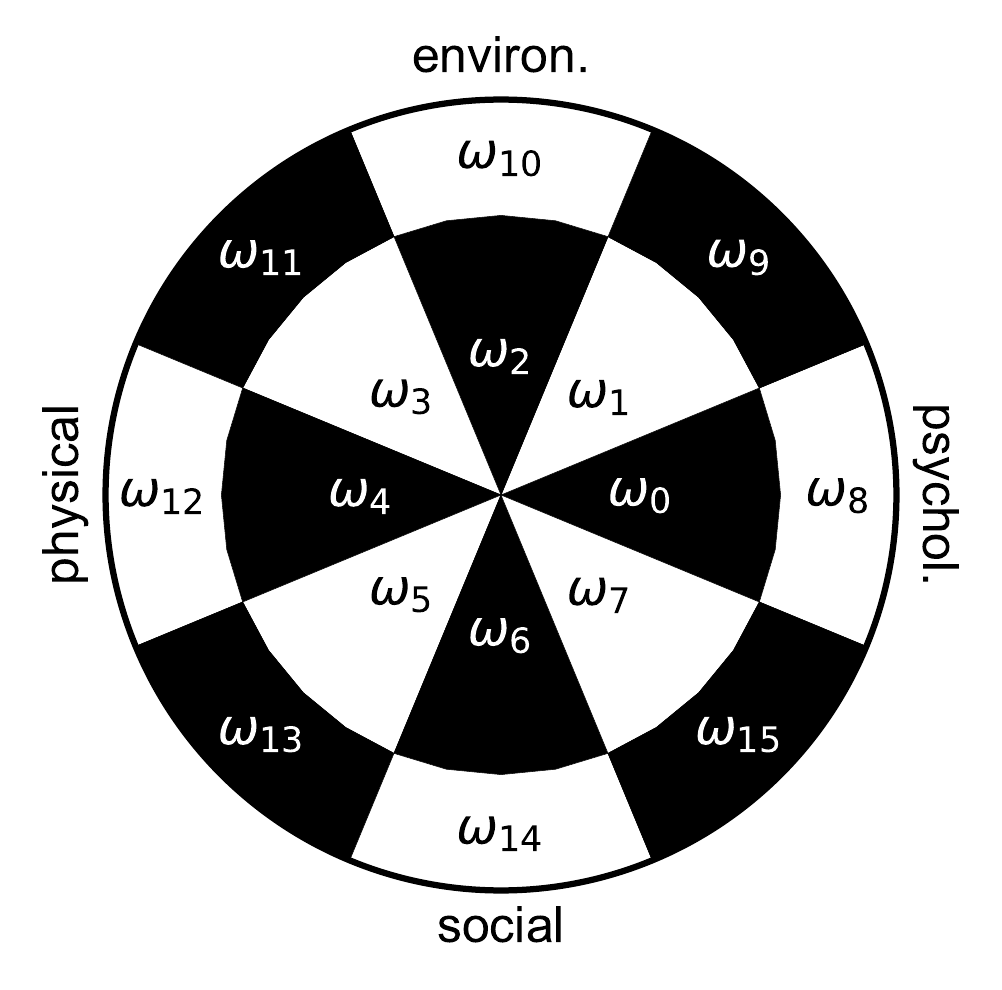}
        \caption{Partition with alternative configuration}
        \Description{}
        \label{fig:proposal:disc2}
    \end{subfigure}
    \label{fig:proposal:discs}
    \caption{The unit disc partitioned into the annular sectors defined by $n_a=2$ annuli and $n_s/d=2$ sectors per domain.
    The instrument has $d=4$ domains, so the partition has $n_s\!=\!8$ sectors.
    (\subref{fig:proposal:disc1}) The widths of the annuli were chosen so that all cells have the same area.
    (\subref{fig:proposal:disc2}) 
    The radius dividing the annuli is $r=0.5$, so cells have different areas.
    In addition, sectors are placed so that each domain axis bisects some sector.
    }
\end{figure}

\paragraph{Step 3}
The third step defines the feature space.
It does so by partitioning the unit disc $\mathbb{\overline{D}}$ into disjoint cells that result from the intersection of disjoint annuli and disjoint sectors of the disc.
The number of annuli, $n_a$, and the number of sectors per domain of the instrument, $n_s/d$, are hyperparameters of the model.
Thus, the disc is partitioned into $n_{as}=n_a \times n_s$ annular sectors.
These cells are enumerated in anticlockwise order, from the origin to the boundary of the disc.
For example, Figure \ref{fig:proposal:disc1} shows a partition of the disc with two annuli and eight sectors.
Their pairwise intersection defines $n_{as} = 16$ cells, which are enumerated in the partition $\Omega = (\omega_0, \ldots, \omega_{15})$.
This operation is detailed in Algorithm \ref{alg:learning:ml:step3}.
In the pseudocode, the annuli are specified in line 3.
Each annulus is described as the difference between two sets: $\{ z \in \overline{\mathbb{D}}: \, \mid\! z \!\mid \, \leq \sqrt{(p\!+\!1) / n_a} \}$, which describes the set of points contained in the closed disc of radius $\sqrt{(p\!+\!1) / n_a}$, excluding the points in $\{ z \in \overline{\mathbb{D}}: \, \mid\! z \!\mid \, \leq \sqrt{p / n_a} \}$, the closed disc of radius $\sqrt{p / n_a}$, both centred at the origin.
We follow a similar approach to specify the sectors.
In line 7, a sector is specified as the set $\{\rho e^{i\theta} \!:\! \rho \in (0,1], \, \theta \in [\, q \, \Delta\theta, (q\!+\!1) \, \Delta\theta \,)\}$, which describes all points in the unit disc whose argument $\theta \in [\, q \, \Delta\theta, (q\!+\!1) \, \Delta\theta \,)$.
Finally, in line 12, the annular sector $\omega_r$ is defined as the intersection between an annulus and a sector.
This partition defines cells that have the same area and sectors that \aspas{miss} the domain axes.
The alternative partition shown in Figure \ref{fig:proposal:disc2} has the same number of cells, but its sectors are placed differently, and the cells do not have the same area.
These are arbitrary but consequential decisions, and alternatives will be explored in Section \ref{section:proposal:ml-task:options}.

Although the pseudocode uses the language of sets to specify regions of the unit disc, in practice, we take advantage of open-source libraries for computational geometry.
These libraries have geometric primitive operations that allow us to use polygonal shapes to approximate annuli and sectors.
This means that, during runtime, each cell $\omega_r$ stores the description of an annular sector as a polygonal shape.
Of course, the goodness of this approximation can be empirically calibrated.
For example, let $\mu(\cdot)$ be a function that \aspas{measures} the area of an arbitrary polygonal shape \cite{younhee2017shoelace}.
Then, a calibration criterion can be $\pi - \sum_r \mu(\omega_r) < \delta$ or $\sum_i  \, \mu(Z_i) - \sum_r \mu(Z_i \cap \omega_r) < \delta$ for a small $\delta > 0$.

\begin{algorithm}
\caption{Partition the unit disc into annular sectors (general)}
\label{alg:learning:ml:step3}
\KwData{integers $n_a, n_s$}
\KwResult{$n_{as}$-array $\Omega$}
\Fn{partition-ud($n_a, n_s$)}{
    \For{$p \in 0 \ldots (n_a-1)$}{
            $annulus_p \gets \{ \, z: \, \mid\! z \!\mid \, \leq \, \sqrt{(p+1) / n_a} \, \} \, \backslash \,  \{ \, z: \, \mid\! z \!\mid \, \leq \, \sqrt{p / n_a} \, \}, \, z \in \mathbb{\overline{D}}$ \;
        }
    $\Delta\theta \gets 2\pi/n_s$ \;
    \For{$q \in 0 \ldots (n_s-1)$}{
            $sector_q \gets \{\, \rho e^{i\theta}: \rho \in (0,1], \, \theta \in [\, q \, \Delta\theta, (q\!+\!1) \, \Delta\theta \,) \, \} $ \;
        }
        
    \For{$p \in 0 \ldots (n_a-1)$}{
        \For{$q \in 0 \ldots (n_s-1)$}{
            $r \gets (p \times n_s) + q$ \Comment*[r]{anticlockwise, origin to boundary}
            $\omega_{r} \gets annulus_p \cap sector_q$ \;
        }    
    }
    $n_{as} \gets n_a \times n_s$ \;
    $\Omega \gets (\omega_0, \, \ldots, \, \omega_{n_{as} - 1})$ \Comment*[r]{enumeration of cells of the unit disc}
    return $\Omega$ \;
}
\end{algorithm}

In Figure \ref{fig:proposal:ml-task:default}, the partition $\Omega$ is not represented directly, but the background of the assignment charts reveals their structure: boundaries of annular sectors are wherever colour transitions occur.
Although the background colours are not determined by the partitioning, there is a mapping from $\Omega$ into the colour palette that represents the weights given to the cells, as we shall see in step 5.

\paragraph{Step 4}
In this step, we extract feature vectors that describe how the area of each polygon $Z_i$ is distributed over the partitioning of the disc $\Omega$.
This operation is detailed in Algorithm \ref{alg:learning:ml:step4}.
The function $\mu(\cdot)$ that appears in line 4 is the same discussed in the previous step: it computes the area of an arbitrary polygonal shape.
Thus, the loop in lines 3-5 describes a procedure in which the intersection between some polygon $Z_i$ and each cell $\omega_r$ is obtained.
The numeric value of the area of each polygonal shape, $\mu(Z_i \cap \omega_r)$, is computed and stored in 
$S_{i} \coloneq \big( \, \mu(Z_i \cap \omega_0), \, \ldots, \, \mu(Z_i \cap \omega_{n_{as}-1}) \, \big)$.
In Figure \ref{fig:proposal:ml-task:default}, the matching charts show the decomposition of an assessment polygon $Z_i$ into its intersections with the partition cells.
For instance, there are $n=2$ matching charts for $Z_{89}$: one for \aspas{Poor QOL}, one for \aspas{Good QOL}.
In both charts, the assessment polygon $Z_{89}$ is decomposed into four polygonal shapes $(Z_{89} \cap \omega_r)$, each painted with the same colour in which the annular sector $\omega_r$ is drawn in the assignment chart associated with the same label.

\begin{algorithm}
\caption{Map polygons to feature vectors (general)}
\label{alg:learning:ml:step4}
\KwData{$(m,d)$-array $Z$, $n_{as}$-array $\Omega$}
\KwResult{$(m,n_{as})$-array $S$}
\Fn{ud-to-fs($Z, \Omega$)}{
    \For{$i \in 0 \ldots (m-1)$}{
        \For{$r \in 0 \ldots (n_{as}-1)$}{
            $s_{ir} \gets \mu(Z_i \cap \omega_r)$ 
            \Comment*[r]{area of polygon $Z_i$ covering cell $\omega_r$}
        }    
    }
    return $S$ \;
}
\end{algorithm}

\paragraph{Step 5}
The last step of the learning pipeline determines the weights for each cell/label so that we can reconstruct the assignments in $Y$ based on the assessments in $X$.
This is done by solving the system of linear equations $S\transpose{W}\!=\!Y$, with $W$ being an $(n, n_{as})$-matrix.
This operation is detailed in Algorithm \ref{alg:learning:ml:step5}.
The weights that reconstruct the $j$-th label of an assignment are computed by the loop in lines 2-4.
Each row of the weight matrix $W$ is a solution to the problem of minimising the difference between $Y_{:j}$ and its reconstruction $\hat{Y}_{:j}=S\transpose{W_j}$.
In lines 5-6, the learned weights are used to obtain the reconstruction $\hat{Y}$, and its extreme values are used to determine the range of candidate values for a classification threshold.
Note that $y_{ij} \in \{0, 1\}$, but this constraint does not apply to $\hat{y}_{ij}$.
To make them comparable, a decision rule with the Iverson bracket is used: $\hat{y}_{ij} \gets \ivbracket{\hat{y}_{ij} \geq \mbox{\textit{threshold}}_j}$.
In line 7, the classification thresholds are determined as a solution to the problem of maximising the similarity between $Y$ and $\hat{Y}$.
This approach consists of breaking up a multilabel classification task into multiple binary classification tasks and using a regression model to solve the latter, which are the two strategies discussed in Section \ref{section:relatedwork:tasks:multilabel}.
In the diagram in Figure \ref{fig:proposal:ml-task:default}, the learned weights are represented as the background colours that appear in the assignment charts.
For example, the weights given to the cells of the assignment chart for \aspas{Good QOL} ($j\!=\!0$) comes from $W_0$, and the ones for \aspas{Poor QOL} ($j\!=\!1$) from $W_1$.
Each assignment \aspas{disc} has four cells, so the model has eight weights (plus one threshold per label in multilabel assignments).
Finally, the colour bar on the right of the diagram represents the map that decodes colours into weights.

\begin{algorithm}
\caption{Learn model weights (multilabel)}
\label{alg:learning:ml:step5}
\KwData{$(m,n_{as})$-array $S$, $(m,n)$-array $Y$, int \textit{granularity}, fn \textit{difference}, fn \textit{similarity}}
\KwResult{$(n,n_{as})$-array $W$, $n$-array $\mbox{\textit{thresholds}}$}
\Fn{learn-weights($S, Y, \mbox{granularity, difference, similarity}$)}{
    
    \For{$j \in 0 \ldots (n-1)$}{
        $W_j \gets \argmin\limits_{W_j} \mbox{\textit{difference}}(Y_{:j}, \, S, \, W_j)$ \Comment*[r]{solves $S\transpose{W_j}=Y_{:j}$ for $W_j$}
    } 

    $\hat{Y} \gets S \, \transpose{W}$ \;

    $T \gets \mbox{\textit{candidates}}(\hat{Y}, \mbox{\textit{granularity}})$ \;
 
    $\mbox{\textit{thresholds}} \gets \argmax\limits_{t \in T} \mbox{\textit{similarity}}(Y, \hat{Y}, t)$ \Comment*[r]{selects best thresholds}
    
    return $(W, \mbox{\textit{thresholds}})$ \;
}
\end{algorithm}

\noindent

\subsubsection{Making Predictions and Generating Explanations}
\label{section:proposal:ml-task:prediction}

The first two steps of the prediction pipeline correspond to the steps 1 and 4 of the learning pipeline, as seen in Figure \ref{fig:proposal:pipelines}.
Given an unseen assessment $x$, its scores are mapped to a polygon $z$, and then $z$ is mapped to a vector $s$ in the feature space.
The last step, which is detailed in Algorithm \ref{alg:ml:prediction}, combines the learned weights and thresholds with the vector $s$ to obtain predictions for all labels.
In line 4, a numeric prediction is obtained for each label and stored in $\hat{y}$.
The categorical prediction $labels$ is derived from $\hat{y}$ in lines 5-7.
In the diagram in Figure \ref{fig:proposal:ml-task:default}, the numeric values $\hat{y}_j$ are shown in the tags of the matching charts.
A green tag indicates that the label should be assigned to the assessment ($labels_j\!=\!1$), and a light-yellow tag indicates otherwise ($labels_j\!=\!0$). 
Thus, the model recommends that Participant 089 be associated with the \aspas{Poor QOL} label, and that Participant 030 be associated with the \aspas{Good QOL} label.

\begin{algorithm}
\caption{Make predictions (multilabel)}
\label{alg:ml:prediction}
\KwData{$(1,d)$-array $x$, $n_{as}$-array $\Omega$, $(n,n_{as})$-array $W$, $n$-array $\mbox{\textit{thresholds}}$}
\KwResult{$(1,n)$-array $\hat{y}, (1,n)$-array \textit{labels}}
\Fn{predict($x, \Omega, W, \mbox{\textit{thresholds}}$)}{

    $(z,\cdot) \gets \mbox{\textit{uh-to-ud}}(x)$ \Comment*[r]{from step 1}
    $s \gets \mbox{\textit{ud-to-fs}}(z, \Omega)$ \Comment*[r]{from step 4}
    $\hat{y} \gets s \transpose{W}$ \Comment*[r]{line 5 from step 5}
    \For{$j \in 0 \ldots (n-1)$}{
        $\mbox{\textit{labels}}_j \gets 1 \mbox{ if } \hat{y_j} \geq \mbox{\textit{thresholds}}_j \mbox{ else } 0$ \;        
    }
    return $(\hat{y}, \mbox{\textit{labels}})$ \;
}
\end{algorithm}

It must be clear by now that all data structures required to render an explanation diagram are unaltered byproducts of the model's pipelines, and all salient data structures handled are depicted in the explanation diagram.
Assessment charts display data from $Z$ (an output of step 1), assignment charts show data from $H$ and $W$ (learning, steps 2 and 5) according to a layout specified by $\Omega$ (learning, step 3), and the matching charts combine data from $Z$ and $W$ (steps 1 and 5) and the layout from $\Omega$.
The content and colour of the matching tags are determined by $\hat{y}$ and $labels$ respectively (from the prediction stage).
\standing{This one-to-one correspondence supports our argument about the high level of faithfulness of explanation diagrams generated by Polygrid.}{faithfulness}
However, the argument that addresses the interpretability of these explanations will have to wait until Section \ref{section:proposal:interpretable}.



\subsubsection{Exploring Alternative Values of the Main Parameters}
\label{section:proposal:ml-task:options}

\begin{figure}[htpb]
    \centering
    \begin{subfigure}{.31\textwidth}
        \centering
        \includegraphics[width=.85\linewidth]{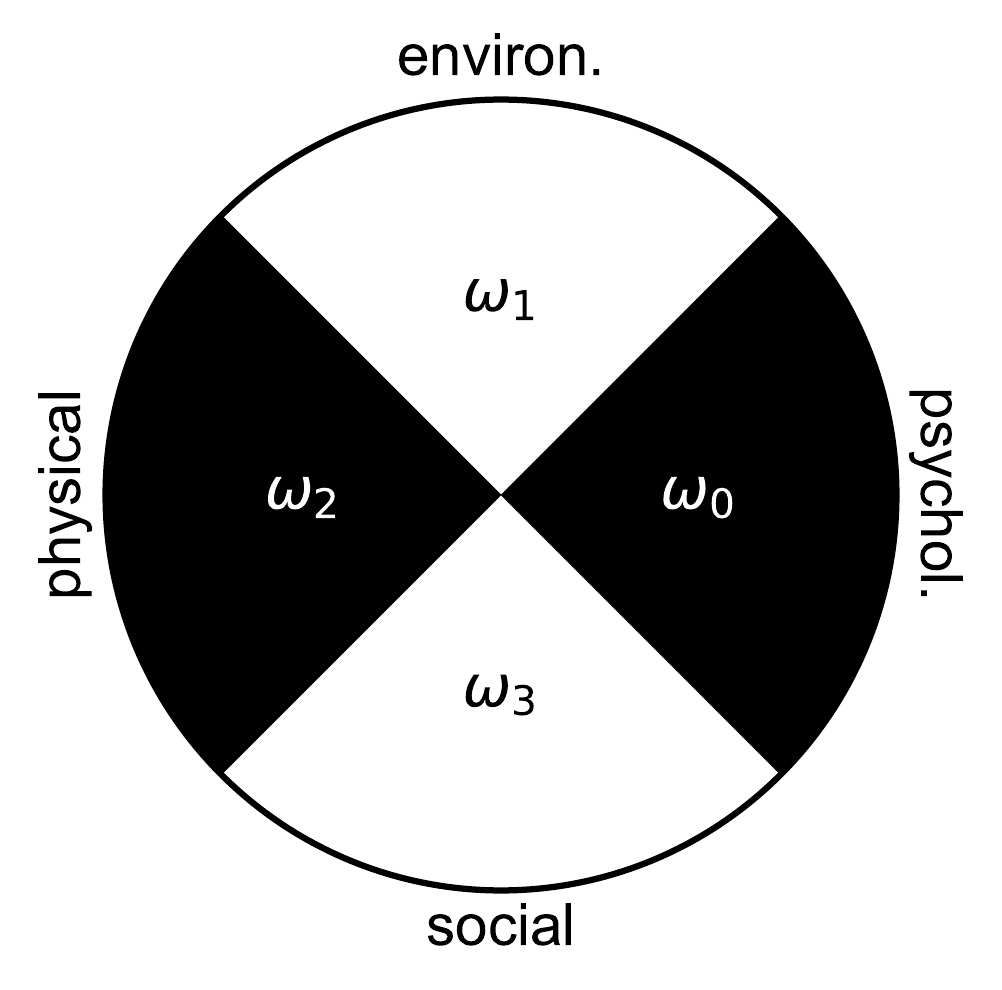}
        \caption{sector = cover}
        \Description{}
        \label{fig:proposal:ml-task:sector.1}
    \end{subfigure}
    \hfill
    \begin{subfigure}{.31\textwidth}
        \centering
        \includegraphics[width=.85\linewidth]{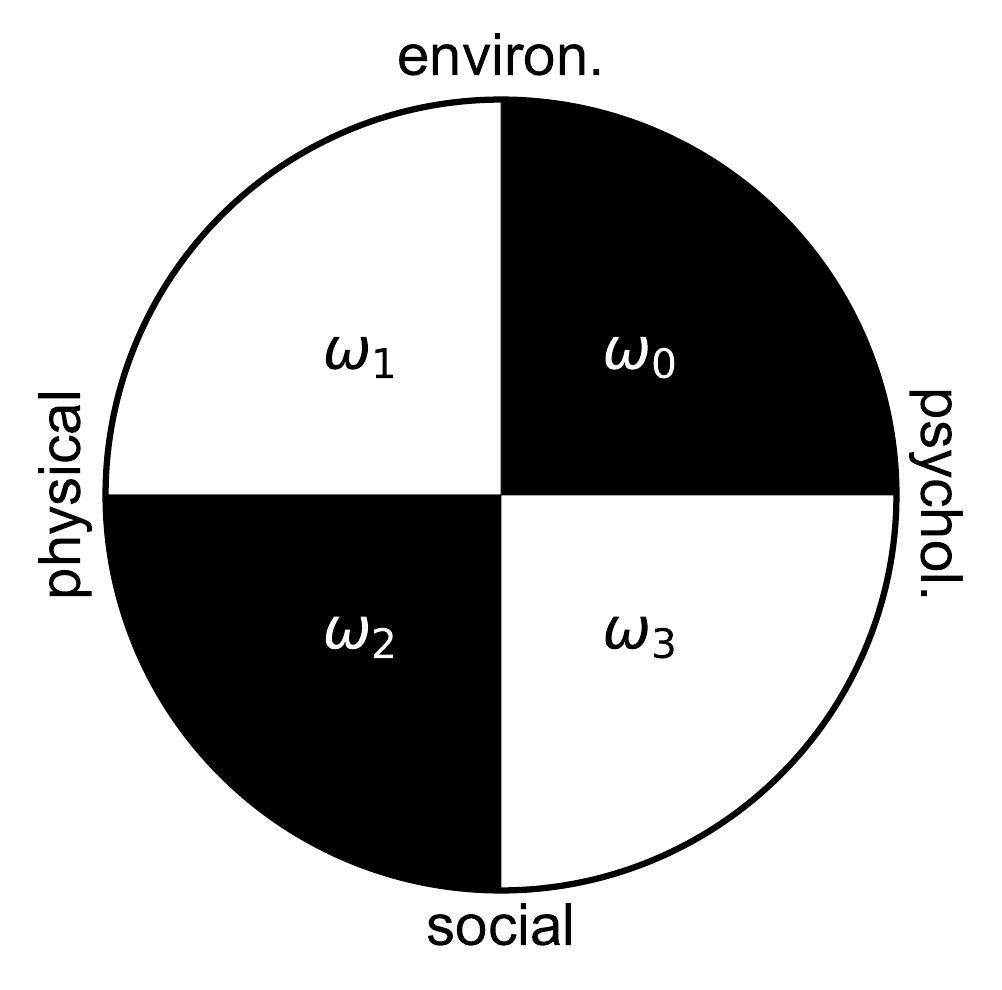}
        \caption{default}
        \Description{}
        \label{fig:proposal:ml-task:disc-default}
    \end{subfigure}
    \hfill
    \begin{subfigure}{.31\textwidth}
        \centering
        \includegraphics[width=.85\linewidth]{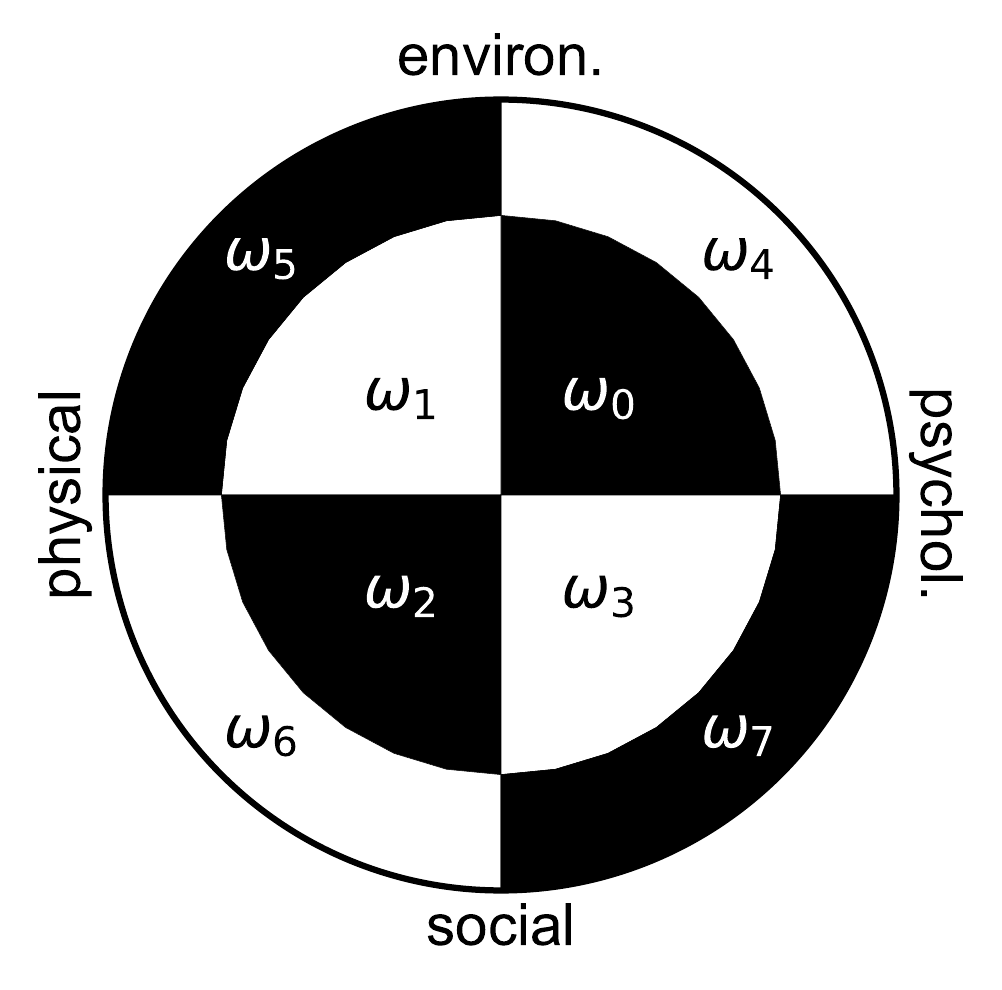}
        \caption{na=2}
        \Description{}
        \label{fig:proposal:ml-task:annulus.1}
    \end{subfigure}
    \vfill
    \begin{subfigure}{.31\textwidth}
        \centering
        \includegraphics[width=.85\linewidth]{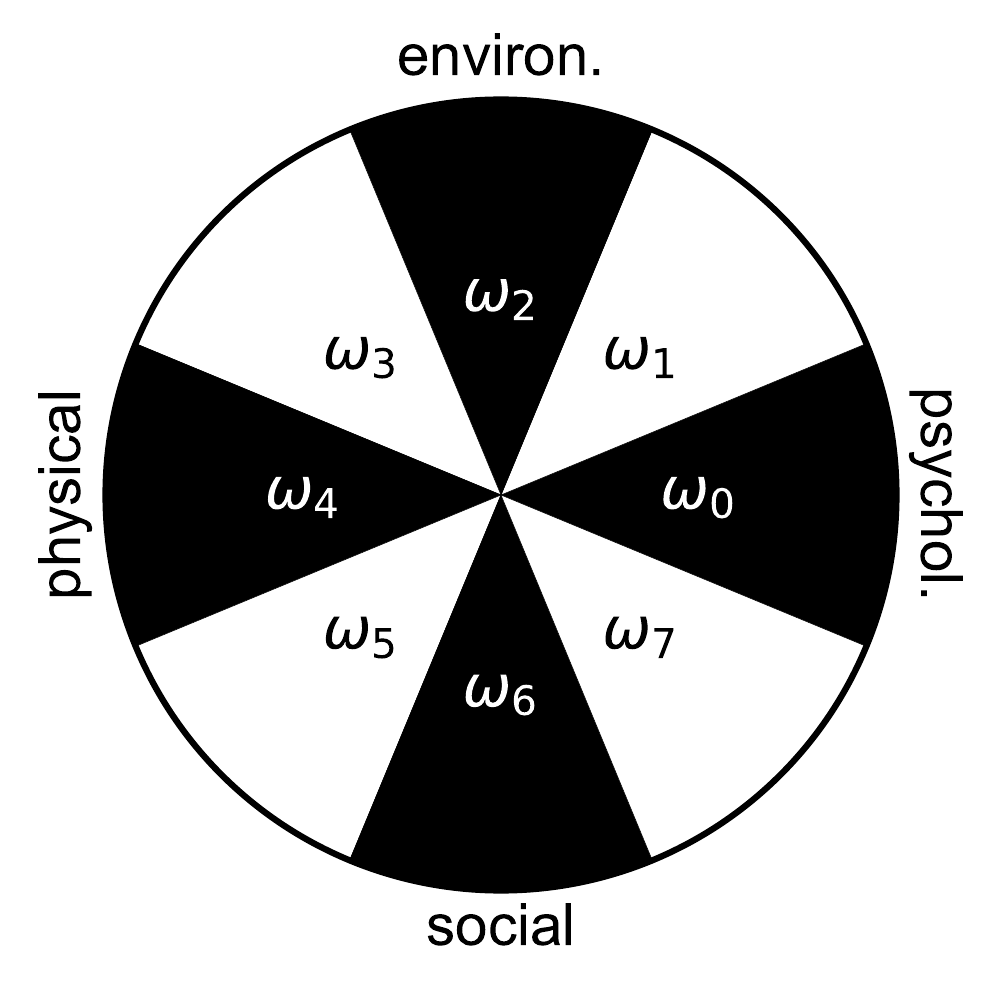}
        \caption{sector = cover, ns/d=2}
        \Description{}
        \label{fig:proposal:ml-task:sector.2}
    \end{subfigure}
    \hfill
    \begin{subfigure}{.31\textwidth}
        \centering
        \includegraphics[width=.85\linewidth]{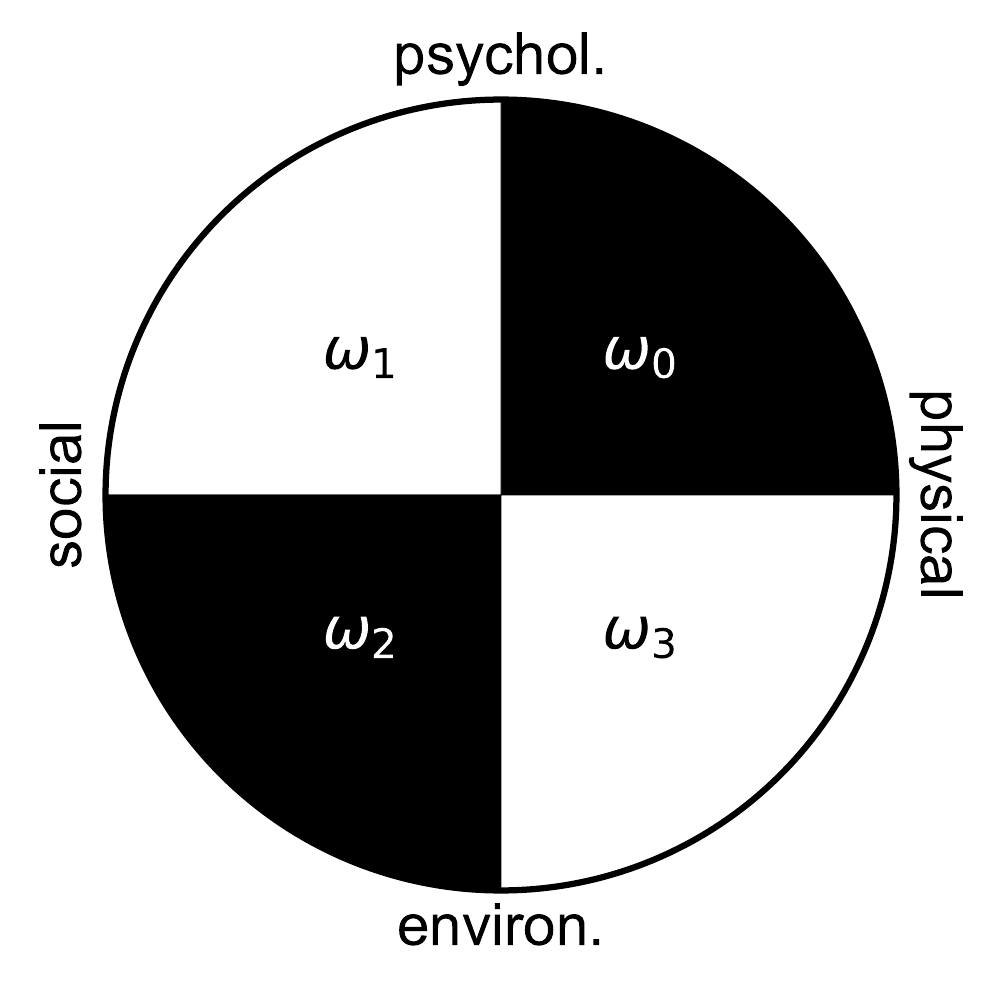}
        \caption{vorder = original}
        \Description{}
        \label{fig:proposal:ml-task:vorder}
    \end{subfigure}
    \hfill
    \begin{subfigure}{.31\textwidth}
        \centering
        \includegraphics[width=.85\linewidth]{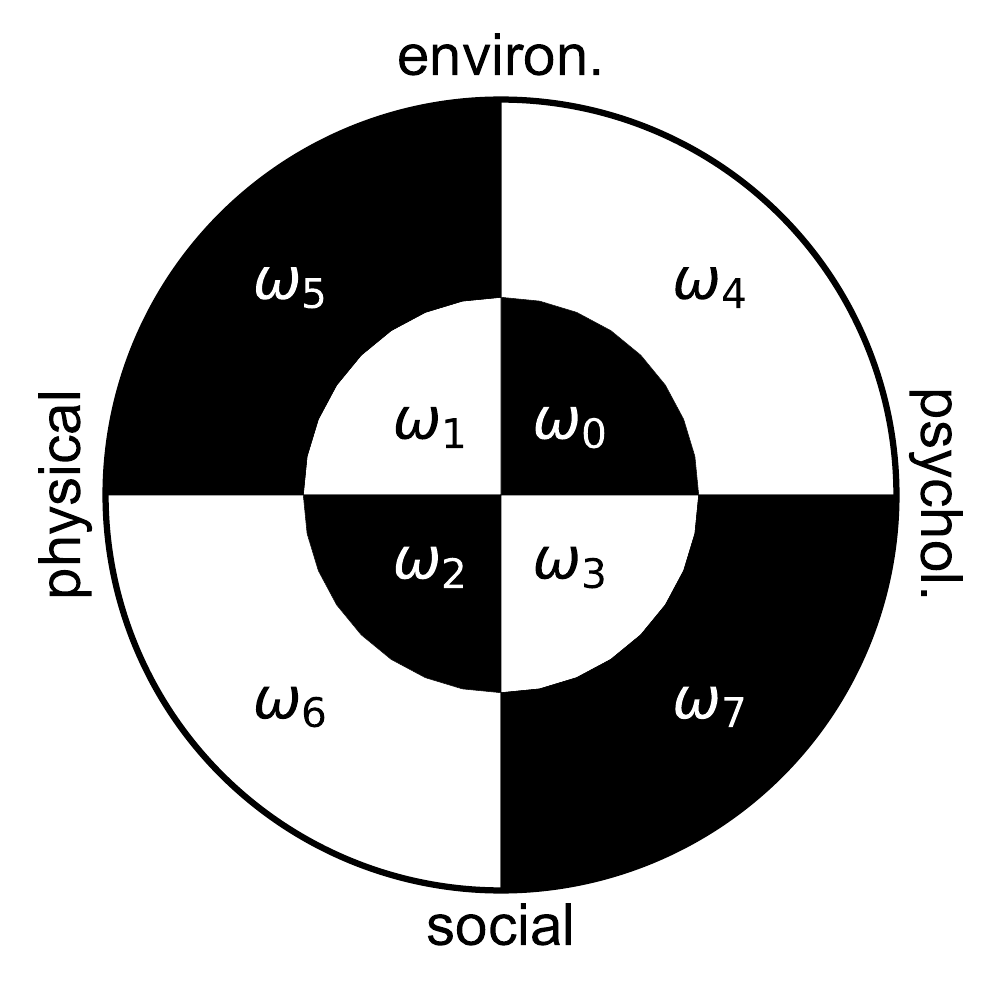}
        \caption{na=2, annulus=r-invariant}
        \Description{}
        \label{fig:proposal:ml-task:annulus.2}
    \end{subfigure}   
    \label{fig:proposal:disc-configs}
    \caption{Partitioning of the disc induced by the configs in Table \ref{tab:performance:ml}.
    Config \ref{fig:proposal:ml-task:disc-default} corresponds to the default config, which was used to produce the explanation diagram in Figure \ref{fig:proposal:ml-task:default}.
    Configs \ref{fig:proposal:ml-task:sector.1}, \ref{fig:proposal:ml-task:annulus.1} and \ref{fig:proposal:ml-task:vorder} differ from the default config by a single parameter, while configs \ref{fig:proposal:ml-task:sector.2} and \ref{fig:proposal:ml-task:annulus.2} differ from \ref{fig:proposal:ml-task:sector.1} and \ref{fig:proposal:ml-task:annulus.1} by a single parameter, respectively.
    }
\end{figure}

The previous section has shown how Polygrid operates based on the default values for its hyperparameters (aka the default config).
However, changing the default values can lead to an increase in performance.
For example, the order in which the instrument's domains are mapped onto the vertices of the polygons is an arbitrary but consequential choice.
In Table \ref{tab:performance:ml}, the difference in performance between configs \ref{fig:proposal:ml-task:disc-default} and \ref{fig:proposal:ml-task:vorder} illustrates this point.
The first corresponds to the default config, and the latter is identical to the first, except that for the mapping of domains onto vertices, which is determined by the parameter \aspas{vorder}.

Similarly, the number and placement of sectors on the disc are also arbitrary choices, which are determined by the parameters \aspas{ns/d} and \aspas{sector}, respectively.
In Table \ref{tab:performance:ml}, the difference in performance between configs \ref{fig:proposal:ml-task:disc-default} (\emph{default}) and \ref{fig:proposal:ml-task:sector.1} illustrates this point.
The latter is identical to the first, except that for the placement of the sectors (\emph{sector=cover}).
Moreover, config \ref{fig:proposal:ml-task:sector.2} is identical to config \ref{fig:proposal:ml-task:sector.1}, except for the doubled number of sectors per domain ($ns/d=2$).
Note that these changes translated into a small increase in performance (e.g., about 8\% increase in accuracy from \ref{fig:proposal:ml-task:disc-default} to \ref{fig:proposal:ml-task:sector.2}).

\begin{table}[htb]
    \caption{\centering Performance under different model parameters on the whoqol dataset}
    \label{tab:performance:ml}
    \begin{tabular}{cccccccccc}
        \toprule
               & \multicolumn{6}{c}{Model Parameters} & \multicolumn{3}{c}{ Performance}\\
               \cmidrule(lr){2-7} \cmidrule(lr){8-10} 
        Config &  ns/d &  na &   vorder &     sector & annulus & solver &  accuracy &  f1.ma &  hamml  \\
        \toprule

         Figure \ref{fig:proposal:ml-task:disc-default} &     1 &   1 &      rho &   miss & s-invariant &  lstsq &     0.592 &     0.580 &     0.409 \\
   Figure \ref{fig:proposal:ml-task:vorder} &     1 &   1 & original &   miss & s-invariant &  lstsq &     0.604 &     0.593 &     0.396 \\
 Figure \ref{fig:proposal:ml-task:sector.1} &     1 &   1 &      rho &  cover & s-invariant &  lstsq &     0.637 &     0.621 &     0.363 \\
 Figure \ref{fig:proposal:ml-task:sector.2} &     2 &   1 &      rho &  cover & s-invariant &  lstsq &     0.642 &     0.629 &     0.357 \\
Figure \ref{fig:proposal:ml-task:annulus.1} &     1 &   2 &      rho &   miss & s-invariant &  lstsq &     0.880 &     0.861 &     0.120 \\
Figure \ref{fig:proposal:ml-task:annulus.2} &     1 &   2 &      rho &   miss & r-invariant &  lstsq &     0.983 &     0.982 &     0.017 \\
   Figure \ref{fig:proposal:ml-task:solver.ridge} &     1 &   1 &      rho &   miss & s-invariant &  ridge &     0.978 &     0.976 &     0.022 \\
     Figure \ref{fig:proposal:ml-task:best} &     1 &   1 &      rho &  cover & s-invariant &  ridge &     0.983 &     0.982 &     0.017 \\
    \bottomrule
    \end{tabular}\\
    \noindent
    \footnotesize \justifying \emph{Legend}: Column \aspas{Config} indicates the figure displaying the partitioning of the disc implied by a model operating with the parameters listed under the Model Parameters heading.
    \aspas{$ns/d$} stands for the number of sectors per domain, \aspas{$na$} stands for the number of annuli, and \aspas{vorder} indicates the heuristic used to select the ordering of the domains.
    Column \aspas{annulus} indicates the annulus type: \aspas{s-invariant} indicates annular sectors with the same area, and \aspas{r-invariant} for annular sectors with the same width. 
    Column \aspas{sector} indicates the sector type: \aspas{miss} means that the initial sector starts at zero radians, and \aspas{cover} means that the first sector is bisected by the first domain axis.
    Column \aspas{solver} indicates the method used to solve systems of linear equations: \aspas{lstsq} means least squares, and \aspas{ridge} indicates ridge regression.
    Under the Performance heading, \aspas{f1.ma} shows the macro-averaged F1 score, and \aspas{hamml} the Hamming loss.
    The performance measurements correspond to the average value obtained from 100 training/test cycles, with identical initial conditions except for differences in parameter values.
    The parameter values for \aspas{Figure \ref{fig:proposal:ml-task:best}} attain the highest performance with less weights.
\end{table}

The number and placement of annuli on the disc are also arbitrary choices, controlled by the parameters \aspas{na} and \aspas{annulus}, respectively.
In Table \ref{tab:performance:ml}, the difference in performance between configs \ref{fig:proposal:ml-task:disc-default} (\emph{default}) and \ref{fig:proposal:ml-task:annulus.1} ($na=2$) illustrates this point.
The latter is identical to the first, except for the number of annuli.
The salient characteristic of both configs is that all cells in a disc have the same area.
This is in contrast with the partitioning in config \ref{fig:proposal:ml-task:annulus.2} (\emph{annulus=r-invariant}), in which the annuli have different areas.
These changes translated into significant improvements in performance (e.g., about 66\% increase in accuracy from \ref{fig:proposal:ml-task:disc-default} to \ref{fig:proposal:ml-task:annulus.2}).
It must be noted that the alternative placement of sectors and annuli results from modifying their specifications in Algorithm \ref{alg:learning:ml:step3}.

\begin{figure}[htpb]
    \centering
    \begin{subfigure}[b]{1\textwidth}
        \centering
        \includegraphics[width=.81\linewidth]{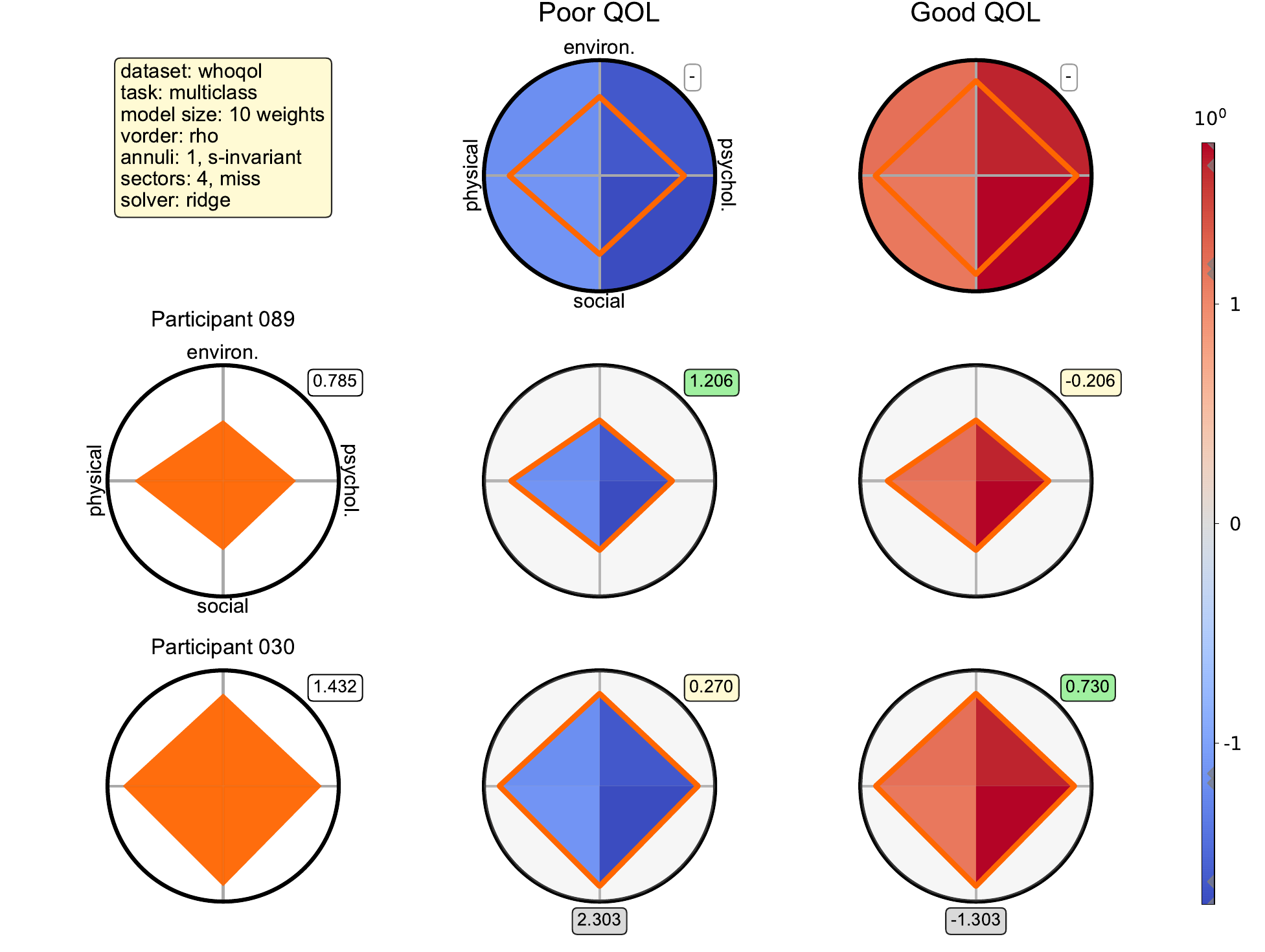}
        \caption{identical to the default config, except for \emph{solver=ridge}}.
        \Description{}
        \label{fig:proposal:ml-task:solver.ridge}
    \end{subfigure}
    \vfill
    \begin{subfigure}[b]{1\textwidth}
        \centering
        \includegraphics[width=.81\linewidth]{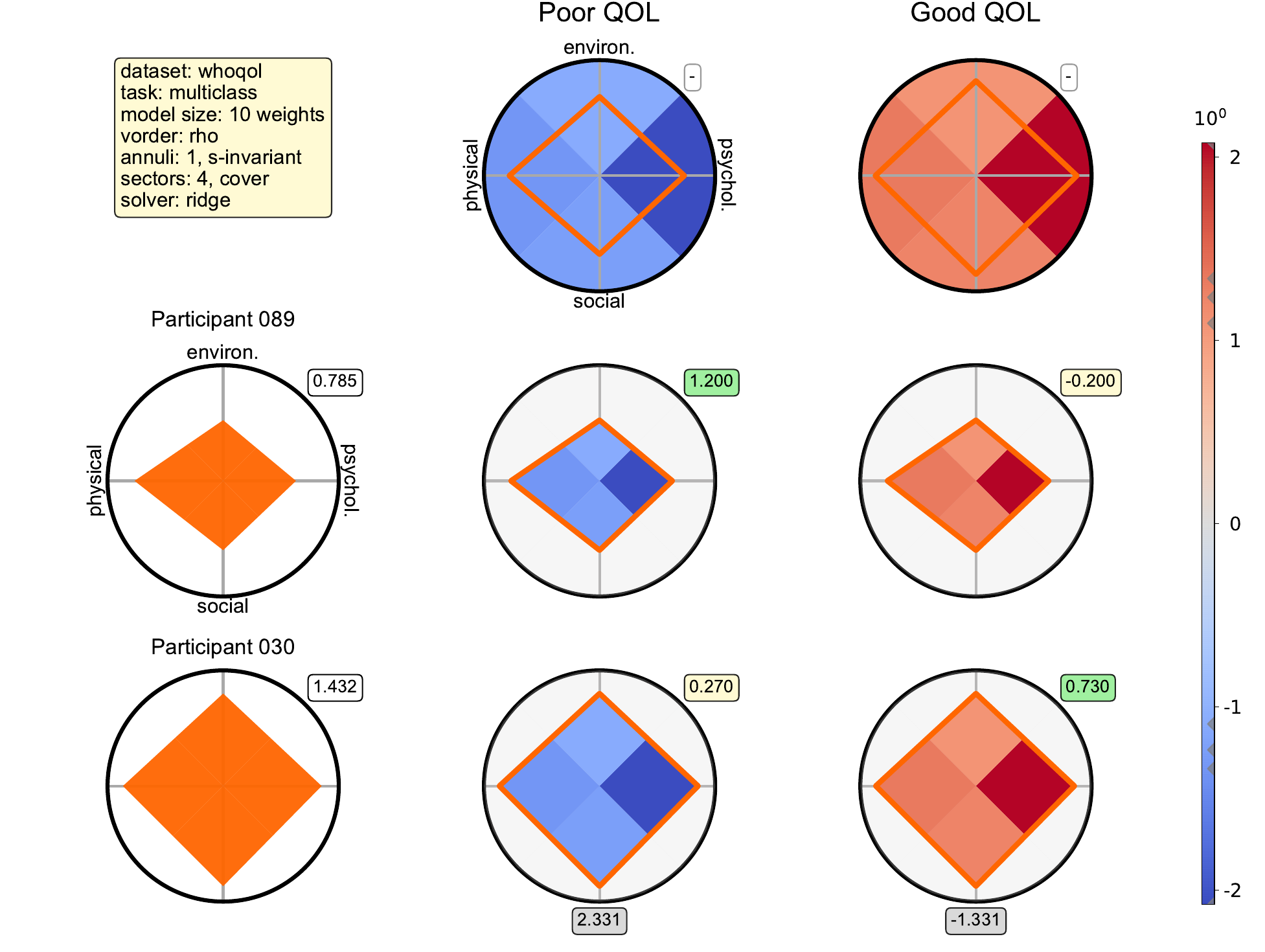}
        \caption{identical to the default config, except for \emph{solver=ridge} and \emph{sector=cover}.} 
        \Description{}
        \label{fig:proposal:ml-task:best}
    \end{subfigure}
    \caption{Examples of an explanation diagram with learned intercepts.
    They appear in the grey tags at the bottom of a column with an assignment chart.
    The value in the matching tag corresponds to the weighted area of the corresponding polygon plus the value in the intercept.
    This results in an increase in instance size.
    }
    \label{fig:proposal:ml-task:solver}
\end{figure}

Finally, we explore alternative ways of solving the system $S\transpose{W}\!=\!Y$ in Algorithm \ref{alg:learning:ml:step5}.
In the default config, this is solved by least squares approximation, which does not implement regularisation.
Besides, recall that the matrix $S$ holds the feature vectors: each element $s_{ir}$ corresponds to the area of the polygon $Z_i$ that is covered by the cell $\omega_r$.
This means that $S$ is a nonnegative matrix (not mean-centred), and it does not include an extra dimension to encode intercepts.
In contrast, intercepts and regularisation are used in config \ref{fig:proposal:ml-task:solver.ridge}: it adds an extra dimension to $S$ and uses ridge regression to solve the resulting system.
This behaviour is imposed by setting \emph{solver=ridge}.
A large difference in performance between configs \ref{fig:proposal:ml-task:disc-default} (\emph{default}) and \ref{fig:proposal:ml-task:solver.ridge} is shown in Table \ref{tab:performance:ml}.
The best performing model, which uses config \ref{fig:proposal:ml-task:best}, is identical to \ref{fig:proposal:ml-task:solver.ridge} except for the choice of sector placement.

The diagrams resulting from the configs in Table \ref{tab:performance:ml} offer quite different explanations as to why an assessment should be assigned to a label.
In light of the decision boundary that was used to create the synthetic assignments for the whoqol dataset (see Section \ref{section:proposal:data}), it makes sense that config \ref{fig:proposal:ml-task:best} achieved the best performance.
This is because (a) the decision boundary was based on the sum-score, (b) there is a monotonic relationship between sum-scores and the area-scores (i.e., the area of the assessment polygon), and (c) the intercepts learned by the model captures differences in size of the assessment polygons.
Both researchers and practitioners would be justified in preferring one model over another depending on the particular context of their application.

\subsection{The Polygrid Model in Label Ranking Tasks}
\label{section:proposal:learning-lr}

\begin{figure}[htpb]
    \centering
    \includegraphics[width=1\linewidth]{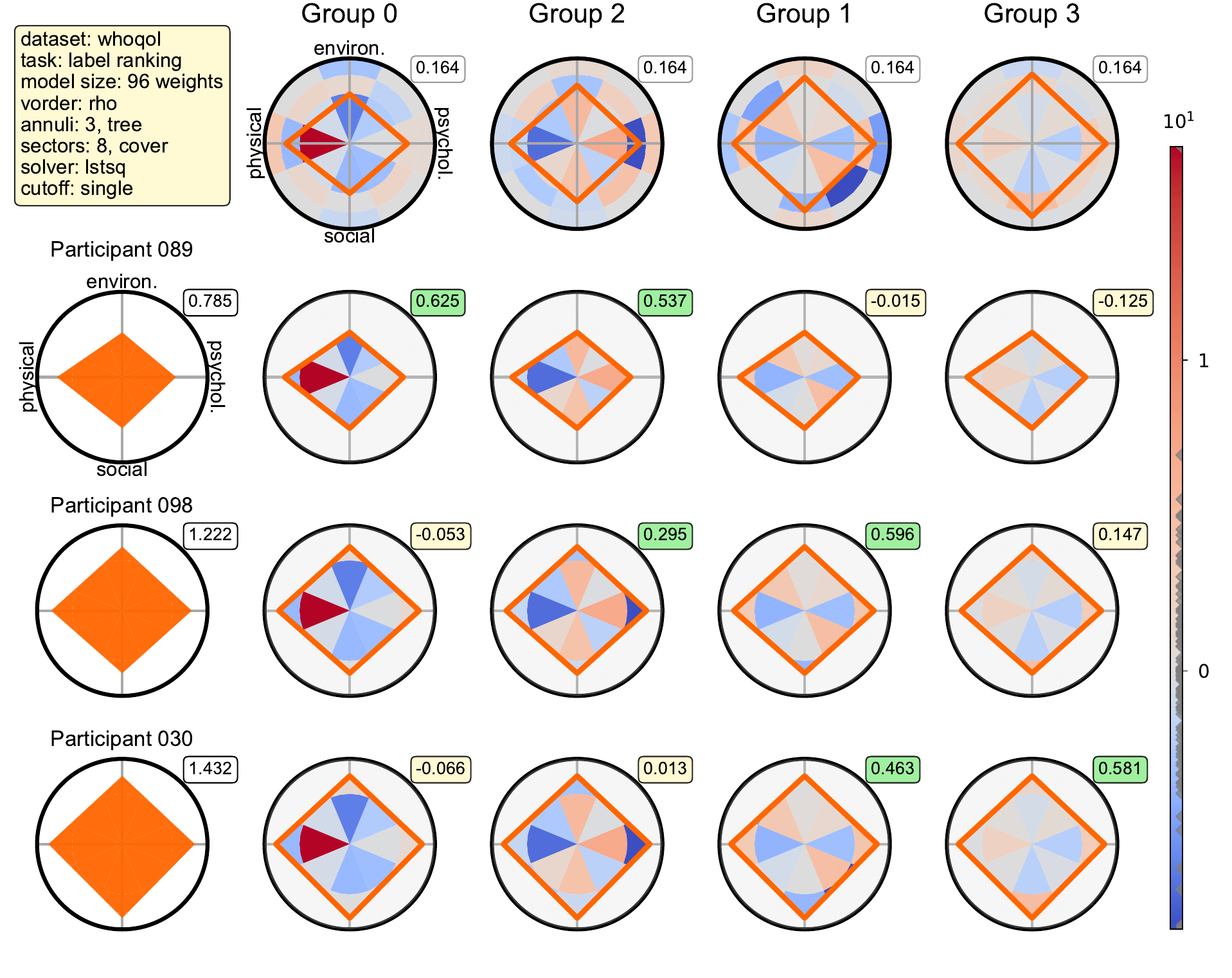}
    \caption{An explanation diagram for the three assessments detailed in Table \ref{tab:whoqol-sample}.
    They are sorted by increasing size of the assessment polygon, from top to bottom.
    The assignment charts are organised in the same way, from left to right.
    The synthetic assignment $Y(LR)$ in Table \ref{tab:whoqol-sample} defines four labels, and each assessment is assigned to two labels.
    In a sense, this choice of layout describes an ordinal multilabel classification of the assessments.
    The config described in the yellow tag at the top left of the diagram shows familiar parameter values, except for the annulus type, which is set to \aspas{tree}.
    In this case, the widths of the annuli are selected according to the thresholds found by a decision tree induced from the whoqol dataset.
    }
    \Description{}
    \label{fig:proposal:lr-task:best}
\end{figure}

To perform label ranking tasks, Polygrid goes through the same steps that were described for multilabel classification tasks, except that three of the algorithms must be adapted to handle the data encoding (see Section \ref{section:proposal:data}).
The first is Algorithm \ref{alg:learning:ml:step2}, which needs to convert a ranking assignment $Y$ into a multilabel assignment $Y^\downarrow$ as a preprocessing step.
The downgrade operator is:
\begin{empheq}[left={Y^{\downarrow} := (y^{\downarrow}_{ij})_{m \times n} \mbox{ such that } y^{\downarrow}_{ij} =  \empheqlbrace}]{equation}
  \begin{aligned}
      \label{eq:relatedwork:downgrade}
      & \, 1,  & & \mbox{if } j \in Y_i, \\[1ex]
      & \, 0,  & & \mbox{otherwise},
  \end{aligned}
\end{empheq}
e.g., if we have an assignment $Y_b=(1,2,-1,-1)$, then it is the case that $Y_b^\downarrow=(0,1,1,0)$.
The second is Algorithm \ref{alg:learning:ml:step5}, which converts a ranking assignment $Y$ to a fuzzy membership matrix $U$ such that:
\begin{empheq}[left={u_{ij} := u(j, Y_i) = \empheqlbrace}]{equation}
  \begin{aligned}
      \label{eq:proposal:logranks}
      & \, \frac{2^{n-1-\pi(j,Y_i)}}{2^n - 1} & & \mbox{if } j \in Y_i, \\[1ex]
      & \, 0                                  & & \mbox{otherwise,}
  \end{aligned}
\end{empheq}
where $\pi(j, Y_i)$ is the position of the label $j$ in the ranking given by $Y_i$.
This conversion uses the position of a label to define an arbitrary degree of membership that is locally consistent.
For example, let $Y_i = (3, 2, -1, -1)$.
Then $u_{ij} \gets u(j, Y_i)$ for $j \in 0 \ldots (n-1)$, and then $U$ is normalised to ensure that $\sum_j U_{ij} = 1$.
As a result, we end up with $U_i = (0, 0, \frac{1}{3}, \frac{2}{3})$.
Finally, Algorithm \ref{alg:ml:prediction} must be adapted to combine label presence and membership data to produce the predicted ranking.
The explanation diagram shown in Figure \ref{fig:proposal:lr-task:best}, which shows the three assessments from Table \ref{tab:whoqol-sample}, was generated by a Polygrid instance trained on the whoqol dataset with the label ranking assignment.

Up to this point in our exposition, we have focused on how Polygrid performs multilabel classification and label ranking tasks.
The problem of recommending referrals is equivalent to learning to solve these tasks, as seen in Section \ref{section:relatedwork:tasks:careplan}.
We framed the operation of the model as a processing pipeline through which the data is progressively transformed to extract numerical representations that can be used later to predict assignments from unseen assessments in the dataset.
Along the way, the algorithmic description offered for each step of the pipeline was systematically complemented by a description of how these numerical representations are graphically displayed on the explanation diagram.
Although this effort demonstrates that explanations generated by Polygrid faithfully reflect the operation of the model, it gives little insight about why the model learns, and 
it does little to justify any claims about its interpretability.
We now turn our attention to these issues.

\subsection{The Learnability of the Polygrid Model}
\label{section:proposal:closerlook}

The results in Table \ref{tab:performance:ml} convey some evidence in support of the idea that the Polygrid model can learn, but stronger empirical evidence will be provided later, in Section \ref{section:offlineval}.
Although the algorithms detailing the learning pipeline give us a clear idea of how Polygrid works, a deeper and more concise explanation of why it learns is missing.
This gap can be filled by noting that the Polygrid model loosely resembles a kernel method \cite{hoffman2008kernel}.
Similarly to the latter, Polygrid maps a $d$-dimensional vector $X_i$, which in our context holds the standardised assessment scores obtained by the $i$-th individual, to a vector $S_i := \Phi(X_i)$, which usually sits in a higher-dimensional space $\mathbb{R}^{n_{as}}$, called feature space in the literature on kernel methods.
In Polygrid, this mapping corresponds to:
\begin{align}
    \nonumber
    \Phi: (0,1]^d &\to (0, \frac{1}{2}\,]^{n_{as}} \\
    X_i &\mapsto \Phi(X_i) := (\text{\textit{ud-to-fs}} \circ \text{\textit{uh-to-ud}})(X_i),
    \label{eq:proposal:phi}
\end{align}
as described in Algorithms \ref{alg:learning:ml:step1} and \ref{alg:learning:ml:step4}.
In other words, a vector $X_i$ in the unit hypercube is mapped to a polygon on the unit disc, and then to a point in the Polygrid's feature space.
The intuition here is that, although real-world data most often encode nonlinear relationships $f(X) = Y$, $\Phi(\cdot)$ may be able to extract features from $X$ such that a linear relationship $\Phi(X) \transpose{W} = Y$ becomes a useful approximation.
However, the resemblance with traditional kernel methods ends there.
In kernel methods, the explicit evaluation of $\Phi(X)$ is bypassed by the \emph{kernel trick}.
In this operation, the inner product in the feature space is replaced by the evaluation of a kernel $k$ in the original domain:
\begin{align}
    \nonumber
    k: \mathbb{R}^d \times \mathbb{R}^d &\to \mathbb{R} \\
    \nonumber
       (X_a, X_b) &\mapsto k(X_a, X_b) = \innerprod{\Phi(X_a)}{\Phi(X_b)}.
    \label{eq:kernel}
\end{align}
In contrast, Polygrid explicitly handles vectors in the feature space, without resorting to a kernel.
The dimension of this space is determined by the hyperparameters $n_a$ and $n_s$, which specify the partitioning of the unit disc described in Algorithm \ref{alg:learning:ml:step3}.
Moreover, Polygrid explores a recurring idea from factorisation methods in recommender systems: it learns an explicit representation in the feature space for each label $j$, namely $W_j$, such that $\innerprod{S_i}{W_j}$ approximates $Y_{ij}$ (see Algorithms \ref{alg:learning:ml:step5}).
In the terminology of recommender systems, the vectors $X_i$ and $S_i$ describe a user, $W_j$ represents an item, and $Y_{ij}$ and the scalar $\innerprod{S_i}{W_j}$ correspond to the relevance of the $j$-th item to the $i$-th user.
All in all, the degree to which Polygrid can solve learning tasks depends on the mapping $\Phi(\cdot)$ being able to extract features from $X$ that makes $\hat{Y} \coloneq \Phi(X) \transpose{W}$ a useful approximation of $Y$.

\subsection{The Interpretability of the Polygrid Model}
\label{section:proposal:interpretable}

On what grounds can we claim that the Polygrid model is interpretable when the very notion of interpretability is still being debated?
If interpretability is conceived as a property, what are the classes of objects (e.g., models, model instances, predictions, explanations) to which interpretability can be ascribed and what is entailed by this?
Is the interpretability of a model dependent on the properties of its inputs (e.g., data cases, hyperparameters) or manifested by the properties of its outputs?
We will use these questions to guide our argument for the Polygrid's interpretability, but before we do that, our usage of some key terms must be made more precise, in order to circumvent the tensions that emerge when using terms that have been inconsistently defined in the literature.

\subsubsection{Preliminaries}
\label{section:proposal:interpretable:definitions}

Let's start by clarifying the distinction between the terms model and model instance.
We stand by their definitions in Section \ref{section:relatedwork:tasks:formalisation}: a model is an abstract structure that spans the hypothesis space $\mathcal{H}$, and a model instance is a hypothesis $h \in \mathcal{H}$ after its parameters have been adjusted to the training data.
However, in the interest of precision, these definitions must be extended to include details that will allow us to speak of \aspas{model size}.

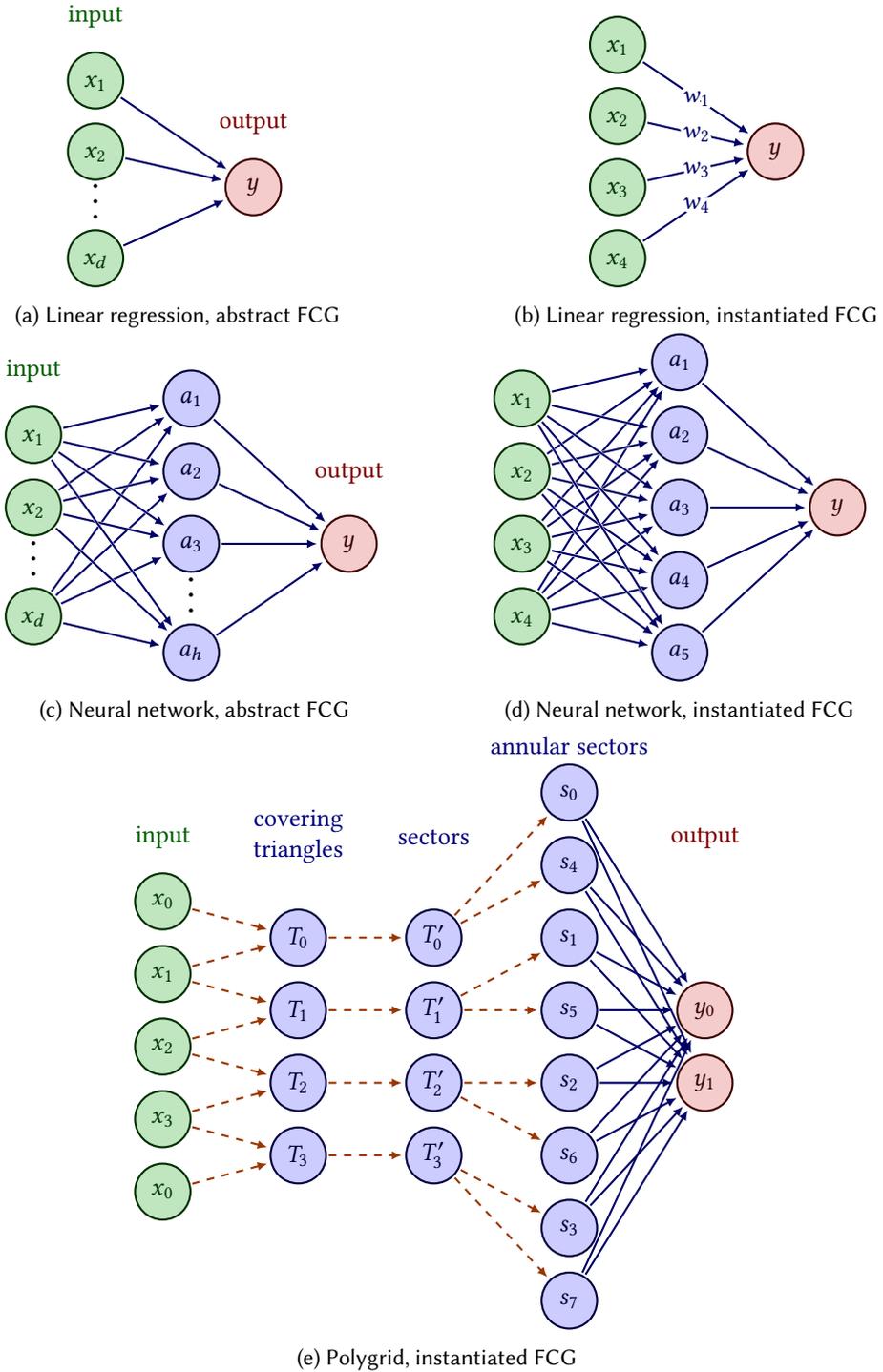
\begin{figure}[htpb]
    \centering
    \begin{subfigure}{.48\textwidth}
        \centering
        \begin{tikzpicture}[x=2.2cm,y=0.98cm]
          \message{^^JLinear Regression, architecture}
          \readlist\Nnod{3,1} 
          \readlist\Nstr{d,} 
          \readlist\Cstr{\strut x,y} 
          \def\yshift{0.5} 
          
          \message{^^J  Layer}
          \foreachitem \N \in \Nnod{ 
            \def\lay{\Ncnt} 
            \pgfmathsetmacro\prev{int(\Ncnt-1)} 
            \message{\lay,}
            \foreach \i [evaluate={\c=int(\i==\N); \y=\N/2-\i-\c*\yshift;
                         \index=(\i<\N?int(\i):"\Nstr[\lay]");
                         \x=\lay; \n=\nstyle;}] in {1,...,\N}{ 
                         
              \node[node \n] (N\lay-\i) at (\x,\y) {$\Cstr[\lay]_{\index}$};
              
              \ifnum\lay>1 
                \foreach \j in {1,...,\Nnod[\prev]}{ 
                  \draw[connect arrow] (N\prev-\j) -- (N\lay-\i);
                }
              \fi
              
            }
            \ifnum\lay<2 
                \path (N\lay-\N) --++ (0,1+\yshift) node[midway,scale=1.5] {$\vdots$};
            \fi      
          }
          
          \node[above=0.2,align=center,mygreen!60!black] at (N1-1.90) {input};
          \node[above=0.2,align=center,myred!60!black] at (N\Nnodlen-1.90) {output};
          
        \end{tikzpicture}
        \caption{Linear regression, abstract FCG}
        \Description{}
        \label{fig:proposal:fcg:lr-abs}
    \end{subfigure}
    \hfill
    \begin{subfigure}{.48\textwidth}
        \centering
        \begin{tikzpicture}[x=2.2cm,y=0.98cm]
          \message{^^JLinear Regression, instantiated}
          \readlist\Nnod{4,1} 
          \readlist\Nstr{4,} 
          \readlist\Cstr{\strut x,y} 
          \def\yshift{0.0} 
          
          \message{^^J  Layer}
          \foreachitem \N \in \Nnod{ 
            \def\lay{\Ncnt} 
            \pgfmathsetmacro\prev{int(\Ncnt-1)} 
            \message{\lay,}
            \foreach \i [evaluate={\c=int(\i==\N); \y=\N/2-\i-\c*\yshift;
                         \index=(\i<\N?int(\i):"\Nstr[\lay]");
                         \x=\lay; \n=\nstyle;}] in {1,...,\N}{ 
                         
              \node[node \n] (N\lay-\i) at (\x,\y) {$\Cstr[\lay]_{\index}$};
              
              \ifnum\lay>1 
                \foreach \j in {1,...,\Nnod[\prev]}{ 
                  \draw[connect,white,line width=1.2] (N\prev-\j) -- (N\lay-\i);
                  \draw[connect arrow] (N\prev-\j) -- (N\lay-\i)
                    node[pos=0.50] {\contour{white}{$w_{\j}$}};
                }
              \fi
            }
          }
          
        \end{tikzpicture}
        \caption{Linear regression, instantiated FCG}
        \Description{}
        \label{fig:proposal:fcg:lr-inst}
    \end{subfigure}
    \hfill
    \begin{subfigure}{.48\textwidth}
        \centering
        \begin{tikzpicture}[x=2.2cm,y=1.0cm]
          \message{^^JNeural network, architecture}
          \readlist\Nnod{3,4,1} 
          \readlist\Nstr{d,h,} 
          \readlist\Cstr{\strut x,a,y} 
          \def\yshift{0.5} 
          
          \message{^^J  Layer}
          \foreachitem \N \in \Nnod{ 
            \def\lay{\Ncnt} 
            \pgfmathsetmacro\prev{int(\Ncnt-1)} 
            \message{\lay,}
            \foreach \i [evaluate={\c=int(\i==\N); \y=\N/2-\i-\c*\yshift;
                         \index=(\i<\N?int(\i):"\Nstr[\lay]");
                         \x=\lay; \n=\nstyle;}] in {1,...,\N}{ 
                         
              \node[node \n] (N\lay-\i) at (\x,\y) {$\Cstr[\lay]_{\index}$};
              
              \ifnum\lay>1 
                \foreach \j in {1,...,\Nnod[\prev]}{ 
                  \draw[connect,white,line width=1.2] (N\prev-\j) -- (N\lay-\i);
                  \draw[connect arrow] (N\prev-\j) -- (N\lay-\i);
                }
              \fi
              
            }
            \ifnum\lay<3 
                \path (N\lay-\N) --++ (0,1+\yshift) node[midway,scale=1.5] {$\vdots$};
            \fi      
          }
          
          \node[above=0.2,align=center,mygreen!60!black] at (N1-1.90) {input};
          \node[above=0.3,align=center,myred!60!black] at (N\Nnodlen-1.90) {output};
          
        \end{tikzpicture}
        \caption{Neural network, abstract FCG}
        \Description{}
        \label{fig:proposal:fcg:nn-abs}
    \end{subfigure}
    \begin{subfigure}{.48\textwidth}
        \centering
        \begin{tikzpicture}[x=2.2cm,y=1.0cm]
          \message{^^JNeural network, instantiated}
          \readlist\Nnod{4,5,1} 
          \readlist\Nstr{4,5,} 
          \readlist\Cstr{\strut x,a,y} 
          \def\yshift{0.0} 
          
          \message{^^J  Layer}
          \foreachitem \N \in \Nnod{ 
            \def\lay{\Ncnt} 
            \pgfmathsetmacro\prev{int(\Ncnt-1)} 
            \message{\lay,}
            \foreach \i [evaluate={\c=int(\i==\N); \y=\N/2-\i-\c*\yshift;
                         \index=(\i<\N?int(\i):"\Nstr[\lay]");
                         \x=\lay; \n=\nstyle;}] in {1,...,\N}{ 
                         
              \node[node \n] (N\lay-\i) at (\x,\y) {$\Cstr[\lay]_{\index}$};
              
              \ifnum\lay>1 
                \foreach \j in {1,...,\Nnod[\prev]}{ 
                  \draw[connect,white,line width=1.2] (N\prev-\j) -- (N\lay-\i);
                  \draw[connect arrow] (N\prev-\j) -- (N\lay-\i);
                }
              \fi
            }
          }
          
          
        \end{tikzpicture}
        \caption{Neural network, instantiated FCG}
        \Description{}
        \label{fig:proposal:fcg:nn-inst}
    \end{subfigure}
    \vfill
    \begin{subfigure}{\textwidth}
        \centering
        \begin{tikzpicture}[x=1.89cm,y=1.0cm,decoration=brace]
          \message{^^Polygrid}
          \readlist\Nnod{5,4,4,8,2} 
          \readlist\Nstr{1,4,4,8,2} 
          \readlist\Cstr{x,T,T',s,y} 
          \def\yshift{0.0} 
          
          \message{^^J  Layer}
          \foreachitem \N \in \Nnod{ 
            \def\lay{\Ncnt} 
            \pgfmathsetmacro\prev{int(\Ncnt-1)} 
            \message{\lay,}
            \foreach \i [evaluate={\c=int(\i==\N); \y=\N/2-\i-\c*\yshift;
                         \index=(\i<\N?int(\i):"\Nstr[\lay]");
                         \x=\lay; \n=\nstyle;}] in {1,...,\N}{ 
                         
              \def\nodeindex{0};
              \tikzmath{
                 \nodeindex = int(\index - 1);
              }
              \ifnum \lay=4
                \ifodd \nodeindex
                    \tikzmath{
                       \nodeindex = int(\nodeindex + (7 - \nodeindex)/2);
                    }
                \else
                    \tikzmath{
                       \nodeindex = int(\nodeindex/2);
                    }
                \fi
              \fi
              \node[node \n] (N\lay-\i) at (\x,\y) {$\Cstr[\lay]_{\nodeindex}$};              
              \ifnum\lay=2 
                  \def\sourcea{0};
                  \def\sourceb{0};
                  \tikzmath{
                     \sourcea = int(\i);
                     \sourceb = int(\i+1);
                  }
                  \draw[connect arrow, dashed, myorange] (N\prev-\sourcea) -- (N\lay-\i);
                  \draw[connect arrow, dashed, myorange] (N\prev-\sourceb) -- (N\lay-\i);
              \fi

              \ifnum\lay=3 
                  \def\sourcea{0};
                  \tikzmath{
                     \sourcea = int(\i);
                  }
                  \draw[connect arrow, dashed, myorange] (N\prev-\sourcea) -- (N\lay-\i);
              \fi
              
              \ifnum\lay=4 
                  \def\sourcea{0};
                  \tikzmath{
                     \sourcea = int((\i+1)/2);
                  }
                  \draw[connect arrow, dashed, myorange] (N\prev-\sourcea) -- (N\lay-\i);
              \fi
              
              \ifnum\lay=5 
                \foreach \j in {1,...,\Nnod[\prev]}{ 
                  \draw[connect arrow] (N\prev-\j) -- (N\lay-\i);
                }
              \fi
            }
          }
          
          \node[above=0.2,align=center,mygreen!60!black] at (N1-1.90) {input};
          \node[above=0.5,align=center,myblue!60!black] at (N2-1.90) {covering\\triangles};
          \node[above=0.75cm,align=center,myblue!60!black] at (N3-1.90) {sectors};
          \node[above=0.0,align=center,myblue!60!black] at (N4-1.90) {annular sectors};
          \node[above=1.7cm,align=center,myred!60!black] at (N\Nnodlen-1.90) {output};


        \end{tikzpicture}
        \caption{Polygrid, instantiated FCG}
        \Description{}
        \label{fig:offline:fcg:ply}
    \end{subfigure}
    \caption{Examples of abstract and instantiated forward computational graphs (FCGs).
    Figure \ref{fig:offline:fcg:ply} corresponds to the instances described in Table \ref{tab:performance:ml} with $n_s/d=1$, $n_a=2$, \emph{sector=miss}, \emph{solver=lstsq}, with 16 weights.
    }
    \label{fig:proposal:fcg}
\end{figure}

Accordingly, we define a model as a quadruple $(C,A,H,O)$ with a coordinator $C$, an architecture $A$, hyperparameters $H$, and an optimisation procedure $O$.
The coordinator organises the operation of the model (e.g., it runs the sequence of steps needed to instantiate the model and fit it to the data, or to produce a prediction for an unseen input).
The architecture $A$ is an \term{abstract forward computational graph}{AFCG}, as illustrated in Figures \ref{fig:proposal:fcg:lr-abs} and \ref{fig:proposal:fcg:nn-abs}.
A \term{forward computational graph}{FCG} is a directed acyclic graph that represents a set of computations applied to the input data to produce an output.
In such a graph, an edge represents a parameter (aka a weight), and a node represents an intermediate variable and possibly any computations required to update that variable \cite{fang2024fcg}.
The coordinator instantiates the architecture $A$ according to the dimensionality of the training data and any relevant hyperparameters in $H$ (e.g., the number of hidden layers, the number of neurons per layer in a multilayer perceptron), and then tunes the weights of the instantiated graph $h$ by calling the optimisation procedure $O$.
The latter maximises the objective function $q(h, D)$ seen in Section \ref{section:relatedwork:tasks:formalisation}.
Two instantiated graphs corresponding to the architectures in Figures \ref{fig:proposal:fcg:lr-abs} and \ref{fig:proposal:fcg:nn-abs} are illustrated in Figures \ref{fig:proposal:fcg:lr-inst} and \ref{fig:proposal:fcg:nn-inst}, respectively.

To obtain a prediction for an unseen input $x$, the coordinator computes $h(x)$ and returns the result to the caller.
To produce an explanation for that prediction, the coordinator returns the instantiated graph $h$, whose nodes' values were updated during the prediction, to the caller.
\standing{This conceptual setup allows us to define fidelity as a relation between a model and the explanations it generates: an explanation is faithful to the degree that the forward computational graph of its corresponding prediction can be extracted from it}{fidelity defined}.
In addition, \standing{we say that a model is transparent to the extent it generates faithful explanations.}{transparency defined}
Note that both definitions are free of human judgement.

\subsubsection{Interpretability as a Multidimensional Property}
\label{section:proposal:interpretable:model}

In the discussion that follows, \standing{we adopt an operational definition of interpretability described by \citeonline{doshivelez2017rigorous}: interpretability is the measure of success participants achieve when asked to perform forward simulation tasks in an experimental context: \aspas{humans are presented with an explanation and an input, and must correctly simulate the model's output.}}{operational interpretability}
This definition clearly implies that interpretability is a graded notion, and we argue in the following that it also implies that it is a multidimensional property.

\paragraph{Data, User, and Interpretability}
Any account of interpretability must agree with the idea that higher dimensional data are less conducive to interpretability than lower dimensional data.
This expectation is reasonable because \aspas{interpretation} demands human cognition, which is a limited resource, and higher dimensional data ultimately implies higher cognitive demands.
Besides, meaningfulness, as the ability to ascribe meaning to each feature of the data, also contributes to interpretability.
Let's perform a thought experiment to bring the latter point home: a layperson is participating in a user study to assess interpretability (using our operational definition).
This is a within-subjects study with two experimental conditions.
In both, the participant is asked to perform five forward simulation tasks for a linear regression model trained on the \href{https://archive.ics.uci.edu/dataset/477/real+estate+valuation+data+set}{UCI Real State Valuation dataset}.
This dataset has six features, which are meaningful to the general public (e.g. location, house age, distance to the nearest metro station), and the participant must come up with the price per unit of area of a house.
In the first condition, the subject is presented with a forward computational graph, similar to Figure \ref{fig:proposal:fcg:lr-inst}, printed on a sheet of paper.
The nodes and edges of the graph are annotated with their corresponding names and values (e.g., \emph{house age = 32} instead of $x_2$).
The only exception is the output node, which is annotated only with its name (\emph{house price of unit area} instead of $y$).
Moreover, the formula for updating the output node is described in the legend.
This setting enables the participant to use her understanding of the problem domain in numerous creative ways, and some of these ways are useful in reviewing her answers before submission.
In the second condition, the subject is presented with the same computational graph, but the names of the features and any contextual elements that could hint at the domain problem are replaced with unrecognisable symbols (e.g., \aspas{0D280} instead of $x_2$).
\premise{The participant cannot review her answers in light of her knowledge of the domain and is more prone to make mistakes because of a larger cognitive demand to process unfamiliar symbols.}{effect of blocking}
In principle, this blocking (of the meaning of the data features) could lead to a lower average rate of success compared to the first experimental condition, which implies lower interpretability in the adopted methodological framework.

\paragraph{Model, User and Interpretability}
With respect to how a model contributes to interpretability, there are (at least) two relevant characteristics which are traceable to the model's architecture: scalability and meaning preservation.
Consider the first experimental condition of our thought experiment.
The linear regression model illustrated in Figure \ref{fig:proposal:fcg:lr-abs} takes up $d+1$ weights, and the meaning of each weight is derived from the meaning of the nodes it connects (i.e., the rate of change of the target variable per unit change in the feature), except for the intercept.
Now suppose that we replaced the linear regression model with the multilayer perceptron shown in Figure \ref{fig:proposal:fcg:nn-abs} with $h=5$ neurons in the hidden layer.
This instance would take up $5d + 11$ weights.
A larger instance implies that the participant is subject to a higher cognitive demand, so there are more opportunities for mistakes and more incentive to defer the task, which could lead to a lower average rate of success in our hypothetical user study.
Even if the linear regression model was replaced with a multilayer perceptron with $h=2$ neurons, which specifies a smaller instance ($2d+5$ weights), \confirm{the meaning of a weight can no more be derived from the nodes it connects because the hidden nodes do not correspond to any concepts in the problem domain.}{is this true? I still have a sense of how the output responds to the input, dont I?}
This produces effects that are similar to the blocking of meaning seen in the previous paragraph.
\standing{In a sense, we are advancing an operational notion of meaning preservation as the ability of the study participant to ascribe meaning to nodes and edges of the forward computational graph, given that the input and output data are meaningful.}{meaning preservation}

In conclusion, \standing{this analysis suggests that interpretability is better described as a multidimensional property that emerges from the interaction among multiple factors}{multidimensional interpretability}\footnote{Here, the term \aspas{interaction} is being used in the sense it is employed in discussions about perspectivism.
As an example from \citeonline{giere2006perspectivism} goes, the colour of a sweater is not a property that is intrinsic to that object.
Many factors determine our perception of colour: characteristics of the sources of light and their placement in the environment, characteristics of the fabric, as well as conditions of the viewer that may affect her perception (e.g., colour blindness).} rather than a unidimensional property (i.e., a linear combination of multiple independent contributions).
Despite this more complex conception, it must be noted that higher levels of interpretability can only be achieved in settings where both the data and the model have characteristics that are conducive to interpretability.

\subsubsection{A Defence of Polygrid's Interpretability}
\label{section:proposal:interpretable:defence}

Assuming the reader agrees with our tactical choice for an operational definition of interpretability and with our analysis about its contributing factors, we now proceed to justify the claim that Polygrid is an interpretable model.
Let's start with the data.
Psychometric data are typically low dimensional and meaningful to our user: the attending care professional was trained to use the instrument for CGA\footnote{As an example, the three psychometric instruments described in Appendix \ref{section:background:cga:instruments} are low dimensional, with $d \leq 5$.}.
Since both of these characteristics are conducive to interpretability, it remains to be shown that Polygrid also has characteristics that are conducive to interpretability.
Given that the Polygrid's scalability is finely controlled by the hyperparameters $n_a$ and $n_s/d$, it remains to be shown that it also preserves meaning.
We argue that it does on the basis of three premises about how Polygrid transforms the input data into predictions, and how the data are depicted in the explanation diagram.
The (informal) argument goes like this:

\begin{enumerate}

    \item [] \logicarg 
    {\makecell[l]{(P1) The assessment scores are visually encoded as the vertices of the assessment polygon.} \\
    (P2) The assessment scores are meaningful to the user.
    }
    {(C1) The vertices of the assessment polygon are meaningful to the user.}
    
\end{enumerate}

\begin{enumerate}

    \item [] \logicarg
    {\makecell[l]{(P3) The area of the assessment polygon is a visual analogue of the measurand.} \\
    (P4) The measurand is meaningful to the user.
    }
    {(C2) The area of the assessment polygon is meaningful to the user.}
    
\end{enumerate}

\begin{enumerate}

    \item [] \logicarg
    {\makecell[l]{\\(P5) The transformation of the assessment polygon into the matching polygon \\ \makebox[4.0ex]{} preserves the meanings of the assessment polygon.} \\
    (C3) The assessment polygon is meaningful to the user (= C1 and C2).
    }
    {(C4) The matching polygon is meaningful to the user.}\\
    
\end{enumerate}

\noindent
The premise P1 is true by construction (Algorithm \ref{alg:learning:ml:step1}), and the premises P2 and P4 are true because the user has been trained to use the instrument.
We argue that P3 is true due to the relationship between the measurand and the area of the assessment polygon.
Recall from Section \ref{section:background:measurement-of-health} that, for instruments based on the congeneric model, the standing of an individual on the latent variable is often estimated using Equation \ref{eq:background:weighted-score} to compute a sum-score.
Under certain assumptions, the area of the assessment polygon induces an ordering of subjects on the measurand that is equal to the ordering induced by the sum-scores.
This result warrants identifying the meaning of the measurand with the meaning of area of the assessment polygon, for the purpose of ordering people.
This relationship is fully detailed in Appendix \ref{appendix:areascore}, but here is a cut version.
Under the assumptions:
\begin{enumerate}
    \item [(A1)] $\mathring{x}_{ik} = \lambda_k \eta_i + \epsilon_{ik}$ (a structural equation from a congeneric factor model, Eq. \ref{eq:background:sem});
    \item [(A2)] $\eta_i > 0, $ for all $i = 0 \ldots m-1$ (the latent variable tracks a capacity of individuals);
    \item [(A3)] $\lambda_k > 0, $ for all $k=0 \ldots d-1$ (items correlate positively with the latent variable);
    \item [(A4)] $\epsilon_{ik} = 0$ (the measurement error is negligible for the purpose of ordering people);
    \item [(A5)] $\max\{\mathring{X}_{:k}\} = C \mbox{ for all } k \in 0 \ldots d-1, C > 0$  (all domain subscales have the same range);
\end{enumerate}
\noindent
whenever $\eta_a > \eta_b$, then it must be the case that:
\begin{align}
    \nonumber
    \eta_a > \eta_b  
    \overset{A3}{\implies}{\implies} \lambda_k \eta_a > \lambda_k \eta_b 
    \overset{A1,A4}{\implies} \mathring{x}_{ak} > \mathring{x}_{bk}
    \overset{A5,A2}{\implies} x_{ak} > x_{bk} > 0 \,\, (\forall k \in 0 \ldots d-1) \\
    \nonumber
    (x_{ak} > x_{bk}) \, \wedge \, (x_{a,k+1} > x_{b,k+1})
    \implies x_{ak} x_{a,k+1} > x_{bk} x_{b,k+1}  \,\, (\forall k \in 0 \ldots d-1) \\
    \nonumber
    \mbox{Fix } 2\nu = sin \big(\frac{2 \pi}{d}\big) > 0. \mbox{ Then } \nu x_{ak} x_{a,k+1} > \nu x_{bk} x_{b,k+1} \implies
    \sigma_{ak} > \sigma_{bk} \,\, (\forall k \in 0 \ldots d-1) \\
    \nonumber
    \sigma_{ak} > \sigma_{bk} \,\, (\forall k \in 0 \ldots d-1) \implies
    \sum_k \sigma_{ak} > \sum_k \sigma_{bk} \implies
    \sigma_a > \sigma_b.
\end{align}

\noindent
The statement $\eta_a > \eta_b$ speaks about two individuals that have distinct levels of the measurand $\eta$ (e.g., quality of life), with person A having a higher position on the latent variable than person B.
Accordingly, the implied statement $\sigma_a > \sigma_b$ indicates that the area of the assessment polygon for person A is greater than that of person B.
In these equations, $\sigma_{ak}$ corresponds to the area of the polygonal shape formed by two consecutive domain scores, namely $\mu(\triangle(0, \, x_{ak} \, \zeta_k, \, x_{a, k+1} \, \zeta_{k+1}))$\footnote{The expression $\triangle(z_0, \ldots, z_{d-1})$ indicates that the tuple $(z_0, \ldots, z_{d-1})$, taken as the closed polygonal chain $(z_0, \ldots, z_{d-1}, z_0)$, specifies a solid polygonal shape. 
In the term $x_{a, k+1} \, \zeta_{k+1}$, indices are taken modulo $d$, as usual.}.

Finally, premise P5 is true because the user can ascribe meaning to the elements of the forward computational graph that maps the assessment polygon to the matching polygon.
For example, the computational graph shown in Figure \ref{fig:offline:fcg:ply} is embedded in every explanation diagram generated by a Polygrid instance with the config in Figure \ref{fig:proposal:ml-task:annulus.2}.
In this graph, the nodes $T_0, \ldots, T_3$ correspond each to a triangle formed by two consecutive vertices of the assessment polygon (and the origin).
In fact, all nodes in blue represent some layer-wise additive component of the assessment polygon.
The edges connecting the two outer layers derive their meaning from the nodes they connect: the rate of change of $y_j$ per unit change in the feature $s_{r}$.
The remainder of the edges are constants.

\section{An Offline Performance Evaluation of the Polygrid Model}
\label{section:offlineval}

In this section, we report on the results of an offline evaluation of the Polygrid model.
We describe its performance against a number of datasets from the healthcare domain.
These datasets contain health-related assessments using standardised instruments (i.e., psychometric data) and include multilabel and label ranking assignments.
Once Polygrid has been fit to the data, several metrics are calculated and compared with a set of alternative models, whose performance were subjected to unusually strict conditions, namely controlling for model size\footnote{The term \aspas{model size} refers to the average number of weights taken up by instances of a model when trained on some dataset.
Thus, it is a context-dependent notion.
The concept of weight was defined in Section \ref{section:proposal:interpretable:definitions}.}.
But why should we control for model size?
The design of the evaluation was crafted with two goals in mind: 
\begin{enumerate}

    \item Considering multilabel classification and label ranking as relevant tasks, we want to show how competitive the Polygrid model is compared to top-performing models.
    
    \item If an alternative model achieves a degree of performance that is markedly superior to that of Polygrid, we want to answer the question: \emph{What did it do differently?}
    In other words, we want to be able to ascribe an observed difference in performance to some observable difference in the constitutive characteristics of the models (e.g., differences in their forward computational graphs).
    The aim is to identify ideas to improve the Polygrid model.
       
\end{enumerate}

These goals are in tension.
The first aims to rank the new model among top performers based on its performance on a collection of datasets.
This corresponds to the standard practice of evaluating a new model in a mature domain and, in such a setting, the size of a model is usually not taken into account directly \cite{demsar2006statistical}.
In contrast, the second goal is to identify potential improvements for the Polygrid model.
In this context, an improvement is an answer to the posed question.
However, we are not interested in answers that ascribe an observed difference in performance to a difference in size of the models being compared.
This is \premise{because interpretability deteriorates as the model size increases}{interpretability vs. scalability} and, since we care for interpretability, 
it seems reasonable to constrain the size of the models during evaluation.
We assume that the Polygrid model is fairly interpretable up to some size and constrain the evaluation to this limit.
As a consequence, a reported ranking such as $M_1 \succ M_2$ must be read as \aspas{for models of a certain size, the performance of $M_1$ dominated that of $M_2$.}

Another reason why this evaluation departs from the standard design is that the application of recommender systems to healthcare is far from being a mature domain, as discussed in Section \ref{section:relatedwork:healthrecsys}.
In fact, the application of recommender systems in gerontological care is in its infancy: there are no standard tasks or metrics, let alone public benchmark datasets.
This means that pursuing the first goal to the detriment of the second would lead us to less useful results.
For example, knowing that a multilayer perceptron with about 2,000 weights dominates the performance of Polygrid in a certain semi-synthetic dataset.
Arguably, this result does not provide useful guidance to model selection, given that it would deliver poorly on interpretability because of its size.
The remainder of this section is organised as follows.
The design of this evaluation is detailed in Section \ref{sec:offlineval}, its results are presented in Section \ref{sec:offlineval:results} and discussed in Section \ref{sec:offlineval:discussion}.

\subsection{The Design of the Offline Performance Evaluation}
\label{sec:offlineval}

In a nutshell, this is an offline evaluation in which we assess nine models across 15 datasets using nine distinct metrics.
We follow a standard protocol to rank the models according to their performance \cite{demsar2006statistical}, but the protocol is adapted to tackle the unmet assumptions of the statistical test.
In the following, we detail how the datasets were collected and curated, how the models and relevant metrics were selected, and how the evaluation process explores the Polygrid's configuration space, generates the performance data, and manages the size of the models.
Finally, the adaptation of the statistical analysis is justified and explained.

\subsubsection{Datasets}
\label{sec:offlineval:datasets}

Ideally, all datasets included in this evaluation should be similar to those collected by research initiatives implementing the WHO's \sigladef{Integrated Care for Older People}{ICOPE} approach to reorient primary care services for older people \cite{tavassoli2022ICOPE,ferrioli2024icopebr}.
These initiatives stand out for their emphasis on the quality of the data:
a large body of health professionals is trained to assess members of a well-defined target population using a standardised instrument, a large number of members of the target population is engaged in the study, and all recommendations made for each participant by their attending primary care professional are recorded.
Unfortunately, we were unable to access these datasets, either due to data protection barriers or because the study is still ongoing.
To overcome this limitation, we secured access to three datasets containing health-related assessments:

\begin{itemize}

    \item WHOQOL: this dataset was collected by researchers in the Department of Gerontology at the Federal University of São Carlos (Brazil) for a study reported in \citeonline{castro2022whoqol}. 
    It contains quality of life assessments of 100 older individuals (50 years or older, average 67 years, 86\% female) who participated in \sigladef{University of the Third Age}{U3A} activities promoted by the \sigladef{Fundação Educacional São Carlos}{FESC} institution in October 2019. 
    Assessments were conducted using the WHOQOL-BREF instrument (Annex \ref{section:background:whoqol}), but no recommendations or referrals were systematically recorded.
    In other words, using the notation defined in Section \ref{section:proposal:data}, this dataset contains only the description matrix, $\mathring{D} = (\mathring{X}, - )$. 
    
    \item ELSIO1: this dataset was derived from a publicly available dataset collected by researchers that contributed to the \sigladef{Brazilian Longitudinal Study of Aging}{ELSI-Brazil}, led by \sigladef{Oswaldo Cruz Foundation}{Fiocruz} \cite{limacosta2018elsi}.
    The researchers conducted interviews and collected physical measurements of 9,412 individuals in a nationwide sample of community-dwelling adults aged 50 years or more over the years 2015-16.
    The original data  were converted into assessments of intrinsic capacity (WHOIC instrument, Annex \ref{section:background:whoic}) following a method adapted from \citeonline{aliberti2022ic}, resulting in 7,175 derived assessments.
    For reasons related to the computational cost of the Polygrid model, \limitation{we decided to use a 10\% random sample of the data (718 of 7,175 data cases)}{why not a stratified sample (by area-score, for example)?}.
    Similar to the WHOQOL dataset, this contains only health assessments.

    \item AMPIAB: this dataset was collected by researchers at the University of São Paulo (Brazil) for a study reported in \citeonline{marcucci2020ampiab}.
    It contains the health assessments of 510 people (60 years or older, average 76 years, 78\% women) who were home-assisted by the \sigladef{Elderly Caregiver Programme}{PAI} of the Health Department of the city of São Paulo.
    Assessments were conducted in November 2018 using the \sigladef{Multidimensional Evaluation of Older People in Primary Care}{AMPI/AB} instrument (Annex \ref{section:background:ampiab}).
    Referrals were recorded for 128 patients.

\end{itemize}

In summary, these datasets contain health assessment data of distinct Brazilian populations (people participating in U3A activities in São Carlos, people receiving health care at home in São Paulo, and community dwellers nationwide).
Data were collected with different instruments (WHOQOL and WHOIC capture health capacity, and AMPIA/AB captures health deficits).
These instruments were developed in the factor-analytic tradition.
Referrals are mostly absent and, to address this issue, \limitation{we augmented the datasets with synthetic assignments}{synthetic assignments}.
The latter were generated based on \premise{a core principle in healthcare that asserts that individuals with similar needs should receive similar care}{principle for stratification} \cite{basil2020ethics}.
Furthermore, these assignments were constrained so that they reproduce key distributional properties that were observed in two real-world datasets, as explained next.

As summarised in Table \ref{tab:offlineval:datasets}, we created five versions of each dataset listed above: one version has a multiclass assignment, two have multilabel assignments, and two have label ranking assignments.
For example, the assignments in the whoqol dataset (used in Section \ref{section:proposal}) stratify the sample into individuals with good and poor quality of life, according to a criterion proposed by \citeonline{silva2014whoqolcut} --- individuals with a score equal to or greater than 60 are assigned to the \aspas{Good QoL} class.
The assignments in the elsio1 dataset were created by computing the Katz index\footnote{The Katz index summarises the ability of a person to perform activities of daily living, such as bathing, dressing, going to toilet, transferring, continence, and feeding \cite{katz1963index}.
} for each individual, as in \citeonline{aliberti2022ic}.
And the ampiab dataset has assignments based on the criterion described in \citeonline{marcucci2020ampiab}, which classifies individuals as healthy, pre-frail, and frail.
The two multilabel versions have identifiers that end with \aspas{-ml-11} or \aspas{-ml-22}.
\future{With the exception of the ampiab-ml-11 dataset}{This was tactical mistake, I guess. We should have ampiab-ml-11 with synthetic assignment, and create another dataset with real-world assignments - ampiab-ml-rw}, their assignments were created by applying fuzzy clustering to the assessment data, subject to a constraint on the $\lambda$-cut to reproduce distributional properties seen in these two studies:

\begin{table}[tpb]
    \centering
    \footnotesize
    \caption{Characteristics of the datasets used in the offline evaluation}
    \begin{tabular}{cccccccccc}
    \toprule
        \textbf{dataset} & \textbf{assignment} & \textbf{\#inst} & \textbf{\#ft} & \textbf{\#lb} & \textbf{card.} & \textbf{imb.} & \textbf{\#ls} & \textbf{\#sls} & \textbf{max l/i} \\ \midrule
        whoqol       & multiclass    & 100 &  4 &  2 & 1.00 & 0.31 & 2 & 0 & 1 \\ \midrule
        whoqol-ml-11 & multilabel    & 100 &  4 & 11 & 1.09 & 0.87 & 16 & 4 & 2 \\ \midrule
        whoqol-ml-22 & multilabel    & 100 &  4 & 22 & 4.54 & 0.72 & 77 & 63 & 8 \\ \midrule
        whoqol-lr-11 & label ranking & 100 &  4 & 11 & 1.09 & 0.87 & 18 & 7 & 2 \\ \midrule
        whoqol-lr-22 & label ranking & 100 &  4 & 22 & 4.54 & 0.72 & 92 & 86 & 8 \\ \midrule
        ampiab       & multiclass    & 510 &  5 &  3 & 1.00 & 0.64 & 3 & 0 & 1 \\ \midrule
        ampiab-ml-11 & multilabel    & 128 &  5 & 11 & 1.08 & 0.99 & 15 & 8 & 2 \\ \midrule
        ampiab-ml-22 & multilabel    & 510 &  5 & 22 & 4.54 & 0.61 & 215 & 113 & 13 \\ \midrule
        ampiab-lr-11 & label ranking & 510 &  5 & 11 & 1.08 & 0.56 & 25 & 6 & 2 \\ \midrule
        ampiab-lr-22 & label ranking & 510 &  5 & 22 & 4.54 & 0.61 & 259 & 168 & 13 \\ \midrule
        elsio1       & multiclass    & 718 &  5 &  6 & 1.00 & 1.00 & 6 & 0 & 1 \\ \midrule
        elsio1-ml-11 & multilabel    & 718 &  5 & 11 & 1.08 & 0.49 & 20 & 2 & 2 \\ \midrule
        elsio1-ml-22 & multilabel    & 718 &  5 & 22 & 4.54 & 0.58 & 234 & 86 & 10 \\ \midrule
        elsio1-lr-11 & label ranking & 718 &  5 & 11 & 1.08 & 0.49 & 25 & 3 & 2 \\ \midrule
        elsio1-lr-22 & label ranking & 718 &  5 & 22 & 4.54 & 0.58 & 431 & 299 & 10 \\ 
    \bottomrule
    \end{tabular}\\
    \noindent
    \footnotesize \justifying \emph{Legend}: Column \aspas{\#inst} shows the number of instances, \aspas{\#ft} of features, and  \aspas{\#lb} of labels. \aspas{card.} stands for cardinality, \aspas{imb.} for imbalance, \aspas{\#ls} is the number of labelsets and \aspas{\#sls} the number of single labelsets. The maximum number of labels per individual in a dataset is shown in \aspas{max l/i}.
    The imbalance score is computed as $1 - MaxIR(Y)^{-1}$ \cite[p.44]{herrera2016multilabel}.
    \label{tab:offlineval:datasets}
\end{table}

\begin{itemize}

    \item A subset of the AMPIAB dataset in which instances have assigned referrals \cite{marcucci2020ampiab}.
    This sample is taken as the \aspas{ampiab-ml-11} dataset, which has 128 instances and 11 labels (referrals).
    The cardinality observed in this sample is about $1.08$.
    This statistic was used to calibrate the $\lambda$-cut when creating the assignments for the whoqol-ml-11 and elsio1-ml-11 datasets.

    \item Although we did not have access to the dataset collected for the study reported in \citeonline{tavassoli2022ICOPE}, we used the summary data in Table 3 of that work to infer that the dataset has 22 labels and a cardinality around $4.54$. 
    This statistic was used to calibrate the constraint when creating assignments for the whoqol-ml-22, ampiab-ml-22, and elsio1-ml-22 datasets.
    
\end{itemize}

\noindent
In other words, a raw dataset $\mathring{D}$ was transformed into a multilabel dataset $D$ by this transformation:
\begin{equation}
    \underset{\substack{ \downarrow \\ original\\dataset}}{\mathring{D}} = (\mathring{X}, -) \mapsto (\underset{\substack{\downarrow \\ \mbox{\tiny \itshape scaled to} \\ \mbox{ \tiny \itshape the unit} \\ \mbox{ \tiny \itshape hypercube}}}{X,} -) \mapsto (X, \underset{\substack{\downarrow \\ \mbox{\tiny \itshape constrained} \\ \mbox{ \tiny \itshape fuzzy} \\ \mbox{ \tiny \itshape clustering}}}{u(X, \lambda)}) = (X,Y) = \underset{\substack{ \downarrow \\ multilabel \\ dataset}}{D}.
    \label{sec:offlineval:dataprep}
\end{equation}

\noindent
Finally, the two label-ranking versions correspond to datasets ending with \aspas{-lr-11} or \aspas{-lr-22}.
Their assignments were also created with the method in Equation \ref{sec:offlineval:dataprep}, with the additional step of using fuzzy membership scores to induce the ordering of the labels in each individual assignment \cite{fuzzy-logic-ross.ch10}.

\subsubsection{Alternative Models}
\label{sec:offlineval:models}
The following models were chosen because they appear at the top of the rankings published in the reference studies that were described in Section \ref{section:relatedwork:tasks}:

\begin{itemize}
    \item Top-performing models for multilabel classification: in the rankings published in \citeonline{bogatinovski2022multilabel}, the Binary Relevance with Random Forests of Decision Trees (BRRF) is the best performing model across multiple metrics among models that follow a problem transformation approach, while the Random Forest of Decision Trees (RF) is among the best models in the ensemble of adapted algorithms approach\footnote{In \citeonline{bogatinovski2022multilabel}, the Binary Relevance with Random Forests of Decision Trees is referred to as RFDTBR, and the Random Forest of Predicting Clustering Trees is referred to as RFPCT, but we refer to them as BRRF and RF, respectively.
    The latter is a Breiman's Random Forest, and since a Predictive Clustering Tree is induced with the ID3 algorithm, we assume it is reasonably similar to a CART tree.
    }.
    
    \item Top-performing models for label ranking: according to a ranking published in the supplementary material of \citeonline{fotakis2022labelranking}, the Random Forest (RF) model is competitive with the best performing models when evaluated against the semi-synthetic benchmark datasets, besides being relatively simple.
    
\end{itemize}

\noindent
We impose the use of the CART Decision Tree (DT) as base learning model on BRRF and RF\footnote{We use implementations from the \href{https://scikit-learn.org/1.2/}{scikit-learn} and \href{https://scikit-multilearn-ng.github.io/scikit-multilearn-ng/}{scikit-multilearn-ng} packages.}.
The DT model is also included in the alternative models, as well as the Binary Relevance with Decision Trees (BRDT).
The aim is to allows us to isolate the contribution of the binary relevance strategy from that of the bagging strategy of Random Forests\footnote{The bagging strategy used in RF consists in (a) ignoring some less relevant features when looking for the best split, and (b) using a distinct random subset of the training data to fit each individual tree in the forest.} by comparing the performance of BRRF and RF with that of BRDT and DT.
The following models were also included:
Least Squares Linear Regression (Linear), Ridge Regression (Ridge), and a Multilayer Perceptron with a single hidden layer and sigmoid activation (MLP).
These models were included because of their structural similarity with the Polygrid model.
For example, a Polygrid model with a single annulus and a single sector per domain, Polygrid$(1,1)$, will take up as many parameters as a Linear or Ridge model.
In this circumstance, differences in performance can be precisely determined: the observed difference can be solely ascribed to the partitioning of the feature space\footnote{For example, setting sector type to \aspas{miss} in Polygrid$(1,1)$, as in Figure \ref{fig:proposal:ml-task:vorder}, replaces the features by their cyclic interactions in a linear regression: $(x_0, x_1, x_2, x_3) \to (x_0 x_1, x_1 x_2, x_2 x_3, x_3 x_0)$.}.
The MLP model was included because it generally projects the input data to a higher-dimensional space, similar to what Polygrid does.
Accordingly, if an MLP instance and a Polygrid instance have the same size but there is a substantial difference in performance, then the difference can be ascribed to differences in how these models project the input data to the feature space and differences in the dimensionality of the target feature space.
Finally, a Random model was included for sanity checking.

\subsubsection{Relevant Metrics}
\label{sec:offlineval:metrics}

\begin{figure}
    \centering
    \includegraphics[width=1\textwidth]{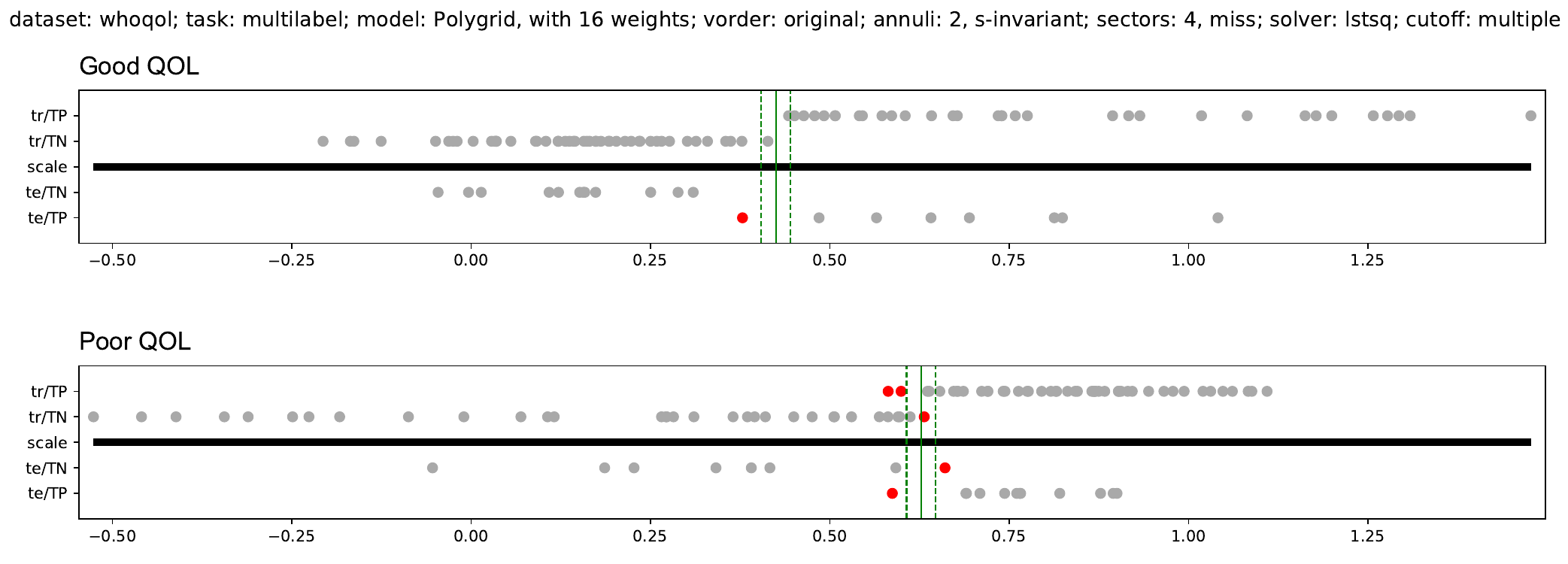}
    \caption{This diagram organises the matching scores (points) in four levels: label, partition, ground case, and predicted case.
        \textbf{(1)} the upper subplot corresponds to the \aspas{Good QOL} label, and the lower one to the \aspas{Poor QOL} label.
        \textbf{(2)} the scores from subjects in the training partition are plotted above the thick line, in the \aspas{tr/*} rows, and the ones from subjects in the test partition appears below that line, in the \aspas{te/*} rows.
        \textbf{(3)} the scores plotted in the \aspas{*/TN} rows correspond to true negative cases, and the scores in the \aspas{*/TP} rows are true positive cases.
        \textbf{(4)} A solid green line represents a threshold value (see Algorithm \ref{alg:learning:ml:step5}).
        Ideally, true negative cases should appear to the left of this line, and true positive cases should appear to its right.
        Cases that fail to meet this condition appear in red.
        Our Jaccard metric assesses the normalised size of the intersection between the minimal intervals defined by the scores in the \aspas{te/TN} and \aspas{te/TP} rows, averaged over the labels.
    }    
    \Description{}
    \label{fig:proposal:scales-introduced}
\end{figure}

To evaluate models in multilabel classification tasks, we selected the following subset of the metrics reported in \citeonline{bogatinovski2022multilabel}: subset accuracy (which is referred to as accuracy for simplicity), micro and macro averaged F1 scores (f1.micro and f1.macro), label-weighted F1 score (f1.weigh), and the Hamming loss (hammingl).
The accuracy metric indicates the share of predicted labelsets that exactly match the ground truth, f1.micro and hamming loss measure success at the level of the cells of the \aspas{prediction table}, and the difference between f1.macro and f1.weigh indicates the impact of the imbalance of the dataset on model learning.
In addition, we also implemented a set-based Jaccard similarity metric that is used to assess the degree to which the model can separate true positive cases from true negative cases, as illustrated in Figure \ref{fig:proposal:scales-introduced}.
Regarding model evaluation in label ranking tasks, we followed \citeonline{fotakis2022labelranking} and selected the traditional Kendall's tau metric.
Moreover, we implemented a metric that is equivalent to subset accuracy, which we called lracc, that computes the fraction of predicted label rankings that exactly match the ground truth, and a Hamming loss equivalent metric, called lrloss, that computes the fraction of predicted label/rank cells that do not match the ground truth.

\subsubsection{Evaluating the Polygrid Model}
\label{sec:offlineval:methodology:stage1}

The first stage of the evaluation consists of a search for the best Polygrid config for each dataset and metric.
In this context, a config consists of a tuple indicating the number of sectors per domain ($1 \ldots 3$), the number of annuli ($1 \ldots 8$), the order of the vertices ($3$ options), the type of sector ($2$ options), the type of annuli ($3$ options), the type of solver ($4$ options), and the cutoff scheme ($2$ options), totalling 3,456 configs.
Each config is evaluated $ss = 50$ times, producing samples of this size for every (dataset, config, metric)-triple.
Therefore, we obtained 186,624 samples for multiclass and multilabel datasets, and 62,208 samples for label ranking datasets, resulting in 248,832 samples in total\footnote{Details: 248,832 = 186,624 + 62,208, with 186,624 $=$ 9 datasets $\times$ 6 metrics $\times$ 3,456 configs, and 62,208 $=$ 6 datasets $\times$ 3 metrics $\times$ 3,456 configs.}.
The sample size $ss$ was empirically chosen so that the evaluation could be repeated on a different computing infrastructure with high consistency.
The ranges for the number of sectors per domain and the number of annuli were chosen so that the workload demanded by the evaluation could be executed within a couple of weeks.

\subsubsection{Evaluating the Alternative Models}
\label{sec:offlineval:methodology:stage2}

The second stage of the evaluation measures the performance of the alternative models in the best configs found in the previous stage.
As mentioned earlier, a size constraint is placed on the instances of the alternative models.
For example, the best Polygrid config found in the previous stage for the whoqol dataset ($d=4$ features, $n=2$ labels), based on the accuracy metric, was (\emph{nspd = 1, na = 2, vorder: averages, annulus: r-invariant, sector: cover, solver: ridge, cutoff: single}).
In this config, a Polygrid instance takes up 18 weights.
This means that, when fitting the MLP model for this config, the number of neurons in the hidden layer is calculated so the number of weights taken up by the instance is close to $18$.
The architecture of the MLP model imposes $h = (target - n)/(d + n + 1) = (18 - 2)/(4 + 2 + 1) = 16/7 = 2.3$, with $h$ being the number of neurons.
Since $h$ must be an integer, the strategy is to round off to the next integer ($h=3$) and track deviations from the target number of weights so that the mean approaches the target value.
If we set the number of repetitions to $ss = 10$, the strategy will produce MLP instances with the following sizes: $[23, 9, 23, 16, 23, 9, 23, 16, 23, 9]$, which averages to $17.4$.
For DT and BRDT models, the average size of their instances is controlled by limiting the depth of their trees\footnote{A split node has two parameters (feature and threshold), whereas a leaf node has just one.}.
The same strategy applies to RF and BRRF instances, which have an additional limit on the number of trees per instance.
For RF instances, the maximum number of trees is set to the number of existing labels. 
This allows for a structural comparison between RF and BRDT models, as both can grow the same number of trees.
In the case of BRRF, each component RF is allowed to grow up to $10$ trees.
In contrast, the number of weights in instances of Linear and Ridge models is determined solely by the number of features of the training data and, therefore, no strategy can be enforced.

\subsubsection{Data Analysis}
\label{sec:offlineval:methodology:stage3}

To support our first goal, which requires the comparison of multiple models based on their performance over multiple datasets, we adopted a popular method in machine learning research, advanced by \citeonline{demsar2006statistical} and \citeonline{benavoli2016should}.
The method consists of applying the Friedman test to the performance data to test for significant differences among models.
If a significant difference is found, then pairwise comparisons of average ranks using the Wilcoxon signed-rank test are performed.
Since multiple comparisons inflate the family-wise error rate, the significance level $\alpha$ is adjusted accordingly using the Bonferroni's or Holm's method.
Luckily, this method has been automated by software such as the \href{https://mirkobunse.github.io/critdd/index.html}{critdd package}.
However, every statistical method comes with assumptions, and a critical one is not met in our case: the datasets must provide independent observations.
The issue is that the three original datasets were augmented with synthetic assignments to build our collection of 15 datasets which, thus, are not independent.
In fact, we found that the method implemented in critdd was not able to discern the accuracy of the Random model from that of the remaining ones, which we take as evidence that the violation of the independence assumption is too severe.

To overcome this issue, we shift from a significance testing approach to a more deterministic one.
Pairwise comparisons of average ranks are still the core of the method but signed rank tests are replaced by dominance tests.
The latter are based on a matrix $A$ which counts the number of times a model dominates another on a given metric.
For example, assume $j_0$ refers to the Polygrid model, and $j_1$ to the Linear model.
The value of $A_{j_0 j_1}$ is increased by one whenever the confidence intervals for accuracy do not overlap, and the point estimate for Polygrid is greater than that of the Linear model.
Finally, to rank the models according to some metric, we combine the average ranks with the dominance data using a method that is based on the following analogy\footnote{We explore the analogy \aspas{models are like persons} to make the explanation more intuitive.}:
\begin{itemize}

    \item The top-ranking model (in terms of average rank) is \aspas{hired} to lead the first echelon.
    
    \item Once hired, the leader must staff the first echelon with competitive models.
    A model $j_1$ is competitive with the leading model $j_0$ if $abs(A_{j_0 j_1} - A_{j_1 j_0}) \leq 1$.
    The value $A_{j_0 j_1}$ is the number of times that model $j_0$ dominates the performance of model $j_1$.
    The upper bound of the inequality accounts for settings with an odd number of datasets.
    
    \item Once all models that are competitive with the leader have been hired, the next free top-ranking model is hired to lead the second echelon, and the process restarts.
    \item The process ends when all models have been hired.
    
\end{itemize}

\noindent
As a result, the ranking of the models is represented by a hierarchy of echelons, with each echelon being represented by a non-empty set of model indices.
Figure \ref{fig:offlineval:results:stage3} shows the ranking produced by this method.
The following statements are true about the hierarchies shown in the figure: the leading model in the first echelon dominates the leading model in the second echelon, which dominates the leading model in the third echelon, and so on.
Moreover, any model in an echelon is competitive with the leading model in that same echelon.
Note that this graphical representation conveys ranking information that is structurally equivalent to that found in critical difference diagrams, which are used in the standard method mentioned earlier \cite{demsar2006statistical}, or in Hasse graphs, which have recently been proposed in emerging evaluation methodologies \cite{jansen2024statistical}.

\subsection{Results}
\label{sec:offlineval:results}

\begin{sidewaysfigure}[htpb]
  \vskip 32.5\baselineskip
  \centering
  \scriptsize
  \begin{subfigure}{0.15\textwidth}
    \includegraphics[width=\linewidth]{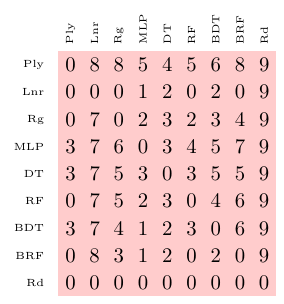}
    \caption{accuracy}
  \end{subfigure}
  \begin{subfigure}{0.15\textwidth}
    \includegraphics[width=\linewidth]{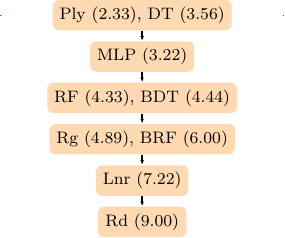}
    \caption*{}
  \end{subfigure}
  \hfill
  \begin{subfigure}{0.15\textwidth}
    \includegraphics[width=\linewidth]{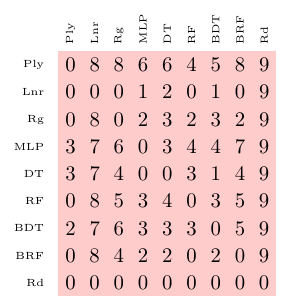}
    \caption{hammingl}
  \end{subfigure}
  \begin{subfigure}{0.15\textwidth}
    \includegraphics[width=\linewidth]{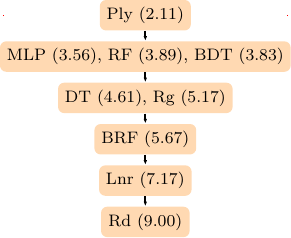}
    \caption*{}
  \end{subfigure}
  \hfill
  \begin{subfigure}{0.15\textwidth}
    \includegraphics[width=\linewidth]{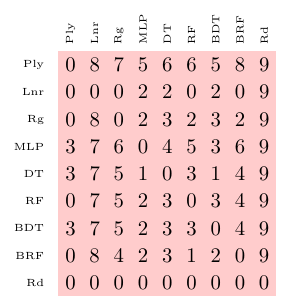}
    \caption{f1.micro}
  \end{subfigure}
  \begin{subfigure}{0.15\textwidth}
    \includegraphics[width=\linewidth]{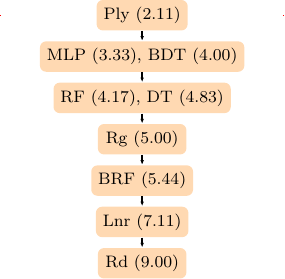}
    \caption*{}
  \end{subfigure}
  \vfill
  \begin{subfigure}{0.15\textwidth}
    \includegraphics[width=\linewidth]{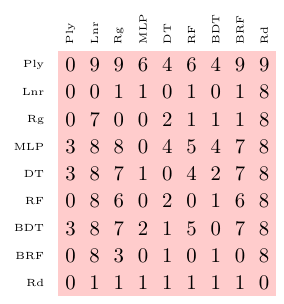}
    \caption{f1.macro}
  \end{subfigure}
  \begin{subfigure}{0.15\textwidth}
    \includegraphics[width=\linewidth]{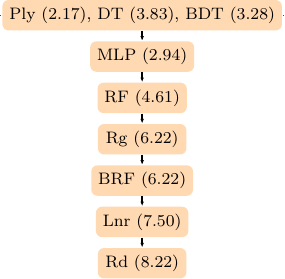}
    \caption*{}
  \end{subfigure}
  \hfill
  \begin{subfigure}{0.15\textwidth}
    \includegraphics[width=\linewidth]{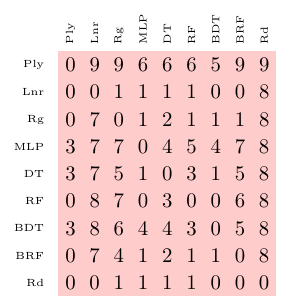}
    \caption{f1.weigh}
  \end{subfigure}
  \begin{subfigure}{0.15\textwidth}
    \includegraphics[width=\linewidth]{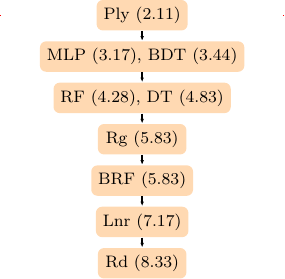}
    \caption*{}
  \end{subfigure}
  \hfill
  \begin{subfigure}{0.15\textwidth}
    \includegraphics[width=\linewidth]{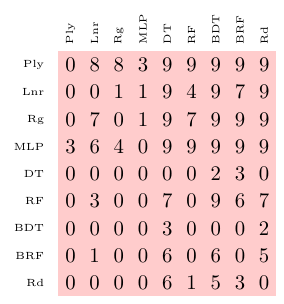}
    \caption{jaccsim}
  \end{subfigure}
  \begin{subfigure}{0.15\textwidth}
    \includegraphics[width=\linewidth]{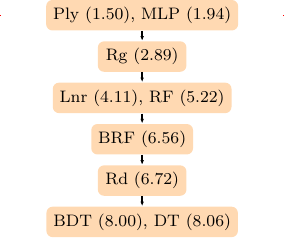}
    \caption*{}
  \end{subfigure}
  \vfill
  \begin{subfigure}{0.15\textwidth}
    \includegraphics[width=\linewidth]{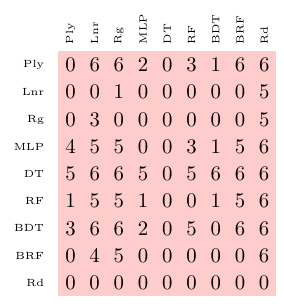}
    \caption{lracc}
  \end{subfigure}
  \begin{subfigure}{0.15\textwidth}
    \includegraphics[width=\linewidth]{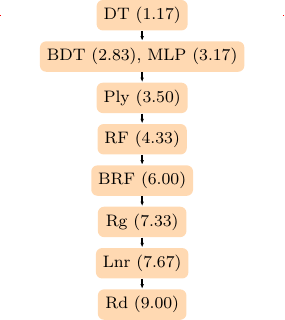}
    \caption*{}
  \end{subfigure}
  \hfill
  \begin{subfigure}{0.15\textwidth}
    \includegraphics[width=\linewidth]{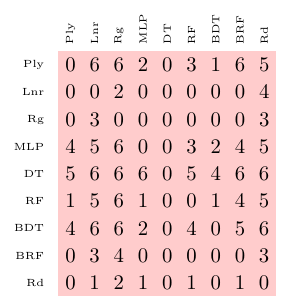}
    \caption{lrloss}
  \end{subfigure}
  \begin{subfigure}{0.15\textwidth}
    \includegraphics[width=\linewidth]{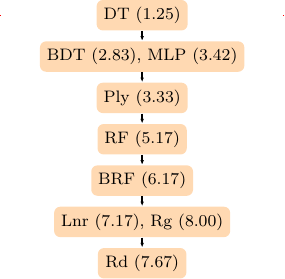}
    \caption*{}
  \end{subfigure}
  \hfill
  \begin{subfigure}{0.15\textwidth}
    \includegraphics[width=\linewidth]{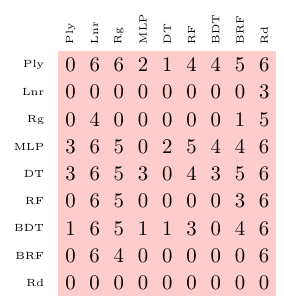}
    \caption{ktau}
  \end{subfigure}
  \begin{subfigure}{0.15\textwidth}
    \includegraphics[width=\linewidth]{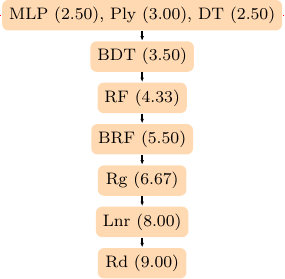}
    \caption*{}
  \end{subfigure}
  \vfill
  \caption{Results from the offline evaluation.
  \emph{Legend:} Models ranked by their performance in each metric.
  For each metric, the respective dominance matrix and ranking are depicted.
  An element $A_{j_0 j_1} = c$ means that model $j_0$ dominated the performance of model $j_1$ in $c$ datasets, with respect to some metric.
  In the corresponding diagram, which depicts the ranking of the models, the number that appears in parenthesis indicates the average rank of the model.
  Finally, the names of some models have been abbreviated: Polygrid (Ply), Linear (Lnr), Ridge (Rg),  BRDT (BDT), BRRF (BRF), and Random (Rd).
  Consult Appendix \ref{appendix:offline-results} for a detailed depiction of these results.
  }
  \Description{}
  \label{fig:offlineval:results:stage3}
\end{sidewaysfigure}

The results of the first stage of the evaluation are shown in Tables \ref{tab:offlineval:results:stage1} and \ref{tab:offlineval:results:stage1:dict}.
The indices of the configs in which Polygrid achieves the highest performance on some dataset and metric are listed on Table \ref{tab:offlineval:results:stage1}.
Of the 3,456 Polygrid configs in the search space, only 47 configs appear in the top ranking for some dataset and metric.
The specification of a config can be consulted by looking up its index in Table \ref{tab:offlineval:results:stage1:dict}.
For example, the config in which Polygrid achieved the highest accuracy on the whoqol dataset has index 166, and corresponds to (\emph{nspd = 1, na = 2, vorder: averages, annulus type: r-invariant, sector type: cover, solver: ridge, cutoff:  single}).
The final results are illustrated in Figure \ref{fig:offlineval:results:stage3},
in which the model rankings for each metric are displayed in a diagrammatic form, sided by the resulting dominance matrix used to group the models into distinct levels of performance (echelons).
In general, these results show that Polygrid was competitive with the alternative models in multiclass and multilabel datasets but was less successful in label ranking datasets.

\begin{table}[!ht]
    \centering
    \footnotesize
    \caption{Indices of the best Polygrid configs per metric and dataset (Stage 1)}
    \begin{tabular}{cccccccccc}
    \toprule
        ~ & \multicolumn{9}{c}{multiclass and multilabel datasets}  \\
        \cmidrule(lr){2-10} 
        ~ & \multicolumn{3}{c}{whoqol} & \multicolumn{3}{c}{ampiab} & \multicolumn{3}{c}{elsio1} \\
        \cmidrule(lr){2-4} \cmidrule(lr){5-7} \cmidrule(lr){8-10}
        metric & base & ml-11 & ml-22 & base & ml-11 & ml-22 & base & ml-11 & ml-22 \\ 
    \toprule
    
        accuracy & 166 & 3168 & 3304 & 1952 & 2806 & 3193 & 2995 & 3429 & 2275 \\ \midrule
        hammingl & 166 & 3168 & 3434 & 1952 & 166 & 3185 & 2995 & 3267 & 2115 \\ \midrule
        f1.micro & 166 & 3424 & 3450 & 1952 & 1494 & 3185 & 2995 & 3267 & 2275 \\ \midrule
        f1.macro & 166 & 3349 & 3402 & 1952 & 1164 & 2993 & 3283 & 3267 & 2275 \\ \midrule
        f1.weigh & 166 & 3349 & 3450 & 1952 & 2313 & 3185 & 3285 & 3429 & 2275 \\ \midrule
        jaccsim & 6 & 3349 & 1830 & 42 & 1884 & 1507 & 121 & 2976 & 2646 \\ \midrule
    
        ~ & ~ & ~ & ~ & ~ & ~ & ~ & ~ & ~ & ~ \\ 
    \toprule
        ~ & \multicolumn{9}{c}{label ranking datasets}  \\
        \cmidrule(lr){2-10} 
        ~ & \multicolumn{3}{c}{whoqol} & \multicolumn{3}{c}{ampiab} & \multicolumn{3}{c}{elsio1} \\
        \cmidrule(lr){2-4} \cmidrule(lr){5-7} \cmidrule(lr){8-10}
        metric & base & lr-11 & lr-22 & base & lr-11 & lr-22 & base & lr-11 & lr-22 \\
    \toprule
        ktau   & - & 3032 & 3081 & - & 3185 & 1457 & - & 2799 & 3281 \\ \midrule
        lracc  & - & 3394 & 2757 & - & 3345 & 2873 & - & 3413 & 2173 \\ \midrule
        lrloss & - & 3394 & 3317 & - & 3184 & 2769 & - & 3408 & 2131 \\ 
    \bottomrule
    
    \end{tabular}\\
    \noindent
    \footnotesize \justifying \emph{Note}: Results from the first stage of the offline evaluation.
    The upper block shows the best configs for multiclass and multilabel datasets, and the lower block for label ranking datasets.
    The specification of each Polygrid config can be recovered by looking up its corresponding index in Table \ref{tab:offlineval:results:stage1:dict}.
    \label{tab:offlineval:results:stage1}
\end{table}

\begin{table}[!ht]
    \centering
    \footnotesize
    \caption{Descriptions of the best Polygrid configs found in stage 1, by index}
    \begin{tabular}{clcl}
    \toprule
        index & description & index & description \\ \midrule

6 & (1, 1, avg, s-invt, cover, ridge, single)      & 2995 & (3, 5, msr, r-invt, cover, lstsqsym, multiple) \\ \midrule
42 & (1, 1, avg, tree, miss, lstsqsym, single)           & 3032 & (3, 6, avg, s-invt, miss, lstsq, single)       \\ \midrule
121 & (1, 1, msr, r-invt, miss, lstsq, multiple)     & 3081 & (3, 6, rho, s-invt, miss, lstsq, multiple)          \\ \midrule
166 & (1, 2, avg, r-invt, cover, ridge, single)      & 3168 & (3, 7, avg, s-invt, cover, lstsq, single)      \\ \midrule
1164 & (2, 1, avg, s-invt, miss, lstsquni, single)    & 3184 & (3, 7, avg, r-invt, cover, lstsq, single)      \\ \midrule
1457 & (2, 3, avg, r-invt, cover, lstsq, multiple)    & 3185 & (3, 7, avg, r-invt, cover, lstsq, multiple)    \\ \midrule
1494 & (2, 3, rho, s-invt, cover, ridge, single)           & 3193 & (3, 7, avg, r-invt, miss, lstsq, multiple)     \\ \midrule
1507 & (2, 3, rho, r-invt, cover, lstsqsym, multiple)      & 3267 & (3, 7, msr, s-invt, cover, lstsqsym, multiple) \\ \midrule
1830 & (2, 5, msr, s-invt, cover, ridge, single)      & 3281 & (3, 7, msr, r-invt, cover, lstsq, multiple)    \\ \midrule
1884 & (2, 6, avg, s-invt, miss, lstsquni, single)    & 3283 & (3, 7, msr, r-invt, cover, lstsqsym, multiple) \\ \midrule
1952 & (2, 6, rho, tree, cover, lstsq, single)                  & 3285 & (3, 7, msr, r-invt, cover, lstsquni, multiple) \\ \midrule
2115 & (2, 7, msr, s-invt, cover, lstsqsym, multiple) & 3304 & (3, 7, msr, tree, miss, lstsq, single)              \\ \midrule
2131 & (2, 7, msr, r-invt, cover, lstsqsym, multiple) & 3317 & (3, 8, avg, s-invt, cover, lstsquni, multiple) \\ \midrule
2173 & (2, 8, avg, s-invt, miss, lstsquni, multiple)  & 3345 & (3, 8, avg, tree, cover, lstsq, multiple)           \\ \midrule
2275 & (2, 8, msr, r-invt, cover, lstsqsym, multiple) & 3349 & (3, 8, avg, tree, cover, lstsquni, multiple)        \\ \midrule
2313 & (3, 1, avg, s-invt, miss, lstsq, multiple)     & 3394 & (3, 8, rho, tree, cover, lstsqsym, single)               \\ \midrule
2646 & (3, 3, rho, s-invt, cover, ridge, single)           & 3402 & (3, 8, rho, tree, miss, lstsqsym, single)                \\ \midrule
2757 & (3, 4, avg, r-invt, cover, lstsquni, multiple) & 3408 & (3, 8, msr, s-invt, cover, lstsq, single)      \\ \midrule
2769 & (3, 4, avg, tree, cover, lstsq, multiple)           & 3413 & (3, 8, msr, s-invt, cover, lstsquni, multiple) \\ \midrule
2799 & (3, 4, rho, s-invt, miss, ridge, multiple)          & 3424 & (3, 8, msr, r-invt, cover, lstsq, single)      \\ \midrule
2806 & (3, 4, rho, r-invt, cover, ridge, single)           & 3429 & (3, 8, msr, r-invt, cover, lstsquni, multiple) \\ \midrule
2873 & (3, 4, msr, tree, miss, lstsq, multiple)            & 3434 & (3, 8, msr, r-invt, miss, lstsqsym, single)    \\ \midrule
2976 & (3, 5, msr, s-invt, cover, lstsq, single)      & 3450 & (3, 8, msr, tree, miss, lstsqsym, single)           \\ \midrule
2993 & (3, 5, msr, r-invt, cover, lstsq, multiple)    & --- & ~ \\ 

    \bottomrule
    \end{tabular}\\
    \vskip 1 \baselineskip
    \noindent
    \footnotesize \justifying \emph{Legend}: This is the order in which hyperparameters appear in a tuple: (number of sectors per domain, number of annuli, ordering of the vertices, type of annulus, type of sector, type of solver, cutoff scheme).
    Some values have been abbreviated: averages (avg), measures (msr), s-invariant (s-invt), and r-invariant (r-invt).
    \label{tab:offlineval:results:stage1:dict}
\end{table}

Before we discuss how the goals of the evaluation are supported by these results, we want to address two surprising patterns that appear in the dominance matrices.
In principle, the performance of the Random model should not dominate any of the alternative models.
This implies that the last row of any dominance matrix in Figure \ref{fig:offlineval:results:stage3} should be filled with zeroes.
Moreover, if we extend this principle to state that the performance of any of the models should dominate the Random model, then the last column of these dominance matrices should be filled with 9 (in multilabel metrics) or 6 (in label ranking metrics), except in the last row.
These two patterns are observed for the accuracy, the Hamming loss, and the micro-averaged F1 score, but they break for the remaining metrics.
The reasons for this vary.
For the macro-averaged F1 score, the Random model dominates the performance of the alternative models in the ampiab-ml-11 dataset.
This is due to the severe imbalance of the dataset, with most instances assigned to the majority label (86 out of 128), and most labels having less than five instances each.
For these low-frequency labels, the models are prone to obtain null precision, while the Random model obtains small precision and recall scores.
The class imbalance also explains the last row/column pattern for the weighted F1 score.
For similar reasons, the Random model dominates the performance of several alternative models on the whoqol-lr-22 dataset, and dominates the BRRF model on the ampiab-lr-22 dataset regarding the lrloss metric.
Finally, the last row/column pattern breaks for the Jaccard similarity metric for a different reason: all tree-based models (DT, BRDT, RF, and BRRF) produce extreme scores.
This is to say that the leaf nodes of the regression trees tend to concentrate around the values assumed by $y_{ij} \in \{0, 1\}$.
As a result, the minimal interval that contains the predicted values $\hat{y}_{ij}$ tends to intersect with the corresponding intervals of the other classes, negatively affecting this metric.

\subsection{Discussion}
\label{sec:offlineval:discussion}

As said before, the statistical analysis summarised in Figure \ref{fig:offlineval:results:stage3} shows that Polygrid is competitive with the alternative models in multiclass and multilabel datasets but was less successful in label ranking datasets.
We remind the reader that the following analyses are made in the spirit of learning how Polygrid can be improved, rather than presenting Polygrid as a general alternative to models that are traditionally employed to solve multilabel and label ranking tasks.

\subsubsection{The Performance of Polygrid on Multiclass Datasets}
\label{sec:offlineval:discussion:multiclass}
The performance of the Polygrid model has dominated the alternative models in multiclass datasets, with a few situations in which some alternative model shows comparable performance across multiple metrics.
For example, BRDT ties with Polygrid on the ampiab dataset across all metrics except (subset) accuracy and jaccsim, and the Linear and Ridge models tie with Polygrid on the elsio1 dataset on several metrics.
So, we are interested in clarifying why Polygrid performed so well.

Recall from Section \ref{sec:offlineval:datasets} that a common application of standardised instruments in gerontological studies regards their use to stratify the target population with respect to some risk.
This is why the multiclass datasets were included in our collection.
The synthetic assignments created for these datasets reflect decision boundaries based on sum-scores.
For example, the cutoff proposed by \citeonline{silva2014whoqolcut} for the whoqol dataset is based on the sum-score: $\sum_j \mathring{x}_{ij} \geq 60$.
Similar statements can be made to the ampiab dataset \cite{marcucci2020ampiab} and, to some extent, to the elsio1 dataset \cite{aliberti2022ic}.
Our hypothesis is that Polygrid is better suited to express decision boundaries with such a structure because of the monotonic relationship between sum-scores and area-scores described in Section \ref{section:proposal:interpretable:defence}.
This relationship is stronger in the whoqol and elsio1 datasets than in the ampiab dataset (see Annex \ref{appendix:areascore}).

We present the following evidence to support this hypothesis.
Polygrid attains top accuracy on the whoqol dataset with config 166, which specifies a small model ($nspd=1, na=2, \mbox{ with } 18 \mbox{ weights}$) based on the ridge solver.
In contrast, its top accuracy on the elsio1 dataset, $0.798 \in [0.792, 0.803]$, is achieved by a much larger model specified by config 2995 ($nspd=3, na=5, \mbox{ with } 450 \mbox{ weights}$), which is based on the lstsqsym solver\footnote{The lstsqsym solver is identical to the lstsq solver, except for a change in how assignments are encoded: $y_{ij} \in \{-1, 1\}$ instead of $y_{ij} \in \{0, 1\}$.
This decision may be consequential for regression-like models for multilabel classification \cite{jia2024regression}.}.
However, a smaller model based on the ridge solver shows a comparable performance: config 46 ($nspd=1, na=1, \mbox{ with } 36 \mbox{ weights}$) achieves $0.794	\in [0.79, 0.799]$.
Recall that the ridge solver includes an intercept parameter, which may play the role of a constant in the decision boundaries such as $\sum_j \mathring{x}_{ij} \geq C$.
As anticipated, a similar analysis fails for the ampiab dataset, on which Polygrid achieves its top accuracy of $0.73 \in [0.718, 0.742]$ with config 1952, ($nspd=2, na=6, \mbox{ with } 180 \mbox{ weights}$) based on the lstsq solver.
The highest accuracy achieved with the ridge solver is $0.697 \in [0.687, 0.706]$ (config 3239, $nspd=3, na=7, \mbox{ with } 318 \mbox{ weights}$).
This is a considerably larger model ($318 > 180$ weights) with a lower performance.

\subsubsection{The Performance of Polygrid on Multilabel Datasets}
\label{sec:offlineval:discussion:multilabel}

The Polygrid model obtained mixed results on multilabel datasets.
More specifically, Polygrid has dominated the alternative models on both the whoqol datasets (whoqol-ml-11 and whoqol-ml-22) and on the ampiab-ml-11 dataset.
However, Polygrid was consistently dominated by MLP, DT, and BRDT on both elsio1 datasets (elsio1-ml-11 and elsio1-ml-22) and on the ampiab-ml-22 dataset.
So, we are now interested in clarifying what the alternative models did differently from Polygrid to grant them higher performance.

As shown in Figure \ref{fig:proposal:fcg:nn-inst}, the architecture of the MLP model used in this evaluation has a single hidden layer with sigmoid activation.
The number of neurons in the hidden layer, $h$, is calculated so that the size of the MLP instance matches that of the competing Polygrid instance (when possible).
The output of a neuron in the hidden layer is the inner product between the input vector $X_i$ and a weight vector associated with that neuron, squashed by the activation function.
This means that the hidden layer projects the input vector $X_i$ into a feature space embedded in $\mathbb{R}^{h}$.
Regarding the output layer, it has $n$ neurons (one for each label), and each represents an inner product between a feature vector in $\mathbb{R}^{h}$ and a weight vector, generating an output space embedded in $\mathbb{R}^n$.

A forward computational graph of the Polygrid model is illustrated in Figure \ref{fig:offline:fcg:ply}.
Its output layer is similar to that of the MLP model in that it performs an inner product between a feature vector (i.e., the output of the \aspas{annular sectors} layer) and the weights of the output layer.
However, the models employ different feature extraction methods.
For example, with $d=4, n=2$, an MLP instance with $h=2$ hidden neurons takes up 16 weights, which is exactly the size of the Polygrid instance shown in Figure \ref{fig:offline:fcg:ply}.
However, the MLP instance extracts feature vectors with $h=2$ dimensions, whereas Polygrid extracts vectors with $n_{as}=8$ dimensions.
Moreover, MLP uses learned weights for feature extraction, while Polygrid's feature extraction method depends solely on the partition of the disc.

To test the first hypothesis, that the observed difference in performance is due to the difference in the dimension of the feature spaces, we compared Polygrid's performance in feature spaces with nearly the same dimensionality:
\begin{itemize}

    \item In the elsio1-ml-11 dataset, Polygrid's highest accuracy was achieved with config 3429.
    The competing MLP instance had 77 neurons in the hidden layer.
    A Polygrid instance with config 2979 ($n_s/d=3, n_a=5)$ extracts feature vectors with 74 dimensions, and with config 2993 ($n_s/d=2, n_a=8)$ extracts vectors with 80 dimensions.
    The accuracy of these instances is much lower than that of the competing MLP instance.

    \item In the elsio1-ml-22 dataset, Polygrid's highest accuracy was achieved with config 2275.
    The competing MLP instance had 63 neurons in the hidden layer.
    A Polygrid instance with config 1995 ($n_s/d=2, n_a=6)$ extracts feature vectors with 60 dimensions, as does an instance with config 2835 ($n_s/d=3, n_a=4)$.
    Again, the accuracy of these instances is much lower than that of the competing MLP instance.

    \item In the ampiab-ml-22 dataset, Polygrid with config 3193 achieves the highest accuracy.
    The competing MLP instance had 81 neurons.
    A Polygrid instance with config 2201 ($n_s/d=2, n_a=8)$ extracts feature vectors with 80 dimensions.
    The accuracy of this instance is much lower than that of the competing MLP instance.
    
\end{itemize}

\noindent
Considering these results, we dismiss the first hypothesis.
The second hypothesis, that the difference in performance should be ascribed to the structure of MLP's feature extraction (inner products with learned weights), remains.
If this is the case, \future{an adaptive version of the Polygrid's feature extractor is promising.
By adaptive we refer to the ability to use both assessment and assignment data to learn optimal boundaries of the annuli (or the sectors).
However, the argument for Polygrid's interpretability relies on the way in which the explanation diagram visually encodes the forward computational graph and the model`s architecture preserves the meaning of the input data, so any adaptations should consider these aspects.}{make Polygrid's feature extractor adaptive}

With respect to the DT model, it obtained the highest accuracy in the three datasets in which Polygrid was dominated by some alternative model.
Basically, the DT model learns how to partition the input space into hypercubes and assign each to a target label.
For this purpose, the model uses both the assessment and assignment data.
The boundaries of these hypercubes are specified by multiple thresholds defined along each dimension of the input space.

This characteristic is interesting because it resembles the way Polygrid's partitions the input space.
The boundaries learned by a decision tree can be used to specify the annuli of a Polygrid instance, as shown in Figure \ref{fig:proposal:lr-task:best}.
In that example, only the top $(n_{a}-1)$ boundaries were used.
We explored this idea further with the elsio1-ml-11 dataset.
A Polygrid instance with config 3429 achieved the highest accuracy in this dataset.
Its competing DT instance learned 43 distinct boundaries.
The average distance between two such boundaries was about 0.02, and the minimum distance was about $0.0002$.
The average distance would require a Polygrid instance with 50 annuli, and the minimum distance implies an instance whose diagram cannot be inspected by the user.
We rounded the boundaries to two decimals and constrained the minimal difference to 0.02, resulting in about 24 annuli.
The results showed that an instance with these boundaries attained modest improvement when compared to the Polygrid for config 3429, and was still dominated by DT.
A similar analysis applied to the elsio1-ml-22 and ampiab-ml-22 datasets lead to comparable results.
\future{Our conclusion is that this idea is promising, but the challenge is to learn to partition the input space that combines both input and target data with no sacrifice to interpretability or scrutability.}{improving the partitioning of the input space}

Finally, the third model that dominated Polygrid's performance on the three datasets is the BRDT model, which fits a DT model for each label separately.
At first sight, since each DT would possibly partition the input space differently, one may consider two paths to adapt ideas from this model: (a) to allow for each label to have its own partitioning, or (b) to create a single partition that combines all separate partitions.
The second one seems more promising, since the first alternative would disrupt the interaction of the user with the Polygrid's diagram.
However, it must be said that the DT model, which fits a single decision tree, is simpler and achieved higher accuracy.
Thus, it seems reasonable to prioritise the ideas for improvement coming from the DT model.

\subsubsection{The Performance of Polygrid on Label Ranking Datasets}
\label{sec:offlineval:discussion:label-ranking}

The Polygrid model was dominated on label ranking datasets.
Exceptions are the whoqol-lr-11 dataset, in which Polygrid is dominant, and elsio1-lr-11 and whoqol-lr-22 datasets on the Kendall's tau metric.
The alternative models that dominated most scenarios are the same as in the multilabel classification: MLP, DT and BRDT, with the DT model standing out on datasets with 22 labels.
Consequently, all recommendations for improving Polygrid to perform multilabel classification would probably also benefit this task.

However, a closer analysis of the results shows a large increase in the number of false negatives and a moderate decrease in false positives.
This analysis was based on a comparison of the Polygrid with config 2757 on the whoqol-ml-22 and the whoqol-lr-22 datasets, controlling for the train/test split.
These datasets were chosen because Polygrid dominated the alternative models on the whoqol-ml-22 dataset, but performed poorly on the whoqol-lr-22 dataset.
More specifically, a case-by-case comparison between the results of the Polygrid model on these datasets shows a large increase in false negatives: from 7 (multilabel) to 22 (label ranking).
Moreover, this increase in false negatives is accompanied by a moderate decrease in false positives: from 28 (multilabel) to 20 (label ranking).

Based on this analysis, we conclude that, when moving from a multilabel classification to a label ranking setting, the Polygrid model may have \aspas{lost} some of its ability to assign proper labels to cases.
\future{We believe that this increase in error may be due to the modifications in Algorithm \ref{alg:learning:ml:step5} to adapt it to the label ranking tasks.}{improve longrank}
More precisely, the introduction of the membership matrix $U$, which is calculated using Equation \ref{eq:proposal:logranks}, probably contributed to this negative effect.
Thus, our recommendation is to focus on improving Algorithm \ref{alg:learning:ml:step5} to provide a better adaptation to label ranking tasks.

\section{A User Study to Assess the Interpretability of the Polygrid Diagram}
\label{section:userstudy}

The conceptual framework introduced in Section \ref{section:proposal:interpretable} defines interpretability as a property that emerges from the interaction among data, model, and users.
According to that framework, the contribution that a model brings to interpretability comes from its transparency, scalability, and ability to preserve meaning.
Applied to Polygrid, we concluded that the model has these properties, which makes it conducive to interpretability.
In fact, the defence of the Polygrid's interpretability presented in that section argued that the forward computational graph of a recommendation is visually encoded in its respective explanation diagram, and that the model's architecture allows the elements of the computational graph to inherit the meaning of the input data.

In this section, we complement that conceptual defence with an empirical evaluation of the model's interpretability.
We report on the results of a user study that combines methodological elements from the literature on the interpretability of machine learning models and visualisation research \cite{doshivelez2017rigorous,saket2019task}: the participant is shown an explanation diagram in which the assessment chart displays measurements taken from a flower specimen.
Of course, the explanation diagram is stripped of its tags.
The participant is asked to classify that specimen into one of three species.

\subsection{The Design of the Interpretability Assessment User Study}
\label{sec:userstudy}

\begin{figure}[htpb]
    \centering
    \begin{subfigure}[b]{\textwidth}
        \centering
        \includegraphics[width=0.927\textwidth]{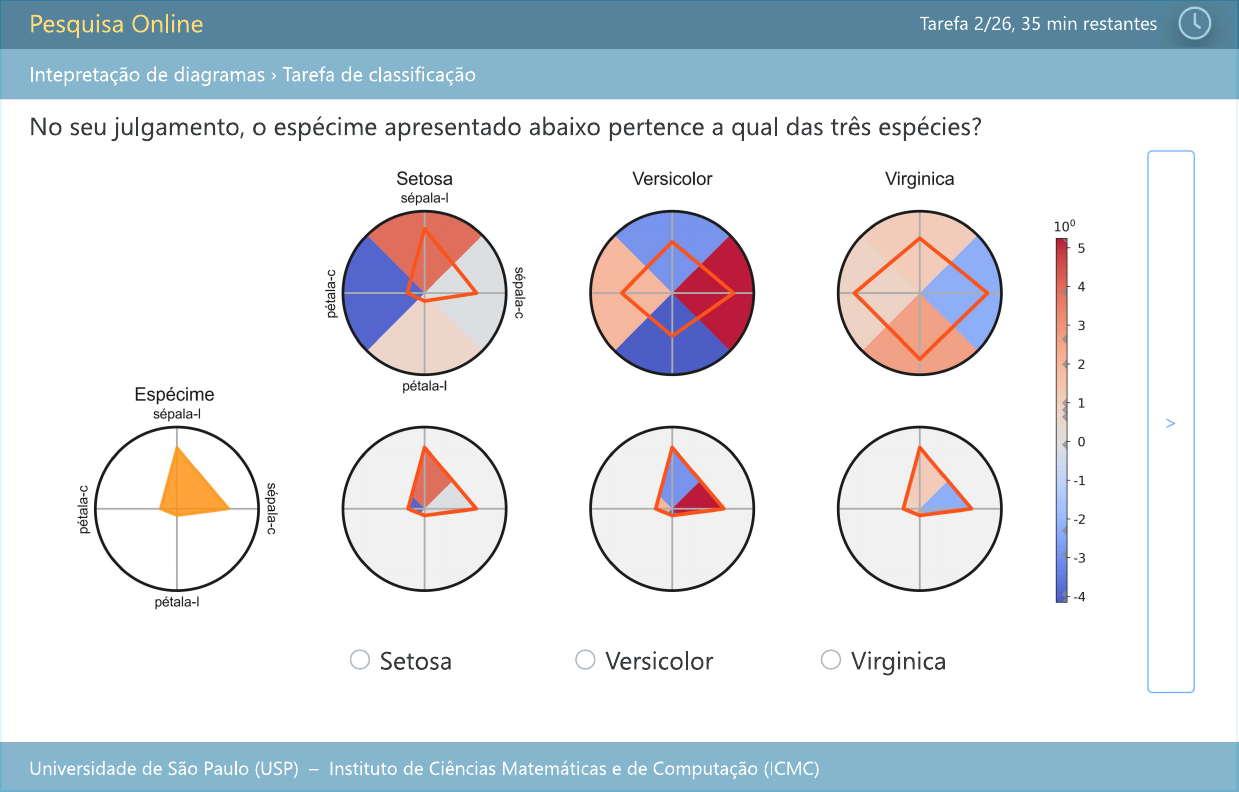}
        \caption{Visual classification task in the Polygrid condition}
        \Description{}
        \label{fig:userstudy:polygrid-condition}
    \end{subfigure}
    \begin{subfigure}[b]{\textwidth}
        \centering
        \includegraphics[width=0.927\textwidth]{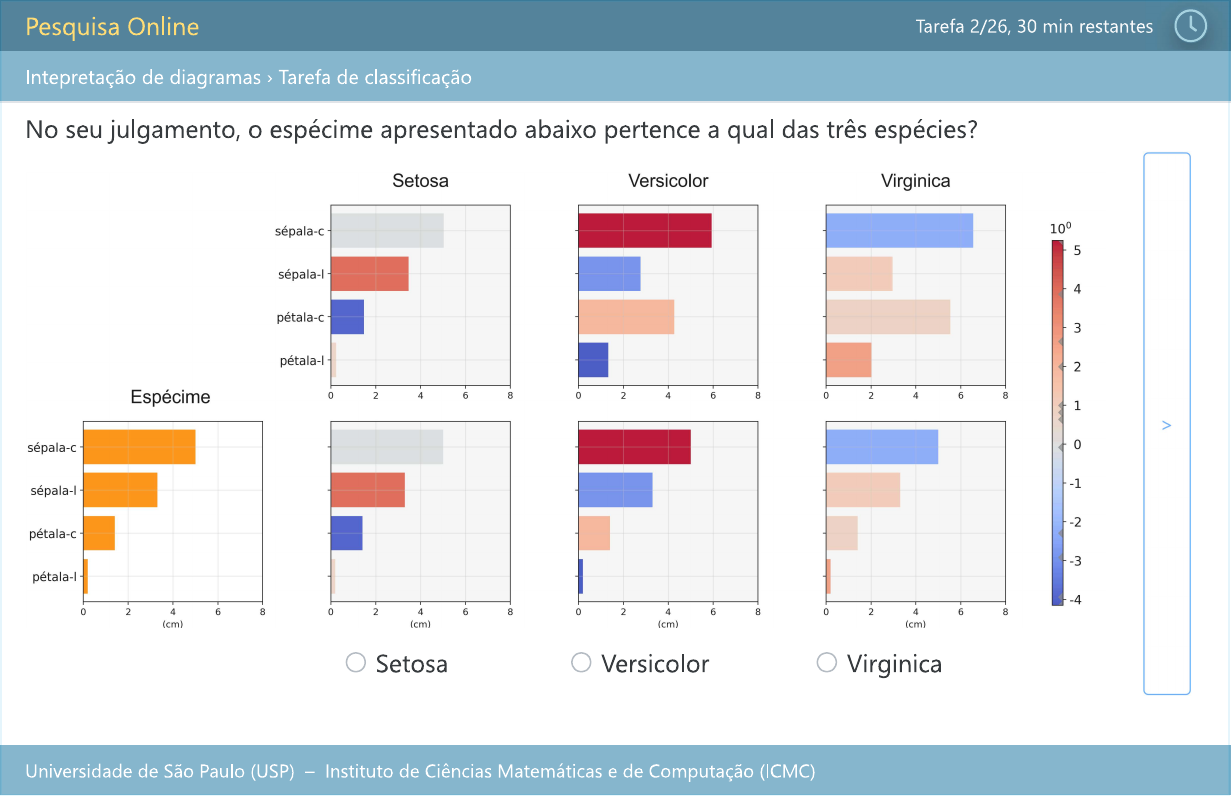}
        \caption{Visual classification task in the Barsgrid condition}
        \Description{}
        \label{fig:userstudy:barsgrid-condition}
    \end{subfigure}
    \caption{Screenshots of webpages for the visual classification task under the Polygrid and  Barsgrid conditions. 
    Both diagrams display the case 49 of the Iris dataset.
    The textual elements are presented in Portuguese, and the main instruction reads \aspas{In your opinion, the specimen shown below belongs to which of the three species?}
    Before landing on this page, the participant watches a video with instructions on how to perform the task, including how to read the diagrams, record their judgement, and move to the next task.
    }
    \label{fig:userstudy:screenshots}
\end{figure}

This study follows a within-subjects design in which each participant is exposed to two experimental conditions.
In one condition, which we call \aspas{Polygrid condition}, the participant is asked to perform a series of visual classification tasks using Polygrid diagrams, such as the one shown in Figure \ref{fig:userstudy:polygrid-condition}.
The diagrams were generated by training a Polygrid instance on the Iris dataset \cite{unwin2021iris}.
The task consists in classifying a specimen into one of the three species: Setosa, Versicolor, or Virginica.
In the other condition, which we call \aspas{Barsgrid condition}, the participant is asked to perform the same task, but using Barsgrid diagrams, which are identical to Polygrid's except for the radar charts being replaced by bar charts, as shown in Figure \ref{fig:userstudy:barsgrid-condition}.
In addition to the participant's response, we collect the time spent to complete each visual classification task.

The collected data allow us to assess the interpretability of Polygrid diagrams by means of a proxy measure based on the definition adopted in Section \ref{section:proposal:interpretable:model}: interpretability is the measure of success participants achieve when asked to perform forward simulation tasks in an experimental context.
Moreover, the data also allow us to assess how Polygrid performs in comparison to the use of bar charts to support visual classification tasks.
In the remainder of this section, we detail how the participant's journey is organised and how the collected data is used to assess the variables that are relevant to this study.

\subsubsection{Experimental Conditions}
\label{sec:userstudy:tasks}

How can we assess the interpretability of Polygrid diagrams by means of a user study in which the participant is asked to perform visual classification tasks?
As said before, we adopted \standing{an operational definition of interpretability as the measure of success participants achieve when asked to perform forward simulation tasks in an experimental context.}{operationalised interpretability}
In this sense, a forward simulation task is a task in which \aspas{humans are presented with an explanation and an input, and must correctly simulate the model's output} \cite{doshivelez2017rigorous}.
In a Polygrid diagram, the assessment chart represents the input, and the explanation is provided by the assignment and matching charts.
In this setting, success depends on the participant's ability to assess the similarity of two visual shapes: the assessment polygon and the class prototypes.
Thus, we take the participant's accuracy in reproducing the Polygrid's output as a proxy measure for its interpretability.

However, this method of evaluating interpretability creates the need to make some adaptations to the Polygrid diagram.
The reader may have noticed that, in Figure \ref{fig:userstudy:polygrid-condition}, the Polygrid diagram was stripped of an element that appears in the diagrams shown in earlier sections: all tags were removed, including the tags informing the degree of similarity between the input and the labels.
The tags were removed because they carry information about the output of the model, which is precisely what the participant must replicate.
Moreover, the interactivity of the Polygrid diagram was disabled, so the participant could not inspect the value assigned to the elements of the diagram.

It is important to highlight that these adaptations preserve an essential feature of the Polygrid diagram: the use of polygons to represent multidimensional data.
In principle, this representation drives a person to resort to their visual intuition (of shape sorting) instead of their analytical abilities (of comparing numerical scores).
Actually, this choice of representation aims to nudge the participant's attention to focus on a holistic view of the latent variable represented by the area of the assessment polygon, rather than on the individual domain scores.
The decision to include the Barsgrid condition in the study design was mainly driven by the need to investigate the presumed impact of this choice of visual representation on the participant's accuracy in solving the tasks.

The Barsgrid diagram was designed to be as similar to Polygrid's as possible.
The substitution of radar charts for bar charts was based on results from the literature on the task-based efficiency of elementary visualisations: empirical evidence suggests that bar charts are among the best basic visualisations to support visual classification tasks\footnote{Based on the taxonomy of \citeonline{amar2005taxonomy}, this task can be described as a combination of filter and sort basic analytic tasks.}.
Moreover, the bar chart naturally drives the participant's attention to focus on domain scores individually.
This creates the contrast needed to estimate the effect of using polygons to represent multidimensional data. 
Furthermore, the bar chart gracefully accommodates a numerical scale that is shared by all domains.
This feature helps us investigate the effect of the absence of numerical scales in Polygrid diagrams, which was pointed out as a deficiency during a pilot study.

Finally, the Iris dataset was chosen because (a) all of its features are measured in the same scale (cm), which satisfies our need to introduce a single numerical scale in the Barsgrid diagram, and (b) it describes generally familiar objects (flowers), which eliminates the need to recruit specialists, as would be the case had we chosen a healthcare dataset.
We trained a Polygrid$(1,1)$ instance with \emph{sector = cover} so that annular sectors can be mapped to bars in the corresponding charts.
This simplifies the video instructions, which guides the participant to focus on the similarity of shapes\footnote{You can watch the instructional videos and try the visual classification tasks in \href{https://marjory.inference.app.br/os2boek/}{this website}.
}.

\subsubsection{Methodology}
\label{sec:userstudy:methodology}

\begin{figure}[t]
    \centering
    \includegraphics[width=1\linewidth]{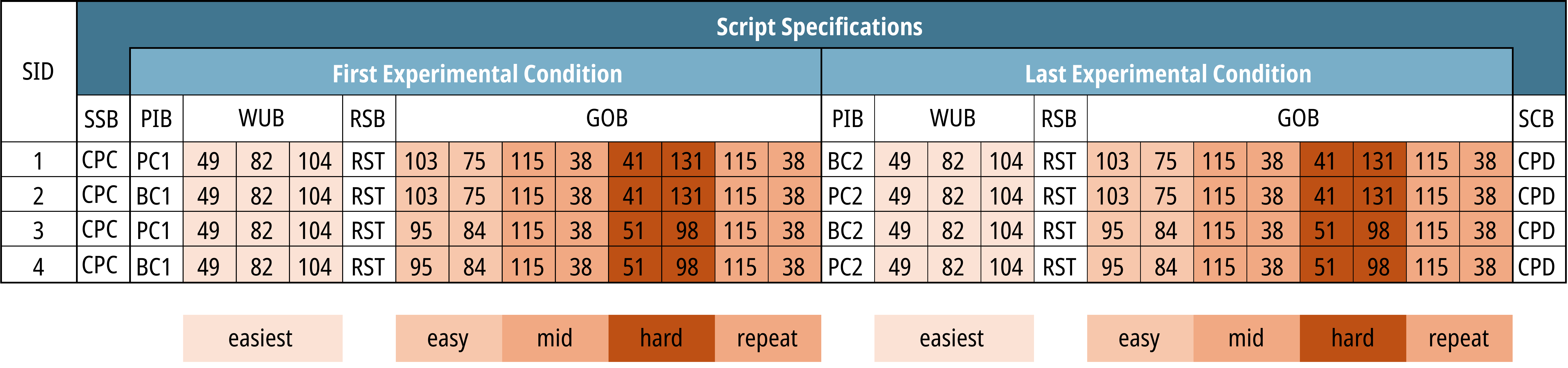}
    \caption{How the participant's journey is specified by a script.
    Each script specifies the journey of a participant during the study.
    A script is identified by a unique SID and is composed of blocks.
    Blocks are sequence of tasks.
    For example, the script with SID$\,=1$ describes a journey that starts with the participant being presented with information about the study, and asked to confirm their consent (task CPC).
    Upon consenting, they watch a video with instructions for the first set of visual classification tasks (PC1 or BC1, depending on the experimental condition to which the participant is exposed first).
    Next, he or she is tasked with three warm-up visual classification tasks, which aim to assert understanding (tasks in the WUB block, numbers identify the dataset instance).
    The participant is shown a positive reinforcement message (task RST) and tasked with eight new visual classification tasks (in the GOB block).
    Once completed, the entire process is then repeated for the remaining experimental condition.
    In the end, the participant is invited to provide contact and demographic information (task CPD).
    \textbf{Blocks}: script starting block (SSB), participant instruction block (PIB), warm-up block (WUB), ready-set block (RSB), go block (GOB), and script closing block (SCB).
    \textbf{Tasks}: collect participant's consent (CPC), collect participant's demographics (CPD), video instruction for Polygrid Condition First (PC1), video instruction for Barsgrid Condition First (BC1), video instruction for Polygrid Condition Last (PC2), video instruction for Barsgrid Condition Last (BC2), and ready-set (RST).
    }
    \Description{}
    \label{fig:userstudy:script-schedule}
\end{figure}

The journey of the participant during the study is described by a script.
A script is made up of blocks, and a block is a sequence of tasks, as illustrated in Figure \ref{fig:userstudy:script-schedule}.
The script starts by presenting information about the study being conducted, such as motivation, objectives, methods employed, and how the collected data will be managed and used.
If the participant gives their consent, then he or she is redirected to a video that explains what is expected from the participant and gives instructions on how to perform the visual classification task corresponding to one of the experimental conditions (e.g., Polygrid condition).
Once the instruction task is completed, the participant is asked to complete three visual classification tasks.
These are warm-up tasks, and their objective is to verify that the participant understands how to perform the task explained in the video.
After completing the warm-up tasks, the participant is shown a positive reinforcement message and is informed that he or she will be asked to complete eight new visual classification tasks that may be more difficult to solve than the ones just completed.
Upon completion, the participant is submitted once again to the same sequence of tasks (starting from the video instruction), but now the tasks are specialised for the remaining experimental condition (e.g., Barsgrid condition).
Finally, the script ends with a closing task, in which the participant is invited to provide contact and demographic information and is informed of the procedure to withdraw their data from the study, as required by local regulations\footnote{Item IV.3.d of the Resolução 466/2012 of the Conselho Nacional de Saúde - Brazil.}.

Before starting recruitment, 100 such scripts were generated and uploaded to the \href{https://marjory.inference.app.br/}{Marjory website}, which can manage scripts in a format equivalent to that of Figure \ref{fig:userstudy:script-schedule}.
The scripts were generated by training a Polygrid$(1,1)$ instance on the Iris dataset split into 120 cases for training and 30 cases for testing.
Only test cases were used to specify visual classification tasks.
These cases were sorted according to their distance from the respective class prototype\footnote{For example, since Case 49 is labelled as Setosa, its distance is computed from the prototype that appears in the assignment chart for the Setosa label (consult Figure \ref{fig:proposal:ml-task:default} as reference).}.
The three cases closest to each prototype were reserved for warm-up tasks (cases 49, 82, and 104).
Of the remaining 27 cases, the first 13 closest to their respective prototype were assigned to the \aspas{left} pool, the 12 cases farther away from their prototype were allocated to the \aspas{right} pool, and the remaining two were reserved for mid and repeat tasks (cases 115 and 38).
The cases in the left pool were randomly drawn to specify easy tasks, and the cases in the right pool were drawn to specify hard tasks.

Participants were invited by email containing a link to the website hosting the study.
When a participant clicks on the link, the next free script in the study is allocated.
To reduce order effects, scripts with an odd identifier (SID) specify the Polygrid condition as the first experimental condition to which the participant is exposed, and the Barsgrid condition as the last experimental condition.
In contrast, scripts with an even identifier specify the opposite order.
This is to say that some participants will start solving visual classification tasks using Polygrid diagrams, and others will start solving tasks using Barsgrid diagrams.
In fact, the script with SID$\,=2n$ with $ (n=1 \ldots 50)$ is a copy of the script with SID$\,=2n-1$ in which the order of the experimental conditions is reversed.
Finally, the number of tasks (26 tasks) was chosen so that the time required to complete a script did not exceed 15 minutes, to avoid the possible effects of visual or cognitive fatigue.

\subsubsection{Data Analysis}
\label{sec:userstudy:analysis}

The core variables of this study are the mean accuracy and the mean completion time.
These variables are aggregated over distinct subsets of the collected data to help us assess the assumptions and relevant hypotheses of the study.
These subsets correspond to data from the different blocks and conditions illustrated in Figure \ref{fig:userstudy:script-schedule}.
For example, once the data are downloaded from the website hosting the study, the first step is to apply a set of exclusion criteria: scripts that were not completed are discarded, as well as completed scripts in which the participant failed the warm-up tasks.
This ensures that the remaining data comes from participants who understood how to perform the tasks.
Next, two other assumptions of the study design are evaluated.
First, we seek evidence that the tasks are indeed arranged from easier to harder cases.
This verification is important for two reasons: (a) the cases in the warm-up block were selected under the assumption that we have the ability to identify the easiest cases to classify, and (b) the cases drawn to assemble the go block are balanced with respect to difficulty, which supports the claim that the mean accuracy is meaningful.
Second, we look for signs that participants experienced fatigue during the experiment.
The effects of fatigue can confound the study variables, and caution is advised when it is detected.

Finally, the proxy measure for interpretability and the relevant hypotheses are assessed.
The latter aim to clarify whether the Barsgrid diagram is more effective than the Polygrid diagram in supporting the tasks in our experiment.
The effectiveness is approached from multiple angles, looking not only for evidence from accuracy and completion time but also from the consistency and consensus of the group of participants.
The consistency measure evaluates whether an individual participant produces the same response when asked to classify the same case a second time.
This measure is computed using the responses for the mid and repeat cases in the go block (cases 115 and 38).
The consensus measure evaluates the degree of agreement among all participants with respect to their responses to the same cases.
It is calculated based on the responses given to cases in the easy, mid, and hard tasks that appear in the go block.
We formalise consistency and consensus as the Cohen's $\kappa$ statistic for intra- and inter-agreement for categorical data \cite{stemler2004comparison}:
\begin{equation}
    \label{eq:userstudy:kappa}
    \kappa(Y_a, Y_b) = \frac{\mathbb{E} \, \ivbracket{y_{aj}=y_{bj}} - \mathbb{E} \, \ivbracket{u_{1j}=u_{2j}}}{1- \mathbb{E} \, \ivbracket{u_{1j}=u_{2j}}} = \frac{p_o - p_e}{1 - p_e}, 
\end{equation}
where $y_{aj}$ is the label assigned to case $j$ by the annotator A, $u_{ij}$ is a label randomly assigned to case $j$, and $\ivbracket{\cdot}$ is the Iverson bracket.
Thus, $p_o$ is the observed agreement between the annotators and $p_e$ is the expected agreement if the labels were assigned randomly.

\subsection{Results}
\label{sec:userstudy:results}

Individuals attending regular courses at the Universidade de São Paulo (USP), Universidade Federal de São Carlos (UFSCar), and Universidade Federal de Ciências da Saúde de Porto Alegre (UFCSPA) were invited by email.
During the period from March 21 to May 28, 2024, a total of $N=56$ people participated in our study.
This group is composed of 40 men, 9 women, plus 7 individuals who chose not to inform their sex; they were aged 17 to 55 years (24 on average); there were 37 undergraduates and 13 graduate students, of varied backgrounds, mainly STEM, applied social sciences, and gerontology.

Of the 56 scripts completed by the participants, ten were discarded because the participant failed (at least one of) the warm-up tasks.
Of the ten discarded scripts, six are Polygrid first and four are Barsgrid first.
Although a relationship between failure to pass the warm-up tasks and the experimental condition seems implausible ($\chi^2(1)=0.68, \, p = 0.41$), the device used by the participant seems to provide a better explanation.
In six of the ten scripts discarded, the participant used a mobile device to access the study website, against the recommendation given in the invitation.
Considering that the proportion of participants who used a mobile device was about 14\% (8 of the 56 participants), the relationship seems likely ($\chi^2(1)=6.57, \, p < 0.015$).
It should be added that the proportion of men and women in the study did not change after the exclusion criteria were applied.

Of the 46 scripts accepted, 24 are paired (i.e., pairs whose SID are $2n$ and $2n-1$) and 22 are single.
Of the latter, 9 are Polygrid first, and 13 are Barsgrid first.
\standing{As will soon be shown, the presence of single scripts and the small imbalance between the study conditions do not threaten the design or the results of the study.}{negligible imbalance between experimental conditions}
Each case from the test partition appears in at least one script, and the number of occurrences by label is closely balanced (332 Setosa, 322 Versicolor, and 358 Virginica).
The results of the data analysis are summarised in Table \ref{tab:userstudy:results:hypotheses}, in which the relevant hypotheses are evaluated at various levels of confidence.
However, before we discuss whether the main hypotheses are supported by the data, the status of two design assumptions must be clarified.

First, let's approach the idea that the tasks are arranged in increasing order of difficulty within the scripts.
The tests whose results are reported in row \aspas{A2a} of Table \ref{tab:userstudy:results:assumptions:a2} are based on two confidence intervals: one around the mean accuracy in easy tasks ($\mu_{acc}^{easy}$), and the other around the mean accuracy in hard tasks ($\mu_{acc}^{hard}$).
To reduce a possible confounding with fatigue, only data from tasks in the first experimental condition are considered.
When the intervals do not overlap, the inequality encoded in the statement of the assumption is evaluated ($\mu_{acc}^{easy} > \mu_{acc}^{hard}$), otherwise the test is declared inconclusive.
At the 80\% confidence level ($\alpha = 0.20$), the statement is evaluated as true ($\mu_{acc}^{easy} = 0.957 \in [0.935, 0.989]$ and $\mu_{acc}^{hard} = 0.804 \in [0.750, 0.859]$, with 92 samples).
However, this result masks a deeper contrast: constrained to the Barsgrid tasks, the statement yields true, but it is inconclusive when constrained to the Polygrid tasks.
\future{This contrast suggests that the scale of difficulty used by the mechanism that assembles the scripts produced clearly noticeable effects on participants solving the Barsgrid tasks, but less noticeable when solving the Polygrid tasks.}{we need to improve the notion of difficulty used to assemble the scripts.}
Moreover, it seems reasonable to assume that participants would take less time to complete easy tasks compared to hard ones.
This assumption is not backed by the completion time data from neither condition, as indicated in row \aspas{A2t}: $\mu_{time}^{easy} = 24.0 \in [21.3, 26.6]$ sec and $\mu_{time}^{hard} = 17.6 \in [15.5, 19.7]$ sec, with 88 and 87 samples respectively.
Our hypothesis is that this decrease is due to learning effects, as participants become more productive as they practice. 

Second, let's seek for signs that the participants experienced fatigue during the experiment.
\standing{When feeling fatigued, participants are less prone to meet cognitive demands that are perceived as low-stakes, as is usually the case in non-paid, voluntary participation in online studies.
In principle, the larger the gap between demanded and volunteered efforts widens, the lower the accuracy and the less time spent on the tasks.}{impact of fatigue on time and accuracy}
\standing{In this case, a contrast between the participant's performance during the first and the last half of the study could unveil the effect of fatigue if it is strong enough.}{empirical test of fatigue}
Moreover, we can constrain this analysis to the data from paired scripts to control for difficulty.
This ensures that the same sequence of tasks appears as the first and last condition in a pair of scripts, as shown in scripts 1 and 2 in Figure \ref{fig:userstudy:script-schedule}.
The remaining confounder, individual differences in ability between participants assigned to the Polygrid first or Barsgrid first scripts, seems to be implausible ($\chi^2(1)=0.07, p=0.79$).
As shown in Table \ref{tab:userstudy:results:assumptions:a3}, the results in row \aspas{A3a} suggest that there is no systematic decrease in accuracy between the first and last experimental conditions ($\mu_{acc}^{first} = 0.917 \in [0.889, 0.944]$ and $\mu_{acc}^{last} = 0.875 \in [0.840, 0.910]$, with 144 samples).
However, constrained to Barsgrid tasks, a systematic decrease is observed at the 80\% confidence level, which it is not accompanied by a decrease in completion time that would be expected in the event of fatigue, as reported in the row \aspas{A3t} for Barsgrid tasks.
Our hypothesis is that this decrease is due to a momentary resistance to meet a cognitive demand that is perceived as larger than previously experienced: participants who start with the Polygrid tasks are likely to experience an increase in cognitive demand when they move on to solve the Barsgrid tasks.
Based on this analysis, we assume that the two design assumptions we singled out have been reasonably secured.

\begin{table}[!ht]
    \centering
    \footnotesize
    \caption{Design assumptions evaluated on data from the accepted scripts - task difficulty}
    \begin{tabularx}{\linewidth}{@{}P{0.5cm}p{7.0cm}ZZZZP{1.7cm}@{}}
    \toprule
        \multirow{2}{\hsize}{} & \multirow{2}{\hsize}{} & \multicolumn{4}{c}{Confidence Level ($\alpha$)} & Estimate \\         
        \cmidrule(lr){3-6}
        ID & Statement of the Assumption & $0.05$ & $0.10$ & $0.15$ & $0.20$ & (at $\alpha=0.20$) \\
    \toprule
 
        \multirow{4}{\hsize}{A2a} & \makecell[l]{Tasks are arranged in increasing order of difficulty \\ (evidence from accuracy, only 1st condition)} & True & True & True  & True  & 92 samples \\ 

            \cmidrule(lr){3-7} 
            & \begin{tikzpicture}[x=1pt, y=1pt, line width=0.1] 
               \fill[gray, opacity=0.3] (0,0)   rectangle ++(200,4); 
               \fill[red, opacity= .4] ( 184, 0) rectangle ++( 16, 4);
               \fill[red, opacity= .8] ( 187, 0) rectangle ++( 11, 4);
              \end{tikzpicture}
                         & \multicolumn{4}{r}{mean accuracy in easy tasks} & $.935 \quad .989$ \\ 
            & \begin{tikzpicture}[x=1pt, y=1pt, line width=0.1] 
               \fill[gray, opacity=0.3] (0,0)   rectangle ++(200,4); 
               \fill[red, opacity= .4] ( 145, 0) rectangle ++( 33, 4);	
               \fill[red, opacity= .8] ( 150, 0) rectangle ++( 22, 4);
              \end{tikzpicture}         
                         & \multicolumn{4}{r}{              in hard tasks} & $.750 \quad .859$ \\

            \cmidrule(lr){2-7} 
            & \makecell[r]{--- just Barsgrid tasks} & True & True & True  & True & 50 samples \\ 
            
            \cmidrule(lr){3-7} 
            & \begin{tikzpicture}[x=1pt, y=1pt, line width=0.1] 
               \fill[gray, opacity=0.3] (0,0)   rectangle ++(200,4); 
               \fill[red] ( 200, 2) circle (1.5);
              \end{tikzpicture}
                         & \multicolumn{4}{r}{mean accuracy in easy tasks} & $1.000$ \\ 
            & \begin{tikzpicture}[x=1pt, y=1pt, line width=0.1] 
               \fill[gray, opacity=0.3] (0,0)   rectangle ++(200,4); 
               \fill[red, opacity= .4] ( 140, 0) rectangle ++( 44, 4);	
               \fill[red, opacity= .8] ( 144, 0) rectangle ++( 32, 4);
              \end{tikzpicture}         
                         & \multicolumn{4}{r}{              in hard tasks} & $.720 \quad .880$ \\

            \cmidrule(lr){2-7} 
            & \makecell[r]{--- just Polygrid tasks} & Inc & Inc & Inc  & Inc  & 42 samples \\ 
            
            \cmidrule(lr){3-7} 
            & \begin{tikzpicture}[x=1pt, y=1pt, line width=0.1] 
               \fill[gray, opacity=0.3] (0,0)   rectangle ++(200,4); 
               \fill[blue, opacity= .4] ( 166, 0) rectangle ++( 34, 4);	
               \fill[blue, opacity= .8] ( 171, 0) rectangle ++( 19, 4);
              \end{tikzpicture}         
                         & \multicolumn{4}{r}{mean accuracy in easy tasks} & $.857 \quad .952$ \\ 
            & \begin{tikzpicture}[x=1pt, y=1pt, line width=0.1] 
               \fill[gray, opacity=0.3] (0,0)   rectangle ++(200,4); 
               \fill[blue, opacity= .4] ( 138, 0) rectangle ++( 48, 4);	
               \fill[blue, opacity= .8] ( 147, 0) rectangle ++( 29, 4);
              \end{tikzpicture}         
                         & \multicolumn{4}{r}{              in hard tasks} & $.738 \quad .881$ \\
            
            \midrule
       
        \multirow{4}{\hsize}{A2t} & \makecell[l]{Tasks are arranged in increasing order of difficulty \\ (based on completion time, only 1st condition)} & Inc & False & False & False & \makecell[c]{88 samples \\ 87 samples}\\ 
            \cmidrule(lr){3-7} 
            & \begin{tikzpicture}[x=1pt, y=1pt, line width=0.1] 
               \fill[gray, opacity=0.3] (0,0)   rectangle ++(200,4); 
               \fill[blue, opacity= .4] ( 65, 0) rectangle ++( 27, 4);	
               \fill[red, opacity= .8] ( 70, 0) rectangle ++( 18, 4);
              \end{tikzpicture}
                         & \multicolumn{4}{r}{mean time to complete easy tasks} & $21.3 \quad 26.6$ \\ 
            & \begin{tikzpicture}[x=1pt, y=1pt, line width=0.1] 
               \fill[gray, opacity=0.3] (0,0)   rectangle ++(200,4); 
               \fill[blue, opacity= .4] ( 47, 0) rectangle ++( 22, 4);	
               \fill[red, opacity= .8] ( 51, 0) rectangle ++( 14, 4);
              \end{tikzpicture}
                         & \multicolumn{4}{r}{                      hard tasks} & $15.5 \quad 19.7$ \\
        
            \cmidrule(lr){2-7} 
            & \makecell[r]{--- just Barsgrid tasks} & Inc & Inc & Inc  & Inc & 47 samples \\ 
            
            \cmidrule(lr){3-7} 
            & \begin{tikzpicture}[x=1pt, y=1pt, line width=0.1] 
               \fill[gray, opacity=0.3] (0,0)   rectangle ++(200,4); 
               \fill[blue, opacity= .4] ( 66, 0) rectangle ++( 40, 4);	
               \fill[blue, opacity= .8] ( 74, 0) rectangle ++( 26, 4);
              \end{tikzpicture}
                         & \multicolumn{4}{r}{mean time to complete easy tasks} & $22.5 \quad 30.0$ \\ 
            & \begin{tikzpicture}[x=1pt, y=1pt, line width=0.1] 
               \fill[gray, opacity=0.3] (0,0)   rectangle ++(200,4); 
               \fill[blue, opacity= .4] ( 50, 0) rectangle ++( 34, 4);	
               \fill[blue, opacity= .8] ( 57, 0) rectangle ++( 23, 4);
              \end{tikzpicture}
                         & \multicolumn{4}{r}{                      hard tasks} & $17.2 \quad 23.8$ \\

            \cmidrule(lr){2-7} 
            & \makecell[r]{--- just Polygrid tasks} & Inc & Inc & False  & False  & 41; 40 smp \\ 
            
            \cmidrule(lr){3-7} 
            & \begin{tikzpicture}[x=1pt, y=1pt, line width=0.1] 
               \fill[gray, opacity=0.3] (0,0)   rectangle ++(200,4); 
               \fill[blue, opacity= .4] ( 52, 0) rectangle ++( 36, 4);	
               \fill[red, opacity= .8] ( 59, 0) rectangle ++( 24, 4);
              \end{tikzpicture}
                         & \multicolumn{4}{r}{mean time to complete easy tasks} & $17.8 \quad 24.7$ \\ 
            & \begin{tikzpicture}[x=1pt, y=1pt, line width=0.1] 
               \fill[gray, opacity=0.3] (0,0)   rectangle ++(200,4); 
               \fill[blue, opacity= .4] ( 36, 0) rectangle ++( 21, 4);	
               \fill[red, opacity= .8] ( 40, 0) rectangle ++( 14, 4);
              \end{tikzpicture}
                         & \multicolumn{4}{r}{                      hard tasks} & $12.2 \quad 16.3$ \\
            
       
    \bottomrule
    \end{tabularx} \\
    \noindent
    \footnotesize \justifying \emph{Legend}: In row A2a, the statement clarifies the scope of the assumption.
    To its right, the results of its evaluation at various levels of confidence are shown, followed by the number of observations in its scope.
    Two scales are plotted below the statement.
    They depict the confidence intervals (CIs) being contrasted: one around the mean accuracy for easy tasks, and the other around the mean accuracy for hard tasks.
    Description and details of each CI appear to the right of each scale.
    An opaque rectangle depicts a CI for $\alpha = 0.20$, and a transparent one depicts a CI for $\alpha = 0.05$.
    If the CIs overlap, they appear in blue, and red otherwise.
    Remaining rows are similarly organised.
    \aspas{Inc} stands for inconclusive, and \aspas{smp} for samples.
    \emph{Note: } In row A2t, we discarded 13 samples with a completion time larger than 90 seconds as outliers.
    \label{tab:userstudy:results:assumptions:a2}
\end{table}

\begin{table}[!ht]
    \centering
    \footnotesize
    \caption{Design assumptions evaluated on data from the accepted scripts - absence of fatigue}
    \begin{tabularx}{\linewidth}{@{}P{0.5cm}p{7.0cm}ZZZZP{1.7cm}@{}}
    \toprule
        \multirow{2}{\hsize}{} & \multirow{2}{\hsize}{} & \multicolumn{4}{c}{Confidence Level ($\alpha$)} & Estimate \\         
        \cmidrule(lr){3-6}
        ID & Statement of the Assumption & $0.05$ & $0.10$ & $0.15$ & $0.20$ & (at $\alpha=0.20$) \\
    \toprule
 
        \multirow{4}{\hsize}{A3a} & \makecell[l]{There is a decrease in accuracy \\ (1st vs. 2nd condition, only paired scripts)} & Inc & Inc & Inc & Inc  & 144 samples \\ 

            \cmidrule(lr){3-7} 
            & \begin{tikzpicture}[x=1pt, y=1pt, line width=0.1] 
               \fill[gray, opacity=0.3] (0,0)   rectangle ++(200,4); 
               \fill[blue, opacity= .4] ( 175, 0) rectangle ++( 18, 4);	
               \fill[blue, opacity= .8] ( 177, 0) rectangle ++( 11, 4);
              \end{tikzpicture}
                         & \multicolumn{4}{r}{mean accuracy in 1st cond.} & $.889 \quad .944$ \\ 
            & \begin{tikzpicture}[x=1pt, y=1pt, line width=0.1] 
               \fill[gray, opacity=0.3] (0,0)   rectangle ++(200,4); 
               \fill[blue, opacity= .4] ( 165, 0) rectangle ++( 21, 4);	
               \fill[blue, opacity= .8] ( 168, 0) rectangle ++( 14, 4);
              \end{tikzpicture}
                         & \multicolumn{4}{r}{              in 2nd cond.} & $.840 \quad .910$ \\

            \cmidrule(lr){2-7} 
            & \makecell[r]{--- just Barsgrid tasks} & Inc & True & True  & True & 72 samples \\ 
            
            \cmidrule(lr){3-7} 
            & \begin{tikzpicture}[x=1pt, y=1pt, line width=0.1] 
               \fill[gray, opacity=0.3] (0,0)   rectangle ++(200,4); 
               \fill[blue, opacity= .4] ( 183, 0) rectangle ++( 17, 4);	
               \fill[red, opacity= .8] ( 186, 0) rectangle ++( 11, 4);
              \end{tikzpicture}
                         & \multicolumn{4}{r}{mean accuracy in 1st cond.} & $.931 \quad .986$ \\ 
            & \begin{tikzpicture}[x=1pt, y=1pt, line width=0.1] 
               \fill[gray, opacity=0.3] (0,0)   rectangle ++(200,4); 
               \fill[blue, opacity= .4] ( 152, 0) rectangle ++( 34, 4);	
               \fill[red, opacity= .8] ( 158, 0) rectangle ++( 23, 4);
              \end{tikzpicture}
                         & \multicolumn{4}{r}{              in 2nd cond.} & $.792 \quad .903$ \\

            \cmidrule(lr){2-7} 
            & \makecell[r]{--- just Polygrid tasks} & Inc & Inc & Inc  & Inc  & 72 samples \\ 
            
            \cmidrule(lr){3-7} 
            & \begin{tikzpicture}[x=1pt, y=1pt, line width=0.1] 
               \fill[gray, opacity=0.3] (0,0)   rectangle ++(200,4); 
               \fill[blue, opacity= .4] ( 161, 0) rectangle ++( 31, 4);	
               \fill[blue, opacity= .8] ( 166, 0) rectangle ++( 20, 4);
              \end{tikzpicture}
                         & \multicolumn{4}{r}{mean accuracy in 1st cond.} & $.833 \quad .931$ \\ 
            & \begin{tikzpicture}[x=1pt, y=1pt, line width=0.1] 
               \fill[gray, opacity=0.3] (0,0)   rectangle ++(200,4); 
               \fill[blue, opacity= .4] ( 169, 0) rectangle ++( 25, 4);	
               \fill[blue, opacity= .8] ( 172, 0) rectangle ++( 17, 4);
              \end{tikzpicture}
                         & \multicolumn{4}{r}{              in 2nd cond.} & $.861 \quad .944$ \\
            
            \midrule
        
        \multirow{4}{\hsize}{A3t} & \makecell[l]{There is a decrease in completion time \\ (1st vs. 2nd condition, only paired scripts)} & Inc & True & True & True & \makecell[c]{190 samples \\ 192 samples} \\ 
            \cmidrule(lr){3-7} 
            & \begin{tikzpicture}[x=1pt, y=1pt, line width=0.1] 
               \fill[gray, opacity=0.3] (0,0)   rectangle ++(200,4); 
               \fill[blue, opacity= .4] ( 43, 0) rectangle ++( 14, 4);	
               \fill[red, opacity= .8] ( 46, 0) rectangle ++( 10, 4);
              \end{tikzpicture}
                         & \multicolumn{4}{r}{mean completion time in 1st cond.} & $13.9 \quad 16.6$ \\ 
            & \begin{tikzpicture}[x=1pt, y=1pt, line width=0.1] 
               \fill[gray, opacity=0.3] (0,0)   rectangle ++(200,4); 
               \fill[blue, opacity= .4] ( 32, 0) rectangle ++( 11, 4);	
               \fill[red, opacity= .8] ( 35, 0) rectangle ++( 8, 4);
              \end{tikzpicture}
                         & \multicolumn{4}{r}{                        2nd cond.} & $10.5 \quad 12.6$ \\
        
            \cmidrule(lr){2-7} 
            & \makecell[r]{--- just Barsgrid tasks} & Inc & Inc & Inc & Inc & 95; 96 smp \\ 
            
            \cmidrule(lr){3-7} 
            & \begin{tikzpicture}[x=1pt, y=1pt, line width=0.1] 
               \fill[gray, opacity=0.3] (0,0)   rectangle ++(200,4); 
               \fill[blue, opacity= .4] ( 45, 0) rectangle ++( 21, 4);	
               \fill[blue, opacity= .8] ( 49, 0) rectangle ++( 14, 4);
              \end{tikzpicture}
                         & \multicolumn{4}{r}{mean completion time in 1st cond.} & $14.7 \quad 18.9$ \\ 
            & \begin{tikzpicture}[x=1pt, y=1pt, line width=0.1] 
               \fill[gray, opacity=0.3] (0,0)   rectangle ++(200,4); 
               \fill[blue, opacity= .4] ( 42, 0) rectangle ++( 18, 4);	
               \fill[blue, opacity= .8] ( 45, 0) rectangle ++( 12, 4);
              \end{tikzpicture}
                         & \multicolumn{4}{r}{                     in 2nd cond.} & $13.6 \quad 17.1$ \\

            \cmidrule(lr){2-7} 
            & \makecell[r]{--- just Polygrid tasks} & True & True & True & True & 95; 96 smp \\ 
            
            \cmidrule(lr){3-7} 
            & \begin{tikzpicture}[x=1pt, y=1pt, line width=0.1] 
               \fill[gray, opacity=0.3] (0,0)   rectangle ++(200,4); 
               \fill[red, opacity= .4] ( 35, 0) rectangle ++( 19, 4);	
               \fill[red, opacity= .8] ( 39, 0) rectangle ++( 12, 4);
              \end{tikzpicture}
                         & \multicolumn{4}{r}{mean completion time in 1st cond.} & $11.9 \quad 15.5$ \\ 
            & \begin{tikzpicture}[x=1pt, y=1pt, line width=0.1] 
               \fill[gray, opacity=0.3] (0,0)   rectangle ++(200,4); 
               \fill[red, opacity= .4] ( 20, 0) rectangle ++( 10, 4);	
               \fill[red, opacity= .8] ( 22, 0) rectangle ++( 6, 4);
              \end{tikzpicture}
                         & \multicolumn{4}{r}{                     in 2nd cond.} & $ 6.8 \quad  8.6$ \\
            
       
    \bottomrule
    \end{tabularx} \\
    \noindent
    \footnotesize \justifying \emph{Legend}: In row A3a, the statement clarifies the scope of the assumption.
    To its right, the results of its evaluation at various levels of confidence are shown, followed by the number of observations in its scope.
    Two scales are plotted below the statement.
    They depict the confidence intervals (CIs) being contrasted: one around the mean accuracy for tasks in the 1st condition, and the other around the mean accuracy for tasks in the 2nd condition.
    Description and details of each CI appear to the right of each scale.
    An opaque rectangle depicts a CI for $\alpha = 0.20$, and a transparent one depicts a CI for $\alpha = 0.05$.
    If the CIs overlap, they appear in blue, and red otherwise.
    Remaining rows are similarly organised.
    \aspas{Inc} stands for inconclusive, and \aspas{smp} for samples.
    \label{tab:userstudy:results:assumptions:a3}
\end{table}

\subsection{Discussion}
\label{sec:userstudy:discussion}

Our proxy measure for the interpretability of Polygrid diagrams is the degree of success that participants achieve in reproducing the output of an instance using the diagram drawn for some case, as shown in the visual classification task in Figure \ref{fig:userstudy:polygrid-condition}.
For the sake of clarity, the instructional videos guided the participant to base their judgement primarily on the similarity between polygons and rely on the weights given to different regions as a tiebreaker.
This is in tension with the Polygrid model, whose output is based on the weighted areas of the matching polygons.
The results suggest that this difference was irrelevant: the overall mean accuracy of participants considering the Polygrid's output as the ground truth is statistically indistinguishable from the same estimate based on the original dataset labels as ground truth (respectively, $0.927 \in [0.910, 0.946]$ and $0.938 \in [0.921, 0.954]$, with 368 samples).
In our view, this is evidence that Polygrid diagrams were easily interpreted by the study participants.
We believe that this conclusion can be generalised to other groups: 
since the critical skill to perform the tasks in this study is (basic) graph comprehension, which is covered by the curricular guidelines for secondary education in many places, the general population would probably find the diagram interpretable.
However, generalising to other contexts (e.g., other datasets, other configs) will require further research, as we briefly discuss in Section \ref{sec:userstudy:concerns}.





We now move on to assess the relative effectiveness of the Polygrid and Barsgrid diagrams.
With respect to accuracy, the results are inconclusive, as summarised in row \aspas{H1a} in Table \ref{tab:userstudy:results:hypotheses} ($\mu_{acc}^{Bars} = 0.909 \in [0.888, 0.931]$ and $\mu_{acc}^{Poly} = 0.917 \in [0.895, 0.938]$, with 276 samples).
In other words, the participant's accuracy when using the Polygrid diagram is comparable to that when using the Barsgrid diagram.
This is evidence that the use of polygons to represent multidimensional data and the absence of a numerical scale in Polygrid diagrams is not associated with a decrease in the participant's accuracy in solving the tasks.
However, it must be added that people made different mistakes when using one diagram or another.
For instance, among the eight participants shown diagrams for case 103, everyone correctly classified the case with the Barsgrid diagram, but only four were able to do so using the Polygrid diagram.
On the other hand, among the nine participants shown diagrams for case 52, everyone correctly classified the case with the Polygrid diagram, but only five were able to do so using the Barsgrid diagram.

\begin{table}[t]
    \centering
    \footnotesize
    \caption{Relevant hypotheses evaluated on data from the accepted scripts}
    
    \begin{tabularx}{\linewidth}{@{}P{0.5cm}p{7.0cm}ZZZZP{1.7cm}@{}}
    \toprule
        \multirow{2}{\hsize}{} & \multirow{2}{\hsize}{} & \multicolumn{4}{c}{Confidence Level ($\alpha$)} & Estimate \\         
        \cmidrule(lr){3-6}
        ID & Statement of the Hypothesis & $0.05$ & $0.10$ & $0.15$ & $0.20$ & (at $\alpha=0.20$) \\
    \toprule
 
        \multirow{4}{\hsize}{H1a} & \makecell[l]{Barsgrid is more effective than Polygrid \\(evidence from accuracy, except repeat tasks)} & Inc & Inc & Inc  & Inc  & 276 samples \\ 

            \cmidrule(lr){3-7} 
            & \begin{tikzpicture}[x=1pt, y=1pt, line width=0.1] 
               \fill[gray, opacity=0.3] (0,0)   rectangle ++(200,4); 
               \fill[blue, opacity= .4] ( 175, 0) rectangle ++( 14, 4);	
               \fill[blue, opacity= .8] ( 177, 0) rectangle ++( 9, 4);
              \end{tikzpicture}
                         & \multicolumn{4}{r}{mean accuracy in Barsgrid tasks} & $.888 \quad .931$ \\ 
            & \begin{tikzpicture}[x=1pt, y=1pt, line width=0.1] 
               \fill[gray, opacity=0.3] (0,0)   rectangle ++(200,4); 
               \fill[blue, opacity= .4] ( 176, 0) rectangle ++( 14, 4);	
               \fill[blue, opacity= .8] ( 179, 0) rectangle ++( 9, 4);
              \end{tikzpicture}         
                         & \multicolumn{4}{r}{              in Polygrid tasks} & $.895 \quad .938$ \\

            \cmidrule(lr){2-7} 
            & \makecell[r]{--- just 1st condition} & Inc & Inc & Inc  & Inc & 150;126 smp \\ 
            
            \cmidrule(lr){3-7} 
            & \begin{tikzpicture}[x=1pt, y=1pt, line width=0.1] 
               \fill[gray, opacity=0.3] (0,0)   rectangle ++(200,4); 
               \fill[blue, opacity= .4] ( 177, 0) rectangle ++( 18, 4);	
               \fill[blue, opacity= .8] ( 180, 0) rectangle ++( 11, 4);
              \end{tikzpicture}
                         & \multicolumn{4}{r}{mean accuracy in Barsgrid tasks} & $.900 \quad .953$ \\ 
            & \begin{tikzpicture}[x=1pt, y=1pt, line width=0.1] 
               \fill[gray, opacity=0.3] (0,0)   rectangle ++(200,4); 
               \fill[blue, opacity= .4] ( 169, 0) rectangle ++( 21, 4);
               \fill[blue, opacity= .8] ( 173, 0) rectangle ++( 13, 4);
              \end{tikzpicture}         
                         & \multicolumn{4}{r}{              in Polygrid tasks} & $.865 \quad .929$ \\

            \cmidrule(lr){2-7} 
            & \makecell[r]{--- just 2nd condition} & Inc & Inc & Inc  & Inc  & 126;150 smp \\ 
            
            \cmidrule(lr){3-7} 
            & \begin{tikzpicture}[x=1pt, y=1pt, line width=0.1] 
               \fill[gray, opacity=0.3] (0,0)   rectangle ++(200,4); 
               \fill[blue, opacity= .4] ( 166, 0) rectangle ++( 23, 4);	
               \fill[blue, opacity= .8] ( 171, 0) rectangle ++( 15, 4);
              \end{tikzpicture}         
                         & \multicolumn{4}{r}{mean accuracy in Barsgrid tasks} & $.857 \quad .929$ \\ 
            & \begin{tikzpicture}[x=1pt, y=1pt, line width=0.1] 
               \fill[gray, opacity=0.3] (0,0)   rectangle ++(200,4); 
               \fill[blue, opacity= .4] ( 178, 0) rectangle ++( 16, 4);	
               \fill[blue, opacity= .8] ( 181, 0) rectangle ++( 11, 4);
              \end{tikzpicture}         
                         & \multicolumn{4}{r}{              in Polygrid tasks} & $.907 \quad .960$ \\
            
            \midrule
       
        \multirow{4}{\hsize}{H1t} & \makecell[l]{Barsgrid is more effective than Polygrid \\(evidence from completion time, all but repeat)} & False & False & False & False & \makecell[c]{267 samples \\ 272 samples}\\ 
            \cmidrule(lr){3-7} 
            & \begin{tikzpicture}[x=1pt, y=1pt, line width=0.1] 
               \fill[gray, opacity=0.3] (0,0)   rectangle ++(200,4); 
               \fill[red, opacity= .4] ( 56, 0) rectangle ++( 13, 4);	
               \fill[red, opacity= .8] ( 58, 0) rectangle ++( 9, 4);
              \end{tikzpicture}
                         & \multicolumn{4}{r}{mean comp. time, Barsgrid tasks} & $17.5 \quad 20.0$ \\ 
            & \begin{tikzpicture}[x=1pt, y=1pt, line width=0.1] 
               \fill[gray, opacity=0.3] (0,0)   rectangle ++(200,4); 
               \fill[red, opacity= .4] ( 33, 0) rectangle ++( 9, 4);	
               \fill[red, opacity= .8] ( 35, 0) rectangle ++( 6, 4);
              \end{tikzpicture}
                         & \multicolumn{4}{r}{                      Polygrid tasks} & $10.6 \quad 12.4$ \\
            
            \midrule

        \multirow{4}{\hsize}{H2} & \makecell[l]{Barsgrid is more effective than Polygrid \\(evidence from intra-agreement, just mid/repeat)} & Inc & Inc & Inc & Inc  & 46 samples \\ 

            \cmidrule(lr){3-7} 
            & \begin{tikzpicture}[x=1pt, y=1pt, line width=0.1] 
               \fill[gray, opacity=0.3] (0,0)   rectangle ++(200,4); 
               \fill[blue, opacity= .4] ( 188, 0) rectangle ++( 12, 4);	
               \fill[blue, opacity= .8] ( 188, 0) rectangle ++( 12, 4);
              \end{tikzpicture}
                         & \multicolumn{4}{r}{mean consistency, Barsgrid tasks} & $.942 \quad 1.000$ \\ 
            & \begin{tikzpicture}[x=1pt, y=1pt, line width=0.1] 
               \fill[gray, opacity=0.3] (0,0)   rectangle ++(200,4); 
               \fill[blue, opacity= .4] ( 194, 0) rectangle ++( 6, 4);	
               \fill[blue, opacity= .8] ( 194, 0) rectangle ++( 6, 4);
              \end{tikzpicture}
                         & \multicolumn{4}{r}{              in Polygrid tasks} & $.971 \quad 1.000$ \\

            \midrule
        
        \multirow{4}{\hsize}{H3} & \makecell[l]{Barsgrid is more effective than Polygrid \\(evidence from inter-agreement, all but repeat)} & False & False & False & False & 1,035 smp\\ 
            \cmidrule(lr){3-7} 
            & \begin{tikzpicture}[x=1pt, y=1pt, line width=0.1] 
               \fill[gray, opacity=0.3] (0,0)   rectangle ++(200,4); 
               \fill[red, opacity= .4] ( 172, 0) rectangle ++( 6, 4);	
               \fill[red, opacity= .8] ( 173, 0) rectangle ++( 4, 4);
              \end{tikzpicture}
                         & \multicolumn{4}{r}{mean consensus, Barsgrid tasks} & $.869 \quad .887$ \\ 
            & \begin{tikzpicture}[x=1pt, y=1pt, line width=0.1] 
               \fill[gray, opacity=0.3] (0,0)   rectangle ++(200,4); 
               \fill[red, opacity= .4] ( 178, 0) rectangle ++( 5, 4);	
               \fill[red, opacity= .8] ( 179, 0) rectangle ++( 4, 4);
              \end{tikzpicture}
                         & \multicolumn{4}{r}{               in Polygrid tasks} & $.898 \quad .914$ \\
        
       
    \bottomrule
        
    \end{tabularx}\\
    \noindent
    \footnotesize \justifying \emph{Legend}: In row H1a, the statement clarifies the scope of the hypothesis.
    To its right, the results of its evaluation at various levels of confidence are shown, followed by the number of observations in its scope.
    Two scales are plotted below the statement.
    They depict the confidence intervals (CIs) being contrasted: one around the mean accuracy in Barsgrid tasks, and the other around the mean accuracy in Polygrid tasks.
    Description and details of each CI appear to the right of each scale.
    An opaque rectangle depicts a CI for $\alpha = 0.20$, and a transparent one depicts a CI for $\alpha = 0.05$.
    If the CIs overlap, they appear in blue, and red otherwise.
    Remaining rows are similarly organised.
    \aspas{Inc} stands for inconclusive. 
    \label{tab:userstudy:results:hypotheses}
\end{table}

Regarding the completion time, the collected data paint a different picture, as reported in row \aspas{H1t} in Table \ref{tab:userstudy:results:hypotheses}.
On average, it took substantially less time to complete a task with a Polygrid diagram than with a Barsgrid diagram ($\mu_{time}^{Bars} = 18.8 \in [17.5, 20.0]$ sec and 
$\mu_{time}^{Poly} = 11.5 \in [10.6, 12.4]$ sec, with 267 and 272 samples, respectively).
Overall, the results indicate that the participants completed the tasks faster and with at least the same level of accuracy when using the Polygrid diagram compared to the Barsgrid diagram.
This is in agreement with comments from several participants who, after finishing the study, found the Polygrid diagram easier to use. 
In fact, nine out of the 15 people who provided feedback explicitly stated a preference for the diagram, citing its ease of use for the tasks.
This finding is consistent with the idea that Polygrid diagrams drive the participants to rely on visual intuition, which is perceived as a less cognitively demanding strategy than the analytical reasoning required by Barsgrid diagrams.

In terms of consistency, equivalent results were observed for both diagrams, as reported in row \aspas{H2} in Table \ref{tab:userstudy:results:hypotheses}.
When presented a case for the second time, most of the participants provided the same response as previously.
Based on the results of a pilot study, cases 115 and 38 were found to be mildly discriminative in terms of the ability of the participants to solve the tasks.
These cases were selected because our aim was to estimate consistency considering cases of average difficulty.

Finally, with respect to consensus, the results suggest a systematic advantage of the Polygrid diagrams.
More precisely, participants provided the same response to a given case more frequently when using Polygrid diagrams compared to Barsgrid diagrams ($\mu_{agree}^{Bars} = 0.878 \in [0.869, 0.887]$ and $\mu_{agree}^{Poly} = 0.906 \in [0.898, 0.914]$, with 1,035 samples).
The large sample size is due to the fact that this measure is computed in an all-versus-all fashion: the responses given by a pair of participants to the set of cases that both evaluated are compared and a single agreement score is computed.
Since all participants evaluate at least two common cases (the mid cases), there are $\binom{46}{2} = 1,035$ such pairs.
Our hypothesis is that this finding is related to individual differences in reasoning strategies.
In other words, if people vary less in their ability to match visual shapes than they do in handling numerical data, then the observed results would be a logical consequence.

\subsection{Concerns and Limitations}
\label{sec:userstudy:concerns}

The design of this study is based on the idea that interpretability depends on the data, models, and users in intricate ways.
Thus, if one needs to assess the contribution of a model to interpretability in an experimental setting, then the other variables should be carefully manipulated so that the study results, poor or good, could be solely ascribed to the model.
According to the framework introduced in Section \ref{section:proposal:interpretable}, this means that we should employ low dimensional data that is meaningful to the candidate participants.

Initially, we considered using the ampiab-ml-11 dataset in this study because this is a healthcare dataset with real assignments.
This would allow us to recruit specialists in the care of older people to participate in the study.
However, in our view, this dataset has three issues that discourage its use in user studies to assess interpretability.
The first issue concerns the huge imbalance of the dataset, as reported in Table \ref{tab:offlineval:datasets}.
The majority class (aka most frequent label), which gathers individuals who were referred to specialised care after a home consultation, occurs in 86 of the 128 cases, while there are seven labels that occur less than five times, and four labels that occur only once.
Moreover, of the 15 labelsets observed in the assignments, eight occur a single time.
In a user study in which participants are asked to perform decision-making tasks\footnote{In the sense discussed in Section \ref{section:relatedwork:visual}, following the typology of \citeonline{brumar2025typology}.}, a balanced dataset is highly preferable because we can compare the participants' responses against a uniform distribution to account for chance effects, as in Equation \ref{eq:userstudy:kappa}, among many other reasons.

%
%

The second issue concerns the level of abstraction of the interventions.
For example, the most frequent label gathers patients who were referred to specialised care, but the specialty of care (e.g., physiotherapy, neurology, cardiology, etc.) is not available in the dataset.
This makes it difficult to the researcher running the user study to hypothesise how participants use the assessment data to select interventions, and also hinders the ability of learning models to combine variance in the assessment data with the assignment data to select good predictive hypotheses.
Relatedly, the third and last issue concerns apparent inconsistencies in the application of referral policies.
For example, one can find individuals with high deficit in the oral health domain that have not been referred to the attention of the oral health team, and also find referrals to that team of patients with low deficit in oral health.

In view of these difficulties, we selected a different dataset.
Our choice for the Iris dataset addresses the requirements for a low dimensional dataset ($d=4$) with attributes that are meaningful to the general public, which simplifies recruitment.
Moreover, the choice also addresses the need to set up a contrast to assess the effect on accuracy of endowing diagrams with scales (no scales vs. shared numerical scales), as mentioned earlier.

The second limitation concerns our choice of \aspas{model size}.
Earlier, we argued that some of the results were probably generalisable because they mainly depend on skills that people in general can learn (e.g., graph comprehension).
However, we reserve judgement on generalising the reported results to contexts in which the diagrams have a more complex structure (e.g., multiple annuli, multiple sectors per domain, four or more labels).
In other words, it is not clear whether the interpretability or the relative advantages of the Polygrid diagram, which stem from reframing a decision-making task into a kind of shape-sorting task, would be preserved in other contexts.

Finally, we also disclaim that the inclusion of the Barsgrid condition in this study did not produce results that are immediately applicable to bar charts.
In Section \ref{sec:userstudy:tasks}, we stated that the bar chart is one of the most effective visualisations for visual classification tasks, which underpins the design of the Barsgrid diagram. 
However, note that the task in our experimental setting differs markedly from that of \citeonline{saket2019task}.

\section{Conclusion}
\label{section:conclusion}


There are challenges that must be overcome to make recommender systems more useful in healthcare settings.
The reasons are varied: the lack of publicly available clinical data, the difficulty that users may have in understanding the reasons why a recommendation was made, the risks that may be involved in following the recommendation, and the uncertainty about its effectiveness.

These challenges were addressed in this work.
We partnered with research groups that have the expertise to collect health assessments using psychometric instruments, and extended these datasets with assignments in a principled way: individuals with similar needs should receive similar care.
We developed a learning model that uses psychometric data to produce recommendations and explanations that are interpretable by care professionals.
Moreover, the expert-in-the-loop use case mitigates the risks of harmful or innefective recommendations.

This work advances the application of the Polygrid model to assist gerontologists in the creation of personalised care plans in primary care.
The Polygrid model, introduced in Section \ref{section:proposal}, performs multilabel classification or label ranking tasks on psychometric data.
The model preserves the meaning of the data attributes to generate explanations that are interpretable by care professionals trained on the psychometric instrument.
The explanations are provided in a diagrammatic format that embeds the forward computational graph of the model, making these explanations faithful and the model transparent.
When the model fits the data well, the Polygrid explanation diagram converts a decision-making task into a shape-matching task.

Adding to the theoretical basis for Polygrid's learnability presented in Section \ref{section:proposal:closerlook}, empirical evidence was presented in Section \ref{section:offlineval}, which reports on the results of an offline evaluation of the Polygrid model.
We adapted the standard evaluation methodology to enforce the comparison between models whose instances have the same size on average, due to the importance of interpretability to our application and its decrease as model instances grow in size.
The results show that Polygrid dominated the competing models on the multiclass datasets, achieved mixed results on the multilabel datasets, and was consistently dominated on the label ranking datasets.
An implication of this outcome is that, if field applications are considered, the researcher is advised to start the project by tackling first the multilabel classification task.

The conceptual defence for Polygrid's interpretability laid out in Section \ref{section:proposal:interpretable:defence} is complemented with the empirical evidence in Section \ref{section:userstudy}, which reports the results of a within-subjects user study that combines methodological elements from the literature on the interpretability of machine learning models and tasks-based effectiveness in visualisation research: the participant is shown an explanation diagram in which the assessment chart displays measurements taken from a flower specimen, and is asked to classify that specimen.
The results show that the participants achieved a mean accuracy of about $0.9$, suggesting that the Polygrid diagram is conducive to interpretability.
The study also suggests that the use of polygons to display multidimensional data and the absence of graduate scales in the Polygrid diagram did not hinder the participants' performance; rather, it reduced the mean completion time and increased the consensus among participants.

\subsection*{Future Work}

Recently, \citeonline{jannach2025rs4good} shared their vision of an initiative in recommender systems research that is more focused on societal concerns (RS4Good).
This work contributes to this initiative, and to invite work related to this project, we highlight its key limitations and point out some directions that we see as the natural evolution of the proposed model.

\noindent
From Section \ref{section:offlineval}:
\begin{itemize}

    \item Researchers in different regulatory contexts (EU's GDPR, US HIPAA) can validate and extend our approach with locally-available healthcare data, testing generalisability across diverse populations and healthcare systems.
    We single out initiatives related to the WHO's ICOPE programme\footnote{\href{https://www.who.int/teams/maternal-newborn-child-adolescent-health-and-ageing/ageing-and-health/integrated-care-for-older-people-icope}{https://www.who.int/teams/maternal-newborn-child-adolescent-health-and-ageing/ageing-and-health/integrated-care-for-older-people-icope}} to adapt health systems for older people as potential research partners \cite{tavassoli2022ICOPE,frikha2024senselife}.
       
\end{itemize}

\noindent
From Section \ref{section:userstudy}:
\begin{itemize}

    \item Assuming access to locally-available healthcare data with real-world assignments, a user study could be conducted with care professionals.
    In this study, a new task should be included in the end of the participants' journey to ask her to fill out questionnaires to assess tolerance for ambiguity and reaction to uncertainty.
    These new variables may be useful for explaining variance in the observed decision-making behaviour, as argues \citeonline{shashar2023referrals}.

    \item Moreover, new user studies can focus on the effects of increasing instance size or the number of labels in the experimental conditions.
    The forward simulation tasks could be replaced with counterfactual simulation tasks, as described in \citeonline{doshivelez2017rigorous}: \aspas{humans are presented with an explanation, an input, and an output, and are asked what must be changed to change the method’s prediction to a desired output.}
    The results may reveal the ways people parse, interpret, and handle the data to make decisions in this setting.

\end{itemize}

\noindent
From Section \ref{section:proposal}:
\begin{itemize}

    \item An adaptive implementation of Polygrid's feature extractor.
    The challenge is to learn to partition the input space (i.e., to design an alternative to Algorithm \ref{alg:learning:ml:step3}) that combines both input and target data with no sacrifice to the Polygrid's interpretability and transparency.

    \item Another promising candidate for Polygrid's performance improvement is the method that converts the assignment matrix $Y$ for label ranking into the membership matrix $U$.
    The analysis of the gap in Polygrid's performance between classification and ranking tasks identifies differences between Algorithms \ref{alg:learning:ml:step5} and its label ranking version as potential causes, and highlights the role played by the logranks function (Eq. \ref{eq:proposal:logranks}) in this gap.


\end{itemize}

Thus, although we point out several opportunities for further work that can be addressed with modest resources, real progress in the direction of deploying the proposed model as an assistive technology is fundamentally in need of quality health care data and partners in healthcare research.

\subsection*{Code and Data Sharing}
The code implementing the Polygrid model and its evaluation environment (Polygrid CLI) is available from our \href{https://github.com/andreplima/polygrid}{project's github repository (click here)}.
The ELSIO1 dataset can be extracted from the ELSI-Brazil Wave 1 dataset with a provided script.
Access to the latter dataset must be requested from the research team that maintains the original data, via \href{https://elsi.cpqrr.fiocruz.br/en/home-english/}{ELSI-Brazil project website}.

\begin{acks}
This study was financed in part by the Coordenação de Aperfeiçoamento de Pessoal de Nível Superior - Brasil (CAPES) - Finance Code 001.
ELSI-Brazil was supported by the Brazilian Ministry of Health: DECIT/SCTIE (Grants: 404965/2012-1 and TED 28/2017); COPID/DECIV/SAPS (Grants: 20836, 22566, 23700, 25560, 25552, and 27510).
\end{acks}

\bibliographystyle{ACM-Reference-Format}
\bibliography{author-references}

@techreport{who2015report,
    title={World report on ageing and health (full report)},
    author={{World Health Organization}},
    institution={WHO},
    year={2015},
    publisher={World Health Organization},
    isbn={978-92-4-069481-1},
    url={https://apps.who.int/iris/handle/10665/186463},
    urlaccessdate = {6/30/2024},
    address={Geneva}
}

@techreport{who2023referrals,
	title = {High-value referrals: learning from challenges and opportunities of the COVID-19 pandemic. Concept paper},
	author = {{World Health Organization}},
    institution={WHO},
	year = {2023},
	publisher = {World Health Organization. Regional Office for Europe},
    isbn={978-92-4-069481-1},
    url={https://iris.who.int/handle/10665/367955},
    urlaccessdate = {7/13/2024},
    address={Copenhagen}
}

@manual{who2012whoqolmanual,
    title={Programme on mental health: {WHOQOL} user manual},
	author = {{World Health Organization}},
    institution={WHO},
    journal={World Health Organization - General Office},
    year={2012},
    url={https://iris.who.int/handle/10665/77932},
    urlaccessdate = {10/6/2025},
    address={Geneva}
}

@article{shashar2023referrals,
	author = {Shashar, Sagi and Ellen, Moriah and Codish, Shlomi and Davidson, Ehud and Novack, Victor},
	title = {Unravelling the determinants of medical practice variation in referrals among primary care physicians: insights from a retrospective cohort study in {Southern Israel}},
	volume = {13},
	number = {8},
    pages={8},
	elocation-id = {e072837},
	year = {2023},
	doi = {10.1136/bmjopen-2023-072837},
	publisher = {British Medical Journal Publishing Group},
	issn = {2044-6055},
	URL = {https://bmjopen.bmj.com/content/13/8/e072837},
	eprint = {https://bmjopen.bmj.com/content/13/8/e072837.full.pdf},
	journal = {BMJ Open}
}

@article {pedersene2015attitude,
	author = {Pedersen, Anette F. and Carlsen, Anders H. and Vedsted, Peter},
	title = {Association of GPs{\textquoteright} risk attitudes, level of empathy, and burnout status with PSA testing in primary care},
	journal = {British Journal of General Practice},
	publisher = {Royal College of General Practitioners},
	volume = {65},
	number = {641},
	pages = {e845--e851},
	year = {2015},
	issn = {0960-1643},
	doi = {10.3399/bjgp15X687649},
	URL = {https://bjgp.org/content/65/641/e845},
}

@article{wennberg2004interview,
    author = {Mullan, Fitzhugh},
    title = {Wrestling With Variation: An Interview With {Jack Wennberg}},
    journal = {Health Affairs},
    volume = {23},
    number = {Supplement 2},
    pages = {VAR-73-VAR-80},
    year = {2004},
    doi = {10.1377/hlthaff.var.73},
    URL = {https://doi.org/10.1377/hlthaff.var.73},
    eprint = {https://doi.org/10.1377/hlthaff.var.73},
}

@article{stemler2004comparison,
    title={A comparison of consensus, consistency, and measurement approaches to estimating interrater reliability},
    author={Stemler, Steven E},
    journal={Practical Assessment, Research, and Evaluation},
    publisher={University of Maryland},
    address={USA},
    issn={1531-7714},
    volume={9},
    number={1},
    pages={4},
    year={2004},
    doi={10.7275/96jp-xz07},
    url={https://doi.org/10.7275/96jp-xz07}
}

@article{nih1987consensus,
    author = {Brown, A. Sue and Brummel-Smith, Kenneth and Burgess, Lavola and D'Agostino, Ralph B. and Goldschmidt, John W. and Halter, Jeffrey B. and Hazzard, William R. and Jahnigen, Dennis W. and Phelps, Charles and Raskind, Murray and Schrier, Robert W. and Sox Jr., Harold C. and Williams, Sankey V. and Wykle, May},
    title = {{National Institutes of Health Consensus Development Conference Statement}: Geriatric assessment methods for clinical decision-making},
    journal = {Journal of the American Geriatrics Society},
    publisher={Wiley-Blackwell},
    address={UK},
    volume = {36},
    number = {4},
    pages = {342-347},
    doi = {10.1111/j.1532-5415.1988.tb02362.x},
    url = {https://doi.org/10.1111/j.1532-5415.1988.tb02362.x},
    year = {1988}
}

@article{welsh2014CGA,
	author = {Welsh, Tomas J. and Gordon, Arnold L. and Gladman, J.R.},
	title = {Comprehensive geriatric assessment - A guide for the non-specialist},
	year = {2014},
	journal = {International Journal of Clinical Practice},
	volume = {68},
	number = {3},
	pages = {290 – 293},
	doi = {10.1111/ijcp.12313},
	url = {https://doi.org/10.1111/ijcp.12313}
}

@article{basil2020ethics,
	author = {Varkey, Basil},
	title = {Principles of Clinical Ethics and Their Application to Practice},
	journal = {Medical Principles and Practice},
	volume = {30},
	number = {1},
	pages = {17-28},
	year = {2020},
	month = {06},
	issn = {1011-7571},
	doi = {10.1159/000509119},
	url = {https://doi.org/10.1159/000509119},
}

@article{gorzoni2017geriatria,
    title={Geriatria: Medicina do S{\'e}culo {XXI}?},
    author={Gorzoni, Milton Luiz},
    journal={Medicina (Ribeir{\~a}o Preto)},
    volume={50},
    number={3},
    pages={144--9},
    year={2017},
    doi={10.11606/issn.2176-7262.v50i3p144-149},
    url={https://doi.org/10.11606/issn.2176-7262.v50i3p144-149},
}

@article{garrard2020cgareview,
    author={Garrard, J.W. and Cox, N.J. and Dodds, R.M. and Roberts, H.C. and Sayer, A.A.},
    title={Comprehensive geriatric assessment in primary care: a systematic review},
    journal={Aging Clinical and Experimental Research},
    publisher={Springer},
    address={Germany},
    year={2020},
    volume={32},
    number={2},
    pages={197-205},
    doi={10.1007/s40520-019-01183-w},
    url={https://doi.org/10.1007/s40520-019-01183-w},
}

@mastersthesis{andrade2019ampiab,
  author = {Andrade, Suzana Carvalho Vaz de},
  title = {An{\'a}lise psicom{\'e}trica da Avalia{\c{c}}{\~a}o Multidimensional da Pessoa Idosa na Aten{\c{c}}{\~a}o B{\'a}sica (AMPI/AB)},
  school = {Escola de Artes, Ci{\^e}ncias e Humanidades, University of S{\~a}o Paulo},
  year = {2019},
  address = {S{\~a}o Paulo, Brazil},
  doi = {10.11606/D.100.2019.tde-25102019-171740},
  url={https://doi.org/10.11606/D.100.2019.tde-25102019-171740}
}

@phdthesis{kappenburg2014comparison,
  author={Kappenburg-ten Holt, Janke Carolien},
  title={A comparison between factor analysis and item response theory modeling in scale analysis},
  school = {University of Groningen},
  year={2014},
  address = {S{\~a}o Paulo, Brazil},
  isbn={9789036770910},
  url={https://research.rug.nl/en/publications/a-comparison-between-factor-analysis-and-item-response-theory-mod}
}

@book{bollen1989semlv,
    title={Structural Equations with Latent Variables},
    author={Bollen, Kenneth A},
    year={1989},
    pages={1-514},
    publisher={John Wiley \& Sons},
    address={USA},
    isbn={0-471-01171-1},
}

@incollection{bollen2013myths,
	author="Bollen, Kenneth A. and Pearl, Judea",
	editor="Morgan, Stephen L.",
	title="Eight Myths About Causality and Structural Equation Models",
	bookTitle="Handbook of Causal Analysis for Social Research",
	year="2013",
	publisher="Springer Netherlands",
	address="Dordrecht",
	pages="301--328",
	isbn="978-94-007-6094-3",
	doi="10.1007/978-94-007-6094-3_15",
	url="https://doi.org/10.1007/978-94-007-6094-3_15"
}

@article{mcneish2018alpha,
  author    = {Daniel McNeish},
  title     = {Thanks coefficient alpha, we’ll take it from here},
  journal   = {Psychological Methods},
  year      = {2018},
  volume    = {23},
  number    = {3},
  pages     = {412--433},
  doi       = {10.1037/met0000144},
  publisher = {American Psychological Association},
  address   = {Washington, DC, USA},
  issn      = {1082-989X},
  eissn     = {1939-1463},
  url       = {https://doi.org/10.1037/met0000144}
}

@article{mcneish2020sumscores,
  author  = {McNeish, Daniel and Wolf, Melissa Gordon},
  title   = {Thinking twice about sum scores},
  journal = {Behavior Research Methods},
  year    = {2020},
  volume  = {52},
  number  = {6},
  pages   = {2287--2305},
  doi     = {10.3758/s13428-020-01398-0},
  url     = {https://doi.org/10.3758/s13428-020-01398-0},
  issn    = {1554-3528}
}

@article{widaman2023sumscore2,
  author = {Widaman, Keith F. and Revelle, William},
  title = {Thinking thrice about sum scores, and then some more about measurement and analysis},
  journal = {Behavior Research Methods},
  year = {2023},
  volume = {55},
  number = {2},
  pages = {788--806},
  issn = {1554-3528},
  doi = {10.3758/s13428-022-01849-w},
  url = {https://doi.org/10.3758/s13428-022-01849-w}
}

@article{mcneish2023sumscore3,
  author = {McNeish, Daniel},
  title = {Psychometric properties of sum scores and factor scores differ even when their correlation is 0.98: A response to {Widaman and Revelle}},
  journal = {Behavior Research Methods},
  year = {2023},
  volume = {55},
  number = {8},
  pages = {4269--4290},
  issn = {1554-3528},
  doi = {10.3758/s13428-022-02016-x},
  url = {https://doi.org/10.3758/s13428-022-02016-x}
}

@article{widaman2024sumscore4,
	author = {Keith F. Widaman and William Revelle},
	title ={Thinking About Sum Scores Yet Again, Maybe the Last Time, We Don’t Know, Oh No...: A Comment on McNeish (2023)},
	journal = {Educational and Psychological Measurement},
	volume = {84},
	number = {4},
	pages = {637-659},
	year = {2024},
	doi = {10.1177/00131644231205310},
	URL = {https://doi.org/10.1177/00131644231205310},
	eprint = {https://doi.org/10.1177/00131644231205310},
}

@article{sijtsma2024sumscore5, 
	title={Recognize the Value of the Sum Score, Psychometrics’ Greatest Accomplishment}, 
	volume={89}, 
	DOI={10.1007/s11336-024-09964-7}, 
    url={https://doi.org/10.1007/s11336-024-09964-7}, 
	number={1}, 
	journal={Psychometrika}, 
	author={Sijtsma, Klaas and Ellis, Jules L. and Borsboom, Denny}, 
	year={2024}, 
	pages={84–117}
}

@article{mcneish2024sumscore6, 
	title={Practical Implications of Sum Scores Being Psychometrics’ Greatest Accomplishment}, 
	volume={89}, 
	DOI={10.1007/s11336-024-09988-z}, 
    url={https://doi.org/10.1007/s11336-024-09988-z}, 
	number={4}, 
	journal={Psychometrika}, 
	author={McNeish, Daniel}, 
	year={2024}, 
	pages={1148–1169}
}

@article{sijtsma2024sumscore8, 
	title={Rejoinder to McNeish and Mislevy: What Does Psychological Measurement Require?}, 
	volume={89}, 
    DOI={10.1007/s11336-024-10004-7}, 
    url={https://doi.org/10.1007/s11336-024-10004-7}, 
	number={4}, journal={Psychometrika}, 
	author={Sijtsma, Klaas and Ellis, Jules L. and Borsboom, Denny}, 
	year={2024}, 
	pages={1175–1185}
}

@book{borsboom2005measuring,
	author = {Borsboom, Denny},
	title = {Measuring the mind: Conceptual issues in contemporary psychometrics},
	year = {2005},
	publisher={Cambridge University Press},
	address={New York, US},
	isbn={978-0-521-84463-5},
	doi = {10.1017/CBO9780511490026},
	url = {http://doi.org/10.1017/CBO9780511490026},
}

@incollection{borsboom2015psychometrics,
	title = {Psychometrics},
	editor = {James D. Wright},
	booktitle = {International Encyclopedia of the Social and Behavioral Sciences (Second Edition)},
	publisher = {Elsevier},
	edition = {Second Edition},
	address = {Oxford},
	pages = {418-422},
	year = {2015},
	isbn = {978-0-08-097087-5},
	doi = {10.1016/B978-0-08-097086-8.43079-5},
	url = {https://doi.org/10.1016/B978-0-08-097086-8.43079-5},
	author = {Denny Borsboom and Dylan Molenaar},
}

@book{michell1999measurement,
  author={Michell, Joel},
  title={Measurement in psychology: A critical history of a methodological concept},
  year={1999},
  publisher={Cambridge University Press},
  address={Cambridge, UK},
  isbn={0-521-62120-8},
  url = {https://www.cambridgebookshop.co.uk/products/measurement-in-psychology}
}

@article{hayes2020omega,
	author = {Andrew F. Hayes and Jacob J. Coutts},
	title = {Use Omega Rather than {Cronbach}’s alpha for Estimating Reliability. But...},
	journal = {Communication Methods and Measures},
	volume = {14},
	number = {1},
	pages = {1--24},
	year = {2020},
	publisher = {Routledge},
	doi = {10.1080/19312458.2020.1718629},
	URL = {https://doi.org/10.1080/19312458.2020.1718629},
}

@techreport{carvalho2017operationalising,
    title={Operationalising the concept of intrinsic capacity in clinical settings},
    author={Carvalho, Islene Araujo and Martin, Finbarr C and Cesari, Matteo and Summi, Yuka and Thiyagarajan, Jotheeswaran A and Beard, John},
    institution={World Health Organization},
    booktitle={WHO Clinical Consortium on Healthy Ageing},
    year={2017},
    month={March},
    url={https://www.who.int/ageing/health-systems/clinical-consortium/CCHA2017-backgroundpaper-1.pdf},
    urlaccessdate={1/8/2021},
    address={Geneva}
}

@article{vergani2004polar,
    author={Vergani, Carlo and Corsi, Maurizio and Bezze, Maria and Bavazzano, Antonio and Vecchiato, Tiziano},
    title={A polar diagram for comprehensive geriatric assessment},
    journal={Archives of Gerontology and Geriatrics},
    publisher={Elsevier},
    address={Ireland},
    issn={0167-4943},
    year={2004},
    volume={38},
    number={2},
    pages={139-144},
    doi={10.1016/j.archger.2003.08.009},
    url={https://doi.org/10.1016/j.archger.2003.08.009},
}

@article{shibata1998development,
    author={Shibata, S. and Tuchiya, A. and Tamura, M. and Sasamori, N. and Yoshida, K. and Iwai, Y. and Hattori, M. and Otsuka, Y. and Tohyama, S. and Uchiyama, A.},
    title={Development of a health promotion system for the elderly: {Committee of Health Evaluation for Elderly Persons} {Council of Japan AMHTS Institutions}},
    journal={Journal of Medical Systems},
    publisher={Springer},
    address={USA},
    issn={1573-689X},
    year={1998},
    volume={22},
    number={1},
    pages={43-49},
    doi={10.1023/A:1022654406018},
    url={http://doi.org/10.1023/A:1022654406018},
}

@article{minemawari1999radar,
    title={The radar chart method and its analysis as a comprehensive geriatric assessment system for elderly disabled patients},
    author={Minemawari, Yoshimori and Kato, Takamasa},
    journal={Japanese Geriatrics Society},
    publisher={Japan Journal of Geriatrics},
    address={Japan},
    volume={36},
    number={3},
    pages={206-212},
    year={1999},
    doi={10.3143/geriatrics.36.206},
    url={https://doi.org/10.3143/geriatrics.36.206},
    issn={0300-9173},
    origlanguage={Japanese},
}

@article{jung2020radar,
    author={Jung, Hee-Won},
    title={Visualizing domains of comprehensive geriatric assessments to grasp frailty spectrum in older adults with a radar chart},
    journal={Annals of Geriatric Medicine and Research},
    publisher={Korean Geriatrics Society},
    address={South Korea},
    year={2020},
    volume={24},
    number={1},
    pages={55-56},
    doi={10.4235/agmr.20.0013},
    url={https://doi.org/10.4235/agmr.20.0013}
}

@article{cavanaugh2025radar,
	title = {Prospective evaluation of comprehensive geriatric assessments in multidisciplinary bladder cancer care and implications for personalized vulnerability phenotyping},
	journal = {Urologic Oncology: Seminars and Original Investigations},
	volume = {43},
	number = {8},
	pages = {468.e7-468.e18},
	year = {2025},
	issn = {1078-1439},
	doi = {10.1016/j.urolonc.2025.03.025},
	url = {https://doi.org/10.1016/j.urolonc.2025.03.025},
	author = {Dana Cavanaugh and Sarah K. Holt and Erin Dwyer and Erin Petersen and John L. Gore and George R. Schade and Petros Grivas and Andrew C. Hsieh and John K. Lee and Bruce Montgomery and Michael T. Schweizer and Todd Yezefski and Evan Y. Yu and Jonathan J. Chen and Jay J. Liao and Emily Weg and Jing Zeng and Samia Jannat and Donna L. Berry and Viraj A. Master and Jose M. Garcia and May J. Reed and Itay Bentov and Jonathan L. Wright and Sarah P. Psutka},
}

@article{tavassoli2022ICOPE,
    title = {Implementation of the {WHO} {Integrated Care for Older People} ({ICOPE}) programme in clinical practice: A prospective study},
    journal = {The Lancet Healthy Longevity},
    volume = {3},
    number = {6},
    pages = {e394-e404},
    year = {2022},
    issn = {2666-7568},
    doi = {10.1016/S2666-7568(22)00097-6},
    url = {https://doi.org/10.1016/S2666-7568(22)00097-6},
    author = {Neda Tavassoli and Philipe {de Souto Barreto} and Caroline Berbon and Celine Mathieu and Justine {de Kerimel} and Christine Lafont and Catherine Takeda and Isabelle Carrie and Antoine Piau and Tania Jouffrey and Sandrine Andrieu and Fatemeh Nourhashemi and John R Beard and Maria Eugenia {Soto Martin} and Bruno Vellas},
}

@article{ferrioli2024icopebr,
    title = {Assessment of intrinsic capacity in the {Brazilian} older population and the psychometric properties of the {WHO/ICOPE} screening tool: A multicenter cohort study protocol},
    journal = {Geriatrics, Gerontolology and Aging},
    volume = {18},
    number = {166},
    numpages={9},
    year = {2024},
    doi = {10.53886/gga.e0000166_EN},
    url = {https://doi.org/10.53886/gga.e0000166_EN},
    author = {Ferriolli, Eduardo and Lourenço, Roberto and Oliveira, Vitor and Mello, Renato and Ferretti-Rebustini, Renata and Pereira, Leani and Busse, Alexandre and Maciel, Álvaro and Leopoldino, Amanda and Lacerda, Ana and Navarro, Anderson and Fattori, Andre and Castro, Carla and Ximenes, Coeli and Fernandes, Daiane and Abreu, Daniela and Rebustini, Flávio and Roschel, Hamilton and Santos, Jair and Filho, Jarbas and Marques, João and Carneiro, José and Pompeu, José and Moriguti, Julio and Pinto, Juliana and Silva, Juscelio and Pfrimer, Karina and Kusumota, Luciana and Pegorari, Maycon and Alves, Natália and Lima, Nereida and Avelar, Núbia and Almeida, Olga and Boas, Paulo and Barreto, Philipe and Júnior, Renato and Guerra, Ricardo and Amorim, Rivia and Corte, Roberta and Rodrigues, Rosalina and Silva, Silvia and Neves, Thiago and Fett, Waleria and Filho, Wilson},
}

@article{hsiao2024icworks,
  author    = {Fei-Yuan Hsiao and Liang-Kung Chen},
  title     = {Intrinsic capacity assessment works—let's move on actions},
  journal   = {The Lancet Healthy Longevity},
  year      = {2024},
  volume    = {5},
  number    = {7},
  pages     = {e448--e449},
  doi       = {10.1016/S2666-7568(24)00110-7},
  url       = {https://doi.org/10.1016/S2666-7568(24)00110-7},
  publisher = {Elsevier}
}

@inproceedings{healthrecsys1,
    author = {Elsweiler, David and Ludwig, Bernd and Said, Alan and Sch{\"a}fer, Hanna and Trattner, Christoph},
    title = {{Engendering Health with Recommender Systems}},
    year = {2016},
    publisher = {ACM},
    address = {New York, USA},
    organization={SIGCHI},
    doi = {10.1145/2959100.2959203},
    url = {https://doi.org/10.1145/2959100.2959203},
    booktitle = {Proceedings of the 10th ACM Conference on Recommender Systems},
    pages = {409–410},
    numpages = {2},
    location = {Boston, USAssachusetts, USA},
    series = {RecSys}
}

@inproceedings{healthrecsys2,
    author = {Elsweiler, David and Hors-Fraile, Santiago and Ludwig, Bernd and Said, Alan and Sch{\"a}fer, Hanna and Trattner, Christoph and Torkamaan, Helma and Calero Valdez, Andr{\'e}},
    title = {{Second Workshop on Health Recommender Systems (HealthRecSys)}},
    year = {2017},
    publisher = {ACM},
    address = {New York, USA},
    organization={SIGCHI},
    doi = {10.1145/3109859.3109955},
    url = {https://doi.org/10.1145/3109859.3109955},
    booktitle = {Proceedings of the 11th ACM Conference on Recommender Systems},
    pages = {374–375},
    location = {Como, Italy},
    series = {RecSys}
}

@inproceedings{healthrecsys3,
    author = {Elsweiler, David and Ludwig, Bernd and Said, Alan and Sch{\"a}fer, Hanna and Torkamaan, Helma and Trattner, Christoph},
    title = {{Third International Workshop on Health Recommender Systems (HealthRecSys)}},
    year = {2018},
    publisher = {ACM},
    address = {New York, USA},
    organization={SIGCHI},
    doi = {10.1145/3240323.3240336},
    url = {https://doi.org/10.1145/3240323.3240336},
    booktitle = {Proceedings of the 12th ACM Conference on Recommender Systems},
    pages = {517–518},
    location = {Vancouver, British Columbia, Canada},
    series = {RecSys}
}

@inproceedings{healthrecsys4,
    author = {Elsweiler, David and Ludwig, Bernd and Said, Alan and Sch{\"a}fer, Hanna and Torkamaan, Helma and Trattner, Christoph},
    title = {{Fourth International Workshop on Health Recommender Systems (HealthRecSys)}},
    year = {2019},
    publisher = {ACM},
    address = {New York, USA},
    organization={SIGCHI},
    doi = {10.1145/3298689.3347053},
    url = {https://doi.org/10.1145/3298689.3347053},
    booktitle = {Proceedings of the 13th ACM Conference on Recommender Systems},
    pages = {554–555},
    location = {Copenhagen, Denmark},
    series = {RecSys}
}

@inproceedings{healthrecsys5,
    author = {Said, Alan and Sch{\"a}fer, Hanna and Torkamaan, Helma and Trattner, Christoph},
    title = {{Fifth International Workshop on Health Recommender Systems (HealthRecSys)}},
    year = {2020},
    publisher = {ACM},
    address = {New York, USA},
    organization={SIGCHI},
    doi = {10.1145/3383313.3411540},
    url = {https://doi.org/10.1145/3383313.3411540},
    booktitle = {Proceedings of the 14th ACM Conference on Recommender Systems},
    pages = {611–612},
    location = {Virtual Event, Brazil},
    series = {RecSys}
}

@inproceedings{sezgin2013review,
    author={Emre Sezgin and Segvi {\"O}zkan},
    booktitle={Proceedings of the E-Health and Bioengineering Conference}, 
    title={A systematic literature review on {Health Recommender Systems}},
    series={EHB},
    year={2013},
    volume={},
    number={},
    pages={1-4},
    doi={10.1109/EHB.2013.6707249},
    url={http://doi.org/10.1109/EHB.2013.6707249},
    isbn={978-1-4799-2372-4},
    publisher={IEEE},
    address={New York, USA},
}

@article{kamran2015survey,
    title={A Survey of Recommender Systems and Their Application in Healthcare.},
    author={Kamran, Malik and Javed, Ali},
    journal={Technical Journal of University of Engineering \& Technology Taxila},
    address={Taxila, Pakistan},
    volume={20},
    number={4},
    year={2015},
    numpages={},
    url={https://tj.uettaxila.edu.pk/older-issues/2015/No4/15.A%20Survey%20of%20Recommender%20Systems%20and%20Their%20Application%20in%20Healthcare.pdf},
}

@inproceedings{ferretto2017review,
    author={Ferretto, Luciano Rodrigo and Cervi, Cristiano Roberto and de Marchi, Ana Carolina Bertoletti},
    booktitle={12th Iberian Conference on Information Systems and Technologies (CISTI)}, 
    title={Recommender systems in mobile apps for health a systematic review}, 
    year={2017},
    volume={},
    number={},
    pages={1-6},
    publisher={IEEE},
    address={New York, USA},
    doi={10.23919/CISTI.2017.7975743},
    url={http://doi.org/10.23919/CISTI.2017.7975743},  
}

@article{horsfraile2018review,
	author = {Hors-Fraile, Santiago and Rivera-Romero, Octavio and Schneider, Francine and Fernandez-Luque, Luis and Luna-Perejon, Francisco and Civit-Balcells, Anton and de Vries, Hein},
	title = {Analyzing recommender systems for health promotion using a multidisciplinary taxonomy: A scoping review},
	year = {2018},
	journal = {International Journal of Medical Informatics},
	volume = {114},
	pages = {143 – 155},
	doi = {10.1016/j.ijmedinf.2017.12.018},
	url = {https://doi.org/10.1016/j.ijmedinf.2017.12.018},
}

@article{azmi2019review,
    author={Azmi, Aini Khairani and Abdullah, Noraswaliza and Emran, Nurul Akmar},
    title={A recommender system model for improving elderly well-being: A systematic literature review},
    journal={International Journal of Advances in Soft Computing and its Applications},
    issn={2074-8523},
    publisher={International Center for Scientific Research and Studies},
    address={Malaysia},
    year={2019},
    volume={11},
    number={2},
    pages={87-108},
}

@article{cheung2019review,
	author = {Cheung, Kei Long and Durusu, Dilara and Sui, Xincheng and de Vries, Hein},
	title = {How recommender systems could support and enhance computer-tailored digital health programs: A scoping review},
	year = {2019},
	journal = {Digital Health},
	volume = {5},
    pages={1--19},
	doi = {10.1177/2055207618824727},
	url = {https://doi.org/10.1177/2055207618824727},
}

@article{ertugrul2020review,
	author = {Celik Ertugrul, Duygu and Elci, Atilla},
	title = {A survey on semanticized and personalized health recommender systems},
	year = {2020},
	journal = {Expert Systems},
	volume = {37},
	number = {4},
    numpages={23},
	doi = {10.1111/exsy.12519},
	url = {https://doi.org/10.1111/exsy.12519},
}

@inproceedings{2019pincayreview,
    author={Jhonny Pincay and L. Ter{\'a}n and Edy Portmann},
    booktitle={Proceedings of the 6th International Conference on eDemocracy \& eGovernment}, 
    title={Health Recommender Systems: A State-of-the-Art Review}, 
    year={2019},
    pages={47-55},
    doi={10.1109/ICEDEG.2019.8734362},
    url={https://doi.org/10.1109/ICEDEG.2019.8734362},
    publisher = {IEEE},
    address={New York, USA},
    location = {Quito, Equador},
    series={ICEDEG}
}

@misc{su2020review,
    title={Do recommender systems function in the health domain: a system review}, 
    author={Jia Su and Yi Guan and Yuge Li and Weile Chen and He Lv and Yageng Yan},
    year={2020},
    eprint={2007.13058},
    archivePrefix={arXiv},
    primaryClass={cs.IR},
    url={https://arxiv.org/abs/2007.13058},
    publisher={Cornell University},
    address={Ithaca, USA}
}

@article{decroon2021review,
	author = {de Croon, Robin and Van Houdt, Leen and Htun, Nyi Nyi and Štiglic, Gregor and Abeele, Vero Vanden and Verbert, Katrien},
	title = {Health recommender systems: Systematic review},
	year = {2021},
	journal = {Journal of Medical Internet Research},
	volume = {23},
	number = {6},
    numpages={21},
	doi = {10.2196/18035},
	url = {https://doi.org/10.2196/18035},
}

@article{tran2021review,
	author = {Tran, Thi Ngoc Trang and Felfernig, Alexander and Trattner, Christoph and Holzinger, Andreas},
	title = {Recommender systems in the healthcare domain: {S}tate-of-the-art and research issues},
	year = {2021},
	journal = {Journal of Intelligent Information Systems},
	volume = {57},
	number = {1},
	pages = {171 – 201},
	doi = {10.1007/s10844-020-00633-6},
	url = {https://doi.org/10.1007/s10844-020-00633-6},
}

@article{cai2022review,
	author = {Cai, Yao and Yu, Fei and Kumar, Manish and Gladney, Roderick and Mostafa, Javed},
	title = {Health Recommender Systems Development, Usage, and Evaluation from 2010 to 2022: A Scoping Review},
	year = {2022},
	journal = {International Journal of Environmental Research and Public Health},
	volume = {19},
	number = {22},
    numpages={},
	doi = {10.3390/ijerph192215115},
	url = {https://doi.org/10.3390/ijerph192215115},
}

@inproceedings{calderon2023review,
    author = {Calder{\'o}n-Blas, Javier A. and Cerd{\'a}n, Mar{\'i}a Ang{\'e}lica and S{\'a}nchez-Garc{\'i}a, {\'A}ngel J. and Dom{\'i}nguez-Isidro, Sa{\'u}l},
	booktitle={2023 Mexican International Conference on Computer Science (ENC)}, 
	title={Medical Recommender Systems: A Systematic Literature Review}, 
	year={2023},
	volume={},
	number={},
	pages={1-8},
	publisher={IEEE},
    address={New York, USA},
	doi={10.1109/ENC60556.2023.10508695},
	url={https://doi.org/10.1109/ENC60556.2023.10508695},  
}

@article{etemadi2023review,
	author = {Etemadi, Maryam and Bazzaz Abkenar, Sepideh and Ahmadzadeh, Ahmad and Haghi Kashani, Mostafa and Asghari, Parvaneh and Akbari, Mohammad and Mahdipour, Ebrahim},
	title = {A systematic review of healthcare recommender systems: Open issues, challenges, and techniques},
	year = {2023},
	journal = {Expert Systems with Applications},
	volume = {213},
    numpages={27},
	doi = {10.1016/j.eswa.2022.118823},
	url = {https://doi.org/10.1016/j.eswa.2022.118823},
}

@article{vieira2023review,
	author = {Vieira, Ana and Carneiro, João and Novais, Paulo and Corchado, Juan and Marreiros, Goreti},
	title = {A systematic review on recommendation systems applied to chronic diseases},
	year = {2023},
	journal = {Intelligent Data Analysis},
	volume = {27},
	number = {5},
	pages = {1223 – 1265},
	doi = {10.3233/IDA-220394},
	url = {https://doi.org/10.3233/IDA-220394},
}

@article{martinez2019pharos,
  title={{PHAROS 2.0—A PHysical Assistant RObot System Improved}},
  author={Martinez Martin, Ester and Costa, Angelo and Cazorla, Miguel},
  journal={Sensors},
  publisher={Multidisciplinary Digital Publishing Institute},
  address={Switzerland},
  issn={1424-8220},
  volume={19},
  number={20},
  pages={4531},
  year={2019},
  doi={10.3390/s19204531},
  url={https://doi.org/10.3390/s19204531},
}

@article{rincon2019new,
  title={A new emotional robot assistant that facilitates human interaction and persuasion},
  author={Rincon, Jaime A. and Costa, Angelo and Novais, Paulo and Julian, Vicente and Carrascosa, Carlos},
  journal={Knowledge and Information Systems},
  publisher={Springer},
  address={UK},
  volume={60},
  number={1},
  pages={363--383},
  year={2019},
  doi={10.1007/s10115-018-1231-9},
  url={https://doi.org/10.1007/s10115-018-1231-9},
}

@inproceedings{silva2017tv4e,
    author={Telmo Silva and Hilma Caravau and Liliana Reis and Carlos Silva and Martinho Mota},
    title={{+TV4E - Delivering} Information about Social Services for Seniors throughout {TV}},
    booktitle={Opportunities and Challenges for European Projects - Volume 1: EPS Portugal 2017/2018},
    year={2017},
    pages={133-149},
    organization={INSTICC},
    doi={10.5220/0008862501330149},
    url={https://doi.org/10.5220/0008862501330149},
    isbn={978-989-758-361-2},
    publisher={SciTePress},
    address = {Portugal},
}

@inproceedings{orte2018dynamic,
    author={Orte, Silvia and Sub{\'\i}as, Paula and Maldonado, Laura Fernandez and Mastropietro, Alfonso and Porcelli, Simone and Rizzo, Giovanna and Boqu{\'e}, Noem{\'\i} and Guye, Sabrina and R{\"o}cke, Christina and Andreoni, Giuseppe and others},
    title={Dynamic decision support system for personalised coaching to support active ageing},
    booktitle={AI*AAL.it 2018 - Fourth Italian Workshop on Artificial Intelligence for Ambient Assisted Living},
    publisher={CEUR Workshop Proceedings},
    address={Trento, Italy},
    year={2018},
    volume={2333},
    pages={16-36},
    url={http://ceur-ws.org/Vol-2333/paper2.pdf},
}

@inproceedings{robinson2017improving,
    author={Robinson, Jon and Appiah, Kofi and Yousaf, Raheem},
    title={Improving the well-being of older people by reducing their energy consumption through energy-aware systems},
    booktitle={Proceedings of the 9th International Conference on eHealth, Telemedicine, and Social Medicine},
    year={2017},
    isbn={9781612085401},
    publisher={International Academy, Research and Industry Association (IARIA)},
    address = {Nice, France},
    numpages={6},
    series={eTELEMED},
    url={http://irep.ntu.ac.uk/id/eprint/30483}
}

@inproceedings{stiller2010demographic,
    author={Stiller, Carsten and Ro{\ss}, Fred and Ament, Christoph},
    title={Demographic recommendations for {WEITBLICK}, an assistance system for elderly},
    booktitle={10th International Symposium on Communications and Information Technologies},
    year={2010},
    pages={406-411},
    doi={10.1109/ISCIT.2010.5664874},
    url={https://doi.org/10.1109/ISCIT.2010.5664874},
    publisher={IEEE},
    location={Tokyo, Japan},
    address={New York, USA}
}

@inproceedings{rist2015care,
    author={Rist, Thomas and Seiderer, Andreas and Hammer, Stephan and Mayr, Marcus and Andr{\'e}, Elisabeth},
    title={{CARE - Extending} a digital picture frame with a recommender mode to enhance well-being of elderly people},
    booktitle={Proceedings of the 9th International Conference on Pervasive Computing Technologies for Healthcare},
    year={2015},
    pages={112-120},
    doi={10.4108/icst.pervasivehealth.2015.259255},
    url={https://ieeexplore.ieee.org/abstract/document/7349386},
    location={Istanbul, Turkey},
    address={New York, USA},
    publisher={IEEE},
    series={PervasiveHealth},
}

@inproceedings{allalouf2020music,
    author={Miriam Allalouf and Avi Cohen and Lea Cohen Sabban and Ayelet Dassa and Sagi Marciano and Stella Melnitzer Beris},
    title={Music Recommendation System for Old People with Dementia and Other Age-related Conditions},
    booktitle={Proceedings of the 13th International Joint Conference on Biomedical Engineering Systems and Technologies},
    volume={Volume 5: HEALTHINF},
    year={2020},
    address = {Set{\'u}bal, Portugal},
    pages={429-437},
    publisher={SciTePress},
    organization={INSTICC},
    doi={10.5220/0008959304290437},
    url={https://doi.org/10.5220/0008959304290437},
    isbn={978-989-758-398-8},
}

@article{gannod2019machine,
    author={Gannod, Gerald C. and Abbott, Katherine M. and Van Haitsma, Kimberly and Martindale, Nathan and Heppner, Alexandra},
    title={A Machine Learning Recommender System to Tailor Preference Assessments to Enhance Person-Centered Care among Nursing Home Residents},
    journal={Gerontologist},
    publisher = {Oxford Academic Press},
    address={UK},
    year={2019},
    volume={59},
    number={1},
    pages={167-176},
    doi={10.1093/geront/gny056},
    url={https://doi.org/10.1093/geront/gny056},
}

@incollection{besik2019rhcs,
    author={Besik, Saliha Irem and Alpaslan, Ferda Nur},
    editor={Gadepally, Vijay and Mattson, Timothy and Stonebraker, Michael and Wang, Fusheng and Luo, Gang and Teodoro, George},
    title={{RHCS - A} Clinical Recommendation System for Geriatric Patients},
    booktitle={Heterogeneous Data Management, Polystores, and Analytics for Healthcare},
    year={2019},
    volume={11470},
    pages={115-132},
    isbn={978-3-030-14177-6},
    doi={10.1007/978-3-030-14177-6_10},
    url = {https://doi.org/10.1007/978-3-030-14177-6_10},
    publisher={Springer},
    address={Cham, Switzerland},
    series={Lecture Notes in Computer Science}
}

@incollection{ponce2015quefaire,
    author={Ponce, Victor and Deschamps, Jean-Pierre. and Giroux, Louis-Philippe and Salehi, Farzad and Abdulrazak, Bessam},
    title={{QueFaire}: Context-Aware in-person social activity recommendation system for active aging},
    booktitle={Inclusive Smart Cities and e-Health},
    publisher={Springer},
    address={Netherlands},
    year={2015},
    volume={9102},
    pages={64-75},
    isbn={978-3-319-19312-0},
    doi={10.1007/978-3-319-19312-0_6},
    url={https://doi.org/10.1007/978-3-319-19312-0_6},
    series={Lecture Notes in Computer Science}
}

@inproceedings{bermingham2013recommending,
    author={Bermingham, Adam and O'Rourke, Julia and Gurrin, Cathal. and Collins, Ronan and Irving, Kate and Smeaton, Alan F.},
    title={Automatically recommending multimedia content for use in group reminiscence therapy},
    booktitle={Proceedings of the 1st ACM International Workshop on Multimedia Indexing and Information Retrieval for Healthcare, Co-located with ACM Multimedia 2013},
    year={2013},
    pages={49-58},
    doi={10.1145/2505323.2505333},
    url={https://doi.org/10.1145/2505323.2505333},
    publisher = {ACM},
    series={MIIRH},
    location = {Barcelona, Spain},
    address = {New York, USA},    
}

@inproceedings{oliva2018health,
    author = {Oliva-Felipe, Luis and Barru\'{e}, Cristian and Cort\'{e}s, Atia and Wolverson, Emma and Antomarini, Marco and Landrin, Isabelle and Votis, Konstantinos and Paliokas, Ioannis and Cort\'{e}s, Ulises},
    title={Health recommender system design in the context of {CAREGIVERSPRO-MMD Project}},
    booktitle={Proceedings of the 11th Pervasive Technologies Related to Assistive Environments Conference},
    year={2018},
    pages={462-469},
    doi={10.1145/3197768.3201558},
    url={https://doi.org/10.1145/3197768.3201558},
    publisher={ACM},
    address={New York, USA},
    location={Corfu, Greece},
    series={PETRA},
}

@inproceedings{luo2010intelligent,
    author = {Luo, Gang and Tang, Chunqiang and Thomas, Selena B.},
    title = {Intelligent Personal Health Record: Experience and Open Issues},
    year = {2010},
    publisher = {ACM},
    address = {New York, USA},
    doi = {10.1145/1882992.1883039},
    url = {https://doi.org/10.1145/1882992.1883039},
    booktitle = {Proceedings of the 1st ACM International Health Informatics Symposium},
    pages = {326–335},
    location = {Arlington, Virginia, USA},
    series = {IHI},
    organization={SIGHIT}
}

@techreport{kaneda2011scl,
    title={SCL/PRB index of well-being in older populations},
    author={Kaneda, Toshiko and Lee, Marlene and Pollard, Kelvin},
    institution={Stanford Center on Longevity},
    year={2011},
    url={https://tinyurl.com/kaneda2011scl},
    urlaccessdate = {10/10/2020},
    address = {Stanford, CA, US}
}

@article{vanhaitsma2013preferences,
    author={Van Haitsma, Kimberly and Curyto, Kimberly. and Spector, Abby and Towsley, Gail and Kleban, Morton and Carpenter, Brian and Ruckdeschel, Katy and Feldman, Penny H. and Koren, Mary Jane},
    title={The preferences for everyday living inventory: Scale development and description of psychosocial preferences responses in community-dwelling elders},
    journal={Gerontologist},
    publisher = {Oxford Academic Press},
    address={UK},
    year={2013},
    volume={53},
    number={4},
    pages={582-595},
    doi={10.1093/geront/gns102},
    url={https://doi.org/10.1093/geront/gns102},
}

@incollection{HandbookRS2022Cap19,
    author="Tintarev, Nava and Masthoff, Judith",
    editor="Ricci, Francesco and Rokach, Lior and Shapira, Bracha",
    title="Beyond Explaining Single Item Recommendations",
    bookTitle="Recommender Systems Handbook",
    year="2022",
    publisher="Springer US",
    address="New York, NY",
    pages="711--756",
    isbn="978-1-0716-2197-4",
    doi="10.1007/978-1-0716-2197-4_19",
    url="https://doi.org/10.1007/978-1-0716-2197-4_19"
    }

@incollection{tal2020measurement,
    author       =	{Tal, Eran},
    title        =	{{Measurement in Science}},
    booktitle    =	{The {Stanford} Encyclopedia of Philosophy},
    editor       =	{Edward N. Zalta},
    url={https://plato.stanford.edu/archives/fall2020/entries/measurement-science/},
    year         =	{2020},
    edition      =	{Fall 2020},
    publisher    =	{Metaphysics Research Lab, Stanford University},
    address = {Stanford, CA, US}
}

@phdthesis{philippi2023challenges, 
    title={On the Challenges of Measurement in the Human Sciences},
    author={Larroulet Philippi, Cristian}, 
    school={Apollo - University of Cambridge Repository}, 
    year={2023}, 
    DOI={10.17863/CAM.102194}, 
    url={https://www.repository.cam.ac.uk/handle/1810/358705} 
}

@book{hausman2015valuing,
    title={Valuing health: Well-being, freedom, and suffering},
    author={Hausman, Daniel M},
    year={2015},
    publisher={Oxford University Press},
    address={New York, NY, USA},
    isbn={978-0-19-0233118-1},
    url={https://global.oup.com/academic/product/valuing-health-9780190233181},
    urlaccessdate = {9/27/2025}
}

@book{krabbe2016measurement,
    title={The measurement of health and health status: Concepts, methods and applications from a multidisciplinary perspective},
    author={Krabbe, Paul},
    year={2016},
    publisher={Academic Press},
    address={London, UK},
    isbn={978-0-12-801504-9},
    url={https://doi.org/10.1016/C2013-0-19200-8},
    urlaccessdate = {9/27/2025}
}

@article{nunes2017systematic,
  title={A systematic review and taxonomy of explanations in decision support and recommender systems},
  author={Nunes, Ingrid and Jannach, Dietmar},
  journal={User Modeling and User-Adapted Interaction},
  publisher={Springer},
  address={Netherlands},
  volume={27},
  number={3-5},
  pages={393--444},
  year={2017},
  doi={10.1007/s11257-017-9195-0},
  url={https://doi.org/10.1007/s11257-017-9195-0},
}

@article{younhee2017shoelace,
    author = {Younhee Lee and Woong Lim},
    journal = {Mathematics Teacher},
    address={USA},
    number = {8},
    pages = {631--636},
    publisher = {National Council of Teachers of Mathematics},
    title = {Shoelace Formula: Connecting the Area of a Polygon with Vector Cross Product},
    volume = {110},
    year = {2017},
    issn={2330-0582},
    doi={10.5951/mathteacher.110.8.0631},
    url={https://doi.org/10.5951/mathteacher.110.8.0631},
}

@article{jia2024regression,
  author  = {Jia, Bin-Bin and Liu, Jun-Ying and Zhang, Min-Ling},
  title   = {Towards exploiting linear regression for multi-class/multi-label classification: An empirical analysis},
  journal = {International Journal of Machine Learning and Cybernetics},
  year    = {2024},
  volume  = {15},
  number  = {9},
  pages   = {3671--3700},
  month   = {sep},
  issn    = {1868-808X},
  doi     = {10.1007/s13042-024-02114-6},
  url     = {https://doi.org/10.1007/s13042-024-02114-6}
}

@manual{orley1996whoqolbref,
    author={Orley, John and Power, Mick and Kuyken, Willem and Sartorius, Norman and Bullinger, Monika and Harper, A.},
    title={{WHOQOL-BREF: Introduction, administration, scoring and generic version of the assessment: field trial version}},
    journal={World Health Organization},
    year={1996},
    url={https://www.who.int/publications/i/item/WHOQOL-BREF},
    urlaccessdate = {7/27/2024},
    address={Geneva}
}

@article{skevington2004whoqol,
  author = {Skevington, Suzanne M. and Lotfy, M. and O'Connell, Kathryn A.},
  title = {The World Health Organization's {WHOQOL-BREF} quality of life assessment: Psychometric properties and results of the international field trial. {A} Report from the {WHOQOL} Group},
  journal = {Quality of Life Research},
  year = {2004},
  volume = {13},
  number = {2},
  pages = {299--310},
  issn = {1573-2649},
  doi = {10.1023/B:QURE.0000018486.91360.00},
  url = {https://doi.org/10.1023/B:QURE.0000018486.91360.00}
}

@article{odonnel2000variation,
    author = {O'Donnell, Catherine A.},
    title = "{Variation in GP referral rates: What can we learn from the literature?}",
    journal = {Family Practice},
    volume = {17},
    number = {6},
    pages = {462-471},
    year = {2000},
    month = {12},
    issn = {0263-2136},
    doi = {10.1093/fampra/17.6.462},
    url = {https://doi.org/10.1093/fampra/17.6.462},
}

@inproceedings{myself2021interpretable,
    author = {de Lima, Andre Paulino and Dantas, Laurentino Augusto and Manzato, Marcelo Garcia and Pimentel, Maria and Orlandi, Brunela and Castro, Paula},
    title = {An Interpretable Recommendation Model for Gerontological Care},
    year = {2021},
    isbn = {9781450384582},
    publisher = {ACM},
    address = {NY, USA},
    url = {https://doi.org/10.1145/3460231.3478850},
    doi = {10.1145/3460231.3478850},
    booktitle = {Proceedings of the 15th ACM Conference on Recommender Systems},
    pages = {620–626},
    numpages = {7},
    location = {Netherlands},
    series = {RecSys '21}
}

@article{castro2022whoqol,
    title={Digital engagement and quality of life of participants at a {University of the Third Age}},
    author={Lorenzi, Lorena Jorge and Alvarez, Pedro A. R. and Bet, Patricia and Castro, Paula Costa},
    journal={Gerontechnology},
    volume={21},
    pages={1-1},
    year={2022},
    doi={10.4017/gt.2022.21.s.567.opp4},
    url={https://doi.org/10.4017/gt.2022.21.s.567.opp4},
}

@article{melo2015desafios,
  title={Desafios da forma{\c{c}}{\~a}o em Gerontologia},
  author={de Melo, Ruth Caldeira and da Silva, Tha{\'\i}s Bento Lima and Cachioni, Meire},
  journal={Revista Kair{\'o}s-Gerontologia},
  volume={18},
  pages={123--147},
  year={2015},
  url = {https://revistas.pucsp.br/index.php/kairos/article/view/27261/19297}
}

@article{shenk2001teaching,
	author = {Shenk, Dena and Rowles, Graham D. and Peacock, James R. and Mitchell, Jim and Fisher, Bradley J.  and Moore, Krista S. and Hare, Lyndall},
	title = {Teaching research in {Gerontology}: Toward a cumulative model},
	journal = {Educational Gerontology},
	volume = {27},
	number = {7},
	pages = {537--556},
	year = {2001},
	publisher = {Routledge},
	doi = {10.1080/036012701753122884},
	URL = {https://doi.org/10.1080/036012701753122884},
	eprint = {https://doi.org/10.1080/036012701753122884},
}

@article{marcucci2020ampiab,  
    title={Health profile of older adults assisted by the Elderly Caregiver Program of Health Care Network of the {City of São Paulo}},  
    volume={18},  
    ISSN={1679-4508},  
    url={https://doi.org/10.31744/einstein_journal/2020AO5263},  
    DOI={10.31744/einstein_journal/2020AO5263},  
    journal={Einstein (São Paulo)},  
    publisher={Instituto Israelita de Ensino e Pesquisa Albert Einstein},  
    author={Andrade, Suzana Carvalho Vaz de and Marcucci, Rosa Maria Bruno and Faria, Lilian de Fátima Costa and Paschoal, Sérgio Márcio Pacheco and Rebustini, Flávio and Melo, Ruth Caldeira de},  
    year={2020},  
    pages={eAO5263} 
}

@article{silva2014whoqolcut,
    title={Cut-off point for {WHOQOL-BREF} as a measure of quality of life of older adults},
    author={Silva, Patrícia Aparecida Barbosa and Soares, Sônia Maria and Santos, Joseph Fabiano Guimarães and Silva, Líliam Barbosa},
    journal={Revista de Saúde Pública},
    volume={48},
    pages={390--397},
    year={2014},
    publisher={SciELO Brasil},
    doi={10.1590/S0034-8910.2014048004912},
    url={https://doi.org/10.1590/S0034-8910.2014048004912},
}

@article{aliberti2022ic,
    title = {Validating intrinsic capacity to measure healthy aging in an upper middle-income country: Findings from the {ELSI-Brazil}},
    journal = {The Lancet Regional Health - Americas},
    volume = {12},
    pages = {100284},
    year = {2022},
    issn = {2667-193X},
    doi = {10.1016/j.lana.2022.100284},
    url = {https://doi.org/10.1016/j.lana.2022.100284},
    author = {Márlon J.R. Aliberti and Laiss Bertola and Claudia Szlejf and Déborah Oliveira and Ronaldo D. Piovezan and Matteo Cesari and Fabíola Bof {de Andrade} and Maria Fernanda Lima-Costa and Monica Rodrigues Perracini and Cleusa P. Ferri and Claudia K. Suemoto},
}

@article{katz1963index,
	author = {Katz, Sidney and Ford, Amasa B. and Moskowitz, Roland W. and Jackson, Beverly A. and Jaffe, Marjorie W.},
	title = {Studies of Illness in the Aged: The Index of {ADL}: A Standardized Measure of Biological and Psychosocial Function},
	journal = {JAMA},
	volume = {185},
	number = {12},
	pages = {914-919},
	year = {1963},
	month = {09},
	issn = {0098-7484},
	doi = {10.1001/jama.1963.03060120024016},
	url = {https://doi.org/10.1001/jama.1963.03060120024016},
}

@article{limacosta2018elsi,
	author = {Lima-Costa, M Fernanda and de Andrade, Fabíola Bof and Souza, Paulo Roberto Borges de and Neri, Anita Liberalesso and Duarte, Yeda Aparecida de Oliveira and Castro-Costa, Erico and de Oliveira, Cesar},
	title = "{The Brazilian Longitudinal Study of Aging (ELSI-Brazil): Objectives and Design}",
	journal = {American Journal of Epidemiology},
	volume = {187},
	number = {7},
	pages = {1345-1353},
	year = {2018},
	month = {01},
	issn = {0002-9262},
	doi = {10.1093/aje/kwx387},
	url = {https://doi.org/10.1093/aje/kwx387},
	eprint = {https://academic.oup.com/aje/article-pdf/187/7/1345/25076724/kwx387.pdf},
}

@InBook{fuzzy-logic-ross.ch10,
    author = {Ross, Timothy J.},
    organization = {Ross, Timothy J.},
	title = {Fuzzy Classification},
	booktitle = {Fuzzy Logic with Engineering Applications},
    edition={3rd},
	year = {2010},
	publisher = {John Wiley \& Sons, Ltd},
    address={UK},
	isbn = {9781119994374},
	chapter = {10},
	pages = {332-368},
	doi = {10.1002/9781119994374.ch10},
	url = {https://doi.org/10.1002/9781119994374.ch10},
}

@book{giere2006perspectivism,
    title = {Scientific Perspectivism},
    author = {Ronald N. Giere},
    publisher = {University of Chicago Press},
    address = {Chicago},
    isbn = {9780226292144},
    url={https://press.uchicago.edu/ucp/books/book/chicago/S/bo4094708.html},
    year = {2006},
}

@misc{doshivelez2017rigorous,
      title={Towards A Rigorous Science of Interpretable Machine Learning}, 
      author={Finale Doshi-Velez and Been Kim},
      year={2017},
      eprint={1702.08608},
      archivePrefix={arXiv},
      primaryClass={stat.ML},
      url={https://arxiv.org/abs/1702.08608}, 
}

@misc{fang2024fcg,
	title={A step-by-step introduction to the implementation of automatic differentiation}, 
	author={Yu-Hsueh Fang and He-Zhe Lin and Jie-Jyun Liu and Chih-Jen Lin},
	year={2024},
	eprint={2402.16020},
	archivePrefix={arXiv},
	primaryClass={cs.LG},
	url={https://arxiv.org/abs/2402.16020}, 
}

@article{rudin2019stop,
  author = {Rudin, Cynthia},
  year = {2019},
  title = {Stop explaining black box machine learning models for high stakes decisions and use interpretable models instead},
  journal = {Nature Machine Intelligence},
  volume = {1},
  number = {5},
  pages = {206--215},
  doi = {10.1038/s42256-019-0048-x},
  url = {https://doi.org/10.1038/s42256-019-0048-x}
}

@article{hoffman2008kernel,
	author = {Thomas Hofmann and Bernhard Sch\"olkopf and Alexander J. Smola},
	title = {{Kernel methods in machine learning}},
	volume = {36},
	journal = {The Annals of Statistics},
	number = {3},
	publisher = {Institute of Mathematical Statistics},
	pages = {1171 -- 1220},
	year = {2008},
	doi = {10.1214/009053607000000677},
	URL = {https://doi.org/10.1214/009053607000000677}
}

@article{demsar2006statistical,
  author  = {Janez Dem{\v{s}}ar},
  title   = {Statistical Comparisons of Classifiers over Multiple Data Sets},
  journal = {Journal of Machine Learning Research},
  year    = {2006},
  volume  = {7},
  number  = {1},
  pages   = {1--30},
  url     = {http://jmlr.org/papers/v7/demsar06a.html}
}

@article{benavoli2016should,
  author  = {Alessio Benavoli and Giorgio Corani and Francesca Mangili},
  title   = {Should We Really Use Post-Hoc Tests Based on Mean-Ranks?},
  journal = {Journal of Machine Learning Research},
  year    = {2016},
  volume  = {17},
  number  = {5},
  pages   = {1--10},
  url     = {http://jmlr.org/papers/v17/benavoli16a.html}
}

@inproceedings{jansen2024statistical,
	author = {Jansen, Christoph and Schollmeyer, Georg and Rodemann, Julian and Blocher, Hannah and Augustin, Thomas},
	booktitle = {Advances in Neural Information Processing Systems},
	editor = {A. Globerson and L. Mackey and D. Belgrave and A. Fan and U. Paquet and J. Tomczak and C. Zhang},
	pages = {98143--98179},
	publisher = {Curran Associates, Inc.},
	title = {Statistical Multicriteria Benchmarking via the {GSD}-Front},
    address = {Vancouver, CA},
	volume = {37},
	year = {2024},
	url = {https://proceedings.neurips.cc/paper_files/paper/2024/file/b1f140eeee243db24e9e006481b91cf1-Paper-Conference.pdf},
}

@article{saket2019task,
  author={Saket, Bahador and Endert, Alex and Demiralp, {\c C}a{\u g}atay},
  journal={IEEE Transactions on Visualization and Computer Graphics}, 
  title={Task-Based Effectiveness of Basic Visualizations}, 
  year={2019},
  volume={25},
  number={7},
  pages={2505-2512},
  doi={10.1109/TVCG.2018.2829750},
  url={https://doi.org/10.1109/TVCG.2018.2829750},
}

@article{romanvillaran2022pites,
	author = {Rom{\'a}n-Villar{\'a}n, Esther and Alvarez-Romero, Celia and Mart{\'i}nez-Garc{\'i}a, Alicia and Escobar-Rodríguez, Germ{\'a}n Antonio and Garc{\'i}a-Lozano, Mar{\'i}a Jos{\'e} and Bar{\'o}n-Franco, Bosco and Moreno-Gavi{\~n}o, Lourdes and Moreno-Conde, Jesus and Rivas-Gonzalez, Jos{\'e} Antonio and Parra-Calder{\'o}n, Carlos Lu{\'i}s},
	title = {A Personalized Ontology-Based Decision Support System for Complex Chronic Patients: Retrospective Observational Study},
	year = {2022},
	journal = {JMIR Formative Research},
	volume = {6},
	number = {8},
	pages = {},
	doi = {10.2196/27990},
	url = {https://doi.org/10.2196/27990},
}

@article{espin2016nutelcare,
	author = {Esp\'in, Vanesa and Hurtado, María V. and Noguera, Manuel},
	title = {Nutrition for Elder Care: A nutritional semantic recommender system for the elderly},
	journal = {Expert Systems},
	volume = {33},
	number = {2},
	pages = {201-210},
	doi = {10.1111/exsy.12143},
	url = {https://doi.org/10.1111/exsy.12143},
	year = {2016}
}

@inproceedings{vercelli2017myaha,
	author={Vercelli, Alessandro and Rainero, Innocenzo and De Rosario, Helios and Summers, Mathew and Wieching, Rainer and Aumayr, Georg and Bandelow, Stephan and Ciferri, Ludovico and Bazzani, Marco},
	booktitle={25th International Conference on Software, Telecommunications and Computer Networks (SoftCOM)}, 
	title={{My-Active and Healthy Ageing (My-AHA)}: An {ICT} platform to detect frailty risk and propose intervention}, 
	year={2017},
	volume={},
	number={},
	pages={1-4},
    publisher={IEEE},
    address={New York, USA},
	doi={10.23919/SOFTCOM.2017.8115505},
	url={https://doi.org/10.23919/SOFTCOM.2017.8115505},
}

@article{azmi2019collaborative,
	author = {Azmi, Aini Khairani and Abdullah, Noraswaliza Binti and Emran, Nurul Akmar},
	title = {A collaborative filtering recommender system model for recommending intervention to improve elderly well-being},
	year = {2019},
	journal = {International Journal of Advanced Computer Science and Applications},
	volume = {10},
	number = {6},
	pages = {131 - 138},
	doi = {10.14569/ijacsa.2019.0100619},
	url = {https://doi.org/10.14569/ijacsa.2019.0100619},
}

@article{zacharaki2020frailsafe,
    author={Zacharaki, Evangelia I. and Deltouzos, Konstantinos and Kalogiannis, Spyridon and Kalamaras, Ilias and Bianconi, Luca and Degano, Cristiana and Orselli, Roberto and Montesa, Javier and Moustakas, Konstantinos and Votis, Konstantinos and Tzovaras, Dimitrios and Megalooikonomou, Vasileios},
    journal={IEEE Journal of Biomedical and Health Informatics}, 
    title={FrailSafe: An ICT Platform for Unobtrusive Sensing of Multi-Domain Frailty for Personalized Interventions}, 
    year={2020},
    volume={24},
    number={6},
    pages={1557-1568},
    doi={10.1109/JBHI.2020.2986918},
    url={https://doi.org/10.1109/jbhi.2020.2986918},
}

@article{angelini2022nestore,
	author =  {Angelini, Leonardo and El Kamali, Mira and Mugellini, Elena and Abou Khaled, Omar and Röcke, Christina and Porcelli, Simone and Mastropietro, Alfonso and Rizzo, Giovanna and Boqué, Noemi and del Bas, Josep Maria and Palumbo, Filippo and Girolami, Michele and Crivello, Antonino and Ziylan, Canan and Subías-Beltrán, Paula and Orte, Silvia and Standoli, Carlo Emilio and Fernandez Maldonado, Laura and Caon, Maurizio and Sykora, Martin and Elayan, Suzanne and Guye, Sabrina and Andreoni, Giuseppe},
	title =   {The {NESTORE e-Coach}: Designing a Multi-Domain Pathway to Well-Being in Older Age},
	journal = {Technologies},
	volume =  {10},
	year =    {2022},
	number =  {2},
    pages={28},
	issn =    {2227-7080},
	doi =     {10.3390/technologies10020050},
	url =     {https://doi.org/10.3390/technologies10020050},
}

@inproceedings{frikha2024senselife,
    author={Frikha, Ghassen and Lorca, Xavier and Pingaud, Hervé and Taweel, Adel and Bortolaso, Christophe and Borgiel, Katarzyna and Lamine, Elyes},
    booktitle={21st International Conference on Computer Systems and Applications (AICCSA)}, 
    title={Leveraging Service Supply Dynamics in Senselife: Building an Explainable Recommender System for Tailored Frailty Prevention}, 
    year={2024},
    volume={},
    number={},
    pages={1-6},
    publisher={IEEE},
    address={New York, USA},
    doi={10.1109/AICCSA63423.2024.10912596},
    url={https://doi.org/10.1109/AICCSA63423.2024.10912596},
}

@article{llorente2024games,
	author = {Llorente, {\'A}lvaro and del Rio, Alberto and Can Semerci, Yusuf and Alfonso Kurano, Jorge and Jim{\'e}nez, David and Menendez, Jose Manuel},
	title = {Assessment of cognitive games to improve the quality of life of {Parkinson’s and Alzheimer’s} patients},
	year = {2024},
	journal = {Digital Health},
	volume = {10},
	pages = {},
	doi = {10.1177/20552076241254733},
	url = {https://doi.org/10.1177/20552076241254733},
}

@article{kolakowski2025careup,
	author = {Kolakowski, Marcin and Lupica, Andrea and Ben Bader, Seif and Djaja-Josko, Vitomir and Kolakowski, Jerzy and Cichocki, Jacek and Ayadi, Jaouhar and Gilardi, Luca and Consoli, Angelo and Mocanu, Irina Georgiana and Cramariuc, Oana Teodora and Ferrazzini, Lionello and Reithner, Eva and Velciu, Magdalena and Borgogni, Barbara and Rivaira, Sofia and Leonzi, Sara and Cucchieri, Giacomo and Stara, Vera},
	title = {{CAREUP}: An Integrated Care Platform with Intrinsic Capacity Monitoring and Prediction Capabilities},
	year = {2025},
	journal = {Sensors},
	volume = {25},
	number = {3},
	pages = {},
	doi = {10.3390/s25030916},
	url = {https://doi.org/10.3390/s25030916},
}

@article{wu2023lime,
	author = {Wu, Yuanyuan and Zhang, Linfei and Bhatti, Uzair Aslam and Huang, Mengxing},
	title = {Interpretable Machine Learning for Personalized Medical Recommendations: A {LIME}-Based Approach},
	year = {2023},
	journal = {Diagnostics},
	volume = {13},
	number = {16},
	pages = {},
	doi = {10.3390/diagnostics13162681},
	url = {https://doi.org/10.3390/diagnostics13162681},
}

@inproceedings{wu2023shap,
    author={Wu, Yuanyuan and Zhang, Linfei and Bhatti, Uzair Aslam and Huang, Mengxing and Zhang, Yu},
    booktitle={6th International Conference on Pattern Recognition and Artificial Intelligence (PRAI)}, 
    title={Interpretable Medical Recommendations Based on {SHAPs}}, 
    year={2023},
    volume={},
    number={},
    pages={977-983},
    publisher={IEEE},
    address={New York, USA},
    doi={10.1109/PRAI59366.2023.10332022},
    url={https://doi.org/10.1109/PRAI59366.2023.10332022},
}

@inproceedings{murakami2010prevention,
	author={Murakami, Kazumasa and Shibano, Tomomi and Fujimoto, Yasunari and Yamaguchi, Torn},
	booktitle={Proceedings of Society of Instrument and Control Engineers (SICE) Annual Conference 2010}, 
	title={Information recommendation system for the care prevention using a communication robot}, 
	year={2010},
	volume={},
	number={},
	pages={388-389},
    publisher={IEEE},
    address={New York, USA},
	doi={},
	url={https://ieeexplore.ieee.org/document/5603071},
}

@inproceedings{sasaki2013walking,
	author={Sasaki, Wataru and Takama, Yasufumi},
	booktitle={Conference on Technologies and Applications of Artificial Intelligence}, 
	title={Walking Route Recommender System Considering {SAW} Criteria}, 
	year={2013},
	volume={},
	number={},
	pages={246-251},
    publisher={IEEE},
    address={New York, USA},
	doi={10.1109/TAAI.2013.56},
	url={https://doi.org/10.1109/TAAI.2013.56},
}

@inproceedings{thakur2018smarthome,
	author={Thakur, Nirmalya and Han, Chia Y.},
	booktitle={9th Annual Information Technology, Electronics and Mobile Communication Conference (IEMCON)}, 
	title={A Context-Driven Complex Activity Framework for Smart Home}, 
	year={2018},
	volume={},
	number={},
	pages={801-806},
    publisher={IEEE},
    address={New York, USA},
	doi={10.1109/IEMCON.2018.8615079},
	url={https://doi.org/10.1109/IEMCON.2018.8615079},
}

@inproceedings{mishra2020smarthome,
	author={Mishra, Prabhat and Busetty, Shiva Murthy and Kumar Gudla, Suresh},
	booktitle={6th World Forum on Internet of Things (WF-IoT)}, 
	title={Enhanced Activity Recognition of the {IoT} Smart Home users through Cluster Analysis}, 
	year={2020},
	volume={},
	number={},
	pages={1-6},
    publisher={IEEE},
    address={virtual},
	doi={10.1109/WF-IoT48130.2020.9221177},
	utl={https://doi.org/10.1109/WF-IoT48130.2020.9221177},
}

@inproceedings{shinde2021iotcars,
	author={Shinde, Komal and Aldaghamin, Areej and Tumar, Iyad and Awad, Abdalkarim and Wolff, Carsten},
	booktitle={11th International Conference on Intelligent Data Acquisition and Advanced Computing Systems: Technology and Applications (IDAACS)}, 
	title={An {IoT}-based Context-Aware Recommender System to Improve the Quality of Life of Elderly People}, 
	year={2021},
	volume={1},
	number={},
	pages={202-206},
    publisher={IEEE},
    address={New York, USA},
	doi={10.1109/IDAACS53288.2021.9660935},
	url={https://doi.org/10.1109/IDAACS53288.2021.9660935},
}

@inproceedings{sitparoopan2021homebridge,
	author={Sitparoopan, Thivyaroopy and Chellapillai, Vajira and Arulmoli, Jayawarman and Chandrasiri, Sanjeevi and Kugathasan, Archchana},
	booktitle={2nd International Informatics and Software Engineering Conference (IISEC)}, 
	title={{Home Bridge} - {Smart} Elderly Care System}, 
	year={2021},
	volume={},
	number={},
	pages={1-5},
    publisher={IEEE},
    address={New York, USA},
	doi={10.1109/IISEC54230.2021.9672411},
	url={https://doi.org/10.1109/IISEC54230.2021.9672411},
}

@inproceedings{dhivakar2022cancer,
	author={Dhivakar, SS and Sivaramakrishnan},
	booktitle={International Conference on Innovative Computing, Intelligent Communication and Smart Electrical Systems (ICSES)}, 
	title={Machine Learning based Cancer Disease Prediction}, 
	year={2022},
	volume={},
	number={},
	pages={1-5},
    publisher={IEEE},
    address={New York, USA},
	doi={10.1109/ICSES55317.2022.9914247},
	url={https://doi.org/10.1109/ICSES55317.2022.9914247},
}

@article{minakata2022walking,
	author = {Minakata, Mayuko and Maruyama, Tsubasa and Tada, Mitsunori and Ramasamy, Priyanka and Das, Swagata and Kurita, Yuichi},
	title = {Safe Walking Route Recommender Based on Fall Risk Calculation Using a Digital Human Model on a 3D Map},
	year = {2022},
	journal = {IEEE Access},
	volume = {10},
	pages = {8424 - 8433},
	doi = {10.1109/ACCESS.2022.3143322},
	url = {https://doi.org/10.1109/ACCESS.2022.3143322},
}

@article{martin2023digihealth,
	author = {Mart{\'i}n, Cristina and Amaya, Isabel and Torres, Jordi and Artola, Garazi and Garc{\'i}a, Meritxell and Garc{\'i}a-Navarro, Teresa and de Ramos, Ver{\'o}nica and Cort{\'e}s, Camilo A. and Kerexeta, Jon and Aguirre, Maia and M{\'e}ndez, Ariane and Unzueta, Luis and del Pozo, Arantza and Larburu, Nekane and Macia, Ivan},
	title = {{DigiHEALTH}: Suite of Digital Solutions for Long-Term Healthy and Active Aging},
	year = {2023},
	journal = {International Journal of Environmental Research and Public Health},
	volume = {20},
	number = {13},
	pages = {},
	doi = {10.3390/ijerph20136200},
	url = {https://doi.org/10.3390/ijerph20136200},
}

@inproceedings{zhou2024recipe,
	author={Zhou, Qianyi},
	booktitle={2nd International Conference on Sensors, Electronics and Computer Engineering (ICSECE)}, 
	title={Recipe Recommendation for the Elderly Based on Collaborative Filtering Algorithm and Neural Network}, 
	year={2024},
	pages={660-663},
    publisher={IEEE},
    address={New York, USA},
	doi={10.1109/ICSECE61636.2024.10729407},
	url={https://doi.org/10.1109/ICSECE61636.2024.10729407}
}

@inproceedings{prasetyo2024recipe,
	author={Prasetyo, Rizky Ferdian and Baizal, Z. K. A.},
	booktitle={International Conference on Communication, Networks and Satellite (COMNETSAT)}, 
	title={Ontology-Based Healthy Food Recommendation for the Elderly}, 
	year={2024},
	pages={76-81},
    publisher={IEEE},
    address={New York, USA},
	doi={10.1109/COMNETSAT63286.2024.10862504},
	url={https://doi.org/10.1109/COMNETSAT63286.2024.10862504}
}

@inproceedings{huang2024smarthome,
    author={Huang, Xiaoping and Palaoag, Thelma Domingo},
    booktitle={2nd International Conference on Control, Electronics and Computer Technology (ICCECT)}, 
    title={Implementation of Intelligent Elderly Care System Based on Cloud Platform and NNB Algorithm}, 
    year={2024},
    pages={841-846},
    publisher={IEEE},
    address={New York, USA},
    doi={10.1109/ICCECT60629.2024.10546115},
    url={https://doi.org/10.1109/ICCECT60629.2024.10546115},
}

@inproceedings{yuan2024shortvideo,
    author={Yuan, Xue and Zhang, Yanxin and Zhou, Qingying},
    booktitle={International Conference on Information Technology, Communication Ecosystem and Management (ITCEM)}, 
    title={Research on Recommendation Algorithm for Short Video Health Content Targeting Elderly Users}, 
    year={2024},
    publisher={IEEE},
    address={New York, USA},
    pages={74-79},
    doi={10.1109/ITCEM65710.2024.00022},
    url={https://doi.org/10.1109/ITCEM65710.2024.00022},
}

@article{bentlage2025mobile,
	author = {Bentlage, Ellen and del R{\'i}o Ponce, Alberto and Ahmed, Mona and Gangas, Pilar and Brach, Michael and Kurano, Jorge Alfonso and Menendez, Jose Manuel},
	title = {Cocreating a Mobile Health App Providing Physical Activity Recommendations for Older People Living With Parkinson Disease or Dementia: User-Centered Pilot Study},
	year = {2025},
	journal = {JMIR Formative Research},
	volume = {9},
	pages = {},
	doi = {10.2196/51831},
	url = {https://doi.org/10.2196/51831},
}

@inproceedings{jiang2025education,
	author = {Jiang, Fangyuan},
	title = {Personalized Recommendation Method of Health Education Resources in Home-based Aged Care Based on Similarity Correlation},
	year = {2025},
	isbn = {9798400713248},
	publisher = {ACM},
	address = {New York, NY, USA},
	doi = {10.1145/3732299.3732313},
	url = {https://doi.org/10.1145/3732299.3732313},
	booktitle = {Proceedings of the 2025 International Conference on Digital Education and Information Technology},
	pages = {69–73},
	location = {},
	series = {DEIT '25}
}

@article{tsoumakas2007multilabel,
	title={Multi-label classification: An overview},
	author={Tsoumakas, Grigorios and Katakis, Ioannis},
	journal={International Journal of Data Warehousing and Mining (IJDWM)},
	volume={3},
	number={3},
	pages={1--13},
	year={2007},
	publisher={IGI Global Scientific Publishing},
}

@inproceedings{cheng09icml,
	author = {Weiwei Cheng and Jens H\"uhn and Eyke H\"ullermeier},
	title = {Decision tree and instance-based learning for label ranking},
	booktitle = {Proceedings of the 26th International Conference on Machine Learning (ICML-09)},
	pages = {161--168},
	year = 2009,
	editor = {L\'{e}on Bottou and Michael Littman},
	address = {Montreal, Canada},
	month = {June},
	publisher = {Omnipress}
}

@incollection{furnkranz2011preference,
	author="F{\"u}rnkranz, Johannes and H{\"u}llermeier, Eyke", 
	editor="F{\"u}rnkranz, Johannes and H{\"u}llermeier, Eyke",
	title="Preference Learning: An Introduction",
	bookTitle="Preference Learning",
	year="2011",
	publisher="Springer Berlin Heidelberg",
	address="Berlin, Heidelberg",
	pages="1--17",
	isbn="978-3-642-14125-6",
	doi="10.1007/978-3-642-14125-6_1",
	url="https://doi.org/10.1007/978-3-642-14125-6_1"
}

@article{madjarov2012multilabel,
	title = {An extensive experimental comparison of methods for multi-label learning},
	journal = {Pattern Recognition},
	volume = {45},
	number = {9},
	pages = {3084-3104},
	year = {2012},
	note = {Best Papers of Iberian Conference on Pattern Recognition and Image Analysis (IbPRIA'2011)},
	issn = {0031-3203},
	doi = {10.1016/j.patcog.2012.03.004},
	url = {https://doi.org/10.1016/j.patcog.2012.03.004},
	author = {Gjorgji Madjarov and Dragi Kocev and Dejan Gjorgjevikj and Sašo Džeroski},
}

@incollection{herrera2016multilabel,
	author="Herrera, Francisco and Charte, Francisco and Rivera, Antonio J. and del Jesus, Mar{\'i}a J.",
	title="Case Studies and Metrics",
	bookTitle="Multilabel Classification : Problem Analysis, Metrics and Techniques",
	year="2016",
	publisher="Springer International Publishing",
	address="Cham",
	pages="33--63",
	isbn="978-3-319-41111-8",
	doi="10.1007/978-3-319-41111-8_3",
	url="https://doi.org/10.1007/978-3-319-41111-8_3"
}

@inproceedings{rivolli2017foodtruck,
	author="Rivolli, Adriano and Parker, Larissa C. and de Carvalho, Andre C. P. L. F.",
	editor="Oliveira, Eug{\'e}nio and Gama, Jo{\~a}o and Vale, Zita and Lopes Cardoso, Henrique",
	title="Food Truck Recommendation Using Multi-label Classification",
	booktitle="Progress in Artificial Intelligence",
	year="2017",
	publisher="Springer International Publishing",
	address="Cham",
	pages="585--596",
	isbn="978-3-319-65340-2"
}

@article{charte2018cometa,
    title = "Tips, guidelines and tools for managing multi-label datasets: The mldr.datasets {R} package and the {Cometa} data repository",
    author = "Francisco Charte and Antonio J. Rivera and David Charte and María J. del Jesus and Francisco Herrera",
    year = "2018",
    journal = "Neurocomputing",
    issn = "0925-2312",
    volume={4},
    pages={68--85},
    doi = "10.1016/j.neucom.2018.02.011",
    url = "https://doi.org/10.1016/j.neucom.2018.02.011",
}

@article{bogatinovski2022multilabel,
	title = {Comprehensive comparative study of multi-label classification methods},
	journal = {Expert Systems with Applications},
	volume = {203},
	pages = {117215},
	year = {2022},
	issn = {0957-4174},
	doi = {10.1016/j.eswa.2022.117215},
	url = {https://doi.org/10.1016/j.eswa.2022.117215},
	author = {Jasmin Bogatinovski and Ljupčo Todorovski and Sašo Džeroski and Dragi Kocev},
}

@inproceedings{fotakis2022labelranking,
  title = {Label Ranking through Nonparametric Regression},
  author = {Fotakis, Dimitris and Kalavasis, Alkis and Psaroudaki, Eleni},
  booktitle = {Proceedings of the 39th International Conference on Machine Learning},
  pages = {6622--6659},
  year = {2022},
  address={Baltimore, USA},
  editor = {Chaudhuri, Kamalika and Jegelka, Stefanie and Song, Le and Szepesvari, Csaba and Niu, Gang and Sabato, Sivan},
  volume = {162},
  series = {Proceedings of Machine Learning Research},
  month = {Jul},
  publisher = {PMLR},
  url = {https://proceedings.mlr.press/v162/fotakis22a.html},
}

@inproceedings{brinker2007ranking,
	author = {Brinker, Klaus and H\"ullermeier, Eyke},
	title = {Case-based multilabel ranking},
	year = {2007},
	publisher = {Morgan Kaufmann Publishers Inc.},
	address = {San Francisco, CA, USA},
	booktitle = {Proceedings of the 20th International Joint Conference on Artifical Intelligence},
	pages = {702–707},
	numpages = {6},
	location = {Hyderabad, India},
	series = {IJCAI'07}
}

@inproceedings{brinker2006unified,
	author = {Brinker, Klaus and F\"urnkranz, Johannes and H\"ullermeier, Eyke},
	title = {A Unified Model for Multilabel Classification and Ranking},
	year = {2006},
	isbn = {1586036424},
	publisher = {IOS Press},
	address = {NLD},
	booktitle = {Proceedings of the 2006 Conference on ECAI 2006: 17th European Conference on Artificial Intelligence August 29 -- September 1, 2006, Riva Del Garda, Italy},
	pages = {489–493},
	numpages = {5}
}

@article{zhou2018ranking,
	title = {Random forest for label ranking},
	journal = {Expert Systems with Applications},
	volume = {112},
	pages = {99-109},
	year = {2018},
	issn = {0957-4174},
	doi = {10.1016/j.eswa.2018.06.036},
	url = {https://doi.org/10.1016/j.eswa.2018.06.036},
	author = {Yangming Zhou and Guoping Qiu},
}

@incollection{legg2013diagram,
	author="Legg, Catherine",
	editor="Moktefi, Amirouche and Shin, Sun-Joo",
	title="What is a Logical Diagram?",
	bookTitle="Visual Reasoning with Diagrams",
	year="2013",
	publisher="Springer Basel",
	address="Basel",
	pages="1--18",
	isbn="978-3-0348-0600-8",
	doi="10.1007/978-3-0348-0600-8_1",
	url="https://doi.org/10.1007/978-3-0348-0600-8_1"
}

@article{dimara2022critical,
	author={Dimara, Evanthia and Stasko, John},
	journal={IEEE Transactions on Visualization and Computer Graphics}, 
	title={A Critical Reflection on Visualization Research: Where Do Decision Making Tasks Hide?}, 
	year={2022},
	volume={28},
	number={1},
	pages={1128-1138},
	doi={10.1109/TVCG.2021.3114813},
	url={https://doi.org/10.1109/TVCG.2021.3114813},
}

@article{brumar2025typology,
  author={Brumar, Camelia D. and Molnar, Sam and Appleby, Gabriel and Potter, Kristi and Chang, Remco},
  journal={IEEE Transactions on Visualization and Computer Graphics}, 
  title={A Typology of Decision-Making Tasks for Visualization}, 
  year={2025},
  volume={31},
  number={10},
  pages={8536-8551},
  doi={10.1109/TVCG.2025.3572842},
  url={https://doi.org/10.1109/TVCG.2025.3572842},
}

@inproceedings{amar2005taxonomy,
  author={Amar, Robert. A. and Eagan, James R. and Stasko, John},
  booktitle={IEEE Symposium on Information Visualization (INFOVIS)}, 
  title={Low-level components of analytic activity in information visualization}, 
  year={2005},
  volume={},
  number={},
  publisher={IEEE},
  address={Minneapolis, USA},
  pages={111-117},
  doi={10.1109/INFVIS.2005.1532136},
  utl={https://doi.org/10.1109/INFVIS.2005.1532136},
}

@article{unwin2021iris,
	author = {Unwin, Antony and Kleinman, Kim},
	title = {The Iris Data Set: In Search of the Source of Virginica},
	journal = {Significance},
	volume = {18},
	number = {6},
	pages = {26-29},
	year = {2021},
	month = {11},
	issn = {1740-9705},
	doi = {10.1111/1740-9713.01589},
	url = {https://doi.org/10.1111/1740-9713.01589},
}

@article{alamdari2020ecommerce,
	author={Alamdari, Pegah Malekpour and Navimipour, Nima Jafari and Hosseinzadeh, Mehdi and Safaei, Ali Asghar and Darwesh, Aso},
	journal={IEEE Access}, 
	title={A Systematic Study on the Recommender Systems in the E-Commerce}, 
	year={2020},
	volume={8},
	number={},
	pages={115694-115716},
	doi={10.1109/ACCESS.2020.3002803},
	url={https://doi.org/10.1109/ACCESS.2020.3002803},
}

@article{jannach2025rs4good,
	author = {Jannach, Dietmar and Said, Alan and Tkalcic, Marko and Zanker, Markus},
	title = {Recommender Systems for Good (RS4Good): Survey of Use Cases and a Call to Action for Research that Matters},
	year = {2025},
	publisher = {Association for Computing Machinery},
    volume={3},
    issue={2},
    numpages={21},
	address = {New York, NY, USA},
	doi = {10.1145/3746648},
	url = {https://doi.org/10.1145/3746648},
	journal = {ACM Trans. Recomm. Syst.},
	month = jun,
}

\appendix
\section{Three Psychometric Instruments Used in Gerontological Research}
\label{section:background:cga:instruments}

In recent years, there has been a shift in focus in health care: patients and their families want to be involved in the decision-making process.
A central piece of this patient-centred care movement, which enables care professionals to gain insight about the patients' perspective about their health, is the development of reliable \sigladef{patient reported outcome measures}{PROMs} and their incorporation into the traditional ways of measuring health and the effects of treatment on the patient \cite{krabbe2016measurement}.
These measures also play a significant role in initiatives to promote early interventions as a more effective means of optimising the health trajectories of individuals from a longitudinal perspective.
The techniques reviewed in Section \ref{section:background:measurement-of-health} are instrumental to the success of these efforts.

In this appendix, the structure of three such instruments is briefly described: the WHOQOL-BREF instrument for quality of life, the WHO's instrument for intrinsic capacity, and the AMPI-AB instrument for fragility.
For each instrument, we briefly review the major concerns that lead to its development, the factor-analytic structure of its underlying model, and its scoring procedure.

\subsection{Measuring Quality of Life}
\label{section:background:whoqol}

The WHOQOL-BREF instrument is a shorter version of the WHOQOL-100, which was developed by the WHO in the 1990s in response to an increasing demand for subjective measures of health and well-being that could be applied to populations worldwide.
The WHO defines quality of life as \aspas{an individual’s perception of their position in life in the context of the culture and value systems in which they live, and in relation to their goals, expectations, standards and concerns} \cite{skevington2004whoqol}.

The instrument has a questionnaire with 26 items that survey four domains: physical health (7 items), psychological well-being (6 items), social relationships (3 items), and environmental quality (8 items).
The remaining two items survey the general perception of health and quality of life \cite{orley1996whoqolbref}.
The respondent rates their agreement with the statement of an item on a five-point Likert scale, whose labels are adapted for each domain, as shown in Table \ref{tab:background:whoqol-bref}.
Responses are encoded as numeric values from one to five.


The instrument was developed using a second order confirmatory factor analysis model similar to the one illustrated in Figure \ref{fig:background:selvm:cfa2}.
The 2012 revision of the WHOQOL manual reports the following effect coefficients: 0.84 for the physical domain, 1.0 for the psychological domain, 0.85 to the social domain, and 0.77 for the environment domain \cite{who2012whoqolmanual}.
The reliability was calculated for each domain separately using the Cronbach's alpha (instead of the McDonald's omega) and found to demonstrate good internal consistency (values varying between 0.66 and 0.84).

\begin{table}[tpb]
    \centering
    \scriptsize
    \caption{Questionnaire of the WHOQOL-BREF instrument}
    
    \begin{tabularx}{\linewidth}{@{}P{0.55cm}P{1.85cm}XP{4.8cm}@{}}
    \toprule
        Ord & Domain & Item phrasing & Scale (extremes)\\
    \toprule

1 & general & How would you rate your quality of life? & Very poor to Very good \\ \midrule
2 & general & How satisfied are you with your health? & Very dissatisfied to Very satisfied \\ \midrule
3* & physical & To what extent do you feel that physical pain prevents you from doing what you need to do? & Not at all to An extreme amount \\ \midrule
4* & physical & How much do you need any medical treatment to function in your daily life? & Not at all to An extreme amount \\ \midrule
5 & psychological & How much do you enjoy life? & Not at all to An extreme amount \\ \midrule
6 & psychological & To what extent do you feel your life to be meaningful? & Not at all to An extreme amount \\ \midrule
7 & psychological & How well are you able to concentrate? & Not at all to Extremely \\ \midrule
8 & environment & How safe do you feel in your daily life? & Not at all to Extremely \\ \midrule
9 & environment & How healthy is your physical environment? & Not at all to Extremely \\ \midrule
10 & physical & Do you have enough energy for everyday life? & Not at all to Completely \\ \midrule
11 & psychological & Are you able to accept your bodily appearance? & Not at all to Completely \\ \midrule
12 & environment & Have you enough money to meet your needs? & Not at all to Completely \\ \midrule
13 & environment & How available to you is the information that you need in your day-to-day life? & Not at all to Completely \\ \midrule
14 & environment & To what extent do you have the opportunity for leisure activities? & Not at all to Completely \\ \midrule
15 & physical & How well are you able to get around? & Very poor to Very good \\ \midrule
16 & physical & How satisfied are you with your sleep? & Very dissatisfied to Very satisfied \\ \midrule
17 & physical & How satisfied are you with your ability to perform your daily living activities? & Very dissatisfied to Very satisfied \\ \midrule
18 & physical & How satisfied are you with your capacity for work? & Very dissatisfied to Very satisfied \\ \midrule
19 & psychological & How satisfied are you with yourself? & Very dissatisfied to Very satisfied \\ \midrule
20 & social & How satisfied are you with your personal relationships? & Very dissatisfied to Very satisfied \\ \midrule
21 & social & How satisfied are you with your sex life? & Very dissatisfied to Very satisfied \\ \midrule
22 & social & How satisfied are you with the support you get from your friends? & Very dissatisfied to Very satisfied \\ \midrule
23 & environment & How satisfied are you with the conditions of your living place? & Very dissatisfied to Very satisfied \\ \midrule
24 & environment & How satisfied are you with your access to health services? & Very dissatisfied to Very satisfied \\ \midrule
25 & environment & How satisfied are you with your transport? & Very dissatisfied to Very satisfied \\ \midrule
26* & psychological & How often do you have negative feelings such as blue mood, despair, anxiety, depression? & Never to Always \\        
        
    \bottomrule
    
    \end{tabularx}
    
    \label{tab:background:whoqol-bref}
\end{table}

The instrument has a formal scoring procedure for the domains but does not specify a single score for quality of life.
This is indicative that the four domain scores should be interpreted as a (multidimensional) profile of quality of life instead of a unidimensional measure.
The scoring rules are: (a) reverse the score of the three negatively phrased items, and (b) average the responses obtained for each domain and multiply the average by four (to produce scores that are compatible with the WHOQOL-100 instrument).
The averaged score is proportional to the sum-score (see Equation \ref{eq:background:weighted-score}) and therefore both methods are equivalent for the purpose of ordering of people on a given scale.
The instrument also has rules to handle missing data, but it suffices to say that that scoring procedure works even if one response per domain is missing.

\subsection{Measuring Intrinsic Capacity}
\label{section:background:whoic}

As we discussed briefly in Section \ref{section:background:cga}, the WHO has been advancing a change in national public health policies in response to the fast demographic changes that are underway worldwide.
This change consists in shifting the focus on disease management to an approach that seeks to preserve health of individuals to the older age.
According to \cite{carvalho2017operationalising}, \aspas{instead of trying to manage an array of diseases and treat specific symptoms in a disjointed fashion, interventions should be prioritised in ways that optimise trajectories of older people’s physical and mental capacities.}
An essential component of this strategy is the development of an instrument to assess intrinsic capacity, so that public policies can be evaluated for their effectiveness in preserving intrinsic capacity of populations served by national health systems across the life course of their members.

In this framework, intrinsic capacity is defined as \aspas{the composite of all the
physical and mental capacities of an individual}, which is essential for their everyday functioning \cite{who2015report}.
This construct has been operationalised by the \sigladef{Integrated Care for Older People}{ICOPE} initiative mentioned earlier \cite{tavassoli2022ICOPE,hsiao2024icworks}, and encompasses five domains: sensory function, cognition, vitality, locomotion, and psychological well-being.

A recent study to assess the psychometric properties of an instrument to assess intrinsic capacity of Brazilian populations was conducted by \cite{aliberti2022ic}.
The authors used data from the baseline assessment of the ongoing \sigladef{Brazilian Longitudinal Study of Ageing}{ELSI-Brazil} project to compute intrinsic capacity profiles for the surveyed populations.
The resulting questionnaire, which is shown in Table \ref{tab:background:whoic}, is composed of items that were submitted to the study participant (e.g., Who is the current president of Brazil?) and physical tests (e.g., time the participant takes to walk three metres at the usual pace).
The first column of the table indicates the code(s) of the item(s) in the original ELSI's questionnaire \cite{limacosta2018elsi}.
An asterisk after the code indicates it must be reversed during the scoring procedure.
Note that several items in this questionnaire come from scales widely used in healthcare research, e.g., the 8-items CES-D instrument for depression (items r2 to r9), and the physical tests.

\begin{table}[tpb]
    \centering
    \scriptsize
    \caption{Questionnaire to assess the domains of intrinsic capacity}
    
    \begin{tabularx}{\linewidth}{@{}P{0.85cm}P{1.91cm}XP{5.2cm}@{}}
    \toprule
        Src & Domain & Item phrasing or Test & Scale \\
    \toprule

q7:10 & \makecell[c]{cognitive} & Orientation for year, month, day, and day of the week. & count of correct responses \\ \midrule
q13 & \makecell[c]{cognitive} & Delayed recall of 10 common words. & count of correct responses \\ \midrule
q18 & \makecell[c]{cognitive} & What do people usually use to cut a paper? & count of correct responses \\ \midrule
q19 & \makecell[c]{cognitive} & What is the plant that has a long and green leaf that gives a yellow and long fruit and that we peel to eat it? & count of correct responses \\ \midrule
q20 & \makecell[c]{cognitive} & Who is the current president of Brazil? & count of correct responses \\ \midrule

q21 & \makecell[c]{cognitive} & Who is the current vice-president of Brazil? & count of correct responses \\ \midrule
q14 & \makecell[c]{cognitive} & Distinct animals in 60 seconds. & count of correct responses \\ \midrule
r2* & \makecell[c]{psychological} & During the last week, have you felt depressed most of the time? & Yes, No \\ \midrule
r3* & \makecell[c]{psychological} & During the last week, have you felt like things were harder? & Yes, No \\ \midrule
r4* & \makecell[c]{psychological} & During the last week, have you felt like your sleep wasn't restful most of the time? & Yes, No \\ \midrule
r5  & \makecell[c]{psychological} & Did you feel happy most of the time? & Yes, No \\ \midrule
r6* & \makecell[c]{psychological} & Did you feel lonely most of the time? & Yes, No \\ \midrule
r7  & \makecell[c]{psychological} & Did you enjoy or enjoy life most of the time? & Yes, No \\ \midrule
r8* & \makecell[c]{psychological} & Did you feel sad most of the time? & yes, no \\ \midrule
r9* & \makecell[c]{psychological} & Did you feel like you couldn't get things done? & Yes, No \\ \midrule

n74* & \makecell[c]{psychological} & How would you evaluate the quality of your sleep? & Excellent to Very poor \\ \midrule
n75* & \makecell[c]{psychological} & During the last month, have you taken any sleeping pill? & No, Less than once a week, Btw 1-2 times a week, 3+ times a week. \\ \midrule
n16* & \makecell[c]{sensory} & How do you evaluate your hearing (even when using a hearing device)? & Excellent to Very poor \\ \midrule
n6* & \makecell[c]{sensory} & How good is your eyesight (even when using glasses or contact lenses) for seeing things at a distance, like recognizing a friend across the street? & Excellent to Very poor \\ \midrule
n7* & \makecell[c]{sensory} & How good is your eyesight (even when using glasses or contact lenses) for seeing things up close like reading ordinary newspaper print?  & Excellent to Very poor \\ \midrule

mf33- mf38* & \makecell[c]{locomotor} & Time to walk three meters at the usual pace with or without assistive devices. & in meter/sec, discretised \\ \midrule
mf30- mf32 & \makecell[c]{locomotor} & Balance test from the Short Physical Performance Battery. & in sec, adjusted per age group, discretised \\ \midrule
mf27- mf29 & \makecell[c]{vitality} & Handgrip strength of the dominant hand evaluated using a Sachan handheld dynamometer. & in kg, adjusted per sex and body mass, discretised \\ \midrule
n69* n70* & \makecell[c]{vitality} & Unintentional weight loss 3kg or more during the last 3 months. & Yes, No \\ \midrule
n72* & \makecell[c]{vitality} & In the past week, how often did you feel that you could not carry things forward? & Never or rarely (less than 1 day), Very few times (1-2 days), Sometimes (3-4 days), Most of the time \\ \midrule

n73* & \makecell[c]{vitality} & In the past week, how often did the routine activities require a major effort to be completed? & Never or rarely (less than 1 day), Very few times (1-2 days), Sometimes (3-4 days), Most of the time \\ 

    \bottomrule
    
    \end{tabularx}
    
    \label{tab:background:whoic}
\end{table}

An early version of this instrument was conceived using a second order confirmatory factor analysis model similar to the one illustrated in Figure \ref{fig:background:selvm:cfa2}.
\cite{carvalho2017operationalising} reported the following coefficients: 
0.45 for sensory function, 
0.64 for cognition, 
0.59 for vitality, 
0.95 for locomotion, 
and 0.57 for psychological well-being.
No assessments of reliability were reported, as is usual in the initial stages of development of a new instrument.
Following a more recent trend, \cite{aliberti2022ic} used a bifactor model to evaluate the instrument.
They concluded that the model revealed satisfactory robust goodness-of-fit indices compared to other factor models.
No standard scoring procedure was found in the surveyed articles, as would be expected as the instrument is still under development.
It must be noted, however, that the wide variety of scales must be taken into account in the scoring procedure, to ensure that indicators of a given domain do not dominate the others because the numeric values (arbitrarily) assigned to its levels vary over different ranges.

\subsection{Measuring Frailty}
\label{section:background:ampiab}

In response to an increasingly older population in Brazil, and in agreement with the need for improvements in national health care systems to provide integrated care for this population, as advanced by the WHO, the Municipal Health Department of the city of São Paulo has instituted the 
\sigladef{Elderly Caregiver Program}{PAI} programme.
This is a home care programme that offers clinically frail and socially vulnerable older people \aspas{the services of health care professionals and professional caregivers aiming at rehabilitation, maintenance/improvement of selfcare, and socialization} \cite{marcucci2020ampiab}.

The \sigladef{Multidimensional Evaluation of Older People in Primary Care}{AMPI-AB} instrument is used by the PAI programme as one of the criteria to include people to be assisted.
This instrument is under development, but a recent work has shown that most of its items cluster around five domains: cognitive (3 items), \sigladef{activities of daily living}{ADL} (4 items), \sigladef{instrumental activities of daily living}{IADL} (4 items), oral health (4 items), and morbidities (3 items) \cite[Table 5]{andrade2019ampiab}.
The respondent rates their agreement with the statement of an item using several scales, as shown in Table \ref{tab:background:ampiab}.
To the best of my knowledge, a confirmatory factor analysis was not performed on this instrument, but \cite{andrade2019ampiab} reports the results of an exploratory factor analysis in which the target construct seems to be some notion of frailty, since the score an individual obtains is used to classify her into one of three classes: healthy (0 to 5 points), pre-frail (6 to 10 points), or frail (more than 10 points).
The instrument uses the sum-score from Equation \ref{eq:background:weighted-score}.
With regard to reliability, \cite{andrade2019ampiab} reports a value of $0.79$ for the Cronbach's $\alpha$ and of $0,78$ for the McDonald's $\omega$, which indicates an acceptable level for early stages of research.
However, several measures of closeness to unidimensionality suggest that the data collected with this instrument will fit poorly to a congeneric model: a value of $0.84$ is reported for UNICO (Unidimensional Congruence), among others.

\begin{table}[tpb]
    \centering
    \scriptsize
    \caption{Questionnaire of the AMPI-AB instrument}
    
    \begin{tabularx}{\linewidth}{@{}P{0.65cm}P{1.85cm}XP{4.8cm}@{}}
    \toprule
        Ord & Domain & Item phrasing & Scale (extremes)\\
    \toprule

2* & morbidities & In general, compared to other people your age, would you say your health is: & Very poor to Very good \\ \midrule
4 & morbidities & Have you had/do you have any of the conditions below? & count of reported items from a predefined list\\ \midrule
5 & morbidities & How many medications do you take daily? & count of reported items \\ \midrule
11a & cognitive & Has a family member or friend told you that you are becoming forgetful? & Yes, No \\ \midrule
11b & cognitive & Has your forgetfulness worsened in recent months? & Yes, No \\ \midrule
11c & cognitive & Is your forgetfulness preventing you from performing any daily activities? & Yes, No \\ \midrule
13a & ADL & Do you need help getting out of bed? & Yes, No \\ \midrule
13b & ADL & Do you need help getting dressed? & Yes, No \\ \midrule
13c & ADL & Do you need help eating? & Yes, No \\ \midrule
13d & ADL & Do you need help bathing? & Yes, No \\ \midrule
10c* & IADL & Can you walk 400 meters (approximately four blocks)? & Yes, No \\ \midrule
10d* & IADL & Can you sit or stand up without difficulty? & Yes, No \\ \midrule
14a & IADL & Do you need help performing activities outside the home? & Yes, No \\ \midrule
14b & IADL & Do you need help managing money (paying bills, checking change, going to the bank, etc.)? & Yes, No \\ \midrule
17a & oral health & If you wear dentures, are they poorly fitted? & Yes, No \\ \midrule
17b & oral health & Do you have trouble chewing? & Yes, No \\ \midrule
17c & oral health & Do you have trouble swallowing? & Yes, No \\ \midrule
17d & oral health & Have you stopped eating any foods because of problems with your teeth or dentures? & Yes, No \\ 

    \bottomrule
    
    \end{tabularx}
    
    \label{tab:background:ampiab}
\end{table}

\section{A Link Between Sum-scores and Area-scores}
\label{appendix:areascore}

The assessment polygon is a visual element of the Polygrid diagram, as indicated in Figure \ref{fig:proposal:ml-task:default}.
Under some assumptions, the area of an assessment polygon has a monotonic relationship with the sum-score of the assessment.
This relationship is important for two reasons: (a) it justifies the equivalence between the sum-score and the area-score for the purpose of ordering subjects with respect to a latent variable, and (b) it warrants the use of the assessment polygon as a visual analogue of the measurand, for the same purpose.

The remainder of this appendix is organised as follows. 
Section \ref{ap:sum-area:congeneric} revisits the congeneric model to show how its sum-scores are used to order subjects with respect to a latent variable.
The notion of area-score is introduced in Section \ref{ap:sum-area:area-score}, and the equivalence between sum- and ara-scores is deduced in Section \ref{ap:sum-area:proof}.
Finally, Section \ref{ap:sum-area:evidence} presents empirical evidence in support of this relationship.

\subsection{The Structure of Data Collected with a Psychometric Instrument}
\label{ap:sum-area:congeneric}

Following the notation established in Section \ref{section:proposal:data}, let $\mathring{X}$ be an $(m,d)$-matrix with the scores obtained by $m$ individuals who were evaluated using a psychometric instrument with $d$ domains, with $d>2$.
Thus, $\mathring{x}_{ik}$ denotes the score obtained by the $i$-th individual on the $k$-th domain surveyed by that instrument.
We assume the data were collected with an instrument whose subscales were validated against a congeneric factor model.
Based on this, I consider valid each of the following statements:
\begin{itemize}

    \item \standing{Each subject is associated with some level $\eta_i$ of the measurand (i.e., the attribute that the latent variable tracks in people), which cannot be directly observed;}{congeneric model}

    \item  \standing{The measurand is assumed to exert causal power on the cognitive processes a person relies on when he or she responds to the items in the instrument's questionnaire.}{congeneric model, causal account}
    Of course, the response given to an item by a subject is an observable variable.
    Formally, $\mathring{x}_{ik} = f_k(\eta_i)$;

    \item \standing{The relationship between a latent variable $\eta_i$ and the response given to each of its respective items is linear.}{congeneric model}
    Moreover, \premise{responses given to distinct items may be more or less influenced by the latent variable.}{congeneric model}
    Thus $f_k(\eta_i) = \lambda_k \eta_i$, where $\lambda_k$ represents the direct effect of $\eta_i$ on $\mathring{x}_{ik}$;

    \item \standing{Responses given to items are subject to noise.}{measurement error has negligible effects on ordering subjects}
    Any deviations from the expected linear response are accounted for as measurement error ($\epsilon_{ik}$).
    Thus, we arrive at the standard measurement equation formulated at the individual level:
    \begin{equation} 
        \label{eq:std-measurement-model}
        \mathring{x}_{ik} = \lambda_k \eta_i + \epsilon_{ik} \,.
    \end{equation}
    
\end{itemize}

The ability to order subjects according to their standing on a latent variable is a valued property of psychometric instruments.
It is useful to stratify a target population with respect to some risk \cite{silva2014whoqolcut,marcucci2020ampiab,aliberti2022ic} or \fix{to assess the effect of interventions in a clinical trial \cite{philippi2023challenges}}{cite a more focused work}.
For instance, in Figure \ref{fig:whoqol-assessments-24-72}, how should Participants 030 and 089 be ordered according to their assessments of quality of life?
Which participant has a \aspas{higher} quality of life?
Essentially, this task has two common approaches.
The first is to sum up the products of all item responses by their respective factor loadings to obtain a weighted score $\hat{\eta}_i = \sum_{k} \lambda_{k} \mathring{x}_{ik}$, as in Equation \ref{eq:background:weighted-score}.
When an instrument' scoring procedure uses this method, this suggests that the measurement error is negligible for the purpose of ordering people.
In this case, Equation \ref{eq:std-measurement-model} is reduced to $\mathring{x}_{ik} \approx \lambda_k \eta_i$, and also $\hat{\eta}_i \approx \sum_{k} \lambda_{k} (\lambda_k \eta_i) = \eta_i \sum_{k} \lambda_{k}^2 = C \eta_i$, with $C$ being an arbitrary positive constant.
Thus, if two subjects $A$ and $B$ obtain weighted scores $\hat{\eta}_a$ and $\hat{\eta}_b$ such that $\hat{\eta}_a > \hat{\eta}_b$, then $C\eta_a > C\eta_b \iff \eta_a > \eta_b$ must hold.
The second approach is what is usually called the sum-score in the literature on psychometrics.
It corresponds to a particular case of the previous approach in which factors $\lambda_j = 1$ for all $j$.
From now on, we will refer to scores obtained from both approaches simply as sum-scores.

\subsection{The Area-score and its Pictorial Representation}
\label{ap:sum-area:area-score}

In a Polygrid diagram, the scores a subject obtains when assessed with a psychometric instrument are represented in an assessment chart.
Figure \ref{fig:whoqol-assessments-24-72} shows assessment charts for two subjects from the whoqol dataset.
Their scores were described in Table \ref{tab:whoqol-sample}.
The yellow polygon, which we refer to as an assessment polygon, is the main component of an assessment chart.
Its vertices are determined by the subject' scores for each domain of the WHOQOL-BREF instrument (see Appendix \ref{section:background:whoqol}).

We define the \emph{area-score of the $i$-th subject} ($\sigma_i$) as the area of the assessment polygon that represents the scores obtained by that subject.
It must be noted that the area of any assessment polygon can be easily evaluated by decomposition.
For example, the assessment polygon for Participant 030 in Figure \ref{fig:whoqol-assessments-24-72} is decomposed into four triangles:
\begin{itemize}
    \item triangle T1, defined by the scores for the psychological and environmental domains;
    \item triangle T2, defined by the scores for the environmental and physical domains;
    \item triangle T3, defined by the scores for the physical and social domains;
    \item triangle T4, defined by the scores for the social and psychological domains.
\end{itemize}

\noindent
The assessment polygon for Participant 089 has been annotated so as to remind us that the area of a triangle with sides $a$ and $b$, connected by an angle $\gamma$, is given by $\frac{\sin{\gamma}}{2}ab$.
In this example, with $\gamma=\pi/2$, the area simplifies to $\frac{1}{2}ab$.
Based on the fact that $\gamma$ depends only on the number of domains surveyed by the instrument, the area of the triangles covering an assessment polygon\footnote{
Recall from Section \ref{section:proposal:ml-task:learning} that the assessment polygon is a polygon whose vertices are scalar multiples of the roots of unity, $\zeta^d=1$,
and they are specified by construction in Algorithm \ref{alg:learning:ml:step1}.
} can be easily generalised.
In other words, since $\gamma = 2\pi/d$ for all covering triangles, the area of any such triangle is $\sigma_{ik} := \nu \, x_{ik} \, x_{i,k+1}$, with $\nu = \frac{\sin \gamma}{2}$.
Clearly, $\nu > 0$ for $d > 2$.
Consequently, the area of an assessment polygon is $\sigma_i := \sum_k \sigma_{ik} = \sum_k \nu \, x_{ik} \, x_{i,k+1} = \nu \sum_k x_{ik} \, x_{i,k+1}$, with $(k+1)$ taken modulo $d$.

\begin{figure}[t]
  \centering
  \includegraphics[width=.75\linewidth]{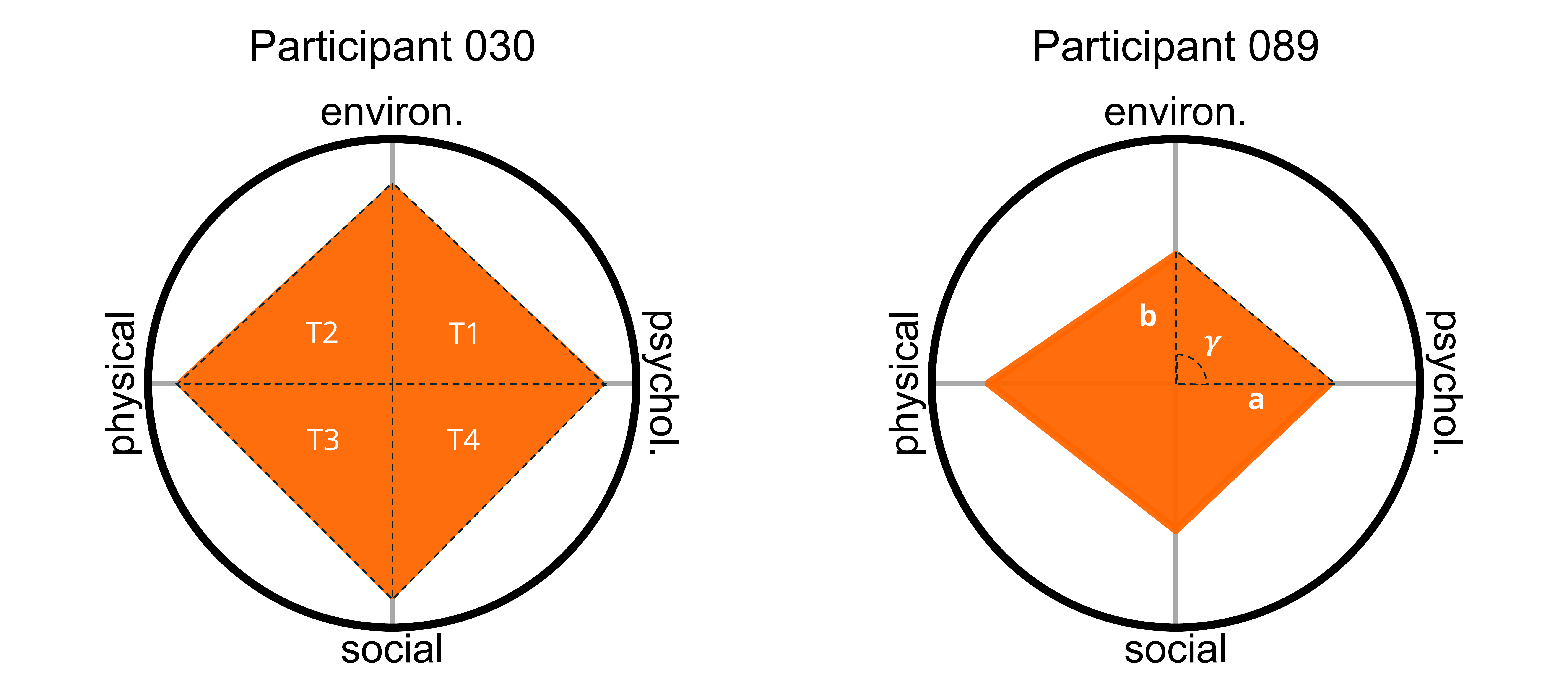}
  \caption{Assessment charts of two subjects evaluated with the WHOQOL-BREF instrument. 
  In the assessment chart for Participant 030, the assessment polygon is decomposed into four triangles, T1 to T4. 
  In the assessment polygon for Participant 089, T1 appears with two of its sides annotated as $a$ and $b$, connected by the angle $\gamma$.
  }
  \Description{}
  \label{fig:whoqol-assessments-24-72}
\end{figure}

\subsection{The Monotonic Relationship Between Sum-scores and Area-scores}
\label{ap:sum-area:proof}

We now demonstrate that, under certain assumptions, whenever $\eta_a > \eta_b$ holds, it is the case that $\sigma_{a} > \sigma_{b}$ also holds.
The assumptions are as follows:
\begin{itemize}

    \item [(a1)] \premise{The data collected with the instrument reliably fit a congeneric factor model}{reliable instrument}.
    This means that the instrument measures a unidimensional latent variable, and the congeneric model is the \aspas{mechanism} that allows us to access the latent variable.
    Formally, $\mathring{x}_{ik} = \lambda_k \eta_i + \epsilon_{ik}$;
    
    \item [(a2)] The latent variable tracks an attribute that is framed as a capacity rather than a deficit.
    \premise{Since the variable admits only unidimensional scalars, it seems a sensible choice to represent it by non-negative real numbers}{capacity is positive}.
    Formally, $\eta_i > 0, $ for all $i = 0 \ldots m-1$;
    
    \item [(a3)] The correlation between the latent variable and each of its indicators is positive, as explained in Equation \ref{eq:background:corrpos}.
    Formally, $\lambda_k > 0, $ for all $k=0 \ldots d-1$;
    
    \item [(a4)] The measurement error is negligible for the purpose of ordering on the measurand, so $\epsilon_{ik} \approx 0$.
    It must be noted that this idealised condition is approached as the instrument's reliability increases. 
    Here, we assume that McDonald's $\omega$ from Equation \ref{eq:background:omega} is the measure of reliability.

    \item [(a5)] The subscales of the instrument have the same range.
    As seen in Section \ref{section:background:whoqol}, the maximum score a person attains in any of the four domains  of the WHOQOL-BREF instrument is 5.
    Recall from Section \ref{section:proposal:data} that $\mathring{x}_{ik} \geq 0$ and $x_{ik} = \mathring{x}_{ik} / max\{\mathring{X}_{:k}\}$.
    If the instrument subscales have the same upper bound $max\{\mathring{X}_{:k}\} = C \mbox{ for all } k \in 1 \ldots d$, with $C > 0$, then it is the case that $\mathring{x}_{ak} > \mathring{x}_{bk} \implies \mathring{x}_{ak}/C > \mathring{x}_{bk}/C \implies x_{ak} > x_{bk}$.

\end{itemize}

\noindent
Under assumptions (a1) to (a5), whenever there are actual individual differences in the attribute represented by $\eta$, namely $\eta_a > \eta_b$ , it must be the case that:
\begin{align}
    \label{eq:sum-area:stt1} 
    \eta_a > \eta_b  
    \overset{a3}{\implies} \eta_a \lambda_k > \eta_b \lambda_k
    \overset{a1,a4}{\implies} \mathring{x}_{ak} > \mathring{x}_{bk}
    \overset{a5,a2}{\implies} x_{ak} > x_{bk} > 0 \,\, (\forall k \in 0 \ldots d-1) \\
    \label{eq:sum-area:stt2}
    (x_{ak} > x_{bk}) \, \wedge \, (x_{a,k+1} > x_{b,k+1})
    \implies x_{ak} x_{a,k+1} > x_{bk} x_{b,k+1}  \,\, (\forall k \in 0 \ldots d-1) \\
    \label{eq:sum-area:stt3}
    \mbox{Fix } 2\nu = sin \big(\frac{2 \pi}{d}\big) > 0. \mbox{ Then } \nu x_{ak} x_{a,k+1} > \nu x_{bk} x_{b,k+1} \implies
    \sigma_{ak} > \sigma_{bk} \,\, (\forall k \in 0 \ldots d-1) \\
    \label{eq:sum-area:stt4}
    \sigma_{ak} > \sigma_{bk} \,\, (\forall k \in 0 \ldots d-1) \implies
    \sum_k \sigma_{ak} > \sum_k \sigma_{bk} \implies
    \sigma_a > \sigma_b
\end{align}

\noindent
The statement \ref{eq:sum-area:stt1} speaks about the scores obtained by two subjects for a single domain $k$.
If subject $A$ has a higher position on the latent variable $\eta$ than subject $B$, namely $\eta_a > \eta_b$, then a similar relation holds for their scores on the $k$-th domain, $x_{ak} > x_{bk}$.
The statement \ref{eq:sum-area:stt2} expands the previous one to a neighbouring domain:
since $x_{ak} > x_{bk} > 0$ holds for the $k$-th domain, and $x_{a,k+1} > x_{b,k+1} > 0$ holds for the $(k+1)$-th domain, the inequality about the area of two rectangles formed by consecutive scores also holds: $x_{ak} x_{a,k+1} > x_{bk} x_{b,k+1}$.
The statement \ref{eq:sum-area:stt3} morphs those rectangles into the covering triangles formed by two consecutive vertices of an assessment polygon.
Using the examples in Figure \ref{fig:whoqol-assessments-24-72} with $k=0$, the statement $\nu x_{ak} x_{a,k+1} > \nu x_{bk} x_{b,k+1}$ means that the area of the triangle T1 in the assessment polygon of Participant 030 is greater than the area of the corresponding triangle in the assessment polygon of Participant 089.
Finally, the statement \ref{eq:sum-area:stt4} starts with a fact about the area of the covering triangles, $\sigma_{ak} > \sigma_{bk}$, and concludes with a fact about area-scores.

\subsection{Empirical Evidence for the Reliability of the Relationship}
\label{ap:sum-area:evidence}

We tested the reliability of the relationship between sum-scores and area-scores on our benchmark datasets.
We used unit scaled data instead of original data (in $\mathring{X}$) because, unlike WHOQOL-BREF, the AMPIAB and ELSIO1 instruments have domain scales whose range vary greatly, which violates the assumption (a5).
The test was designed as follows:
(a) for each pair of assessments $(x_a, x_b) \in X^2$, we compute their unit weighted scaled sum-scores $(s_a, s_b) := (\sum_k x_{ak}, \sum_k x_{bk})$;
(b) If $(s_a = s_b)$, the pair is discarded from the analysis.
Otherwise, the respective area-scores are computed using the same data: $(\sigma_a, \sigma_b) := (\nu \sum_k x_{ak} x_{a,k+1}, \, \nu \sum_k x_{bk} x_{b,k+1})$, the test $\ivbracket{sgn(\sigma_a - \sigma_b) = sgn(s_a - s_b)}$ is performed, and failures are recorded as violations of the relationship being investigated;
(c) The latter step is repeated once for each arrangement of the instrument's domains (or vertices order, as presented in Section \ref{section:proposal:ml-task:options}).
These are the results we found:
\begin{itemize}

    \item Dataset WHOQOL, with $m=100, d=4$ (3 arrangements).
    This setting generates $3\, \binom{100}{2} = 14,850$ pairs, of which $1266$ were discarded.
    No violations occur.

    \item Dataset ELSIO1, with $m=718, d=5$ (12 arrangements).
    This setting generates $12\, \binom{718}{2} = 3,088,836$ pairs, from which $138,552$ were discarded.
    The violation rates observed in all arrangements ranged from 0.6\% to 0.9\%, with weighted average of 0.8\%.
    Example of a pair in which a violation occurs: Participants 359 and 1111 in Figure \ref{fig:sum-area-violation:elsio1}, with $(s_a, s_b) = (4.042, 4.067) $ and $ (\sigma_a, \sigma_b) = (1.548, 1.543)$.
    This pair violates the relationship in 7 of the 12 arrangements.
    No pair violates the relationship in all 12 arrangements, but violations occur in each arrangement.
        
    \item Dataset AMPIAB, with $m=510, d=5$ (12 arrangements).
    This setting generates $12\, \binom{510}{2} = 1,557,540$ pairs, of which $63,708$ were discarded.
    Violations occur in each arrangement, and the observed violation rates ranged from 4.9\% to 7.7\%, with weighted average of 6.6\%.
    Example of a pair in which a violation occurs: Patients 8 and 26 in Figure \ref{fig:sum-area-violation:ampiab}, with $(s_a, s_b) = (1.217, 1.117)$ and $(\sigma_a, \sigma_b) = (0.059, 0.095)$.
    Unlike the example in the previous case, this pair violates the relationship in all arrangements.
    
\end{itemize}

\noindent
Based on these results, we conclude that the relationship between sum-score and area-scores is reliable in the WHOQOL and ELSIO1 datasets but less so in the AMPIAB dataset.
One main hypothesis for the larger average violation rate in AMPIAB compared to WHOQOL and ELSIO1 is that the data fit a congeneric model poorly.
In Section \ref{section:background:ampiab}, we recall that the AMPI/AB instrument achieved low scores in standard unidimensionality tests.

\begin{figure}[htpb]
  \scriptsize
  \begin{subfigure}{1\textwidth}
    \centering
    \includegraphics[width=0.75\linewidth]{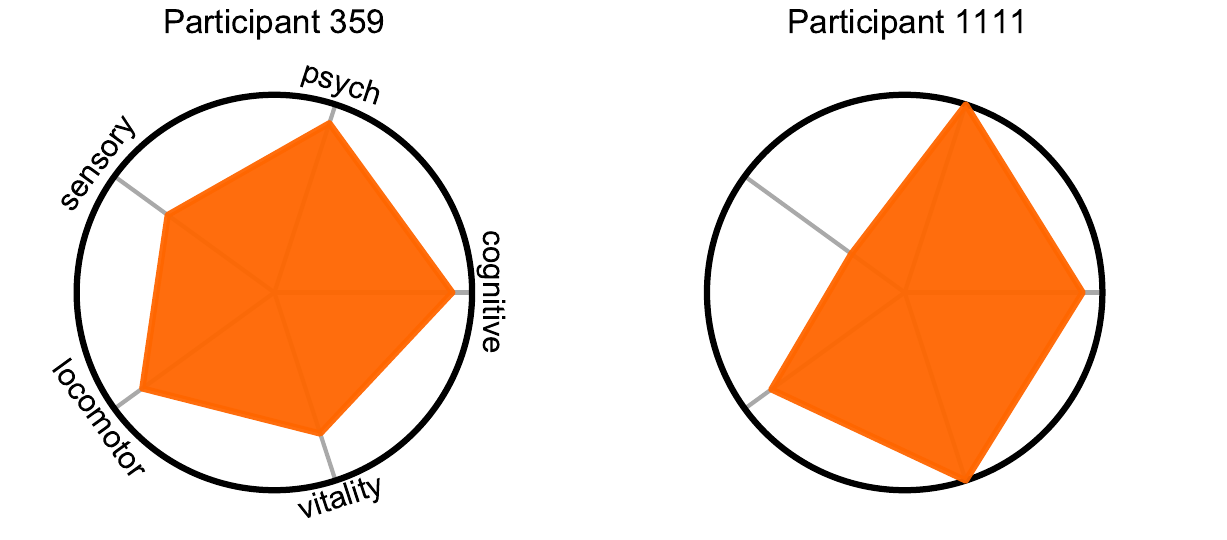}
    \caption{Two assessments violating the sum-area relationship in ELSIO1 dataset}
    \Description{}
    \label{fig:sum-area-violation:elsio1}
  \end{subfigure}
  \vfill
  \begin{subfigure}{1\textwidth}
    \centering
    \includegraphics[width=0.75\linewidth]{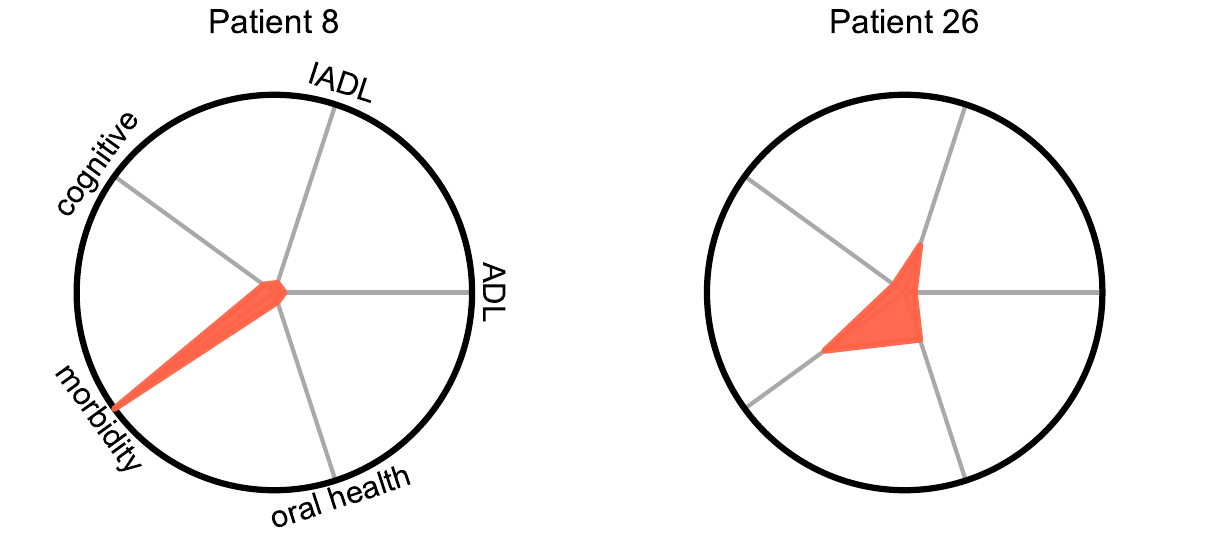}
    \caption{Two assessments violating the sum-area relationship in AMPIAB dataset}
    \Description{}
    \label{fig:sum-area-violation:ampiab}      
  \end{subfigure}
\end{figure}

\section{Detailed Results of the Offline Evaluation}
\label{appendix:offline-results}

\begin{figure}[htpb]
    \centering
    \includegraphics[width=0.86\linewidth]{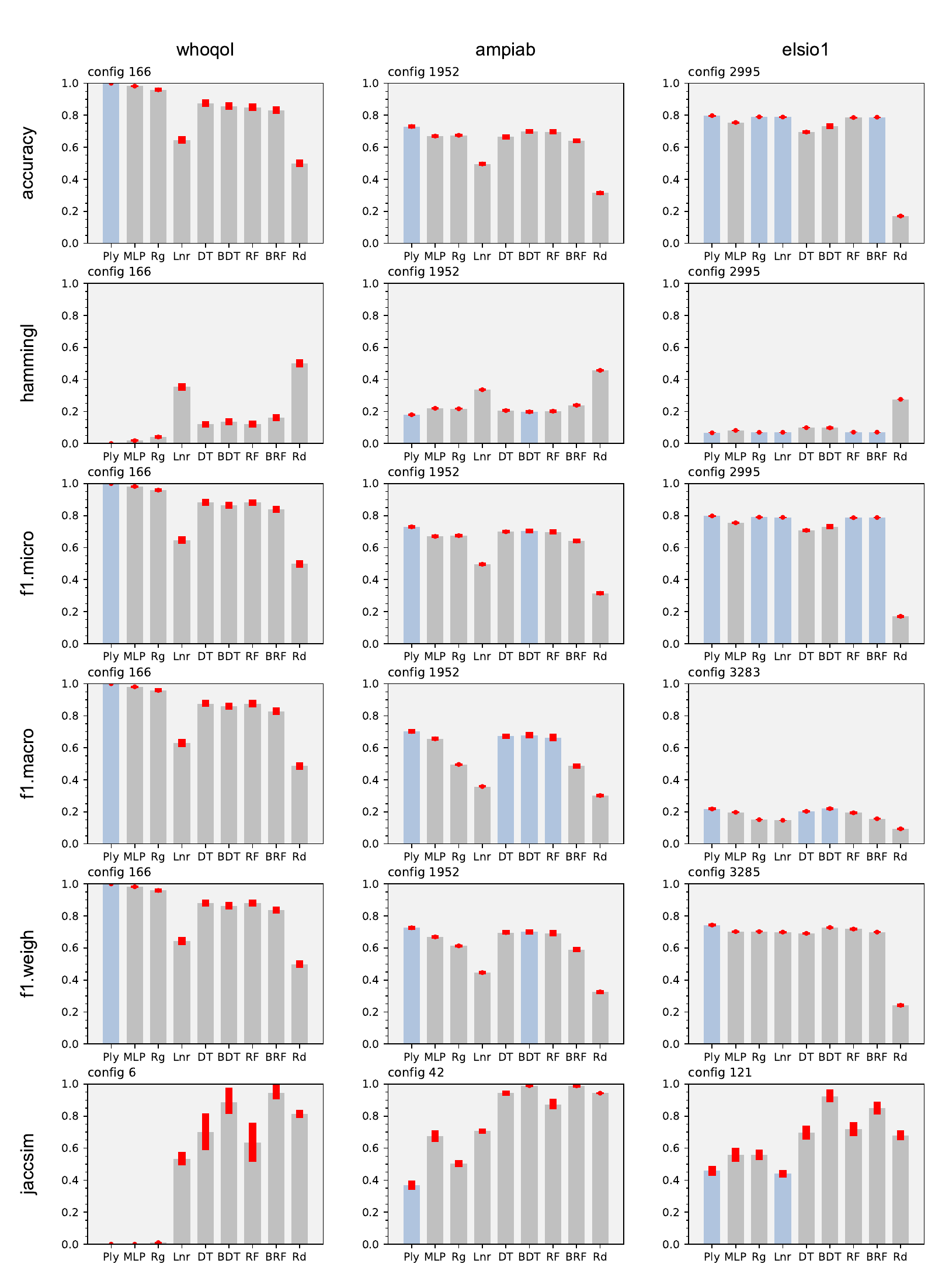}
    \caption{Results from the offline evaluation on the multiclass datasets.
    The config is identified at the top left of each chart.
    The names of some models have been abbreviated: Polygrid (Ply), Linear (Lnr), Ridge (Rg), BRDT (BDT), BRRF (BRF), and Random (Rd).
    A blue bar indicates that the measurement's confidence interval overlaps with Polygrid's.
    A grey bar means lower performance than Polygrid, and an orange bar means higher performance.
    The red rectangle at the top of a bar represents the confidence interval for that measurement.
    A worksheet with the results of the evaluation is available from our \href{https://github.com/andreplima/polygrid}{project's github repository (click here)}.
    }
    \Description{}
    \label{fig:offlineval:results:stage2:panel1}
\end{figure}

\begin{sidewaysfigure}[htpb]
    \vskip 32.5\baselineskip
    \centering
    \includegraphics[width=1.\textwidth]{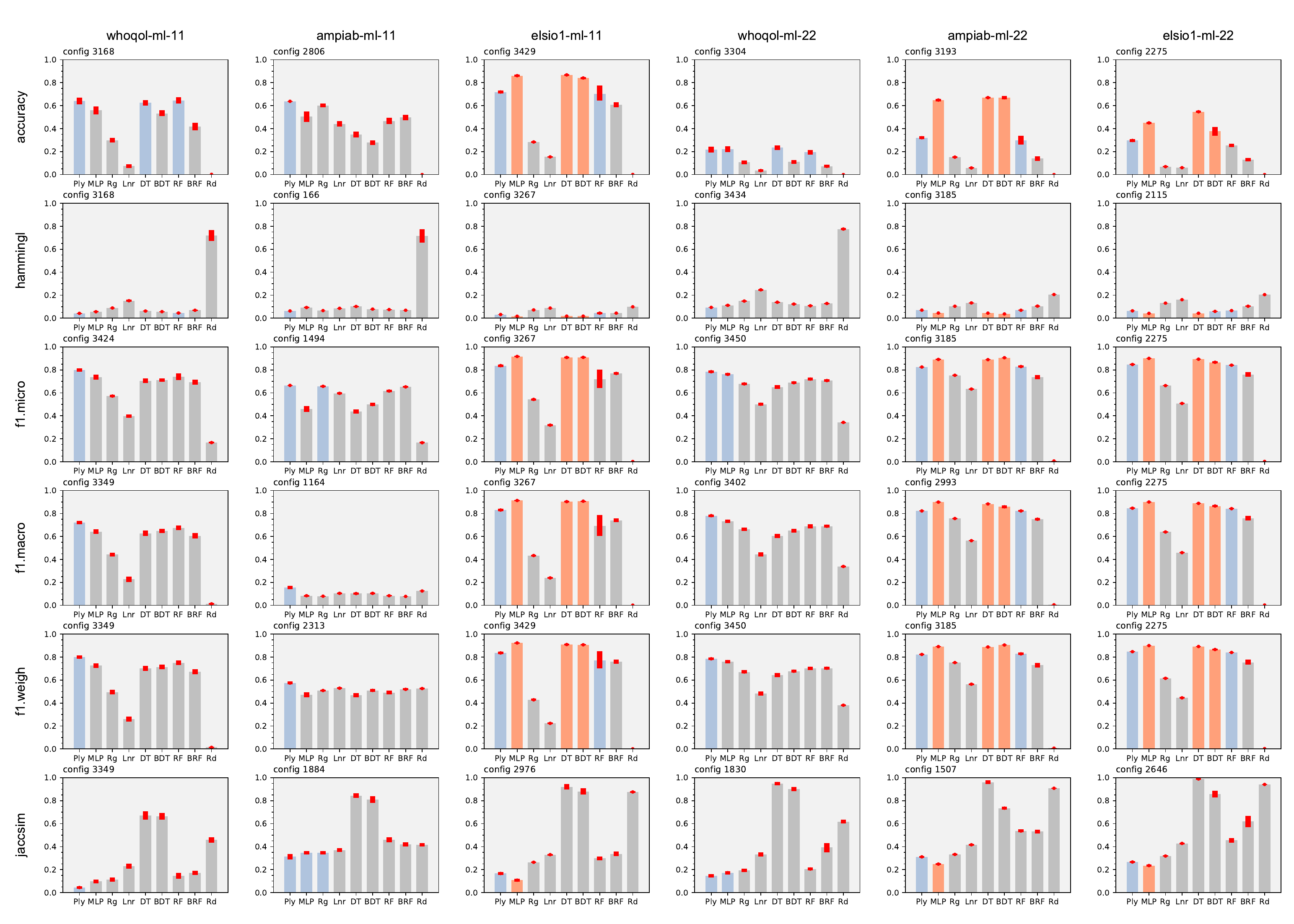}
    \caption{Results from the offline evaluation on the multilabel datasets}
    \Description{}
    \label{fig:offlineval:results:stage2:panel3}
\end{sidewaysfigure}

\begin{sidewaysfigure}[htpb]
    \vskip 19.5\baselineskip
    \centering
    \includegraphics[width=1\textwidth]{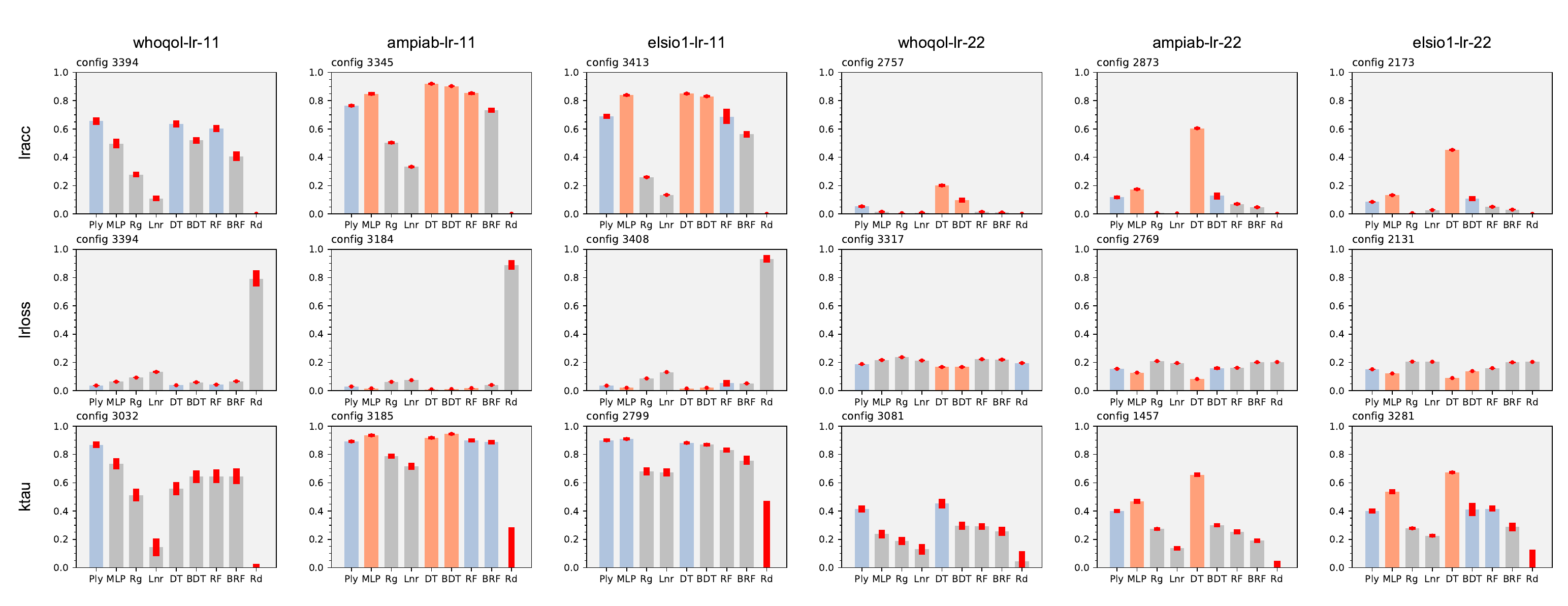}
    \caption{Results from the offline evaluation on the label ranking datasets.
    Each chart shows a comparison between Polygrid and the alternative models regarding their performance on some dataset and with respect to some metric.
    The position of a chart encodes the related dataset and metric by the column and row it occupies in the grid, respectively.
    The config is identified at the top left of each chart.
    The names of some models have been abbreviated: Polygrid (Ply), Linear (Lnr), Ridge (Rg), BRDT (BDT), BRRF (BRF), and Random (Rd).
    A blue bar indicates that the measurement's confidence interval overlaps with Polygrid's.
    A grey bar means lower performance than Polygrid, and an orange bar means higher performance.
    The red rectangle at the top of a bar represents the confidence interval for that measurement.
    A worksheet with the results of the evaluation is available from our \href{https://github.com/andreplima/polygrid}{project's github repository (click here)}.
    }
    \Description{}
    \label{fig:offlineval:results:stage2:panel5}
\end{sidewaysfigure}

\end{document}